\documentclass[10pt,twocolumn,letterpaper]{article}
\usepackage[utf8]{inputenc}
\usepackage[cjk]{kotex}

\usepackage{cvpr}              
\usepackage{graphicx}
\usepackage{amsmath}
\usepackage{amssymb}
\usepackage{booktabs}
\usepackage{tablefootnote}
\usepackage{enumitem}
\usepackage{floatrow}
\usepackage{caption}
\usepackage{adjustbox}
\usepackage{floatrow}
\usepackage{subfiles}

\usepackage[pagebackref,breaklinks,colorlinks]{hyperref}

\usepackage[capitalize]{cleveref}
\crefname{section}{Sec.}{Secs.}
\Crefname{section}{Section}{Sections}
\Crefname{table}{Table}{Tables}
\crefname{table}{Tab.}{Tabs.}

\usepackage[accsupp]{axessibility}
\usepackage{multirow}
\usepackage{makecell}
\usepackage{xcolor}
\usepackage{pifont}
\usepackage{tabularx}
\newcommand{\cmark}{\ding{51}}%
\newcommand{\xmark}{\ding{55}}

\def\cvprPaperID{10299} 
\def\confName{CVPR}
\def\confYear{2023}

\begin{document}

\title{Towards Practical Plug-and-Play Diffusion Models}

\author{Hyojun Go\thanks{Co-first autor.} \quad Yunsung Lee\footnotemark[1] \quad JinYoung Kim\footnotemark[1] \quad Seunghyun Lee \\  Myeongho Jeong\quad  Hyun Seung Lee \quad Seungtaek Choi\thanks{Corresponding author.}\vspace{-0.3mm}\\
Riiid AI Research \vspace{-0.3mm}\\
\tt\small \{hyojun.go, yunsung.lee, jinyoung.kim, seunghyun.lee\vspace{-0.3mm}\\  
\tt\small myeongho.jeong,  hyunseung.lee, seungtaek.choi\}@riiid.co
}

\newcommand{\todoc}[2]{{\textcolor{#1}{\textbf{#2}}}}
\newcommand{\todoblue}[1]{\todoc{blue}{\textbf{#1}}}
\newcommand{\todored}[1]{\todoc{red}{#1}}
\newcommand{\todobrown}[1]{\todoc{brown}{#1}}
\newcommand{\hyun}[1]{\todoc{purple}{\textbf{hyunseung:} #1}}
\newcommand{\hist}[1]{\todored{\textbf{seungtaek:} #1}}
\newcommand{\hyojun}[1]{\todoblue{\textbf{hyojun:} #1}}
\newcommand{\mh}[1]{\todobrown{\textbf{myeongho:} #1}}
\newcommand{\yunsung}[1]{\todoc{green}
{\textbf{yunsung:} #1}}

\newcommand\Tstrut{\rule{0pt}{2.2ex}}       
\newcommand\Bstrut{\rule[-0.6ex]{0pt}{0pt}} 
\newcommand{\TBstrut}{\Tstrut\Bstrut} 
\newcommand\DTstrut{\rule{0pt}{5.2ex}}       
\newcommand\DBstrut{\rule[-1.6ex]{0pt}{0pt}} 
\newcommand{\DTBstrut}{\DTstrut\DBstrut} 

\maketitle

\begin{abstract}
Diffusion-based generative models have achieved remarkable success in image generation.
Their guidance formulation allows an external model to plug-and-play control the generation process for various tasks without fine-tuning the diffusion model.
However, the direct use of publicly available off-the-shelf models for guidance fails due to their poor performance on noisy inputs.
For that, the existing practice is to fine-tune the guidance models with labeled data corrupted with noises. 
In this paper, we argue that this practice has limitations in two aspects: (1) performing on inputs with extremely various noises is too hard for a single guidance model; (2) collecting labeled datasets hinders scaling up for various tasks.
To tackle the limitations, we propose a novel strategy that leverages multiple experts where each expert is specialized in a particular noise range and guides the reverse process of the diffusion at its corresponding timesteps.
However, as it is infeasible to manage multiple networks and utilize labeled data, we present a practical guidance framework termed \textbf{P}ractical \textbf{P}lug-\textbf{A}nd-\textbf{P}lay (\textbf{PPAP}), which leverages parameter-efficient fine-tuning and data-free knowledge transfer.
We exhaustively conduct ImageNet class conditional generation experiments to show that our method can successfully guide diffusion with small trainable parameters and no labeled data.
Finally, we show that image classifiers, depth estimators, and semantic segmentation models can guide publicly available GLIDE through our framework in a plug-and-play manner. 
Our code is available at~\url{https://github.com/riiid/PPAP}.
\end{abstract}
\section{Introduction}
\label{sec:intro}
\begin{figure}[t]
    \centering
    \includegraphics[width=0.9\columnwidth]{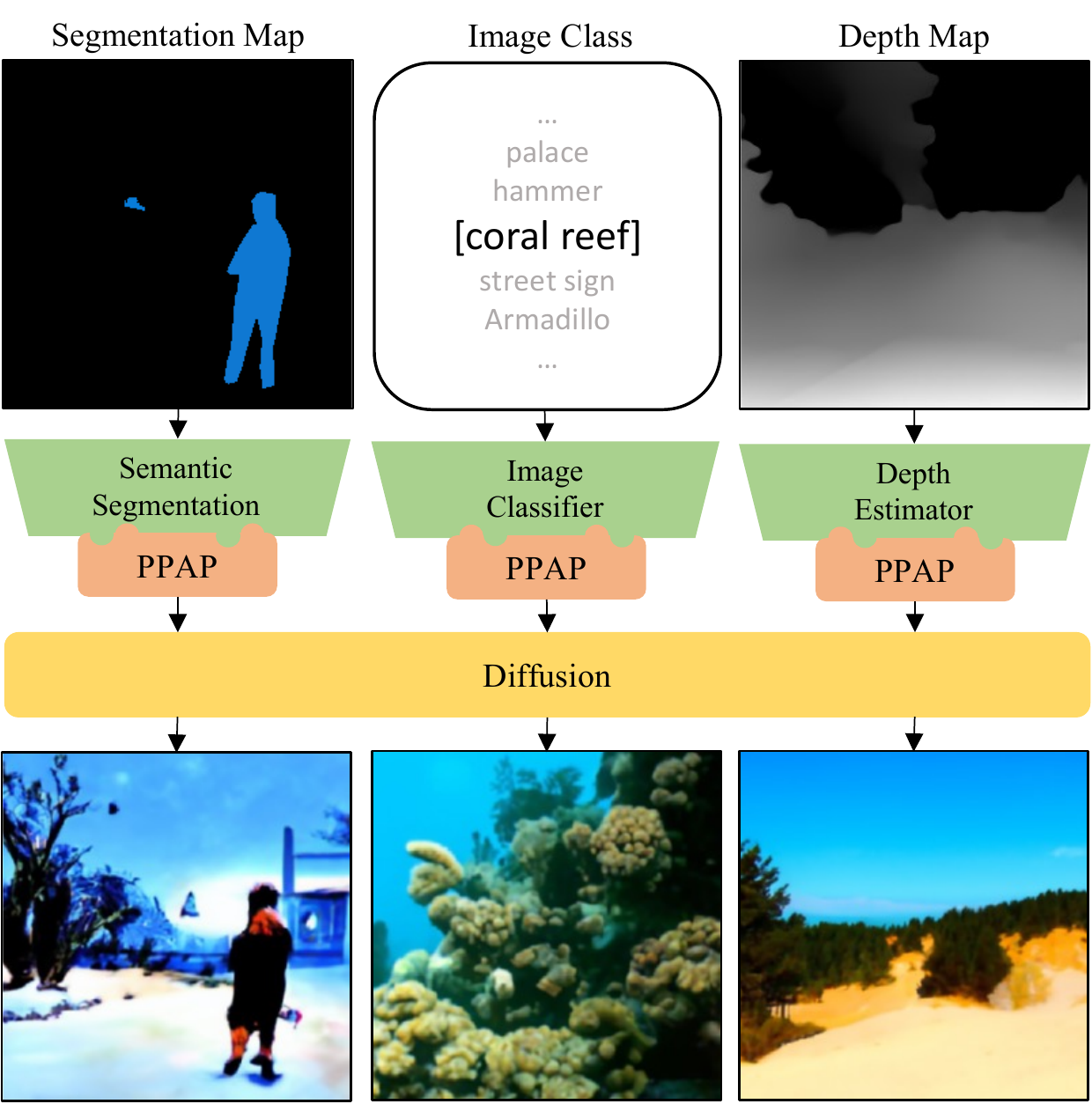}
    \vspace{-0.2cm}
    \caption{Overview of our framework. Practical Plug-And-Play (PPAP) enables the diffusion model to be guided by leveraging off-the-shelf models. Images shown below are generated by guiding the unconditional GLIDE~\cite{nichol2021glide} with DeepLabV3~\cite{chen2017rethinking}, ResNet50~\cite{he2016deep}, and MiDaS~\cite{ranftl2020towards} in a \textit{plug-and-play} manner.}
    \label{fig:my_label}
\end{figure}

Recently, diffusion-based generative models~\cite{sohl2015deep} have shown great success in various domains, including image generation~\cite{rombach2022high, gu2022vector,ruiz2022dreambooth}, text-to-speech~\cite{popov2021grad, jeong2021diff}, and text generation~\cite{li2022diffusion}. 
Specifically, for image generation, recent works have shown that diffusion models are capable of generating high-quality images comparable to those generated by GANs~\cite{dhariwal2021diffusion, goodfellow2020generative}, while not suffering from mode collapse or training instabilities~\cite{nichol2021improved}.

In addition to these advantages, their formulation allows the external model guidance~\cite{sohl2015deep,dhariwal2021diffusion, song2020score}, which guides the generation process of diffusion models towards the desired condition.
Since guided diffusion leverages external guidance models and does not require further fine-tuning of the diffusion model, it holds the potential for cheap and controllable generation in a \textit{plug-and-play} manner. 
For example, previous approaches use an image classifier for class-conditional image generation~\cite{dhariwal2021diffusion, song2020score}, a fashion understanding model for fashion image editing~\cite{kong2022leveraging}, and a vision-language model for text-based image generation~\cite{nichol2021glide,avrahami2022blended}.
From these, if the publicly available off-the-shelf model can be used for guidance, one can easily apply one diffusion to various generation tasks.

For this purpose, an existing practice is to fine-tune the external off-the-shelf model on a noisy version of the training dataset~\cite{dhariwal2021diffusion,goodfellow2020generative}, to adapt the model on the noisy latent images encountered during the diffusion process.
However, we argue that such a practice has two challenges for \textit{plug-and-play} generation:
(1) A single guidance model is insufficient to make predictions on inputs corrupted with varying degrees of noise, namely a too difficult task; and 
(2) It requires a labeled training dataset, which becomes a major hurdle whenever leveraging the off-the-shelf model.

In this paper, we first investigate the behaviors of classifiers by varying degrees of noise to understand the first challenge.
On one hand, guidance models trained on corrupted images with heavy noise categorize images based on coarse structures. 
As a result, such a model would guide the diffusion model to generate essential skeletal features. 
Meanwhile, guidance models trained on cleaner images capture finer details in the images, guiding the diffusion model to work on finishing touches.

Based on these key observations, we propose a novel multi-experts strategy that uses multiple guidance models, each fine-tuned to specialize in a specific noise region.
Despite the effectiveness of the multi-experts strategy, it should manage multiple networks and utilize the labeled data whenever applying new off-the-shelf models for various generation tasks.

For more practical $\textit{plug-and-play}$ guidance of the diffusion model with multi-experts strategy, we introduce the framework called \textbf{P}ractical \textbf{P}lug-\textbf{A}nd-\textbf{P}lay (PPAP).
First, to prevent the size of guidance models from growing prohibitively large due to the multi-experts strategy, we leverage a parameter-efficient fine-tuning scheme that can adapt off-the-shelf models to noisy images while preserving the number of parameters. 
Second, we transfer the knowledge of the off-the-shelf model on clean diffusion-generated data to the expert guidance models, thereby circumventing the need for collecting labeled datasets.

Our empirical results validate that our method significantly improves performance on conditional image generation with off-the-shelf models with only small trainable parameters and no labeled data. 
We also showcase various applications with the publicly available diffusion model, GLIDE~\cite{nichol2021glide}, by leveraging off-the-shelf image classifiers, depth estimators, and semantic segmentation models in a \textit{plug-and-play} manner.

\section{Related Work}

\noindent\textbf{Diffusion models}
Diffusion models~\cite{dhariwal2021diffusion, ho2020denoising, kingma2021variational, nichol2021improved,sohl2015deep} and score-based models~\cite{song2019generative, song2020score} are families of the generative model that generate samples from a given distribution by gradually removing noise.
Unlike other likelihood-based methods such as VAEs~\cite{kingma2013auto} or flow-based models~\cite{dinh2014nice, dinh2016density}, diffusion models have shown superior generation capabilities comparable to GANs~\cite{brock2018large, goodfellow2020generative, karras2019style}.
Although diffusion models suffer from slow generation, previous works such as DDIM~\cite{song2020denoising}, A-DDIM~\cite{bao2022analytic}, PNDM~\cite{liu2022pseudo}, and DEIS~\cite{zhang2022fast} have achieved significant acceleration in the generation process.

For conditional generation in diffusion models, classifier guidance~\cite{dhariwal2021diffusion, song2020score} and classifier-free guidance~\cite{ho2022classifier} are widely applied to various tasks~\cite{kim2022guided, nichol2021glide,rombach2022high, ho2022imagen, kreuk2022audiogen}.
Classifier guidance uses gradients of the external classifier, whereas classifier-free guidance interpolates between predictions from a diffusion model with and without labels.
However, for classifier-free guidance, diffusion models should be learned as labeled data because it requires the prediction of labels.
In this paper, we focus on the classifier guidance that freezes the unconditional diffusion model and guides it with the external model to conduct various conditional generations without labeled data in \textit{plug-and-play} manner.

\noindent\textbf{Plug-and-play generation}
Following \cite{nguyen2017plug}, we use the term \textit{plug-and-play} to refer to the capability of generating images at test time based on a condition given by a replaceable condition network without training it and generative model jointly. 
There have been various attempts for plug-and-play conditional generation in both image generation~\cite{nguyen2017plug,engel2018latent,song2021solving,kawar2021snips,kadkhodaie2021stochastic} and text generation~\cite{mai2020plug,shen2020educating,dathathri2019plug}, by binding constraints to the unconditional models, such as GAN~\cite{goodfellow2020generative}, VAE~\cite{kingma2013auto}.
These methods allow the single unconditional generative model to perform various tasks by changing the constraint model.

Most similar work to ours, Graikos~\etal~\cite{graikos2022diffusion} attempted plug-and-play on diffusion models for various tasks by directly optimizing latent images with the off-the-shelf model.
However, it fails to generate meaningful images in complex distribution as ImageNet.
Contrary to this, our method successfully guidance in complex datasets by introducing small parameters into the off-the-shelf model and making it suitable for the latent.

\section{Motivation}
\label{sec:motivation}
\subsection{Preliminaries}
\label{subsec:preliminaries}
\noindent\textbf{Diffusion models} 
Diffusion models~\cite{sohl2015deep,rombach2022high, gu2022vector,ruiz2022dreambooth} are a class of generative models that sample data by gradually denoising a random noise. 
The diffusion model comprises two stages, namely, forward and backward processes.
Forward diffusion process $q$ gradually adds noise to the data $x_{0}\sim q(x_{0})$ with some variance schedule $\beta_{t}$, as follows:
\vspace{-0.2cm}
\begin{equation}
\vspace{-0.2cm}
\label{eq:forward}
q(x_{t}|x_{t-1})=\mathcal{N}(x_{t};\sqrt{1-\beta _{t}}x_{t-1},\beta _{t}\mathbf{I}). 
\end{equation}
We repeat the forward process until reaching the maximum timestep $T$. 
Given $x_{0}$, we can directly sample $x_{t}$ as:
\vspace{-0.2cm}
\begin{equation} 
\vspace{-0.2cm}
\label{eq:diffusion}
    x_{t} = \sqrt{\alpha_{t}}x_{0} + \sqrt{1-\alpha_{t}}\epsilon, \quad \epsilon \sim \mathcal{N}(0, \mathbf{I}).
\end{equation}
where $\alpha_{t}:=\prod_{s=1}(1-\beta_{s})$. 
Note that $\sqrt{\alpha_t}$ decreases as $t$ grows, such that $\sqrt{\alpha_{T}} \approx 0$.

Reverse diffusion process starts with a random noise $x_T \sim \mathcal{N}(0, \mathbf{I})$, and produces gradually denoised samples $x_{T-1}, x_{T-2}, \dots $, until reaching the final sample $x_{0}$.
With noise predictor model $\epsilon_\theta(x_t, t)$, reverse process iteratively denoises $x_t$ with $z \sim \mathcal{N}(0, \mathbf{I})$ as follows:
\vspace{-0.2cm}
\begin{equation}
\vspace{-0.2cm}
\label{eq:reverse}
x_{t-1}= \frac{1}{\sqrt{1-\beta_{t}}}(x_{t}-\frac{\beta_{t}}{\sqrt{1-\alpha _{t}}}\epsilon _{\theta}(x_{t},t))+\sigma_{t} z, 
\end{equation}
where $\sigma_{t}^2$ is the variance of reverse process.

\noindent\textbf{Guided diffusion with external models}
Guided diffusion steers sample generation of the diffusion model by leveraging an external model~\cite{dhariwal2021diffusion, song2020score,sohl2015deep}. 
To elaborate, suppose that we have some external guidance model $f_{\phi}$ that predicts certain traits of the input, e.g., a classifier that predicts the image class. 
At each timestep $t$ of reverse diffusion, we use the guidance model to calculate the gradient on $x_{t}$ in the direction that increases the probability of $x_{t}$ having a certain desired trait $y_{target}$. 
We then update $x_{t}$ to take a step in that direction, in addition to the usual denoising update.
More formally, the reverse diffusion process (Eq.~\ref{eq:reverse}) is modified as follows:
\vspace{-0.2cm}
\begin{equation}
\vspace{-0.2cm}
\label{eq:guidance}
\begin{aligned}
    x_{t-1} = &\frac{1}{\sqrt{1-\beta _t}}(x_{t}-\frac{\beta_t}{\sqrt{1-\alpha _t}}\epsilon_{\theta}(x_{t},t))\\ & +\sigma_{t} z - s\sigma_{t}\nabla_{x_t}\mathcal{L}_{guide}(f_\phi(x_{t}), y_{target}),
\end{aligned}
\end{equation}
where $\mathcal{L}_{guide}$ and $s$ denote guidance loss and strength, respectively. 
This formulation enables external models to guide the diffusion for various tasks of interest.
For instance, for class-conditional image generation, $f_{\phi}$ is an image classifier that outputs $P_\phi(y_{target}|x_{t})$, and $\mathcal{L}_{guide}$ is given by $-\log(p_\phi(y_{target}|x_t))$.

\subsection{Observation}
\label{subsec:observation}
This section asks how na\"ive diffusion guidance schemes fail. 
Specifically, we show that when used for guided diffusion, off-the-shelf models fail due to low-confidence prediction, while models trained on data corrupted with a vast range of noise fail. 
Then, we report our major observation that classifiers trained on input corrupted with different noise levels exhibit different behaviors. 
We show that this directly affects diffusion guidance, i.e., having an expert guidance model specialized in different noise regions is crucial for successful guidance.

\noindent\textbf{Setup.} 
Our observational study involves a diffusion model and various classifiers fine-tuned on noise-corrupted data. 
For guidance classifiers, we used ResNet50~\cite{he2016deep} pre-trained on ImageNet and fine-tuned them when necessary.
We use the diffusion model trained on ImageNet $256 \times 256$ with max timesteps $T = 1000$ in~\cite{dhariwal2021diffusion}.
To generate noisy versions of the data, we perform a forward diffusion process, i.e., given an input $x_{0}$, we obtain $x_{t}$ (Eq.~\ref{eq:diffusion}) for $t=1,\dots, T$. 
We use the DDIM sampler~\cite{song2020denoising} with 25 steps, using the classifier under consideration to guide the diffusion model. 
More details are provided in Appendix. \textcolor{red}{B}.

\begin{figure}[t]
    \centering
    \includegraphics[width=1.0\columnwidth]{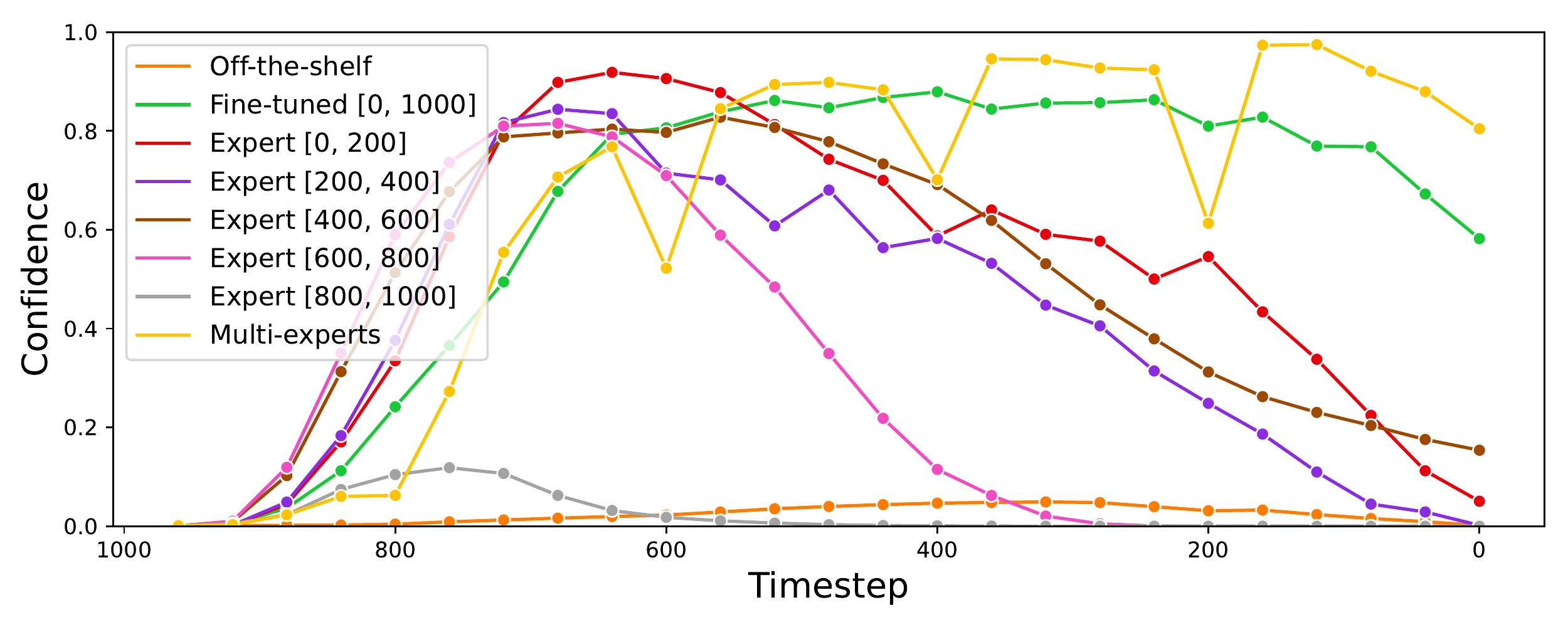}
    \vspace{-0.8cm}
    \caption{Classifier confidence during the reverse process.
    The off-the-shelf model does not increase confidence, showing that it cannot guide the diffusion to its confident region.
    }
    \label{fig:confidence}
\end{figure}

\begin{figure}[t]
    \centering
    \begin{subfigure}[b]{0.15\columnwidth}
        \includegraphics[clip,width=\textwidth]{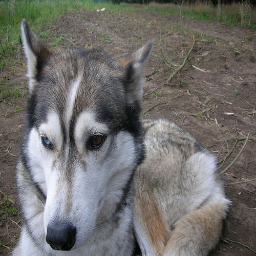}
        \caption{Clean}
        \label{subfig:grad_off-the-shelf}
    \end{subfigure}
    \begin{subfigure}[b]{0.15\columnwidth}
        \includegraphics[clip,width=\textwidth]{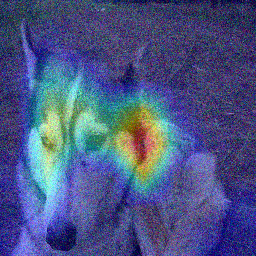}
        \caption{$f_{\phi_1}$}
        \label{subfig:grad_f_phi_1}
    \end{subfigure}
    \begin{subfigure}[b]{0.15\columnwidth}
        \includegraphics[clip,width=\textwidth]{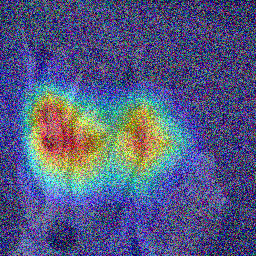}
        \caption{$f_{\phi_2}$}
        \label{subfig:grad_f_phi_2}
    \end{subfigure}
    \begin{subfigure}[b]{0.15\columnwidth}
        \includegraphics[clip,width=\textwidth]{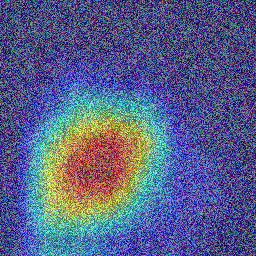}
        \caption{$f_{\phi_3}$}
        \label{subfig:grad_f_phi_3}
    \end{subfigure}
    \begin{subfigure}[b]{0.15\columnwidth}
        \includegraphics[clip,width=\textwidth]{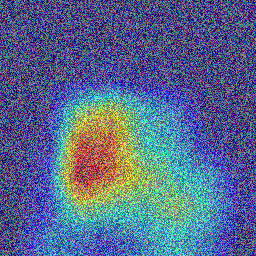}
        \caption{$f_{\phi_4}$}
        \label{subfig:grad_f_phi_4}
    \end{subfigure}
    \begin{subfigure}[b]{0.15\columnwidth}
        \includegraphics[clip,width=\textwidth]{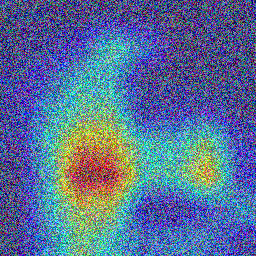}
        \caption{$f_{\phi_5}$}
        \label{subfig:grad_f_phi_5}
    \end{subfigure}
    \caption{
    Grad-CAM visualization of each expert on corrupted images by forward process.
    Experts trained on larger and smaller timestep tend to focus on coarse and fine features, respectively.}
    \label{fig:gradcam}
    
    \vspace{-0.2cm}
\end{figure}

\begin{figure}[t]
    \centering
    \begin{tabular}{@{}c}
        \raisebox{0.65\height}{\rotatebox{90}{\scriptsize Na\"ive}} 
        \raisebox{0\height}{\rotatebox{90}{\scriptsize off-the-shelf}}
        \adjincludegraphics[clip,width=0.92\textwidth,trim={0 0 0 0}]{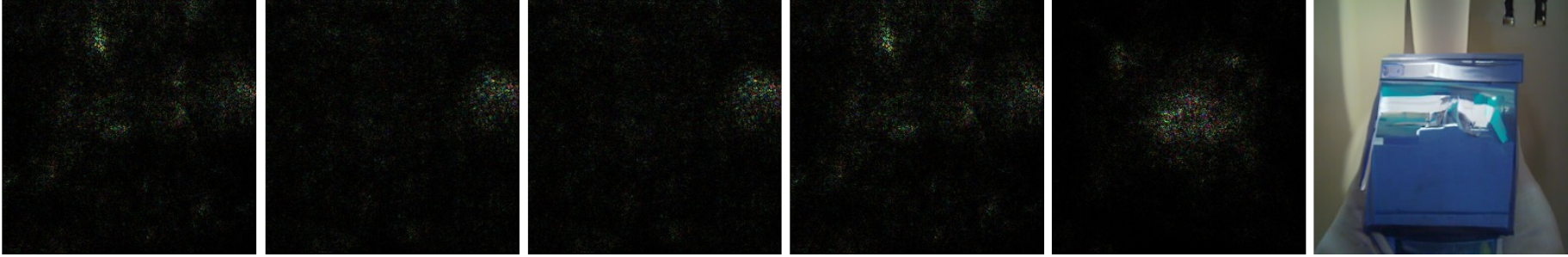} \\
        
        \raisebox{0.7\height}{\rotatebox{90}{\scriptsize Single}}
        \raisebox{0.05\height}{\rotatebox{90}{\scriptsize noise-aware}} 
        \adjincludegraphics[clip,width=0.92\textwidth]{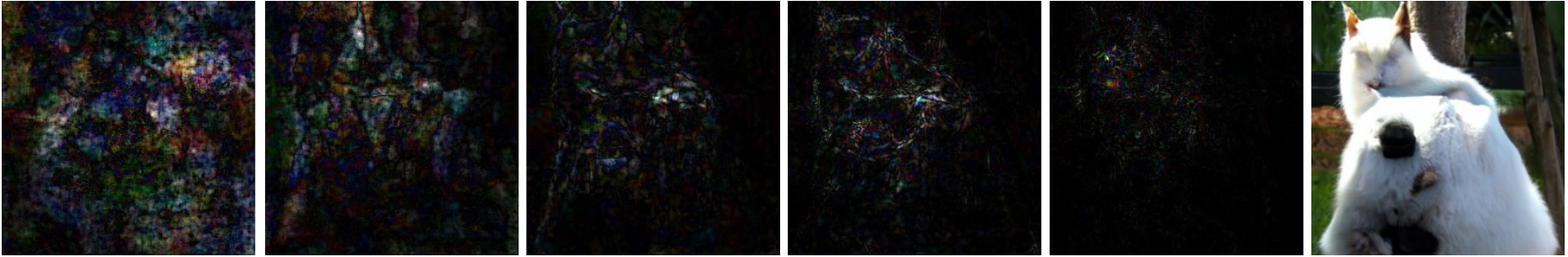} \\
        
        \raisebox{0.45\height}{\rotatebox{90}{\scriptsize Expert}} 
        \raisebox{0.4\height}{\rotatebox{90}{\scriptsize [0,200]}} 
        \adjincludegraphics[clip,width=0.92\textwidth]{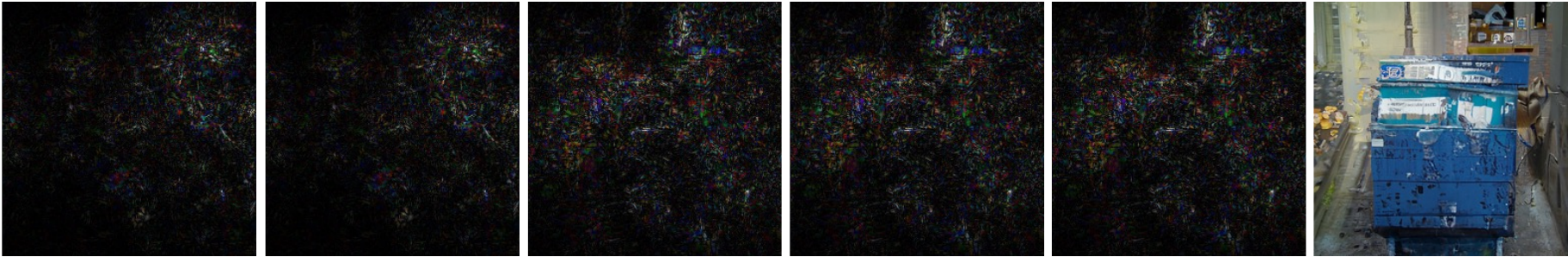} \\
        
        \raisebox{0.45\height}{\rotatebox{90}{\scriptsize Expert}} 
        \raisebox{0.2\height}{\rotatebox{90}{\scriptsize  [200,400]}} 
        \adjincludegraphics[clip,width=0.92\textwidth]{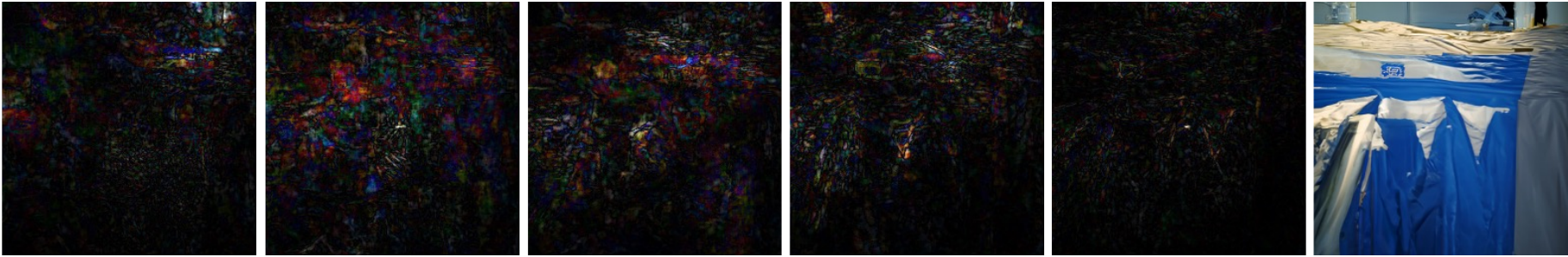} \\
        
        \raisebox{0.45\height}{\rotatebox{90}{\scriptsize Expert}} 
        \raisebox{0.2\height}{\rotatebox{90}{\scriptsize  [400,600]}} 
        \adjincludegraphics[clip,width=0.92\textwidth]{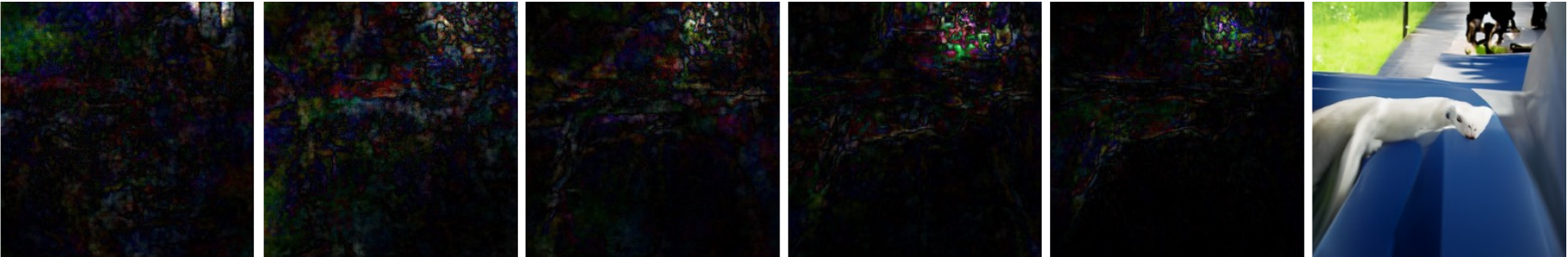} \\
        
        \raisebox{0.45\height}{\rotatebox{90}{\scriptsize Expert}} 
        \raisebox{0.2\height}{\rotatebox{90}{\scriptsize  [600,800]}} 
        \adjincludegraphics[clip,width=0.92\textwidth]{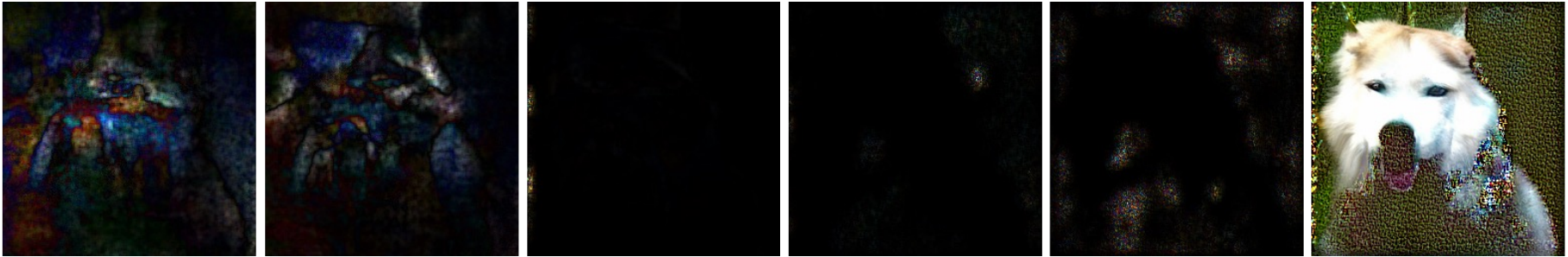} \\
        
        \raisebox{0.6\height}{\rotatebox{90}{\scriptsize Expert}} 
        \raisebox{0.1\height}{\rotatebox{90}{\scriptsize  [800,1000]}} 
        \adjincludegraphics[clip,width=0.92\textwidth]{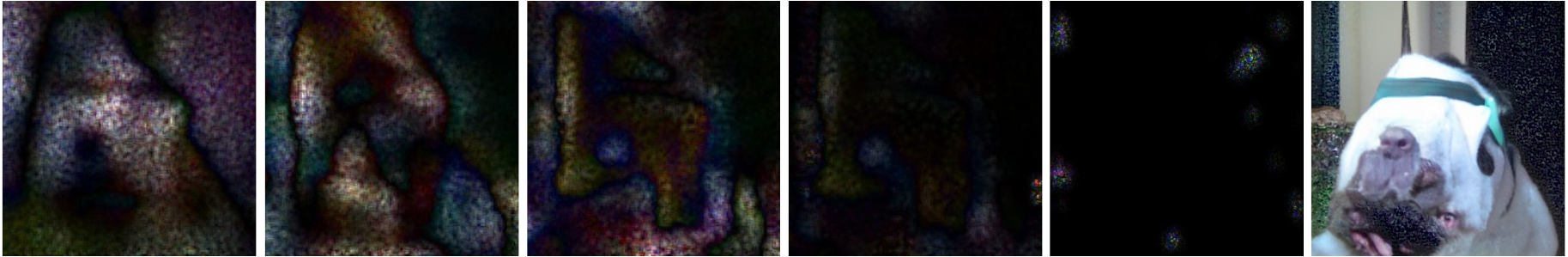} \\
        
        \raisebox{0.75\height}{\rotatebox{90}{\scriptsize Multi}} 
        \raisebox{0.25\height}{\rotatebox{90}{\scriptsize  experts-5}} 
        \adjincludegraphics[clip,width=0.92\textwidth]{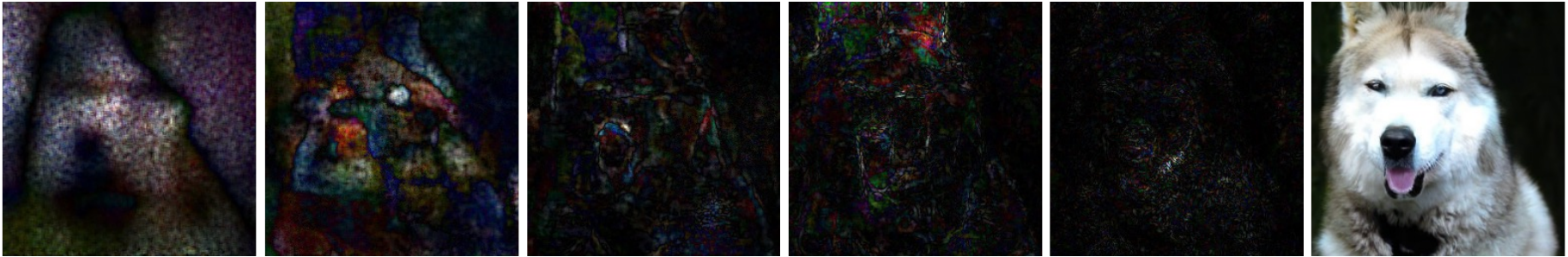}
    \end{tabular}

\vspace{-0.2cm}
    \caption{
    Gradient maps for $x_t$ on $t \in [920, 720, 520, 320, 120]$ (left 5) and generated images (rightmost) when the reverse process is guided to husky from the same initial noise.
    Classifier trained on smaller noise and larger noise tends to modify finer details and coarser structure, respectively.} 
    \label{fig:gradient}
\end{figure}

\noindent\textbf{Na\"ive external model guidance is not enough.}
Here, we investigate failure cases of na\"ive diffusion guidance. 
First, we try diffusion guidance with an off-the-shelf ResNet50 classifier toward an ImageNet class label.
As we can see in the first row of  Fig.~\ref{fig:gradient}, the model fails to provide a meaningful gradient for diffusion guidance.
This is because an off-the-shelf model outputs low-confidence, high-entropy prediction on the out-of-distribution, noisy latent encountered throughout the reverse process as in Fig.~\ref{fig:confidence}.

We also experimented with a ResNet50 classifier fine-tuned on data encountered throughout the forward diffusion process, i.e., $x_{t}$ (Eq.~\ref{eq:forward}) for all $t \in [1, \dots, 1000]$, where $x_{0}$ corresponds to a clean image.
This works better than using a na\"ive off-the-shelf model, as observed in the improved FID ($38.74\rightarrow30.42$) and IS ($33.95\rightarrow43.05$) scores (see Section~{\ref{sec:Exp}} for more details). 
However, again, as seen in Fig~\ref{fig:confidence}, classifier confidence drops for cleaner images ($t\approx 200$), leading to failure cases as in the second row of Fig.~\ref{fig:gradient}.

\noindent\textbf{Behavior of classifier according to learned noise}
To understand the failure of the single noise-aware model, we investigate the behavior of classifiers fine-tuned on specific noise level, i.e., $x_{t}$ for $t\in [a, b] \subset [0, T]$ for some suitable $a > 0$ and $b < T$.

Specifically, we fine-tuned five ResNet50 classifiers $f_{\phi_{i}}$, $i \in \{1, \dots, 5\}$, where $f_{\phi_{i}}$ is trained on noisy inputs $x_{t}$, $t \in \{(i-1)\cdot 200, \dots, i\cdot 200\}$.
We first observe that each $f_{\phi_{i}}$ behave differently via Grad-CAM~\cite{selvaraju2017grad}.
For example, as shown in Fig.~\ref{fig:gradcam}, $f_{\phi_{1}}$ and $f_{\phi_{2}}$ trained on cleaner images predict ‘husky’ based on distinctive canine features such as its eyes or coat pattern.
Meanwhile, $f_{\phi_{4}}$ and $f_{\phi_{5}}$ trained on noisy images make predictions based on overall shape (albeit imperceptible to human eyes).

Such behavior difference manifests when we guide diffusion using different $f_{\phi_{i}}$.
For example, $f_{\phi_{5}}$ trained on noisy images initially generates a husky-like shape but fails to fill in finer details.
On the other hand, $f_{\phi_{1}}$ trained on cleaner images seems to be focusing on generating specific details such as hairy texture but fails to generate a husky-like image due to lack of an overall structure.

These classifiers' behaviors coincide with the previous perspective; the unconditional diffusion focuses on the overall structure and finer details in larger and smaller noise, respectively~\cite{choi2022perception}.
Considering this, we hypothesize that the classifier can guide diffusion at the specific noise level by learning that noise level.

\subsection{Multi-Experts Strategy}
From the above observation, we propose a multi-experts strategy that each expert is fine-tuned to specialize in a specific noise range.
Suppose we are given clean dataset $\{(x_{0}, y)\}$ and maximum diffusion timestep $T$. 
We train $N$ expert guidance model, where the $n$-th expert $f_{\phi_{n}}$ is trained to predict the ground-truth label $y$ given noisy data $x_{t}$, $t \in \{ \frac{n-1}{N}T, \dots, \frac{n}{N}T \}$.
Then, during the reverse diffusion process, we assign an appropriate guidance model depending on the timestep, i.e., $n$ for which $t\in \{\frac{n-1}{N}T + 1, \dots, \frac{n}{N}T  \}$.
More formally, model guidance (Eq.~\ref{eq:guidance}) can be rewritten as:
\begin{equation}
\label{eq:multi_expert_guidance}
\begin{aligned}
    x_{t-1} = &\frac{1}{\sqrt{1-\beta _t}}\big(x_t-\frac{\beta_t}{\sqrt{1-\alpha _t}}\epsilon _{\theta}(x_t,t)\big)\\ 
    & +\sigma _t z - s\sigma_t\nabla_{x_t}\mathcal{L}_{guide}
    \big(f_{\phi_n}(x_t), y \big), 
\end{aligned}
\end{equation}
where $n$ is such that $t\in \{\frac{n-1}{N}T + 1, \dots , \frac{n}{N}T  \}$. Note that this strategy does not incur additional model inference time costs, since only one external guidance model is used depending on the reverse process timestep $t$.

In our observational study in Section~\ref{subsec:observation}, multi-experts guide coarse structure in larger timestep and fine-details in smaller time step, resulting in successful generation for the husky image as shown in Fig.~\ref{fig:gradient}.

\section{Practical Plug-and-Play Diffusion}
\label{Sec:Methodology}

\begin{figure}[t]
    \centering
    \includegraphics[width=\columnwidth]{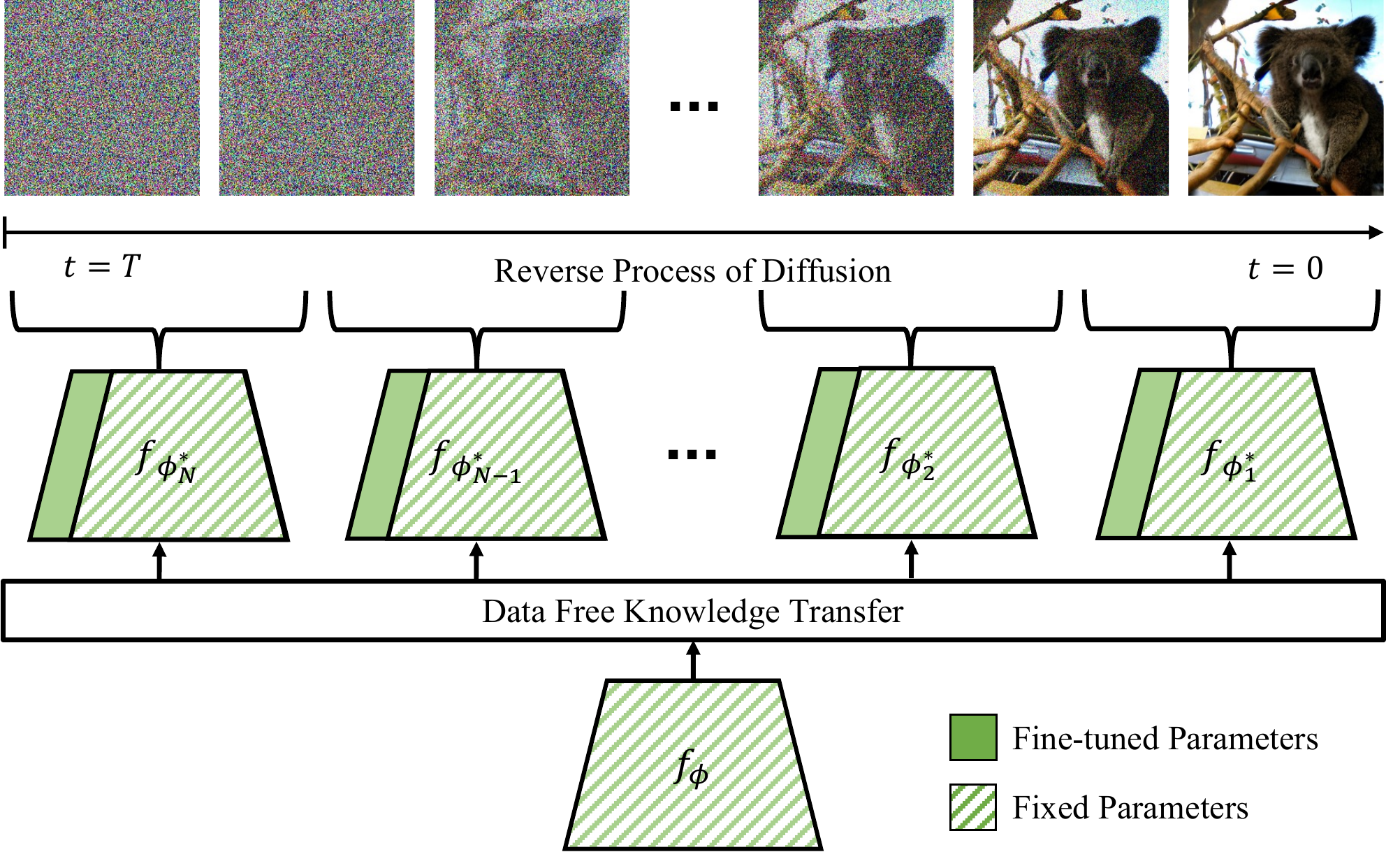}
    \caption{Overview of our method.
    We use parameter-efficient multi-experts that each expert is specialized in a specific noise range.
    We transfer the knowledge of the off-the-shelf model to each expert, thereby bypassing the need for labeled data.
    During the reverse process, we only need to switch the added training parameters accordingly depending on the noise region.
    }
    \label{fig:method}
\end{figure}

Whenever applying a new off-the-shelf model, the multi-experts strategy must utilize multiple networks and collect the labeled dataset.
To deal with this impracticality, we propose a plug-and-play diffusion guidance framework: \textbf{P}ractical \textbf{P}lug-\textbf{A}nd-\textbf{P}lay (PPAP), which takes a multi-experts strategy with the following two components as shown in Fig.~\ref{fig:method}: 
\textbf{(1)} We introduce a parameter-efficient fine-tuning scheme based on parameter sharing in order to prevent the size of guidance models from growing prohibitively large; 
\textbf{(2)} We propose to use a knowledge transfer scheme that transfers an off-the-shelf model's knowledge on clean diffusion-generated data to expert guidance models, thereby bypassing the need for a labeled dataset.

\subsection{Parameter Efficient Multi-Experts Strategy}
One limitation of the proposed multi-experts strategy is that, as the number of guidance models increases $N$-fold, the number of parameters to fine-tune increases $N$-fold.
To tackle this issue, we use a parameter-efficient strategy that only fine-tunes a small number of parameters while reusing most of the frozen off-the-shelf model.
Specifically, we fine-tune bias and batch normalization, and apply LORA~\cite{hu2021lora} to certain weight matrices of the off-the-shelf model.
Since this method does not change architecture such as extending model depth, we do not introduce additional inference time cost.
We denote $n$-th expert as $f_{\phi_n^*}$ to distinguish it from off-the-shelf model $f_\phi$.

During the reverse process of the diffusion model, we only need to switch the added training parameters accordingly depending on the noise region, while reusing the off-the-shelf backbone model.
More architectural details are provided in Appendix. \textcolor{red}{C}.

\subsection{Data Free Knowledge Transfer}
\label{subsec:data-free}
So far, we have assumed that we can access the dataset $\{(x_{0}, y)\}$ that was used to train the guidance model.
For a practical \textit{plug-and-play} generation, applying guidance with an off-the-shelf model should be possible without obtaining a labeled dataset suitable for each task.

We here propose to generate the clean dataset $\{ \tilde{x}_{0} \}$ using the diffusion model, then use it to train expert guidance models.
Our underlying assumption is that, by mimicking the prediction of an off-the-shelf model on a clean image, the expert can operate in the noise domain to some extent.
Namely, we treat the off-the-shelf model $f_{\phi}$ as a teacher, and use its prediction on clean data to serve as labels when training expert guidance models $f_{\phi_{n}}$. 
Formally, we formulate the knowledge transfer loss as:
\vspace{-0.1cm}
\begin{equation}
\begin{aligned}
  \mathcal{L}_{KT} &= \mathbb{E}_{t\sim \mathrm{unif}\{ \frac{n-1}{N}T, \dots, \frac{n}{N}T \}}\big[\mathcal{L}\big(\texttt{sg}(f_\phi(\tilde{x}_0)), f_{\phi_n^*}(\tilde{x}_{t})\big)\big],
\end{aligned}
\end{equation}
where $\texttt{sg}(\cdot)$ is the stop-gradient operator and $\mathcal{L}$ is a task-specific loss function. 
With this formulation, we can easily adapt our method to various tasks of interest, including image classification, monocular depth estimation, and semantic segmentation, by just using different loss functions. 
Due to space limitations, here we describe how we plug and play an image classifier only. 
Further details for other tasks can be found in Appendix. \textcolor{red}{D}.

\noindent\textbf{Image classification}
An image classifier takes an image as input and outputs a logit vector of the form $f(x)\in\mathbb{R}^C$, where $C$ is the number of image classes.
We formulate knowledge transfer loss $\mathcal{L}_{clsf}$ for classifiers as:
\begin{equation}
    \mathcal{L}_{clsf} = D_{KL}\big(\texttt{sg}\big(\texttt{s}(f_\phi(\tilde{x}_0) / \tau)\big), \texttt{s}(f_{\phi_n^*}(\tilde{x}_{t}))\big),
\end{equation}
where $\texttt{s}$ is the softmax operator, $\tau$ the temperature hyperparmeter, and $D_{KL}(\cdot)$ the Kullback-Leibler divergence.

\section{Experiments}
\label{sec:Exp}

In this section, we validate the effectiveness of our framework PPAP by showing that it can guide unconditional diffusion without collecting labeled data.
Specifically, we first conducted various experiments in image classifier guidance to unconditional diffusion model trained on ImageNet dataset~\cite{deng2009imagenet}.
Then, we present the applicability of the image classifier, depth estimation model, and semantic segmentation model to the unconditional version of GLIDE~\cite{nichol2021glide} trained on a large dataset containing various domains.

\subsection{ImageNet Classifier Guidance}
\label{sec:imagenet_guidance}

We conduct experiments on ImageNet class conditional generation to validate the effectiveness of our framework.

\begin{table*}[t!]
\begin{center}
\resizebox{0.94\textwidth}{!}{
\begin{tabular}{c c lcc cccc}
  \hline
  \noalign{\hrule height0.8pt} 
  \multicolumn{1}{l}{Architecture} & \multicolumn{1}{l}{Sampler} & \multicolumn{1}{l}{Guidance} & \multicolumn{1}{c}{\shortstack{Trainable \\ Parameters}} & \multicolumn{1}{c}{Supervision} & \multicolumn{1}{c}{FID ($\downarrow$)} & \multicolumn{1}{l}{IS ($\uparrow$)} & \multicolumn{1}{c}{Precision ($\uparrow$)} & \multicolumn{1}{c}{Recall} \DTBstrut \\
  \hline\noalign{\hrule height0.8pt} 
  \multicolumn{1}{l}{\multirow{14}{*}{ResNet50}} 
  & \multicolumn{1}{l}{\multirow{7}{*}{\shortstack{DDIM \\ (25 Steps)}}} 
    & No & - & None & 40.24 & 34.53 & 0.5437 & 0.6063   \\
  & & Na\"ive off-the-shelf & - & None & 38.74 & 33.95 & 0.5192 & 0.6152   \\
  & & Gradients on $\hat{x}_0$ & - & None & 38.14 & 33.77 & 0.5277 & 0.6252   \\
  & & Single noise aware & 25.5M (100\%) & ImageNet ($\approx$ 1.2M) & 30.42 & 43.05 & 0.5509 & 0.6187   \\
  \cline{3-9}
  & & Multi-experts-5 & 127.5M (500\%) & ImageNet ($\approx$ 1.2M) & \textbf{19.98} & \textbf{74.78} & \textbf{0.6476} & 0.5887   \\
  & & PPAP-5 & 7.3M (28.6\%) & Data-free ($\approx$ 0.5M) & 29.65 & 44.23 & 0.5872 & 0.6012   \\
  & & PPAP-10 & 14.6M (57.2\%) & Data-free ($\approx$ 0.5M) & \underline{27.86} & \underline{46.74} & \underline{0.6079} & 0.5925   \\
  
  \cmidrule[1pt]{2-9}
  & \multicolumn{1}{l}{\multirow{7}{*}{\shortstack{DDPM \\ (250 Steps)}}} 
    & No & - & None & 28.97 & 40.34 & 0.6039 & 0.6445   \\
  & & Na\"ive off-the-shelf & - & None & 29.03 & 39.79 & 0.6042 & 0.6474   \\
  & & Gradients on $\hat{x}_0$ & - & None & 28.81 & 39.80 & 0.6095 & 0.6475   \\
  & & Single noise aware & 25.5M (100\%) & ImageNet ($\approx$ 1.2M) & 38.15 & 31.29 & 0.5426 & 0.6321   \\
    \cline{3-9}
  & & Multi-experts-5 & 127.5M (500\%) & ImageNet ($\approx$ 1.2M) & \textbf{16.37} & \textbf{81.47} & \textbf{0.7216} & 0.5805   \\

  & & PPAP-5 & 7.3M (28.6\%) & Data-free ($\approx$ 0.5M) & 22.70 & 52.74 & 0.6338 & 0.6187   \\
  & & PPAP-10 & 14.6M (57.2\%) & Data-free ($\approx$ 0.5M) & \underline{21.00} & \underline{57.38} & \underline{0.6611} & 0.5996   \\

   
  \hline\noalign{\hrule height0.8pt} 
  \multicolumn{1}{l}{\multirow{14}{*}{DeiT-S}} 
  & \multicolumn{1}{l}{\multirow{7}{*}{\shortstack{DDIM \\ (25 Steps)}}} 
    & No & - & None & 40.24 & 34.53 & 0.5437 & 0.6063   \\
  & & Na\"ive off-the-shelf & - & None & 37.51 & 33.74 & 0.5293 & 0.6186   \\ 
  & & Gradients on $\hat{x}_0$ & - & None & 38.10 & 33.75 & 0.5288 & 0.6212   \\
  & & Single noise aware & 21.9M (100\%) & ImageNet ($\approx$ 1.2M) & 44.13 & 28.31 & 0.4708 & 0.6030   \\
    \cline{3-9}
  & & Multi-experts-5 & 109.9M (500\%) & ImageNet ($\approx$ 1.2M) & \textbf{17.06} & \textbf{80.85} & \textbf{0.7001} & 0.5810   \\
  & & PPAP-5 & 4.6M (21.3\%) & Data-free ($\approx$ 0.5M) & 25.98 & 48.80 & 0.6128 & 0.5984   \\
  & & PPAP-10 & 9.3M (42.6\%) & Data-free ($\approx$ 0.5M) & \underline{24.77} & \underline{50.56} & \underline{0.6220} & 0.5990   \\

  \cmidrule[1pt]{2-9}
  & \multicolumn{1}{l}{\multirow{7}{*}{\shortstack{DDPM \\ (250 Steps)}}} 
    & No & - & None & 28.97 & 40.34 & 0.6039 & 0.6445   \\
  & & Na\"ive off-the-shelf & - & None & 29.41 & 39.55 & 0.6032 & 0.6320   \\
  & & Gradients on $\hat{x}_0$ & - & None & 30.26 & 37.75 & 0.6043 & 0.6407   \\
  & & Single noise aware & 21.9M (100\%) & ImageNet ($\approx$ 1.2M) & 36.01 & 31.90 & 0.5461 & 0.6479   \\
  \cline{3-9}
  & & Multi-experts-5 & 109.9M (500\%) & ImageNet ($\approx$ 1.2M) & \textbf{14.95} & \textbf{83.26} & \textbf{0.7472} & 0.5686   \\
  & & PPAP-5 & 4.6M (21.3\%) & Data-free ($\approx$ 0.5M) & 22.30 & 53.62 & 0.6368 &  0.6074   \\
  & & PPAP-10 & 9.3M (42.6\%) & Data-free ($\approx$ 0.5M) & \underline{20.07} & \underline{60.62} & \underline{0.6734} & 0.5963   \\
   
  \hline\noalign{\hrule height0.8pt} 
\end{tabular}}
\end{center}
\vspace{-0.3cm}
\caption{Overall Results on ResNet50 and DeiT-S.
Note that ``No'' Guidance shows the same performance as it is unconditional generation. \textbf{Bold} denotes the best performance and \underline{underline} indicates the second best performance. Multi-experts-5 and PPAP-10, which showed the best and second-best results in all cases, are both our proposed models. 
}
\vspace{-2mm}
  
\label{tab:main}
\end{table*}

\noindent\textbf{Experimental Setup}
Based on ImageNet pre-trained unconditional ADM~\cite{dhariwal2021diffusion} with 256$\times$256 size, we used two mainstream architectures: (1) CNN-based classifier ResNet50~\cite{he2016deep} and 2) transformer-based classifier DeiT-S~\cite{touvron2021training}.
For each architecture, we used the following variants to serve the guidance model:
\begin{itemize}[noitemsep, nolistsep]
    \item \textbf{Na\"ive off-the-shelf}: ImageNet pre-trained model is used without further training on noise.
    \item \textbf{Single noise aware}: The model is fine-tuned on corrupted data in the whole noise range.
    \item \textbf{Multi-experts-$N$}: We fine-tune $N$ expert guidance models in a supervised manner without applying parameter-efficient tunning. 
    \item \textbf{PPAP-$N$}: $N$ experts are parameter-efficiently knowledge transferred with generated images.
\end{itemize}
For data-free knowledge transfer (Section~\ref{subsec:data-free}), we generate 500k images, which is much less than the ImageNet dataset.
We use DDPM sampler~\cite{ho2020denoising} with 250 steps and DDIM sampler~\cite{song2020denoising} with 25 steps as in~\cite{dhariwal2021diffusion}.
We set the guidance scale $s$ as 7.5 since it achieves good results for most variants.

We also compare other methods which can be applied to diffusion guidance with the off-the-shelf model.
We use two baselines: 
(1) \textit{plug-and-play priors}~\cite{graikos2022diffusion}: starting from the random noised image, they first optimize it to close good mode and desired condition.
(2) \textit{gradients on $\hat{x}_0$}: as in~\cite{avrahami2022blended}, we estimate clean images $\hat{x}_0$ from noisy images $x_t$ as $\hat{x}_0=\frac{x_t}{\sqrt{\alpha_t}} - \frac{\sqrt{1-\alpha_t}\epsilon_\theta(x_t,t)}{\alpha_t}$. We calculate the gradients on $\hat{x}_0$ using it for guidance.
In summary, we observe that they fail to guide the diffusion.
We analyze detailed limitations of their method in Appendix \textcolor{red}{A}.


As in~\cite{dhariwal2021diffusion}, if the model well guides the generation process of diffusion to ImageNet class mode, the fidelity of generated images is improved by sacrificing diversity. Therefore, FID~\cite{heusel2017gans} becomes lower and Inception Score (IS)~\cite{salimans2016improved} becomes higher than its unconditional generation.
Accordingly, precision~\cite{kynkaanniemi2019improved} increases and recall~\cite{kynkaanniemi2019improved} decreases.
We calculate these metrics with the same reference images in~\cite{dhariwal2021diffusion} for quantitative comparison.
For further details on experiments and implementation, refer to Appendix. \textcolor{red}{E.1}. 

\begin{figure}[t]
    \centering
    \begin{tabular}{@{}c@{\hspace{0.5mm}}c@{\hspace{0.5mm}}|@{\hspace{0.5mm}}c@{\hspace{0.5mm}}c}
        \raisebox{1.1\height}{\rotatebox{90}{\scriptsize Na\"ive}} 
        \raisebox{0.15\height}{\rotatebox{90}{\scriptsize Off-the-shelf}} 
        \adjincludegraphics[clip,width=0.20\textwidth,trim={0 0 0 0}]{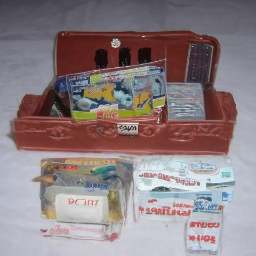} &
        \adjincludegraphics[clip,width=0.20\textwidth,trim={0 0 0 0}]{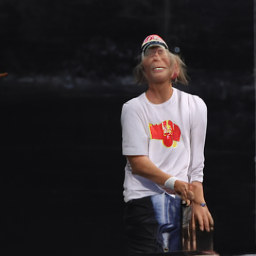} &
        \adjincludegraphics[clip,width=0.20\textwidth,trim={0 0 0 0}]{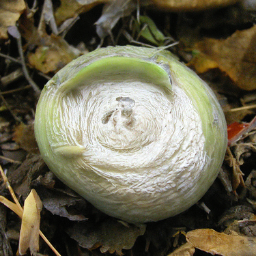} &
        \adjincludegraphics[clip,width=0.20\textwidth,trim={0 0 0 0}]{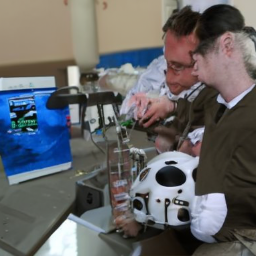} \\
        
        \raisebox{1.0\height}{\rotatebox{90}{\scriptsize Single}} 
        \raisebox{0.2\height}{\rotatebox{90}{\scriptsize Noise-aware}} 
        \adjincludegraphics[clip,width=0.20\textwidth,trim={0 0 0 0}]{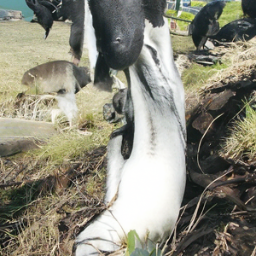} &
        \adjincludegraphics[clip,width=0.20\textwidth,trim={0 0 0 0}]{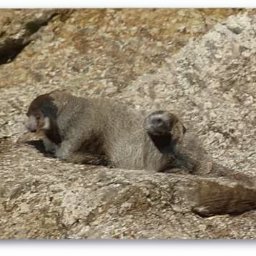} &
        \adjincludegraphics[clip,width=0.20\textwidth,trim={0 0 0 0}]{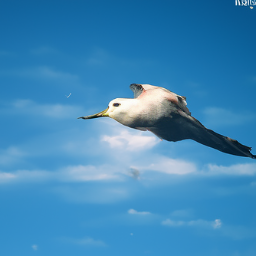} &
        \adjincludegraphics[clip,width=0.20\textwidth,trim={0 0 0 0}]{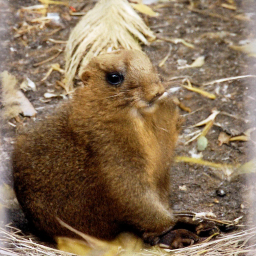} \\
        
        \raisebox{0.35\height}{\rotatebox{90}{\scriptsize Supervised}} 
        \raisebox{0.15\height}{\rotatebox{90}{\scriptsize Multi-experts}} 
        \adjincludegraphics[clip,width=0.20\textwidth,trim={0 0 0 0}]{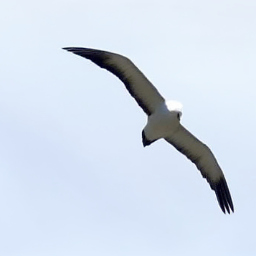} &
        \adjincludegraphics[clip,width=0.20\textwidth,trim={0 0 0 0}]{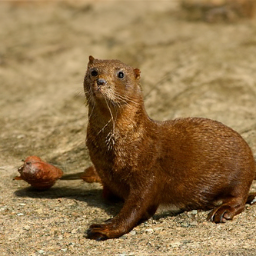} &
        \adjincludegraphics[clip,width=0.20\textwidth,trim={0 0 0 0}]{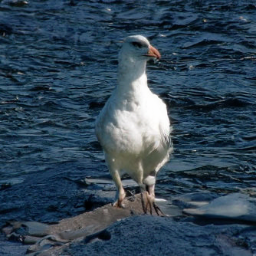} &
        \adjincludegraphics[clip,width=0.20\textwidth,trim={0 0 0 0}]{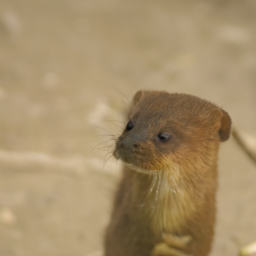} \\
        
        \raisebox{1.1\height}{\rotatebox{90}{\scriptsize PPAP }}
        \raisebox{0.5\height}{\rotatebox{90}{\scriptsize ($N=5$)}}
        \adjincludegraphics[clip,width=0.20\textwidth,trim={0 0 0 0}]{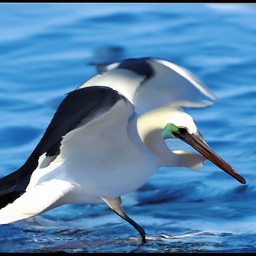} &
        \adjincludegraphics[clip,width=0.20\textwidth,trim={0 0 0 0}]{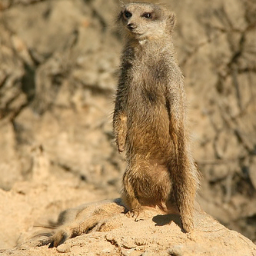} &
        \adjincludegraphics[clip,width=0.20\textwidth,trim={0 0 0 0}]{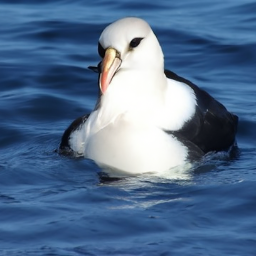} &
        \adjincludegraphics[clip,width=0.20\textwidth,trim={0 0 0 0}]{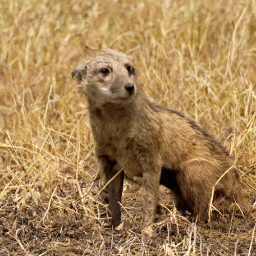} \\
        
        \raisebox{1.1\height}{\rotatebox{90}{\scriptsize PPAP}} 
        \raisebox{0.35\height}{\rotatebox{90}{\scriptsize ($N=10$)}}
        \adjincludegraphics[clip,width=0.20\textwidth,trim={0 0 0 0}]{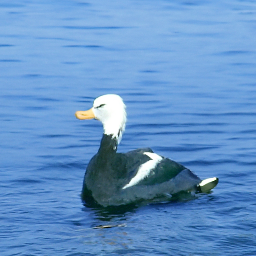} &
        \adjincludegraphics[clip,width=0.20\textwidth,trim={0 0 0 0}]{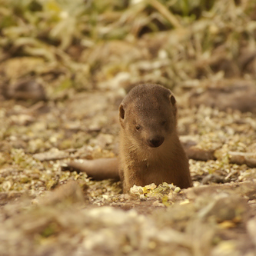} &
        \adjincludegraphics[clip,width=0.20\textwidth,trim={0 0 0 0}]{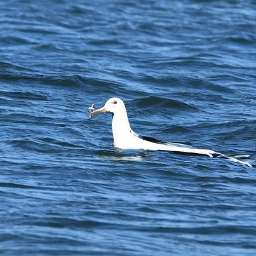} &
        \adjincludegraphics[clip,width=0.20\textwidth,trim={0 0 0 0}]{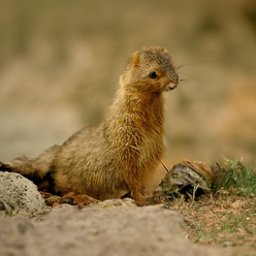} \\
        
        \footnotesize ~~~~~~Albatross & \footnotesize  Mongoose & \footnotesize  Albatross & \footnotesize  Mongoose \\
        
        \multicolumn{2}{c}{\small ~~~~~~~ResNet50} & \multicolumn{2}{c}{\small DeiT-S}
    \end{tabular}
\vspace{-0.2cm}

    \caption{Qualitative results of on ImageNet class conditional generation with DDPM 250 steps.
    Guidance using an off-the-shelf model produces irrelevant images to given classes. On contrary, our multi-experts and PPAP generate well-guided images. 
    More qualitative results are shown in Appendix. \textcolor{red}{G}.
    }
    \vspace{-1mm}
    \label{fig:imagenet_qualitative_resnet}
\end{figure}



\noindent\textbf{Main results}
We first compare the variants of ResNet50 and DeiT-S quantitatively (Table~\ref{tab:main}). Overall, we empirically confirmed the following observations to our advantage: 1) the baselines fail to guide the diffusion process, 2) the multi-experts strategy can boost performance, and 3) our practice is effective in more realistic settings. 

Specifically, 1) Naive off-the-shelf and Gradients on $\hat{x}_0$ do not significantly improve any of the metrics from the unconditional generation, even worsening with the DDPM sampler.
2) Multi-experts-5 achieves the best performance of 14.95 FID, and 83.26 IS scores with DeiT-S, which significantly outperforms the baselines. 
Moreover, 3) PPAP-5 and PPAP-10 show superior performance (20.07 FID and 60.62 IS with DeiT-S) than the Single noise-aware of using a single expert. 
These results indicate that, with our framework, off-the-shelf classifiers can effectively guide the unconditional diffusion model without ImageNet supervision. 

An interesting point is, PPAP outperforms the model fine-tuned with ImageNet supervision in guidance. 
Furthermore, it is noteworthy that our models use only small trainable parameters (21\% $\sim$ 57\%) and fewer unlabeled datasets (500k $<$ 1.2M). This suggests PPAP can successfully guide the diffusion models even in real-world scenarios. We will conduct an ablation study over the varying training data size to further discuss this point.

For further analysis, we show qualitative examples in Fig.~\ref{fig:imagenet_qualitative_resnet} and confirm that the overall results are consistent with the above observations. 
Specifically, we observe that directly using the off-the-shelf models does not effectively guide the diffusion model, generating irrelevant images to the given classes (Albatross and Mongoose), but leveraging multi-experts shows more powerful guidance capabilities. 

We also observe that DeiT-S tends to produce better guidance from Table~\ref{tab:main}.
Considering that transformer-based architectures tend to capture low-frequency components and shape-based feature~\cite{lee2023towards,naseer2021intriguing,park2022vision} than CNN-based architectures, we conjecture that these properties produce better guidance by focusing on shape.

\begin{figure}
    \begin{subfigure}[b]{0.45\columnwidth}
        \includegraphics[width=\columnwidth]{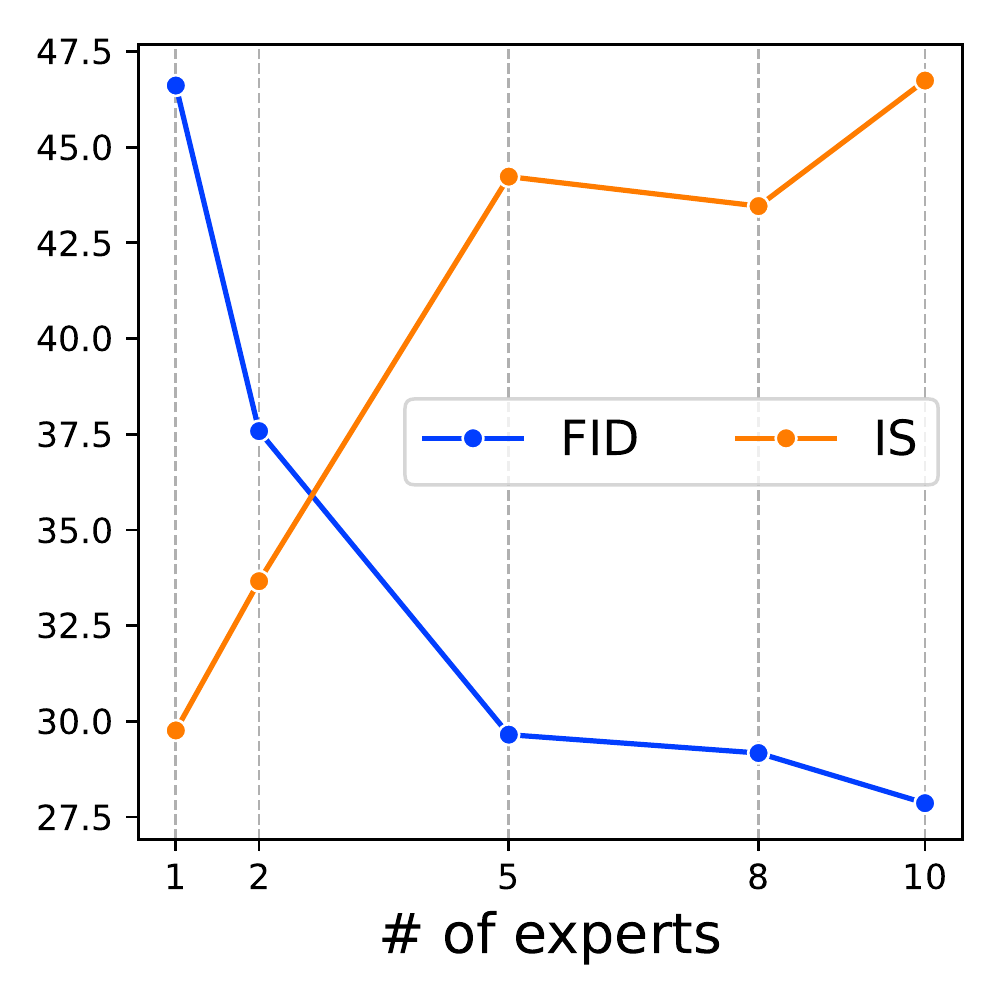}

\vspace{-0.2cm}
        \caption{FID and Inception Score}
        \label{fig:ablation1-number-of-experts-fid-and-is}
    \end{subfigure}
    \begin{subfigure}[b]{0.45\columnwidth}
        \includegraphics[width=\columnwidth]{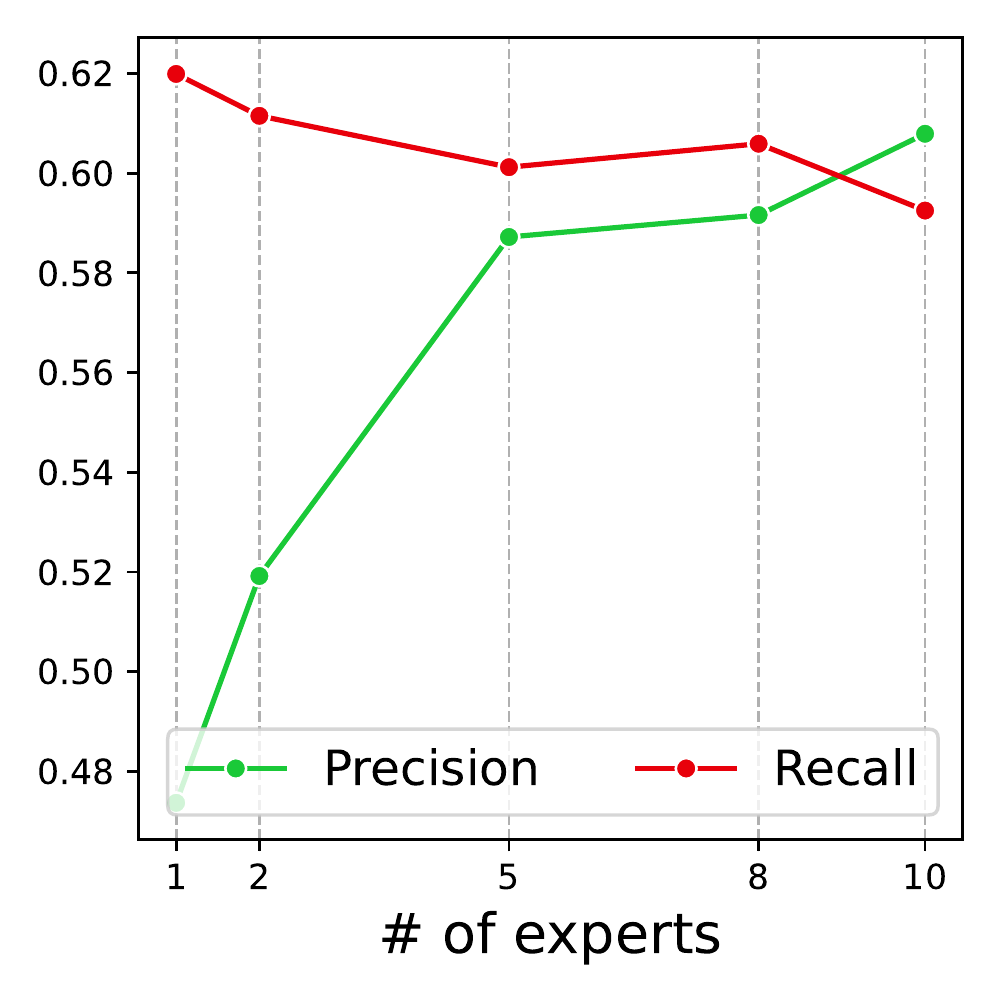}

\vspace{-0.2cm}
        \caption{Precision and Recall}
        \label{fig:ablation1-number-of-experts-precision-and-recall}
    \end{subfigure}

\vspace{-0.1cm}
    \caption{Ablation study for the number of experts.
    We plot quantitative results from the DDIM sampler with 25 steps by varying the number of experts.
    The results show leveraging more experts boosts the performance of guidance. 
    }
    \vspace{-1mm}
    \label{fig:number-of-experts}
\end{figure}

\begin{figure}
    \centering
    \begin{subfigure}[b]{0.45\columnwidth}
        \includegraphics[width=\columnwidth]{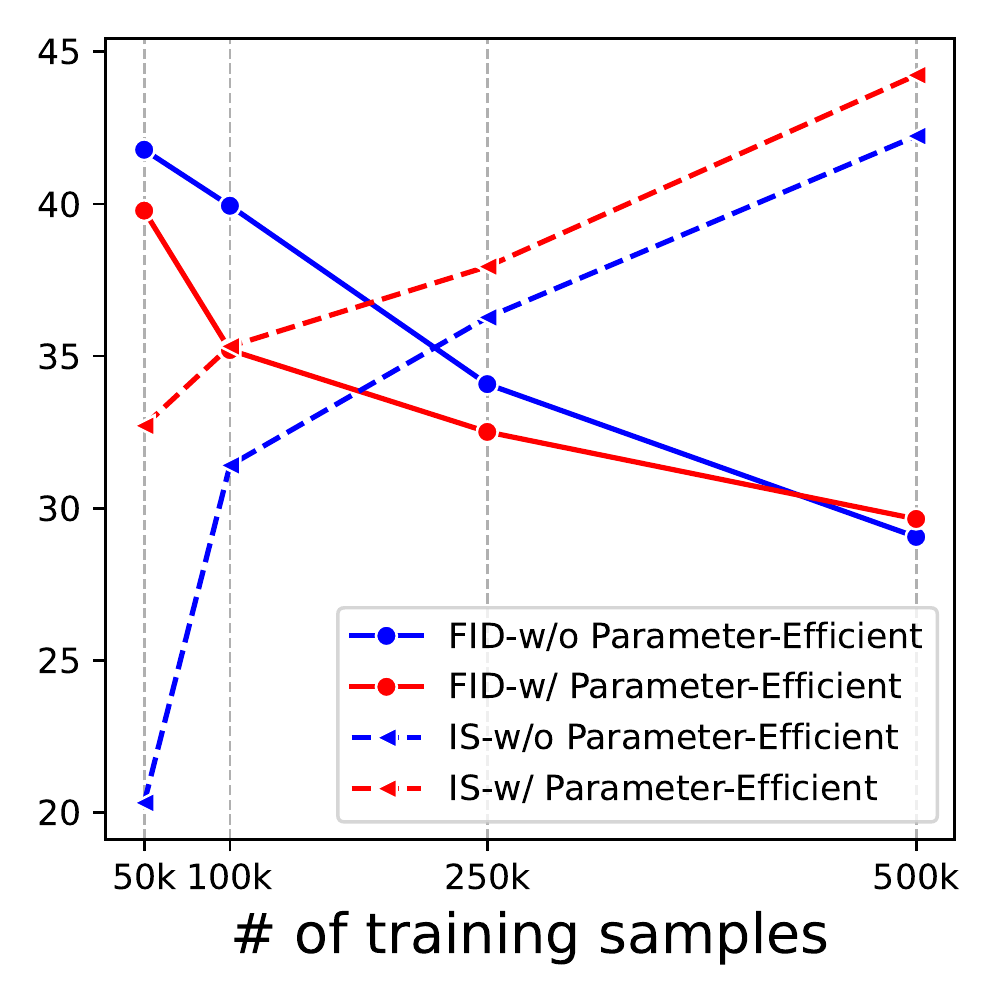}

\vspace{-0.2cm}
        \caption{FID and Inception Score}
        \label{fig:ablation2-data-efficiency-fid}
    \end{subfigure}
    \begin{subfigure}[b]{0.45\columnwidth}
        \includegraphics[width=\columnwidth]{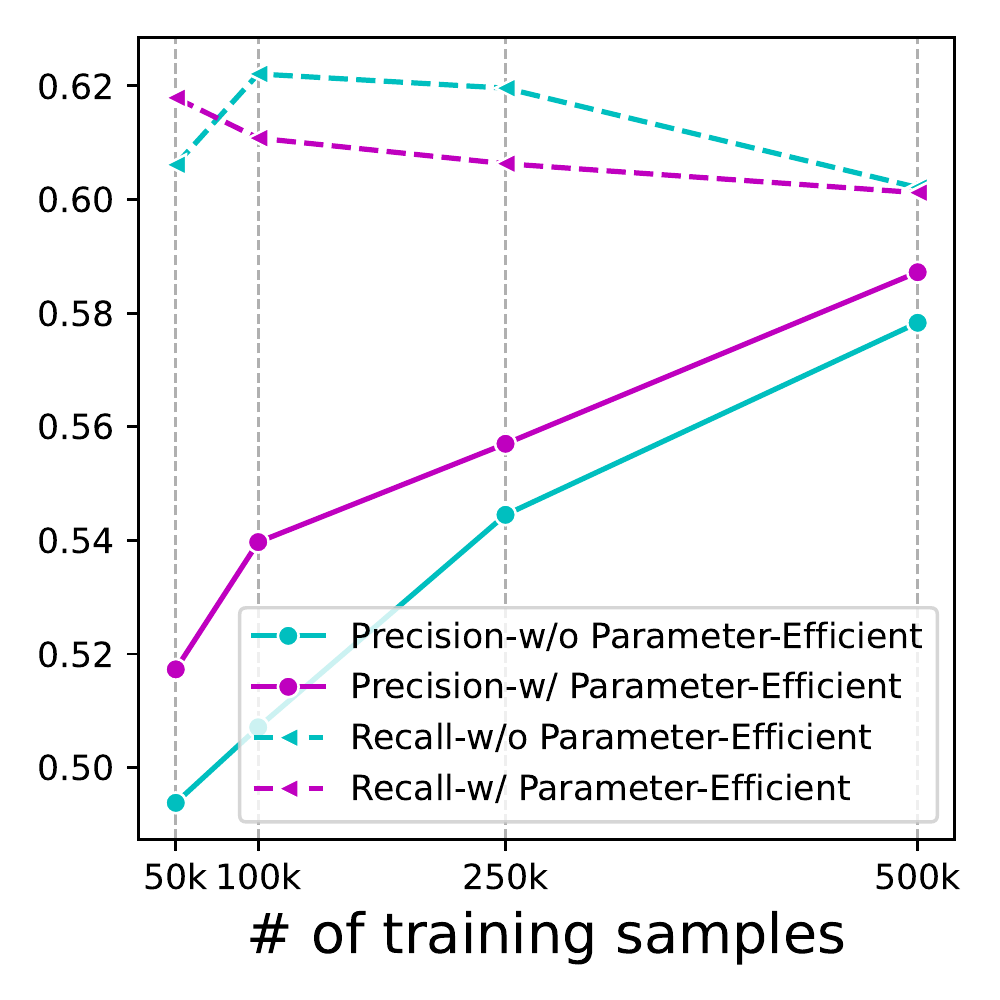}

\vspace{-0.2cm}
        \caption{Precision and Recall}
        \label{fig:ablation2-data-efficiency-is}
    \end{subfigure}

\vspace{-0.2cm}
    \caption{Ablation study for parameter-efficient tunning and data efficiency.
    We plot quantitative results of full-finetuning and parameter-efficient tunning for five experts on the varying size of generated datasets by sampling with DDIM 25 steps.
    Guidance with parameter-efficient experts outperforms fully fine-tuned experts in low-data regimes.
    Increasing the generated data for knowledge transfer improves performance. 
    }
    \label{fig:num_of_dataset}
\end{figure}

\noindent\textbf{Ablation study: Effect of Multi-Experts}
To understand the effect of using multi-experts, we compare several variants of PPAP with a varying number of parameter-efficient experts [1, 2, 5, 8, 10]. 
As shown in Fig.~\ref{fig:number-of-experts}, leveraging more experts effectively improves diffusion guidance, and using 10 experts achieves the best performance.
These results support that using multi-expert to make the noise interval more subdivided helps the guidance.





\noindent\textbf{Ablation study: Effect of Parameter-Efficient Tunning and data efficiency}
Here we analyze the effectiveness of the parameter-efficient fine-tuning strategy, in comparison with full fine-tuning, on the varying sizes of the training dataset. The results are shown in Fig.~\ref{fig:num_of_dataset}.
We observe that using the more generated data is an effective way of improving performance.
Also, the parameter-efficient tunning strategy achieves comparable results to full fine-tuning with 500k generated data, showing that fine-tuning for small parameters is enough.
An important finding is that the parameter-efficient experts even outperform the full fine-tuning variants in low data regimes (50k $\sim$). 
We posit that fully fine-tuning models with only a small dataset would ruin the representation power of the off-the-shelf models.
We present more ablation studies to deeply understand the effectiveness of each component in our method in Appendix~\textcolor{red}{F}.


\begin{figure}[t]
    \centering
    \begin{tabular}{@{}c@{\hspace{0.5mm}}c@{\hspace{0.5mm}}c@{\hspace{0.5mm}}c@{\hspace{0.5mm}}c@{\hspace{0.5mm}}c@{}}
        \raisebox{0.1\height}{\rotatebox{90}{\scriptsize PPAP (Ours)}} 
        \adjincludegraphics[clip,width=0.18\textwidth,trim={0 0 0 0}]{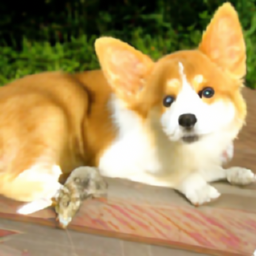} &
        \adjincludegraphics[clip,width=0.18\textwidth,trim={0 0 0 0}]{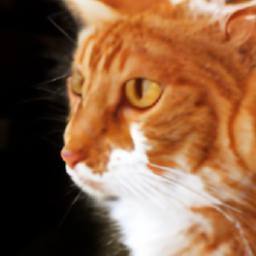} &
        \adjincludegraphics[clip,width=0.18\textwidth,trim={0 0 0 0}]{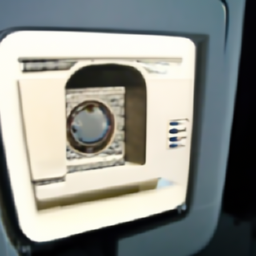} &
        \adjincludegraphics[clip,width=0.18\textwidth,trim={0 0 0 0}]{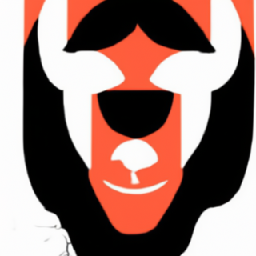} &
        \adjincludegraphics[clip,width=0.18\textwidth,trim={0 0 0 0}]{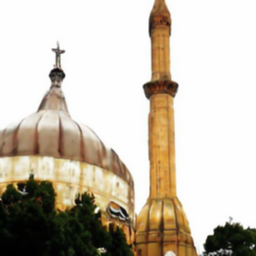} \\
        
        \raisebox{0.9\height}{\rotatebox{90}{\scriptsize Na\"ive}} 
        \adjincludegraphics[clip,width=0.18\textwidth,trim={0 0 0 0}]{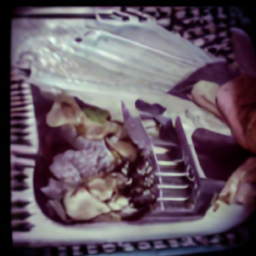} &
        \adjincludegraphics[clip,width=0.18\textwidth,trim={0 0 0 0}]{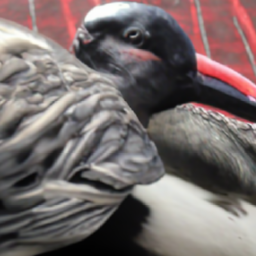} &
        \adjincludegraphics[clip,width=0.18\textwidth,trim={0 0 0 0}]{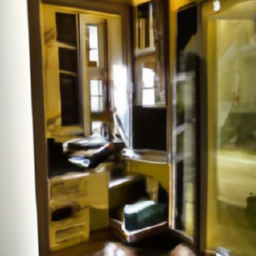} &
        \adjincludegraphics[clip,width=0.18\textwidth,trim={0 0 0 0}]{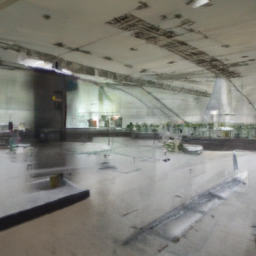} &
        \adjincludegraphics[clip,width=0.18\textwidth,trim={0 0 0 0}]{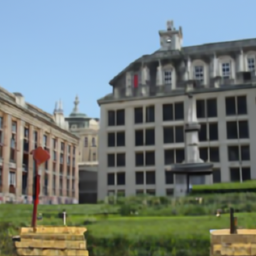}\vspace{-0.1cm} \\
        \footnotesize ~~~~Corgi & \footnotesize  Tiger cat & \footnotesize  Safe & \footnotesize  Mask & \footnotesize Mosque
    \end{tabular}

\vspace{-0.2cm}
    \caption{Generated images by guiding GLIDE with ResNet50. Our framework PPAP-5 succeeds in the guidance of diffusion, but our na\"ive off-the-shelf model fails.
    More qualitative results are presented in Appendix. \textcolor{red}{H.1}.
    }
    \vspace{-1mm}
    \label{fig:glide_image_classifier}
\end{figure}

\subsection{Guiding GLIDE for Various Downstream Tasks}
Here, we will show several practical applications by guiding GLIDE~\cite{nichol2021glide}, which is trained on a largescale unreleased CLIP~\cite{radford2021learning}-filtered dataset~\cite{nichol2021glide}.
With our framework, we can apply an ImageNet~\cite{deng2009imagenet} pretrained classifier~\cite{he2016deep}, zero-shot depth estimator~\cite{ranftl2020towards}, and pretrained semantic segmentation model~\cite{chen2017rethinking} as the guidance model. 

\noindent\textbf{Experimental Setup}
GLIDE~\cite{nichol2021glide} generates images from given text inputs.
GLIDE consists of two diffusion models: 1) Generator:  generates $64\times64$ images from given text inputs, and 2) Upsampler:  upsamples generated $64\times64$ images to $256\times256$ images.
To make GLIDE unconditional, we give the empty token as input of generator diffusion.
Since the upsampler of GLIDE just changes image resolution, we aim to guide generator diffusion.
All guidance models used in GLIDE experiments exploit 5 experts.
For data-free knowledge transfer in our framework, we generate 500k unconditional $64\times64$ images from generator diffusion with the 25-step DDIM sampler.
For generating guidance images, we use a DDPM sampler with 250 steps for generator diffusion and a fast27 sampler~\cite{nichol2021glide} for upsampler diffusion.
We refer to experimental and implementation details for guiding GLIDE in Appendix. \textcolor{red}{E.2}.



\begin{figure}[t]

    \centering
    \begin{tabular}{@{\hspace{0.5mm}}c@{\hspace{0.5mm}}c@{\hspace{0.5mm}}c@{\hspace{0.5mm}}c@{\hspace{0.5mm}}c}
        \raisebox{0.4\height}{\rotatebox{90}{\scriptsize Original}} 
        \adjincludegraphics[clip,width=0.18\textwidth,trim={0 0 0 0}]{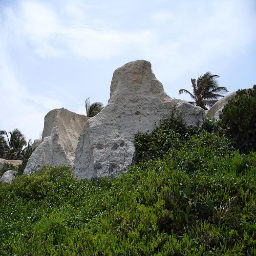} &
        \adjincludegraphics[clip,width=0.18\textwidth,trim={0 0 0 0}]{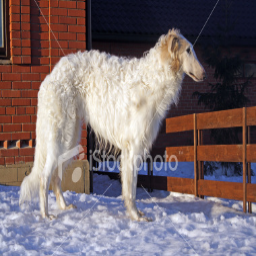} &
        \adjincludegraphics[clip,width=0.18\textwidth,trim={0 0 0 0}]{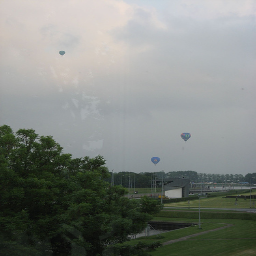} &
        \adjincludegraphics[clip,width=0.18\textwidth,trim={0 0 0 0}]{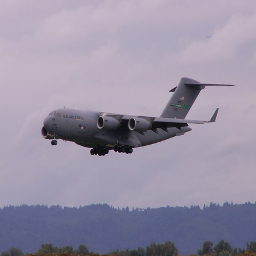} & 
        \adjincludegraphics[clip,width=0.18\textwidth,trim={0 0 0 0}]{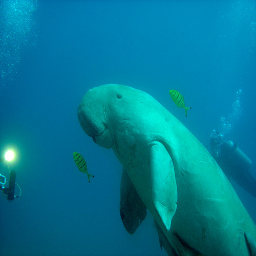} \\

        \raisebox{0.2\height}{\rotatebox{90}{\scriptsize Depth map}} 
        \adjincludegraphics[clip,width=0.18\textwidth,trim={0 0 0 0}]{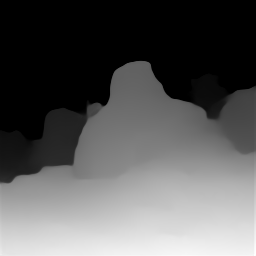} &
        \adjincludegraphics[clip,width=0.18\textwidth,trim={0 0 0 0}]{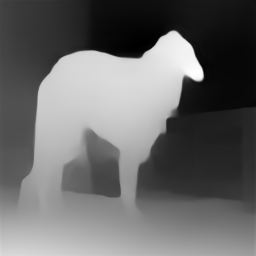} &
        \adjincludegraphics[clip,width=0.18\textwidth,trim={0 0 0 0}]{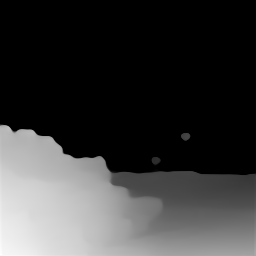} &
        \adjincludegraphics[clip,width=0.18\textwidth,trim={0 0 0 0}]{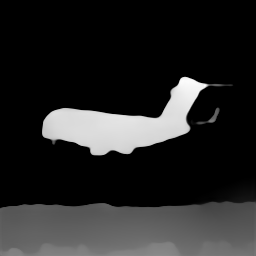} & 
        \adjincludegraphics[clip,width=0.18\textwidth,trim={0 0 0 0}]{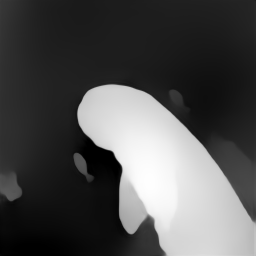} \\

        \raisebox{0.1\height}{\rotatebox{90}{\scriptsize PPAP (Ours)}} 
        \adjincludegraphics[clip,width=0.18\textwidth,trim={0 0 0 0}]{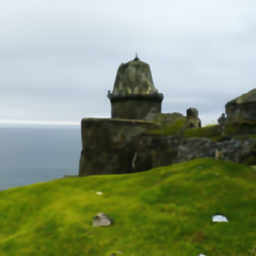} &
        \adjincludegraphics[clip,width=0.18\textwidth,trim={0 0 0 0}]{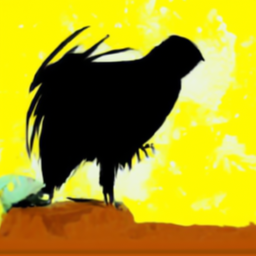} &
        \adjincludegraphics[clip,width=0.18\textwidth,trim={0 0 0 0}]{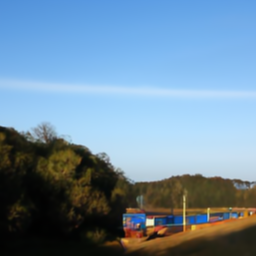} &
        \adjincludegraphics[clip,width=0.18\textwidth,trim={0 0 0 0}]{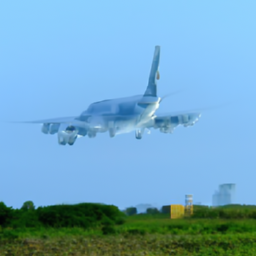} & 
        \adjincludegraphics[clip,width=0.18\textwidth,trim={0 0 0 0}]{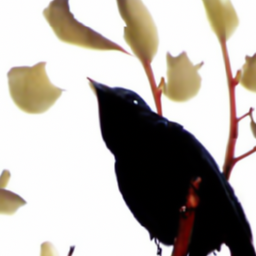} \\

        \raisebox{0.8\height}{\rotatebox{90}{\scriptsize Na\"ive}} 
        \adjincludegraphics[clip,width=0.18\textwidth,trim={0 0 0 0}]{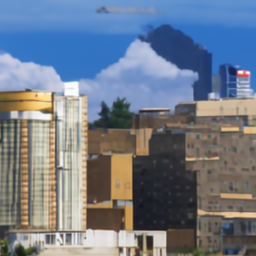} &
        \adjincludegraphics[clip,width=0.18\textwidth,trim={0 0 0 0}]{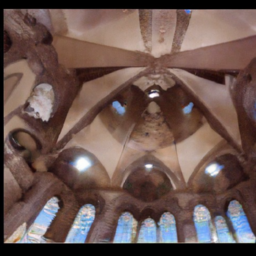} &
        \adjincludegraphics[clip,width=0.18\textwidth,trim={0 0 0 0}]{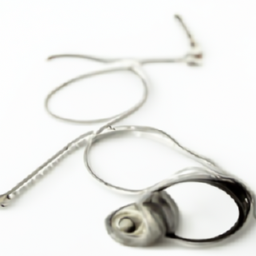} &
        \adjincludegraphics[clip,width=0.18\textwidth,trim={0 0 0 0}]{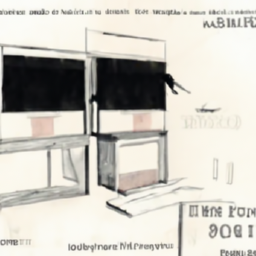} & 
        \adjincludegraphics[clip,width=0.18\textwidth,trim={0 0 0 0}]{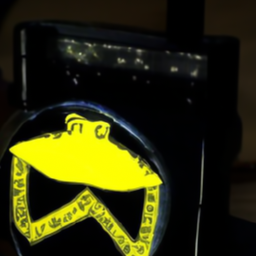}
    \end{tabular}
    \vspace{-0.3cm}
    \caption{
    Generated images by guiding GLIDE with the MidaS depth estimator.
    Our framework PPAP succeeds in the guidance of diffusion, but our off-the-shelf model fails. More qualitative results are presented in Appendix. \textcolor{red}{H.2}.
    }
    \label{fig:glide_depth}

\end{figure}


\noindent\textbf{Guiding GLIDE with ImageNet Classifier}
\label{par:glide_classifier}
We used ImageNet~\cite{deng2009imagenet} pre-trained ResNet50~\cite{he2016deep} for guiding GLIDE to conduct class-conditional image generation.
Figure~\ref{fig:glide_image_classifier} shows generated images by naive off-the-shelf guidance and PPAP ($N=5$) guidance.
The results are consistent with the results in Fig.~\ref{fig:imagenet_qualitative_resnet} as the off-the-shelf model does not generate guided images and our method can guide GLIDE to generate well-guided images.
Notably, our method can semantically guide the GLIDE with varying styles of images, such as cartoon-style images (4th image by ours), which is interesting because ResNet50 has never seen cartoon-style images in the ImageNet dataset.
It shows that PPAP can obtain both the generation ability of GLIDE in various domains and the semantic understanding ability of the classifier.

\noindent\textbf{Guiding GLIDE with Depth Estimator}
\label{par:glide_depth}
MiDaS~\cite{ranftl2020towards,ranftl2021vision} is a monocular depth estimation model designed for zero-shot cross-dataset transfer.
We leverage their zero-shot superiority for guiding GLIDE to depth-map-to-image generation.
As it is hard to provide an arbitrary depth map, we first estimate depth maps from images in the ImageNet dataset.
Then, we feed the depth maps as desirable inputs for guiding the diffusion. 
As shown in Fig.~\ref{fig:glide_depth}, guidance with the naive off-the-shelf model generates images unrelated to the desired depth maps. On the contrary, with our proposed framework, we observe that the generated images are well-aligned with the edges in the depth map.

\begin{figure}[t]

    \centering
    \begin{tabular}{@{\hspace{0.5mm}}c@{\hspace{0.5mm}}c@{\hspace{0.5mm}}c@{\hspace{0.5mm}}c@{\hspace{0.5mm}}c}
        \raisebox{0.4\height}{\rotatebox{90}{\scriptsize Original}} 
        \adjincludegraphics[clip,width=0.18\textwidth,trim={0 0 0 0}]{figure_files/glide_seg/1/base_image.png} &
        \adjincludegraphics[clip,width=0.18\textwidth,trim={0 0 0 0}]{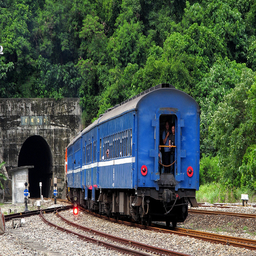} &
        \adjincludegraphics[clip,width=0.18\textwidth,trim={0 0 0 0}]{figure_files/glide_seg/3/base_image.png} &
        \adjincludegraphics[clip,width=0.18\textwidth,trim={0 0 0 0}]{figure_files/glide_seg/4/base_image.png} & 
        \adjincludegraphics[clip,width=0.18\textwidth,trim={0 0 0 0}]{figure_files/glide_seg/5/base_image.png} \\

        \raisebox{0.1\height}{\rotatebox{90}{\scriptsize Segmentation}} 
        \adjincludegraphics[clip,width=0.18\textwidth,trim={0 0 0 0}]{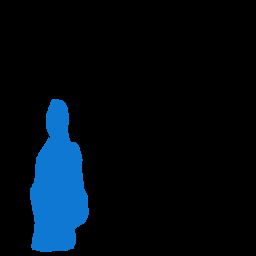} &
        \adjincludegraphics[clip,width=0.18\textwidth,trim={0 0 0 0}]{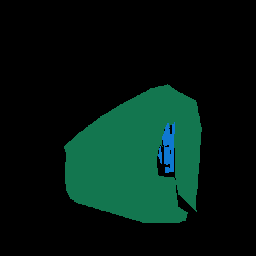} &
        \adjincludegraphics[clip,width=0.18\textwidth,trim={0 0 0 0}]{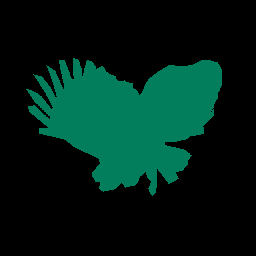} &
        \adjincludegraphics[clip,width=0.18\textwidth,trim={0 0 0 0}]{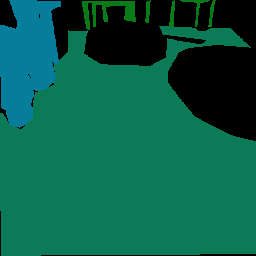} & 
        \adjincludegraphics[clip,width=0.18\textwidth,trim={0 0 0 0}]{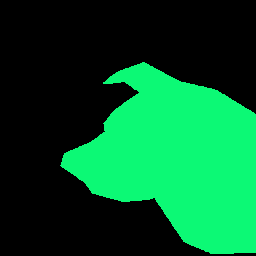} \\

        \raisebox{0.1\height}{\rotatebox{90}{\scriptsize PPAP (Ours)}} 
        \adjincludegraphics[clip,width=0.18\textwidth,trim={0 0 0 0}]{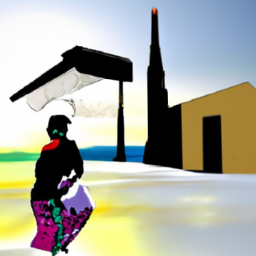} &
        \adjincludegraphics[clip,width=0.18\textwidth,trim={0 0 0 0}]{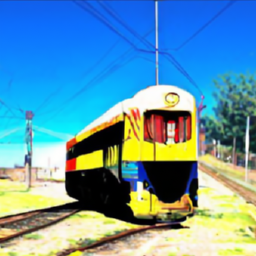} &
        \adjincludegraphics[clip,width=0.18\textwidth,trim={0 0 0 0}]{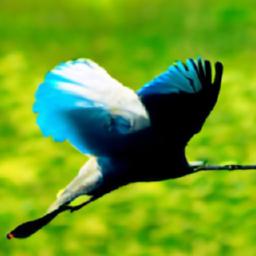} &
        \adjincludegraphics[clip,width=0.18\textwidth,trim={0 0 0 0}]{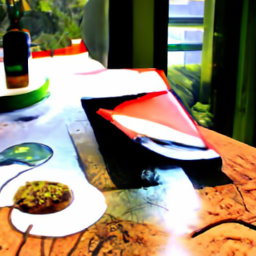} & 
        \adjincludegraphics[clip,width=0.18\textwidth,trim={0 0 0 0}]{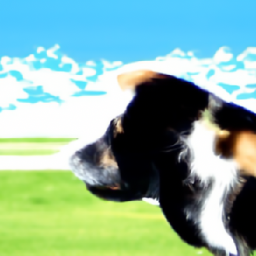} \\

        \raisebox{0.8\height}{\rotatebox{90}{\scriptsize Na\"ive}} 
        \adjincludegraphics[clip,width=0.18\textwidth,trim={0 0 0 0}]{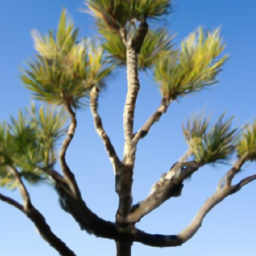} &
        \adjincludegraphics[clip,width=0.18\textwidth,trim={0 0 0 0}]{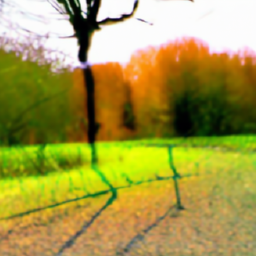} &
        \adjincludegraphics[clip,width=0.18\textwidth,trim={0 0 0 0}]{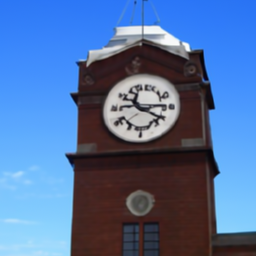} &
        \adjincludegraphics[clip,width=0.18\textwidth,trim={0 0 0 0}]{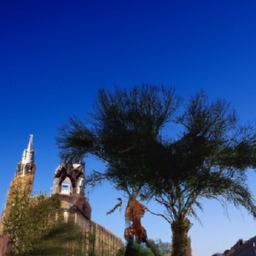} & 
        \adjincludegraphics[clip,width=0.18\textwidth,trim={0 0 0 0}]{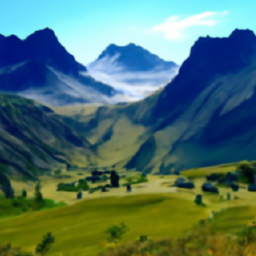}
    \end{tabular}
     \vspace{-0.3cm}
    \caption{
    Generated images by guiding GLIDE with DeepLabv3 semantic segmentation. Our framework PPAP succeeds in the guidance of diffusion, but the naive off-the-shelf guidance fails. More qualitative results are shown in Appendix. \textcolor{red}{H.3}.
    }
    \label{fig:glide_seg}

\end{figure}

\noindent\textbf{Guiding GLIDE with Semantic Segmentation}
Our experiments mentioned above have shown that the PPAP framework is capable of both semantic-level guidance from the ImageNet classifier and pixel-level guidance from the depth estimator. 
Based on the results, we validate whether our PPAP could apply both capabilities together and join into a semantic segmentation task guidance.
Specifically, we used DeepLabv3~\cite{chen2017rethinking} for guidance.
The results in Fig.~\ref{fig:glide_seg} indicate that generated images from our PPAP framework tend to be aligned with the segmentation map, while generated images from naive off-the-shelf guidance fail.

\section{Conclusion}

In this paper, we studied how we can achieve a \textit{practical} plug-and-play diffusion guidance.
From an observation that a classifier would behave differently in varying degrees of noise, we propose the multi-expert strategy of leveraging different experts for different diffusion steps.
Our experiments validate that our proposed framework makes it possible to easily utilize publicly available off-the-shelf models for guiding the diffusion process without requiring further training datasets.
We deal with limitations and future work in Appendix \textcolor{red}{I}.



{\small
\bibliographystyle{ieee_fullname}
\bibliography{custom}
}


\appendix
\onecolumn

\title{Supplementary Material: Towards Practical Plug-and-Play Diffusion Models}


\maketitle
\tableofcontents

\setcounter{figure}{11} 
\setcounter{table}{1}

\twocolumn

\begin{figure}
    \centering
    \begin{subfigure}{0.48\columnwidth}
    \includegraphics[width=1.0\columnwidth]{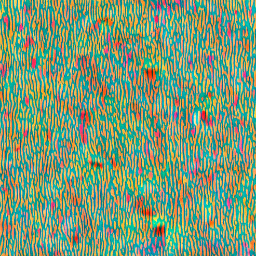}
    \caption{Label: vulture}
    \end{subfigure}
    \begin{subfigure}{0.48\columnwidth}
    \includegraphics[width=1.0\columnwidth]{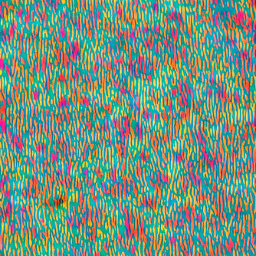}
    \caption{Label: ice lolly}
    \end{subfigure}
    \caption{Generated images from plug-and-play priors~\cite{graikos2022diffusion} with ImageNet pre-trained ResNet50. On the ImageNet dataset, we observe that plug-and-play priors fail to generate realistic images.}
    \label{fig:PNPP}
\end{figure}


\section{Detailed Comparison with Previous Methods}
\label{apx:sec:limit-prev}
We compare our method to two baselines 1) \textit{plug-and-play priors}~\cite{graikos2022diffusion} and 2) gradients on $\hat{x}_0$ by using the approach in~\cite{avrahami2022blended}.
This section describes the details of the comparison and their limitation.
We note that these methods do not require training, but our method requires finetuning the off-the-shelf model.

First, we observe that plug-and-play priors fail to produce realistic images on the ImageNet dataset.
We generate images with unconditional 256$\times$256 ADM~\cite{dhariwal2021diffusion} and pre-trained ResNet50~\cite{he2016deep} using their official code\footnote{\url{https://github.com/AlexGraikos/diffusion_priors}}.
In the same setting as FFHQ experiments in their work, we try to generate images by guiding them to random class labels.
As shown in Fig.~\ref{fig:PNPP}, the generated images are not realistic.
We also increase the optimization step of their method, but there are no significant differences.
Correspondingly, their FID and IS are 358.20 and 1.55, respectively.

\begin{figure*}
    \centering
    \includegraphics[width=\textwidth]{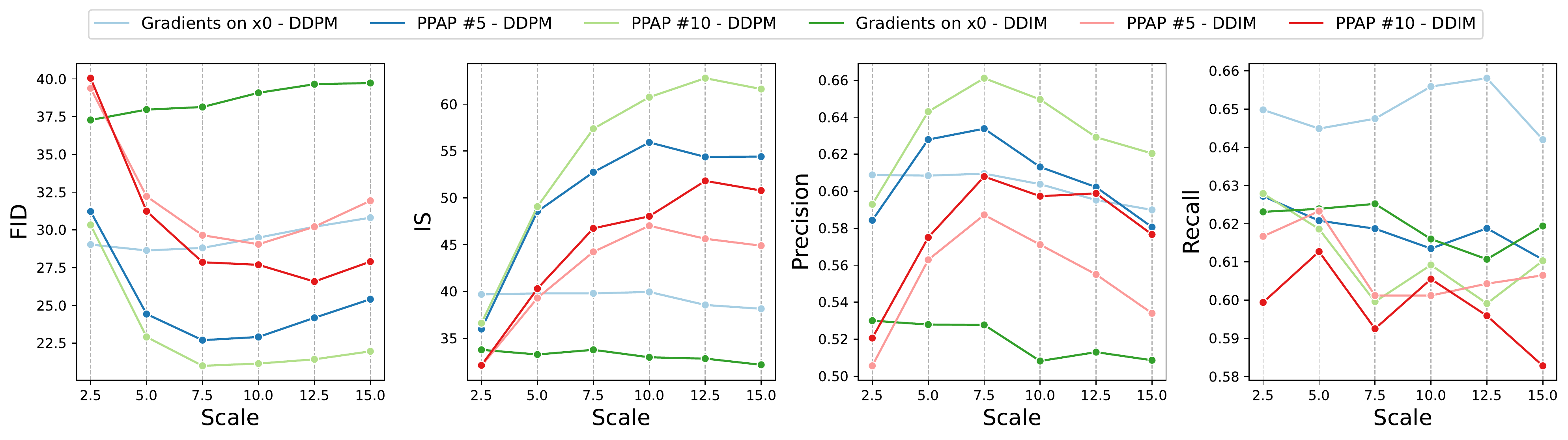}
    \caption{Quantitative comparison between PPAP and gradients on $\hat{x}_0$ according to guidance scale $s$.}
    \label{fig:gradient_x0}
\end{figure*}

We provide quantitative results of gradients on $\hat{x}_0$ in Table~\textcolor{red}{1} when $s=7.5$.
To investigate in further depth, we compare our method with theirs while adjusting the guidance scale $s$ and utilizing the ResNet50 classifier.
For implementing gradients on $\hat{x}_0$, we use the official code\footnote{\url{https://github.com/omriav/blended-diffusion}} of Blended Diffusion~\cite{avrahami2022blended}.
As shown in Fig.~\ref{fig:gradient_x0}, gradients on $\hat{x}_0$ do not significantly improve FID, IS, and Precision, indicating that it does not guide the diffusion model to the class label.
From this, we conjecture that it is effective in image editing with the assistance of several techniques~\cite{avrahami2022blended} but not for guiding the diffusion without these techniques.

As the concurrent work, eDiff-I~\cite{balaji2022ediffi} replicates copies of a single diffusion model and specializes each copy on different time intervals.
However, they apply the idea to the diffusion model, not the guidance model.
Also, while eDiff-I applied its idea to the diffusion model by focusing on training efficiency through a branching scheme, we applied our idea to the guidance model to efficiently increase the number of experts with the parameter-efficient fine-tuning strategy.

\section{Details of Observational Study}
\label{apx:sec:detail-observ}
We use the ResNet50 model and unconditional diffusion model to illustrate our observation in Section \textcolor{red}{3.2}.
In this section, we discuss the details of the experimental setup.

For off-the-shelf ResNet50, we exploit Imagenet pre-trained ResNet50\footnote{We use publicly available torchvision~\cite{paszke2019pytorch} ResNet50}. 
On the ImageNet training dataset, we fine-tune the off-the-shelf model for a single noise-aware ResNet50 that learns the entire timestep $t\in[1,...,1000]$.
During 300k iterations, AdamW~\cite{loshchilov2017decoupled} with a learning rate of 1e-4 and weight decay of 0.05 is utilized as the optimizer.
The batch size is set to 256.

For each expert ResNet50, we use the same optimizer as the single noise-aware model above. 
In order to make the total iterations for training five experts equal to those of the single noise-aware model, we fine-tune off-the-shelf ResNet50 with a batch size of 256 and 60k iterations.

During the reverse process with guidance, the guidance scale $s$ is set as 7.5 since it achieves good results for most variants.

\section{Details of Parameter Efficient Multi-Experts}
\label{apx:sec:peft}
\begin{figure}
    \centering
    \includegraphics[width=1.0\columnwidth]{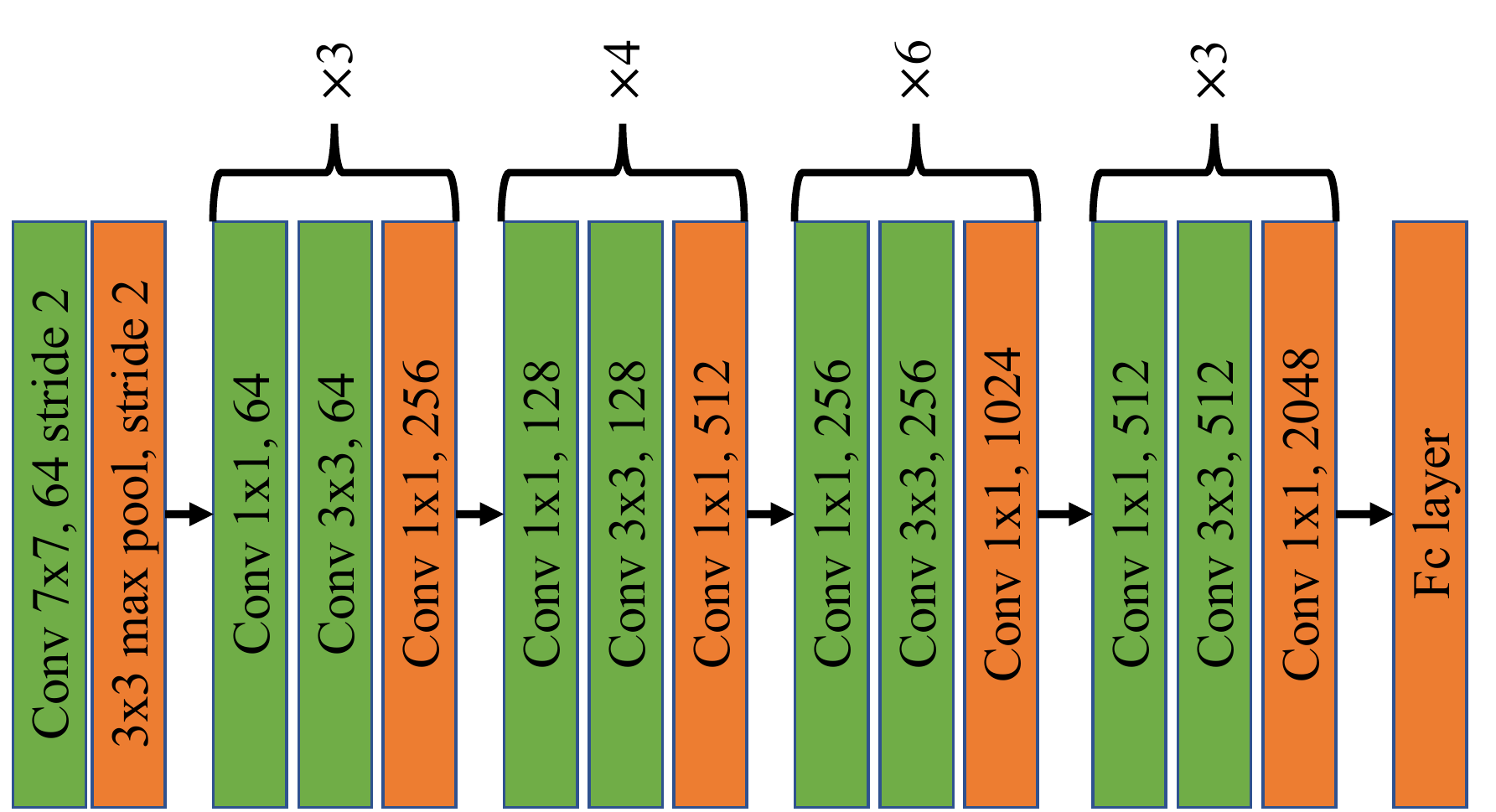}
    \caption{ResNet50 model with LORA. The green color is the layer where LORA is applied.}
    \label{fig:Lora_resnet50}
\end{figure}

For the parameter-efficient multi-experts strategy described in Section~\textcolor{red}{4.1}, we only fine-tune a small number of parameters while reusing most of the frozen off-the-shelf parameters.
Here, we report how parameter efficient tunning is applied to each architecture used in our experiments.

We first apply LORA~\cite{hu2021lora} to certain weight matrix of the off-the shelf model $W_0 \in \mathbb{R}^{d \times k}$, which updates it with low-rank decomposition as $W_0 + \alpha/rBA$ where $B\in \mathbb{R}^{d\times r}$ and $A \in \mathbb{R}^{r \times k}$.
During fine-tuning, $A$ and $B$ are updated, while $W_0$ is frozen.
When inferencing the guidance model in the reverse process, the weight matrix $W_0$ is simply updated as $W=W_0 + BA$, resulting in no additional inference cost.
We note that $\alpha$ is fixed as 8.

If batch normalization and bias terms are used in the architecture, we tune these as well.
This incurs no additional inference costs because it does not alter the model design, such as layer expansion.
We note that such parameter efficient strategy is applicable to various architectures.

\subsection{ADM Guidance Models}
\label{apx:sec:archi_image_class}

We use ResNet-50~\cite{he2016deep} and DeiT-S~\cite{touvron2021training} architecture for the guidance model in ImageNet class conditional generation.
For ResNet-50 architecture, LORA is applied to the first and the second convolutions of each block, and also to the first 7$\times$7 convolution as shown in Fig.~\ref{fig:Lora_resnet50}.
We set the LORA rank as $r=16$, and also tune the bias and batch normalization layers.

DeiT-S is a transformer-based model comprised of a self-attention module and an MLP module.
We only apply LORA to self-attention module weights with a rank of $r=32$.
Since DeiT-S does not have batch normalization layers, we only fine-tune bias terms.

\subsection{GLIDE Guidance Models}

\paragraph{Image classifier}
We use the ResNet50~\cite{he2016deep} classifier for guiding GLIDE to conduct class conditional image generation.
All configuration of parameter-efficient multi-expert is identical to those described in Section~\ref{apx:sec:archi_image_class}.

\paragraph{Depth estimator}
MiDaS-small~\cite{ranftl2020towards} is utilized for guiding GLIDE.
MiDaS is a monocular depth estimation, which is used for various tasks such as the single-image view synthesis~\cite{rombach2021geometry, park2022bridging, liu2021infinite} in a frozen state.
MiDaS-small is compromised of two components: 1) Backbone network which is based on EfficientNet lite3~\cite{tan2019efficientnet} and 2) CNN-based decoder network.
We apply LORA~\cite{hu2021lora} with $r=8$ to point-wise convolution layers of the backbone network as well as convolution layers of the CNN-based decoder network.
Also, bias terms and batch normalization layers are fine-tuned.
The total number of trainable parameters used per expert is 0.7M, which is 3.73\% compared to the MiDaS-small parameter of 21M.

\paragraph{Semantic segmentation}
DeepLabv3-ResNet50~\cite{chen2017rethinking} is exploited for guiding GLIDE. DeepLabv3-ResNet50 is compromised of two components: 1) Backbone network which is based on ResNet-50~\cite{he2016deep} and 2) Atrous Spatial Pyramid Pooling (ASPP) segmentation classifier decoder. We apply LORA to the ResNet-50 backbone. All configuration of LORA is the same as in Section ~\ref{apx:sec:archi_image_class}.
As a result, we introduce 0.88M trainable parameters, which is 2.15\% compared to the DeepLabv3 parameter of 41.95M.

\section{Guidance Formulation of PPAP for Other Tasks}
\label{apx:sec:other-tasks}
Here, we explain how the monocular depth estimator and the semantic segmentation model can be incorporated into the guidance model.

\paragraph{Monocular Depth Estimation}
A monocular depth estimation model takes an image as input and outputs a depth map $f(x) \in\mathbb{R}^{H\times W}$, where $H$ and $W$ represent the height and width of the input image, respectively.
We formulate knowledge transfer loss $\mathcal{L}_{de}$ for depth estimation models as:
\begin{equation}
    \mathcal{L}_{de} = \lVert \texttt{sg}\big(f_{\phi}(\tilde{x}_{0})\big) - f_{\phi_n}(\tilde{x}_{t})\rVert_{1}.
\end{equation}
Then, we guide image generation so that the image has some desired depth map $D_{target} \in \mathbb{R}^{H \times W}$ as follows:
\begin{equation}
    \mathcal{L}_{gd} = \lVert f_{\phi_n}({x}_{t}) - D_{target}\rVert_{1}.
\end{equation}

\paragraph{Semantic Segmentation} 

A semantic segmentation model takes an image as input and outputs a segmentation map, having classification logit vector at the pixel level, $f(x) \in\mathbb{R}^{C \times H \times W}$. We formulate knowledge transfer loss $\mathcal{L}_{ss}$ for semantic segmentation models as:
\begin{equation}
    \mathcal{L}_{ss} = \lVert \texttt{sg}\big(f_{\phi}(\tilde{x}_{0})\big) - f_{\phi_n}(\tilde{x}_{t})\rVert_{1}.
\end{equation}
Then, we guide image generation so that the image is aligned with the segmentation map $S_{target} \in \mathbb{R}^{C \times H \times W}$ as follows:
\begin{equation}
    \mathcal{L}_{gs} = \lVert \texttt{s}(f_{\phi_n}({x}_{t})) - S_{target}\rVert_{1}.
\end{equation}

\section{Experimental Details}
\label{apx:sec:main-details}

\begin{figure*}
    \centering
    \includegraphics[width=0.98\textwidth]{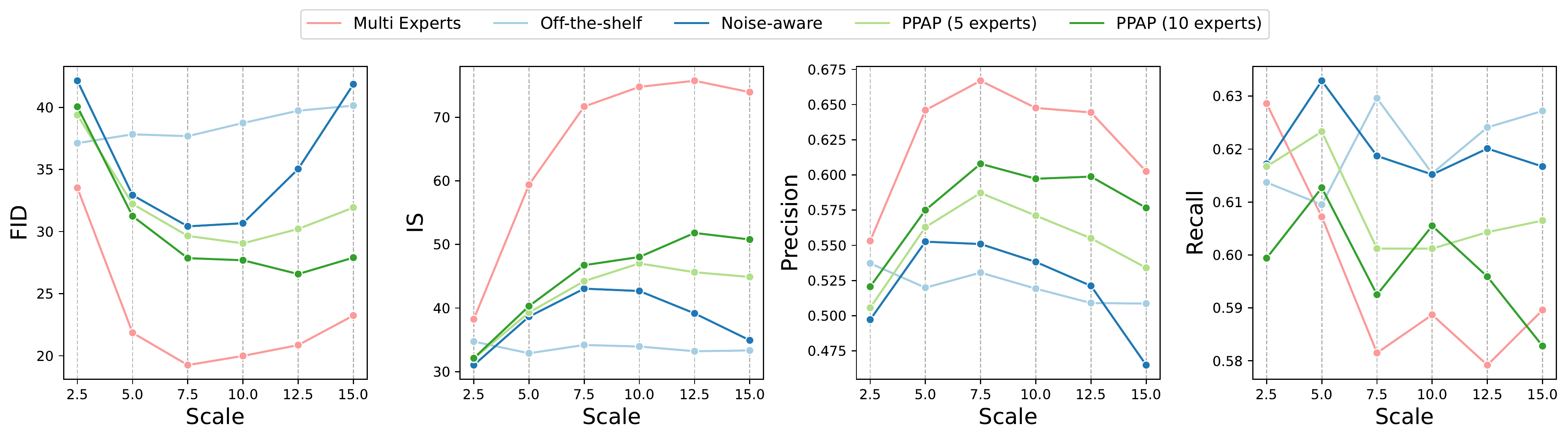}
    \caption{Quantitative results on guiding unconditional ADM with ResNet50 and DDIM 25 steps according to guidance scale $s$.}
    \label{fig:scale}
\end{figure*}

\subsection{Guiding ADM for ImageNet Class Conditional Generation}
\label{apx:subsec:adm-main-details}
When we train the models, AdamW optimizer~\cite{loshchilov2017decoupled} is commonly used with a learning rate of 1e-4, weight decay of 0.05, and batch size of 256.
All variants are trained with the same total iterations of 300k.
For implementing LORA, we use the official code of LORA\footnote{\url{https://github.com/microsoft/LoRA}} (See the details of parameter-efficient multi-expert in Section~\ref{apx:sec:archi_image_class}).
We utilize torchvision pre-trained ResNet50 and timm~\cite{rw2019timm} pre-trained DeiT-S for the off-the-shelf model.
We calculate FID, IS, Precision, and Recall with 10k generated samples with random class labels.
All experiments are conducted on 8$\times$A100 GPUs.

\subsection{Guiding GLIDE for Various Downstream Tasks}
\label{apx:subsec:glide-main-details}
Same with the ImageNet classifier guidance setting, we use an AdamW optimizer with a learning rate of 1e-4 and weight decay of 0.05 to train the models.

In image classifier guidance for GLIDE, we use the same setting as in ImageNet class conditional generation with ADM~\cite{dhariwal2021diffusion}.
For depth estimation and semantic segmentation models, we use MIDAS-small and DeepLabv3-ResNet50, publicly available in torch-hub~\cite{paszke2019pytorch}.
We train these with a batch size of 128 and 300k iterations.

For guidance scale $s$, we use norm-based scaling for guidance gradients~\cite{kim2022guided}.
We set the gradient ratio as 0.3.

\section{More Ablation Study on ImageNet Conditional Generation}
\label{apx:sec:ablations}

%

\subsection{Effect of Guidance Scale $s$}
Here, we change the guidance scale $s$ from 2.5 to 15.0 for identifying performance according to the guidance scale $s$.

Figure~\ref{fig:scale} shows the results according to the guidance scale $s$ where DDIM sampler with 25 steps is used.
From these results, we can see that multi-expert and PPAP greatly outperform the single-noise aware model in most guidance scales.
Second, it can be seen that there is a sweet spot around $s=7.5$, where the guidance scale is neither too large nor too small, and where neither the image quality nor the guidance capability is compromised.
Therefore, we set the default guidance scale as 7.5 in our experiments.

\subsection{Effect of Guidance Model Size}
\begin{table}[]
\resizebox{\columnwidth}{!}{
\begin{tabular}{c|l|rrrr}
Sampler                                                                 & Method                         & \multicolumn{1}{l}{FID} & \multicolumn{1}{l}{IS} & \multicolumn{1}{l}{Precision} & \multicolumn{1}{l}{Recall} \\ \hline
\multirow{4}{*}{\begin{tabular}[c]{@{}c@{}}DDIM\\ 25 step\end{tabular}}  & ResNet152 + PPAP               & \textbf{26.78}                   & \textbf{49.59}                 & \textbf{0.5976}                        & 0.6208                     \\
                                                                         & ResNet50 + PPAP                & 29.65                   & 44.23                  & 0.5872                        & 0.6012                     \\ \cline{2-6} 
                                                                         & ResNet152 + single noise aware & 29.63                   & 42.09                  & 0.5601                        & 0.6305                     \\
                                                                         & ResNet152 + off-the-shelf      & 40.15                   & 33.33                  & 0.5074                        & 0.6175                     \\ \hline
\multirow{4}{*}{\begin{tabular}[c]{@{}c@{}}DDPM\\ 250 step\end{tabular}} & ResNet152 + PPAP               & \textbf{20.34}   & \textbf{61.01}  & \textbf{0.6401}         & 0.6164      \\
                                                                         & ResNet50 + PPAP                & 22.70                   & 52.74                  & 0.6338                        & 0.6187                     \\ \cline{2-6} 
                                                                         & ResNet152 + single noise aware & 22.75                   & 53.38                  & 0.6261                        & 0.6363                     \\
                                                                         & ResNet152 + off-the-shelf      & 31.44                   & 38.28                  & 0.5861                        & 0.6529                     \\ \hline
\end{tabular}
}
\caption{Quantitative results on ADM guidance by increasing the size of classifiers. Increasing the size of the model from ResNet50 to ResNet152 improves the performance of guidance, and PPAP with ResNet152 outperforms single noise-aware ResNet152. Although PPAP is trained in an unsupervised manner, it outperforms the single noise-aware model.
}
\label{apx:table_increasing_model_size}
\end{table}

We also analyze the guidance when the size of the model increases.
Instead of using ResNet50, we use ResNet152 for training the single noise-aware model and PPAP with five experts.
Specifically, we implement LORA with the same configuration in ResNet50, resulting in 8.6\% trainable parameters per expert compared to the parameter of ResNet152.
Training settings such as the optimizer and iterations are the same as ADM guidance with ResNet50.

As shown in Table~\ref{apx:table_increasing_model_size}, we observe that 1) increasing the classifier size from ResNet50 to ResNet152 improves the performance of guidance and 2) PPAP also outperforms the single noise-aware model even when ResNet152 is used.

\subsection{Effect of Multi-Experts}
\label{apx:subsec:multi-experts}
In Section~\textcolor{red}{5.1}, we validate the efficacy of the multi-experts strategy by comparing the results with a range of expert numbers [1, 2, 5, 8, 10].
Here, we present more results for supporting the effectiveness of the multi-experts.

\begin{table}[t]
\centering

\resizebox{\columnwidth}{!}{
\begin{tabular}{c|l|c|cccc}
Sampler                                                                  & Method             & \begin{tabular}[c]{@{}c@{}}Trainable\\ Parameters\end{tabular} & FID   & IS     & Precision & Recall \\ \hline
\multirow{4}{*}{\begin{tabular}[c]{@{}c@{}}DDIM\\ 25 step\end{tabular}}  & ResNet18$\times$5  & 58.4M                                                          & \textbf{20.38} & \textbf{63.74}  & \textbf{0.6638}    & 0.5898 \\
                                                                         & ResNet152$\times$1 & 60M                                                            & 29.63 & 42.09  & 0.5601    & 0.6305 \\ \cline{2-7} 
                                                                         & ResNet50$\times$5  & 109.9M                                                         & \textbf{19.98} & \textbf{74.77} & \textbf{0.6476}    & 0.5887 \\
                                                                         & ResNet50$\times$1  & 25.5M                                                          & 30.42 & 43.05  & 0.5509    & 0.6187 \\ \hline
\multirow{4}{*}{\begin{tabular}[c]{@{}c@{}}DDPM\\ 250 step\end{tabular}} & ResNet18$\times$5  & 58.4M                                                          & \textbf{16.65} & \textbf{76.68}  & \textbf{0.7182}    & 0.579  \\
                                                                         & ResNet152$\times$1 & 60M                                                            & 22.75 & 53.38  & 0.6261    & 0.6363 \\ \cline{2-7} 
                                                                         & ResNet50$\times$5  & 109.9M                                                         & \textbf{16.37} & \textbf{81.47}  & \textbf{0.7216}& 0.5805 \\
                                                                         & ResNet50$\times$1  & 25.5M                                                          & 38.15 & 31.29  & 0.5426    & 0.6321 \\ \hline
\end{tabular}
}
\caption{Quantitative comparison between multi-expert strategy and single noise-aware model. $\times5$ represents using five experts and $\times1$ represents using single noise-aware model. A multi-expert configuration with the same architecture and parameter fair significantly outperforms a single noise-aware model.}
\label{apx:table_resnet18vsresnet152}
\end{table}

We compare the multi-expert strategy with 5 experts and the single-noise-aware model in a setting where the trainable parameters are fair.
For fair trainable parameters, we fine-tune 5 experts ResNet18 and single noise-aware ResNet152~\cite{he2016deep}.
Both models are trained with a total of 300k iterations on the ImageNet train dataset using the AdamW optimizer, a learning rate of 1e-4, and a weight decay of 0.05.

Table~\ref{apx:table_resnet18vsresnet152} shows the quantitative comparison between the multi-expert strategy and a single noise-aware model.
Through these results, we empirically confirm that the multi-expert strategy with the same architecture and parameter fair setting significantly outperforms a single noise-aware model.
Specifically, in parameter fair setting, five experts with ResNet18 significantly outperform single noise-aware ResNet152.
We note that a multi-expert strategy does not incur additional inference time costs, but increasing the size of a single noise-aware model can incur additional inference time costs.
Considering these, it seems much better to construct a multi-expert rather than increase the size of a single noise-aware model.

\begin{table}[t]
    \centering
    \resizebox{0.99\columnwidth}{!}{
    \begin{tabular}{cc||cccc}
    \toprule
         Method & $t$-conditioned &FID & IS & Precision & Recall\\ \hline

        \multirow{2}{*}{Multi-expert-5 } & \cmark & 19.98 & 74.78 & 0.6476 & 0.5887 \\
        
        & \xmark & 19.54 & 73.23 & 0.6509 &  0.5916 \\ \hline
   
        \multirow{2}{*}{Single noise aware} & \cmark & 30.08 & 43.00 & 0.6017 & 0.6059 \\
        & \xmark & 30.42 &43.05 & 0.6009 &0.6187 \\ \hline
        
    \toprule 
    \end{tabular}
    }
    \caption{Impacts of $t$-condition on ResNet50 with DDIM 25 step sampler.}
    \label{tab:t-condition}
\end{table}
\subsection{Effect of Timestep Conditioned Guidance Model}
Multi-experts strategy can be seen as a piecewise function w.r.t time step $t$.
From this point of view, one can wonder about the advantage of multi-experts over $t$-conditioned models.
To answer this and show the effectiveness of the multi-expert strategy, we compare the performance of the model when it is conditioned on $t$ and it is not.
We add the $t$-embeddings of~\cite{dhariwal2021diffusion} to the input of the ResNet50 model for conditioning and generate images with DDIM 25-step sampler. 
As shown in Table~\ref{tab:t-condition}, naively conditioning the model on $t$ does not improve the performance.

\subsection{Effect of Data-Free Knowledge Transfer}
\begin{table}[t]
    \centering
    \resizebox{0.98\columnwidth}{!}{
    \begin{tabular}{ccc||cccc}
    \toprule
         \multirow{2}{*}{Method} & Parameter- & \multirow{2}{*}{Dataset} & \multirow{2}{*}{FID} & \multirow{2}{*}{IS}  & \multirow{2}{*}{Precision} & \multirow{2}{*}{Recall}\\
         & efficient & & & &\\ \hline
         
          & \multirow{2}{*}{\xmark} & ImageNet & 19.98& 74.78& 0.6476& 0.5887\\
         
          Multi-&  & Data-Free & 29.87 & 44.10 & 0.5837 & 0.5912 \\ \cline{2-7}

         experts-5 & \multirow{2}{*}{\cmark} & ImageNet & 20.32 & 72.30 & 0.6333 & 0.5898
        \\
         
       & & Data-Free & 29.65 & 44.23 & 0.5872 & 0.6012 \\ \toprule
    \end{tabular}
    }
    \caption{Impacts of the data-free strategy on ResNet50 with DDIM 25 step sampler.}
    \label{tab:data-free}
\end{table}
We conducted an ablation study for understanding the impacts of the data-free strategy.
We used five ResNet50 experts with and without parameter efficient strategy applied in them.
Table~\ref{tab:data-free} shows the quantitative results when these five experts were supervisely trained on ImageNet and trained on data-free knowledge transfer.
The results indicate that the models trained on labeled data produce better guidance than those trained with the data-free strategy.

\section{More Qualitative Results on ADM}
\label{apx:sec:adm-more-qual}
\subsection{Guided ADM with ResNet50}
\label{apx:subsec:adm+res}

\begin{figure*}[h]

    \centering
    \begin{tabular}{@{\hspace{2mm}}c@{\hspace{2mm}}c@{\hspace{2mm}}c@{\hspace{2mm}}c@{\hspace{2mm}}c@{\hspace{2mm}}c@{\hspace{2mm}}c@{\hspace{2mm}}c}
    
        \raisebox{0.2\height}{\rotatebox{90}{\scriptsize Na\"ive Off-the-shelf}} 
        \adjincludegraphics[clip,width=0.15\textwidth,trim={0 0 0 0}]{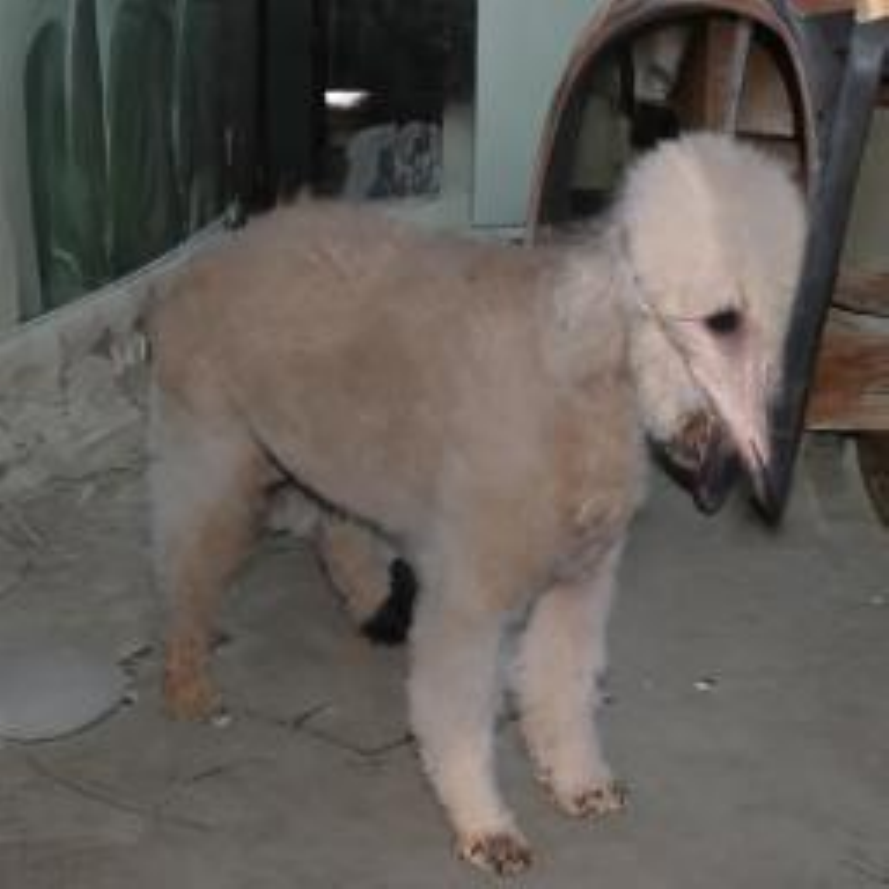} &
        \adjincludegraphics[clip,width=0.15\textwidth,trim={0 0 0 0}]{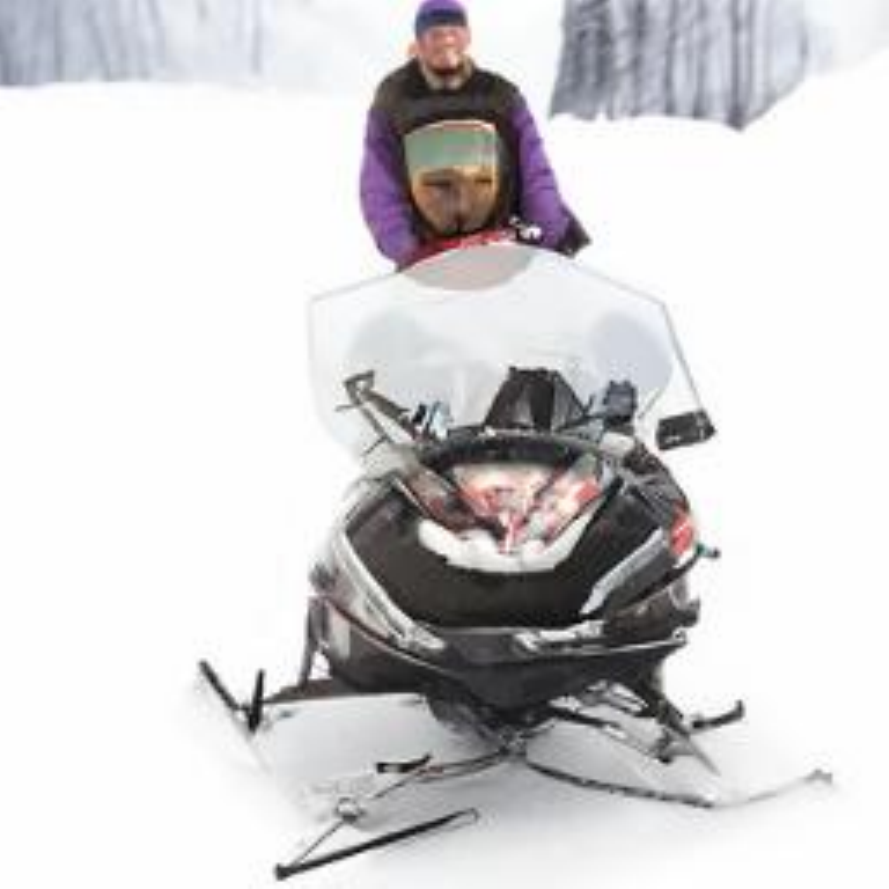} &
        \adjincludegraphics[clip,width=0.15\textwidth,trim={0 0 0 0}]{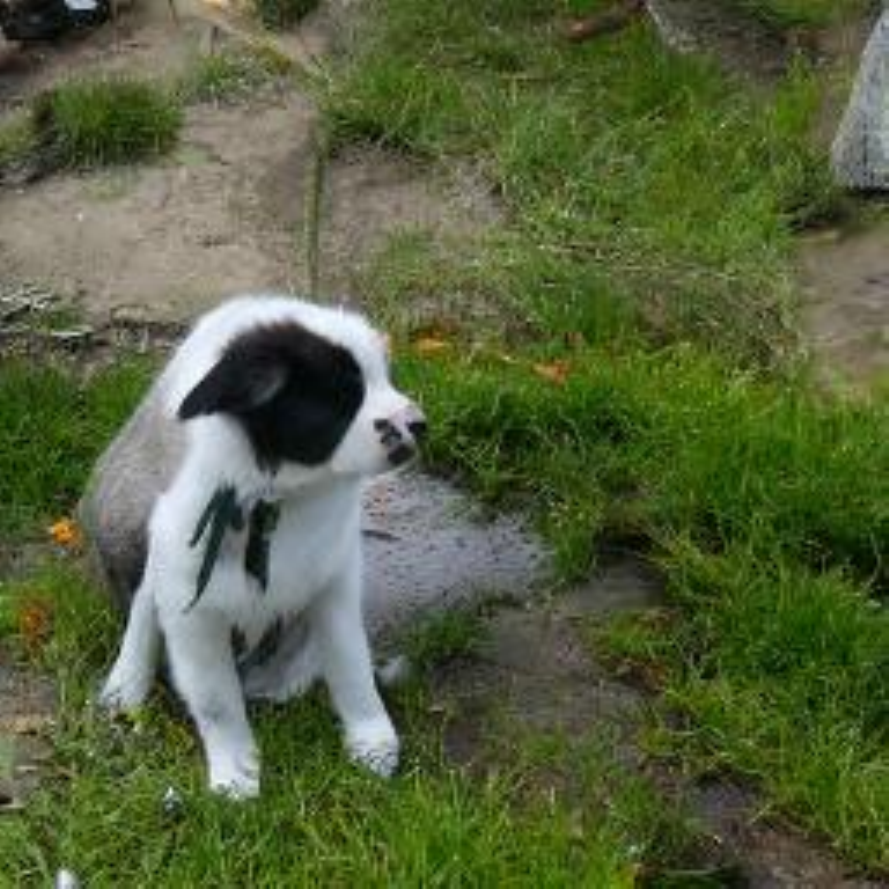} &
        \adjincludegraphics[clip,width=0.15\textwidth,trim={0 0 0 0}]{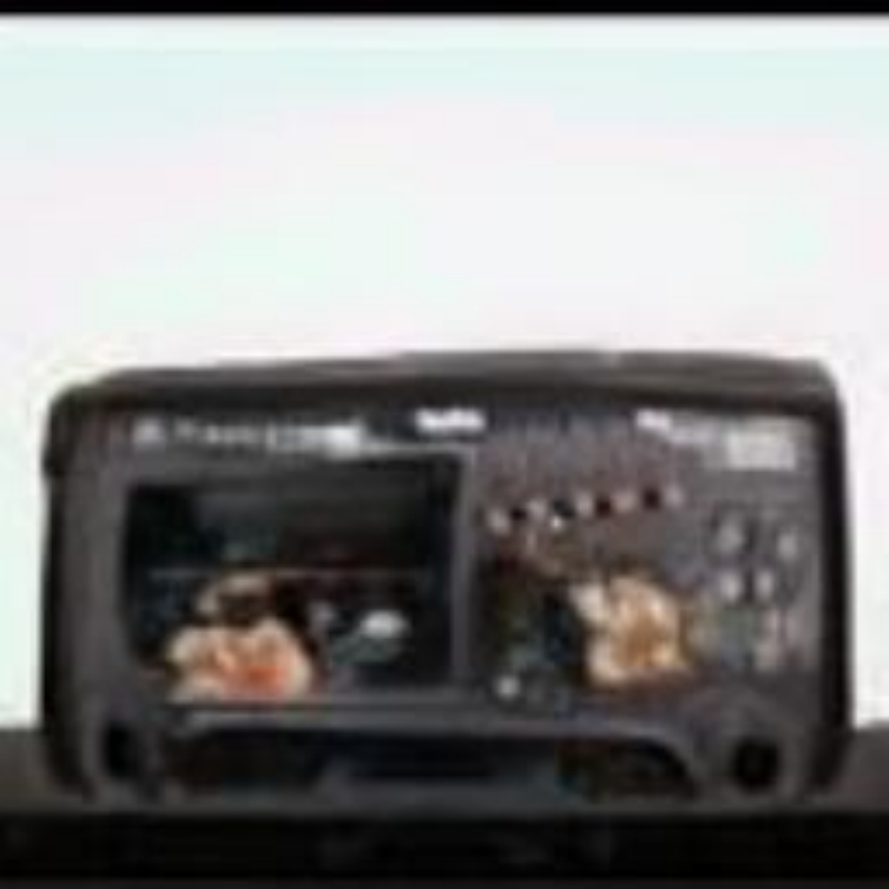} &
        \adjincludegraphics[clip,width=0.15\textwidth,trim={0 0 0 0}]{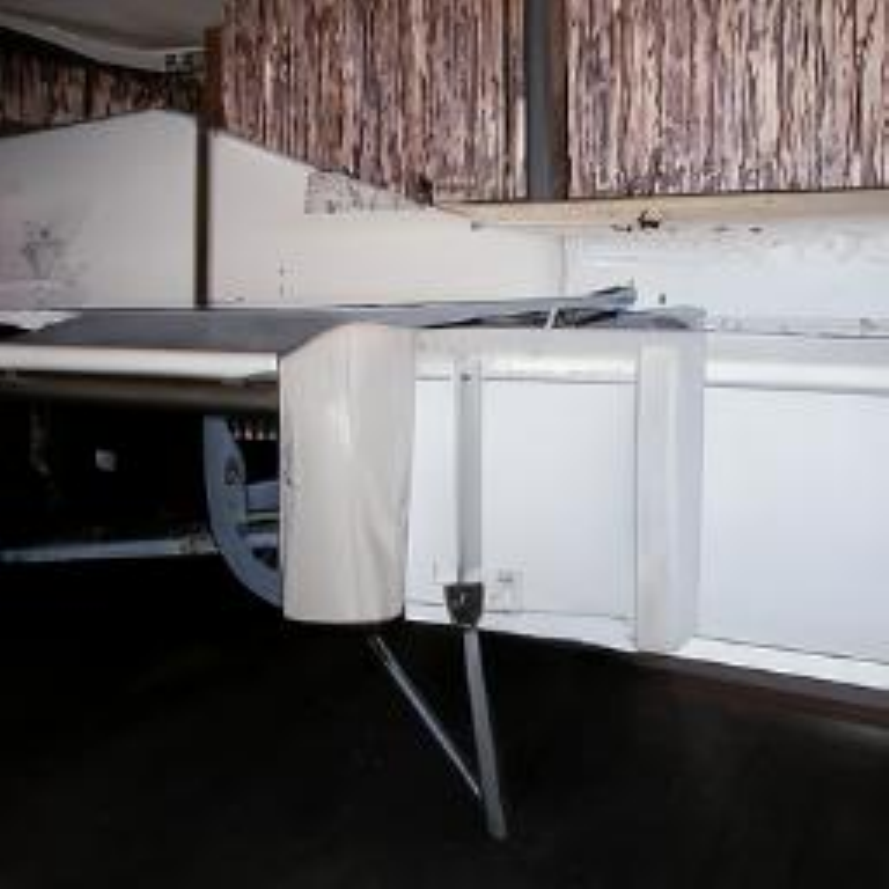} &
        \adjincludegraphics[clip,width=0.15\textwidth,trim={0 0 0 0}]{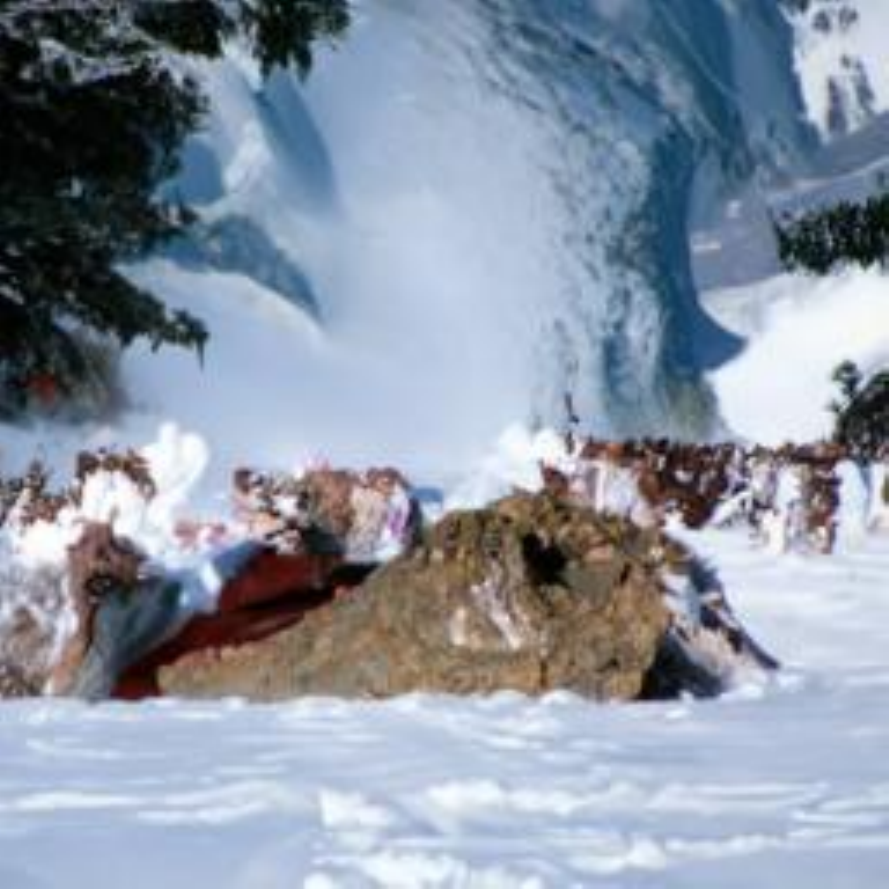} \\

            \raisebox{0.2\height}{\rotatebox{90}{\scriptsize Single Noise-aware}} 
        \adjincludegraphics[clip,width=0.15\textwidth,trim={0 0 0 0}]{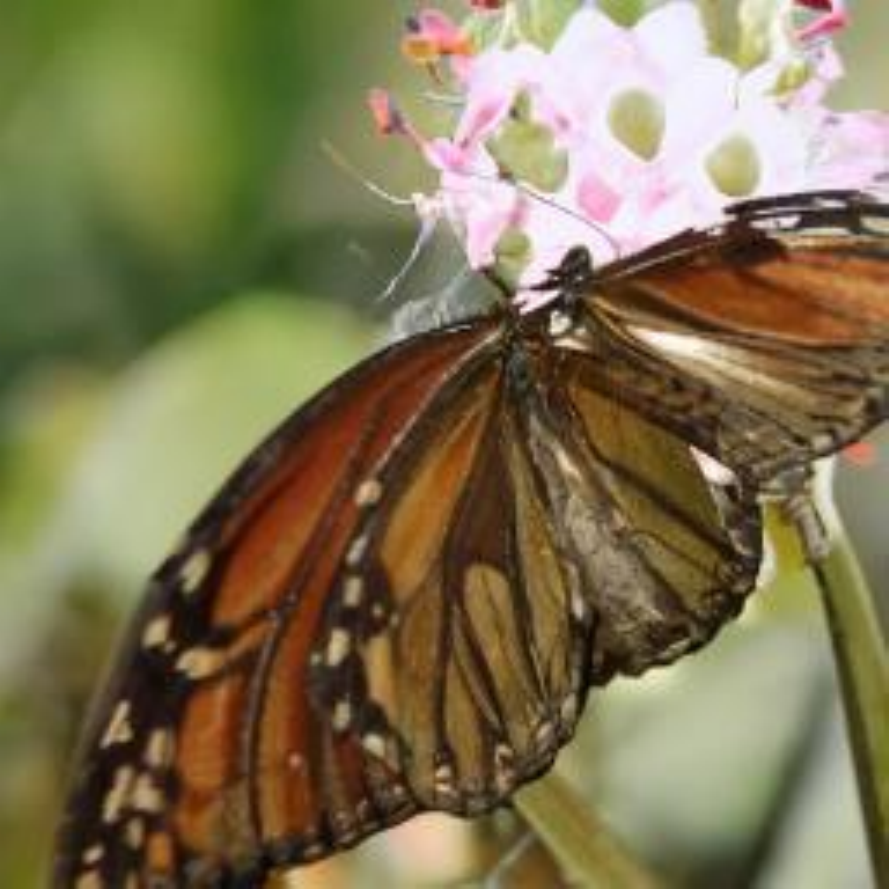} &
        \adjincludegraphics[clip,width=0.15\textwidth,trim={0 0 0 0}]{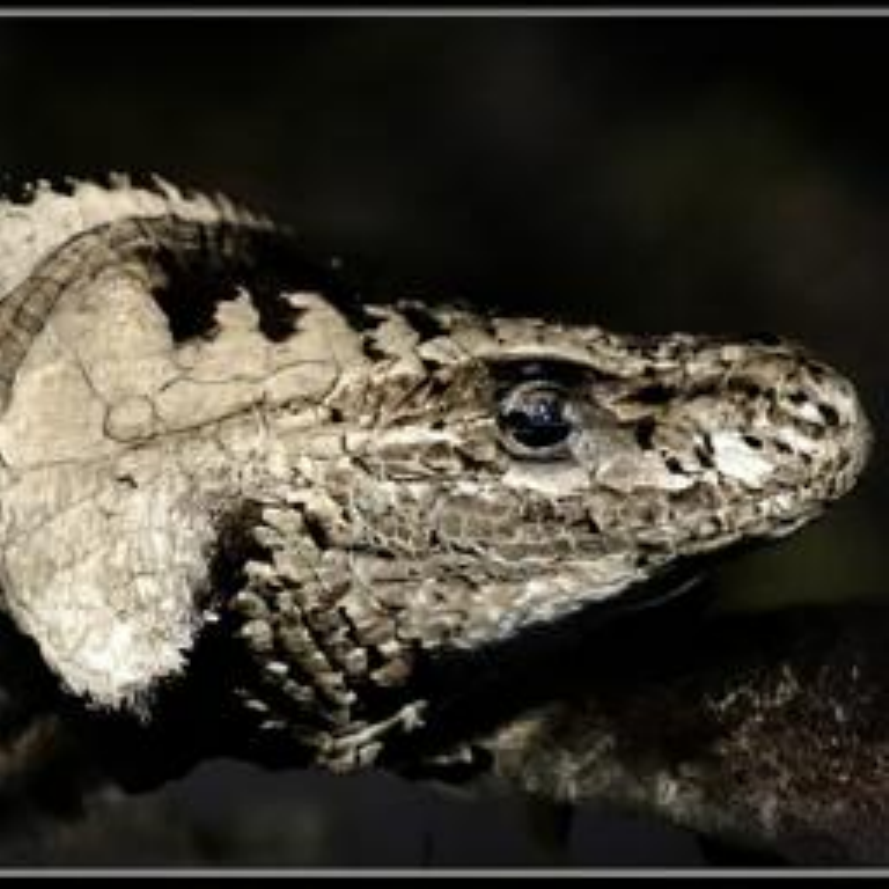} &
        \adjincludegraphics[clip,width=0.15\textwidth,trim={0 0 0 0}]{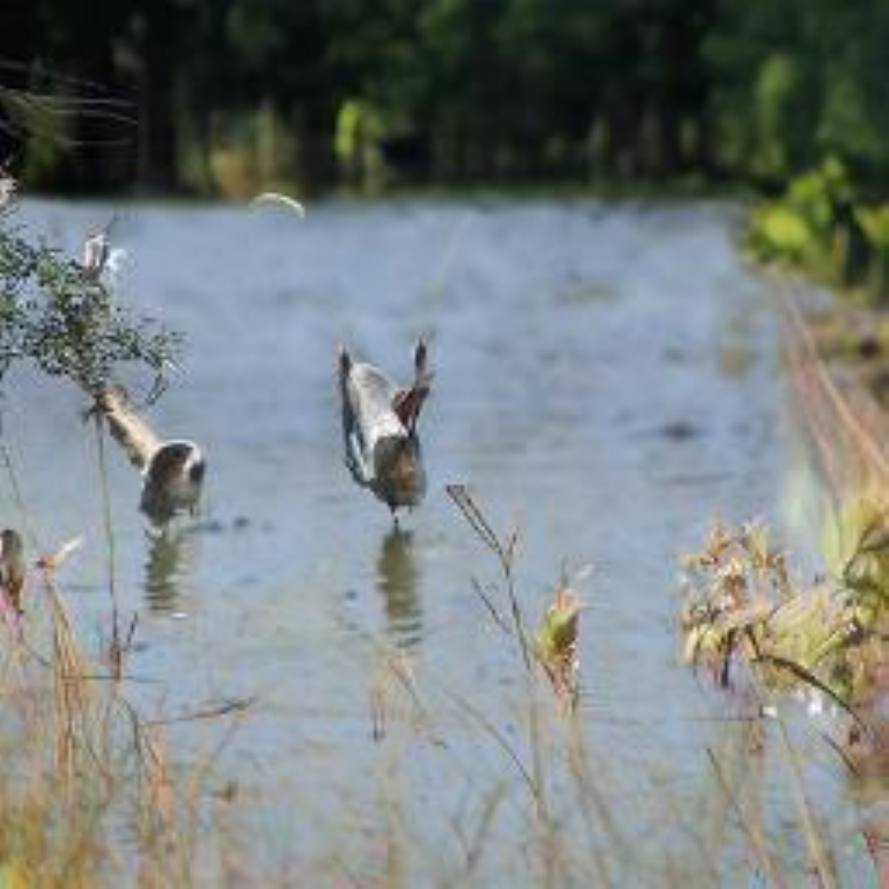} &
        \adjincludegraphics[clip,width=0.15\textwidth,trim={0 0 0 0}]{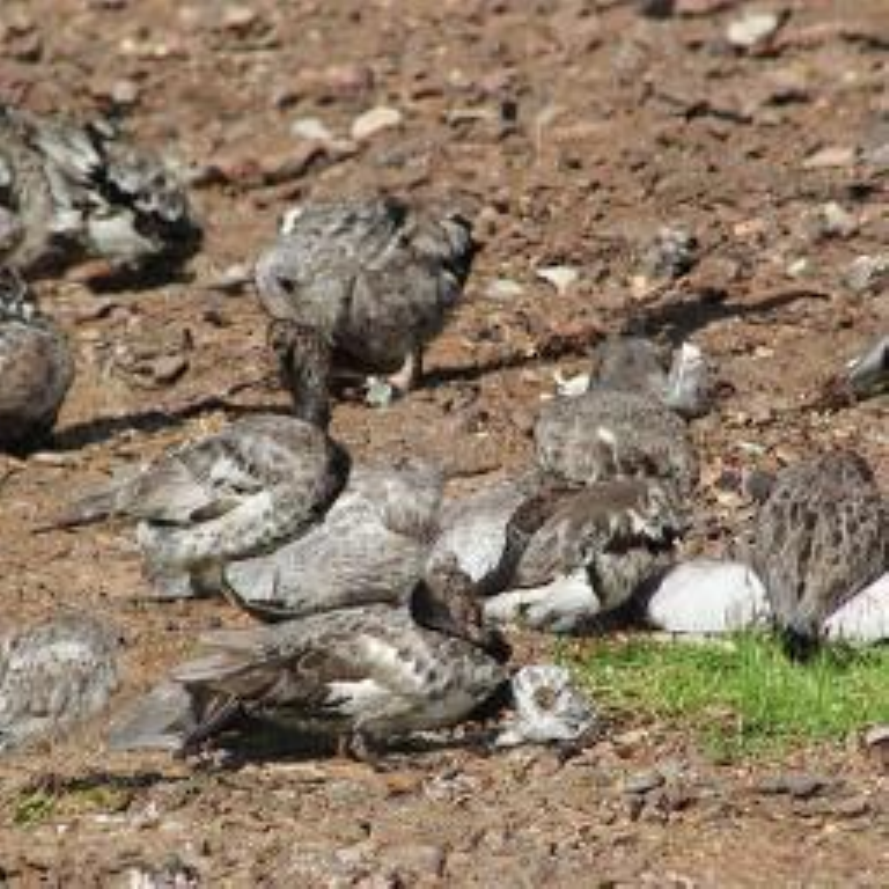} &
        \adjincludegraphics[clip,width=0.15\textwidth,trim={0 0 0 0}]{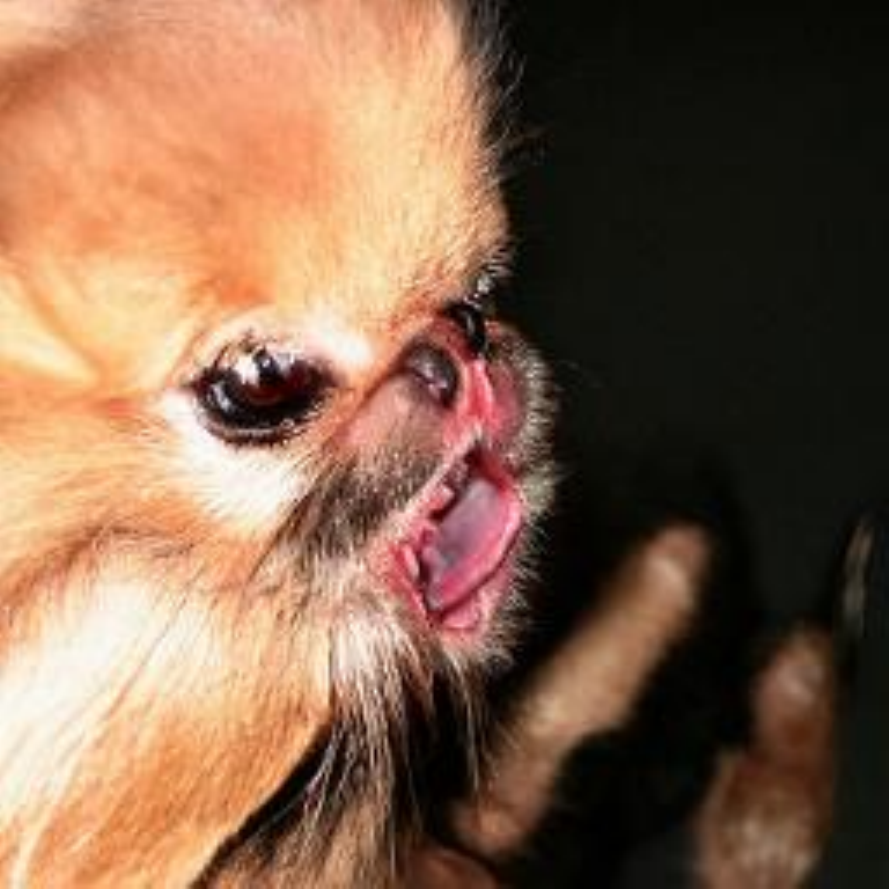} &
        \adjincludegraphics[clip,width=0.15\textwidth,trim={0 0 0 0}]{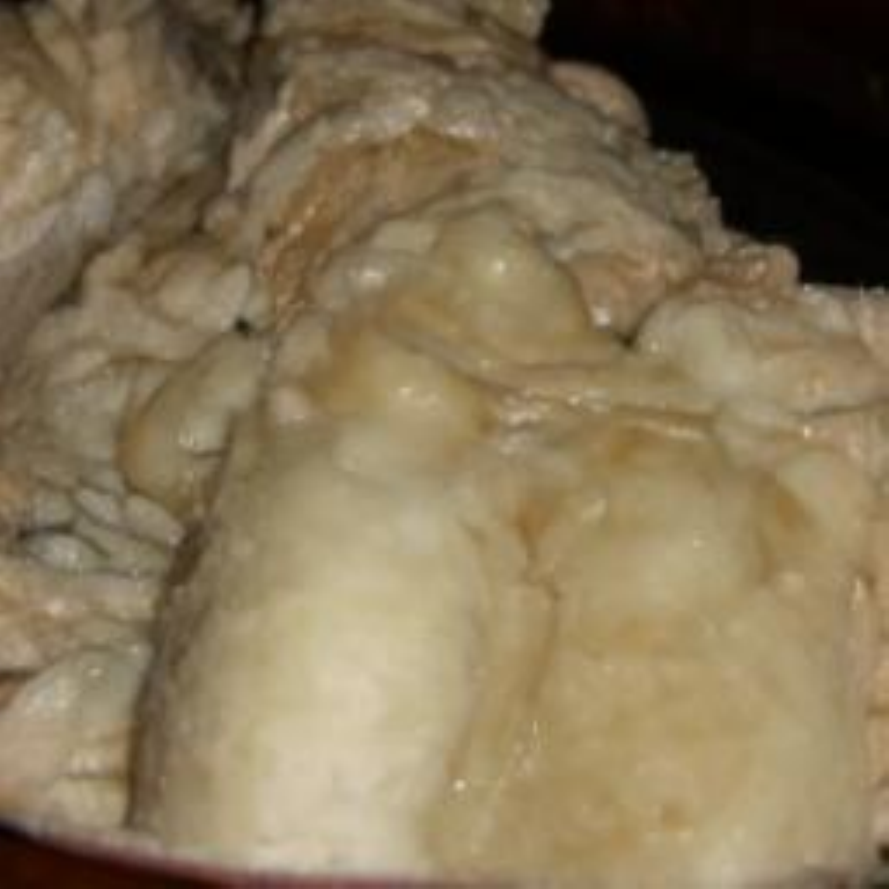} \\

        \raisebox{0\height}{\rotatebox{90}{\scriptsize Supervised Multi-Experts}} 
        \adjincludegraphics[clip,width=0.15\textwidth,trim={0 0 0 0}]{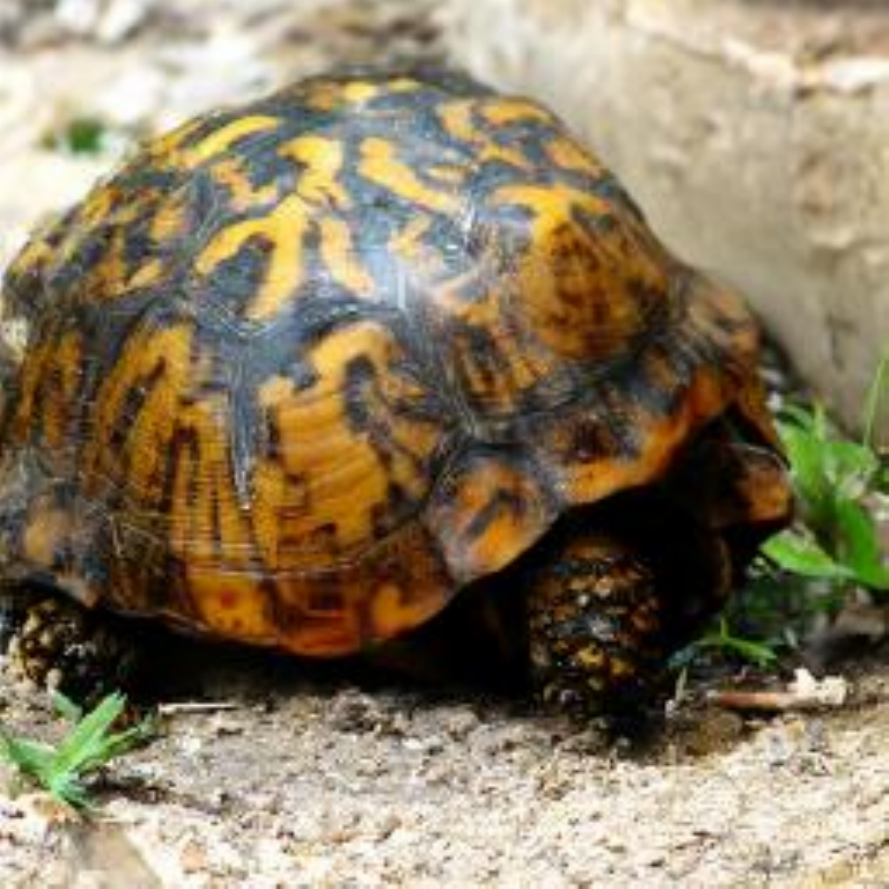} &
        \adjincludegraphics[clip,width=0.15\textwidth,trim={0 0 0 0}]{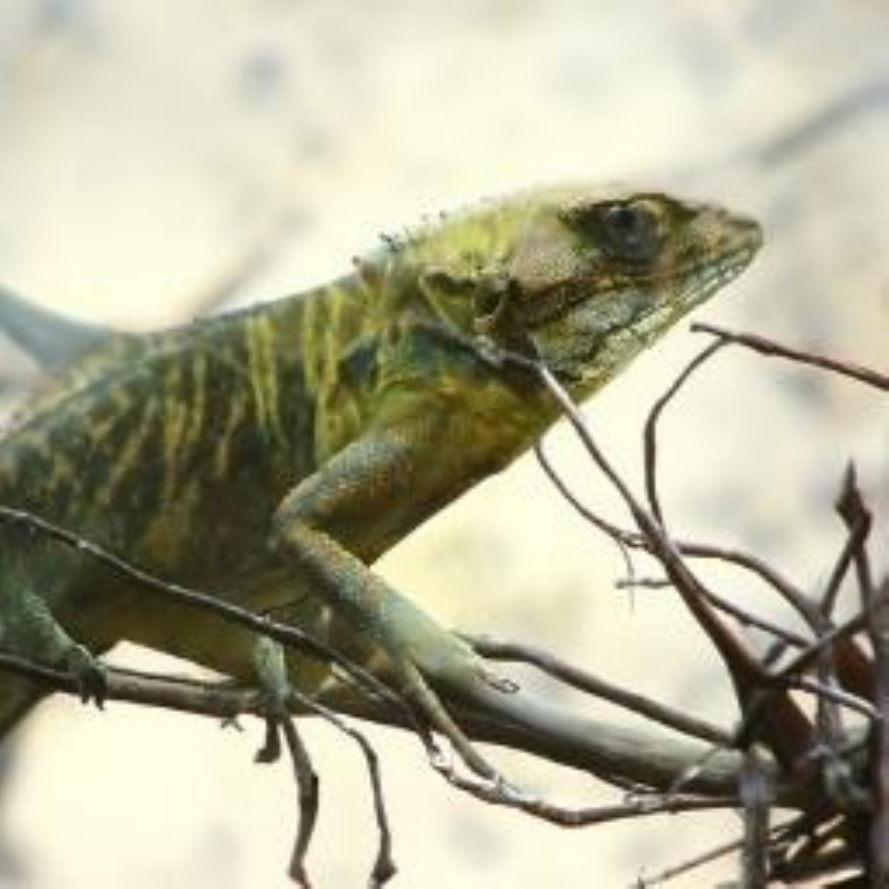} &
        \adjincludegraphics[clip,width=0.15\textwidth,trim={0 0 0 0}]{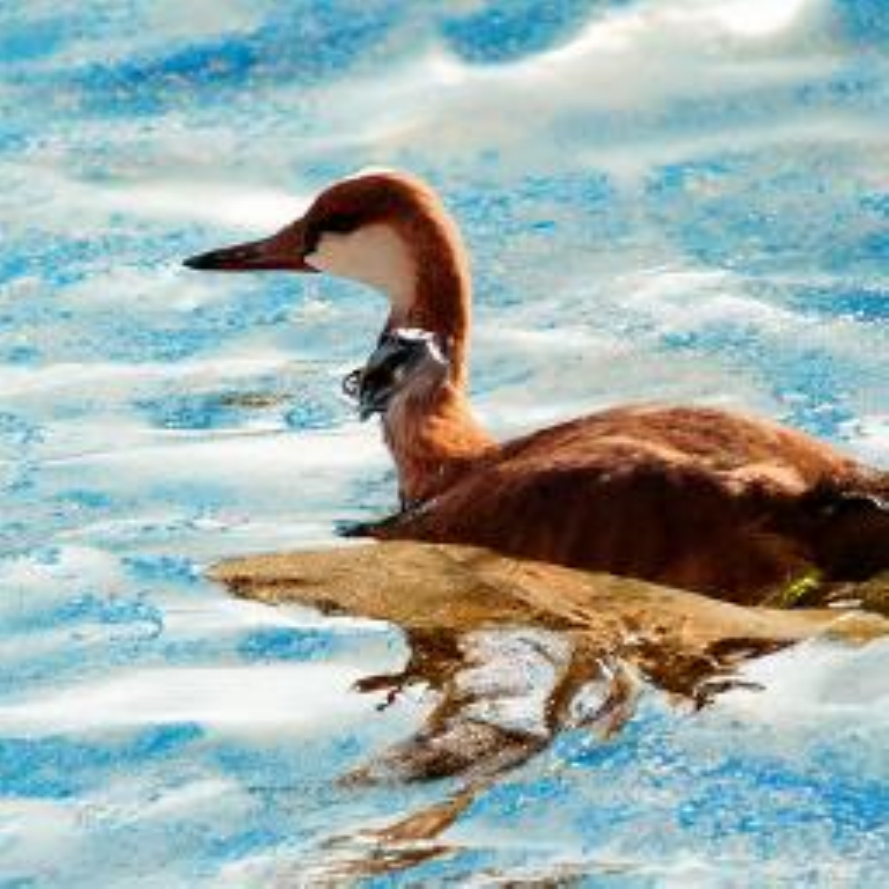} &
        \adjincludegraphics[clip,width=0.15\textwidth,trim={0 0 0 0}]{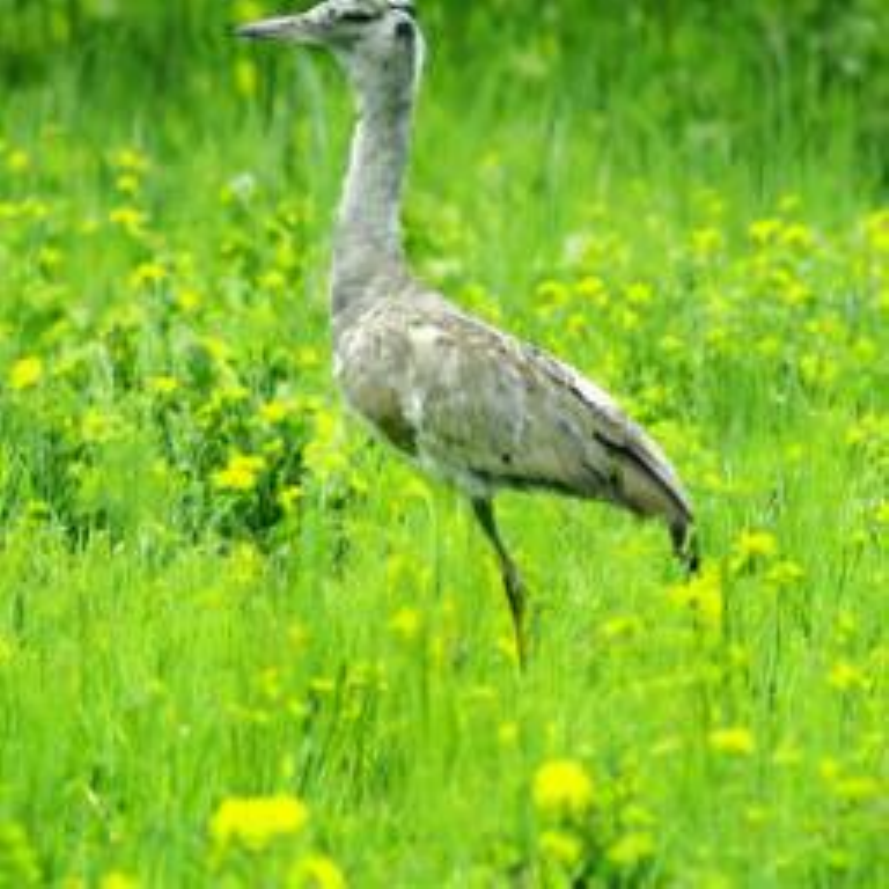} &
        \adjincludegraphics[clip,width=0.15\textwidth,trim={0 0 0 0}]{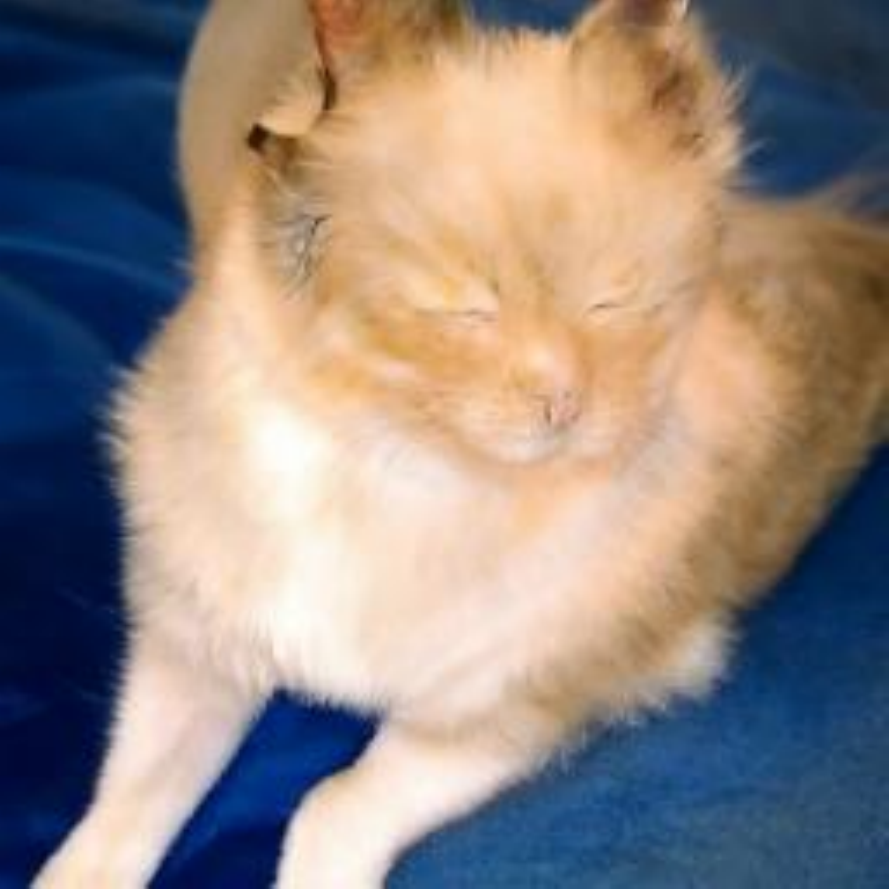} &
        \adjincludegraphics[clip,width=0.15\textwidth,trim={0 0 0 0}]{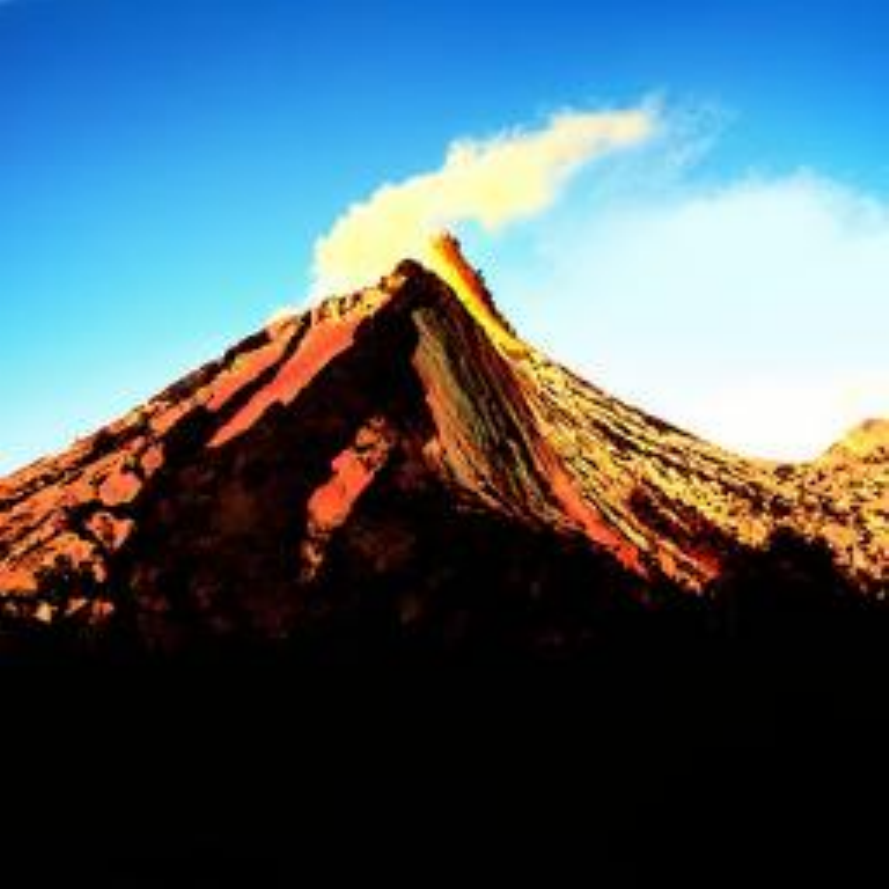} \\

        \raisebox{0.35\height}{\rotatebox{90}{\scriptsize PPAP ($N=5$)}} 
        \adjincludegraphics[clip,width=0.15\textwidth,trim={0 0 0 0}]{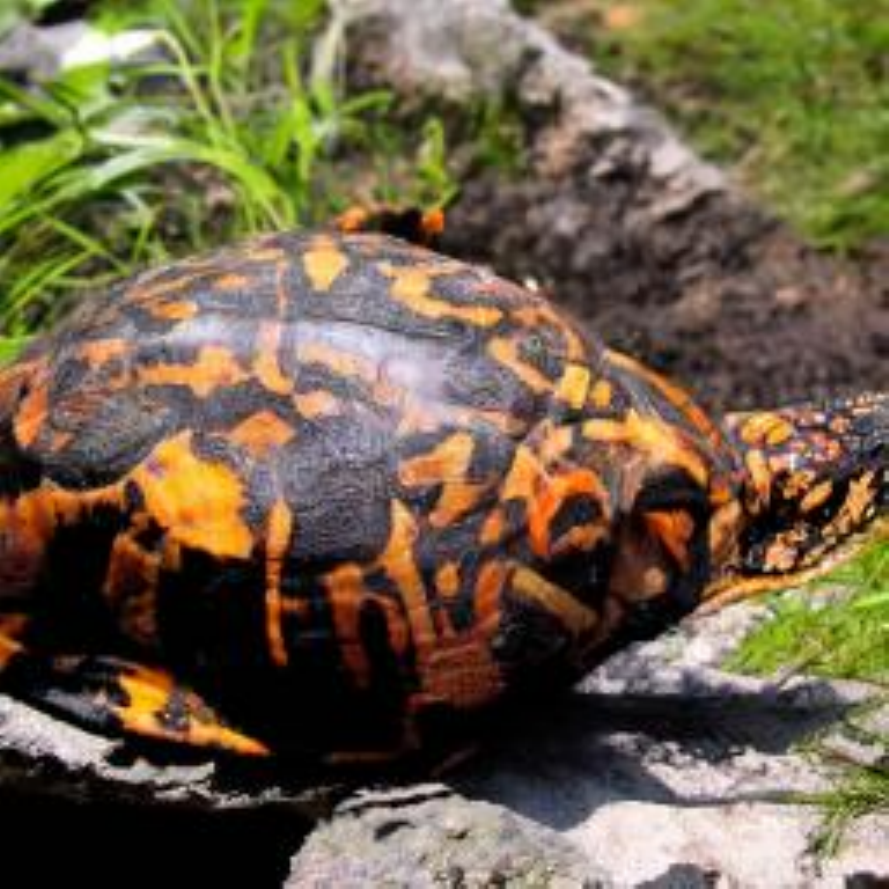} &
        \adjincludegraphics[clip,width=0.15\textwidth,trim={0 0 0 0}]{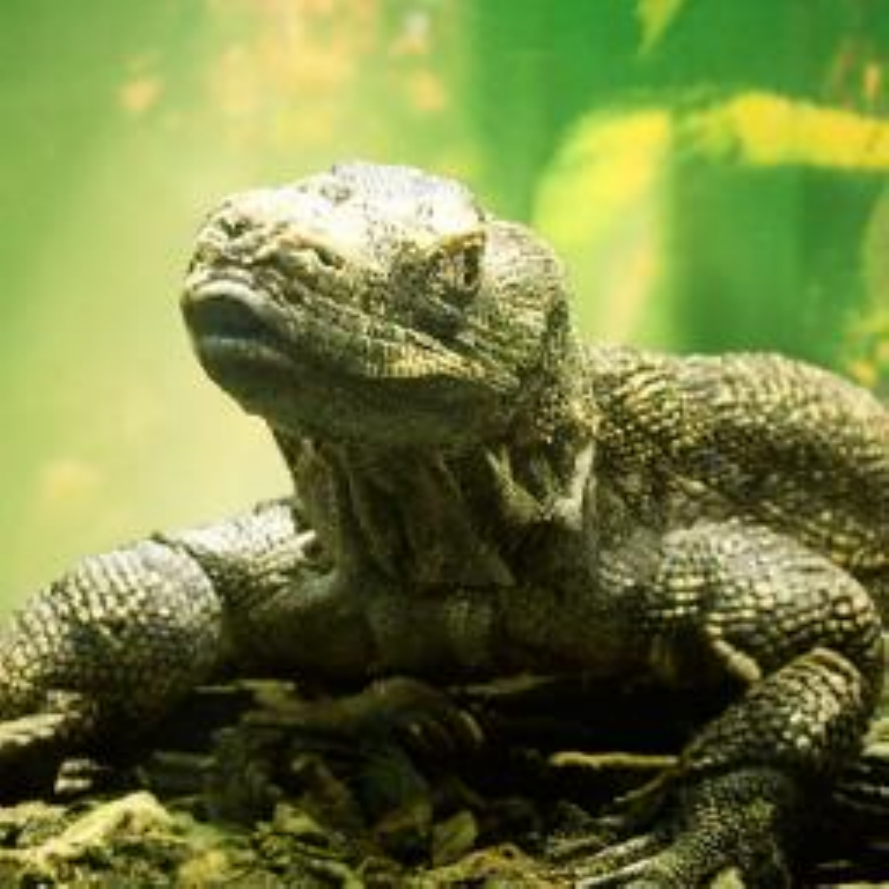} &
        \adjincludegraphics[clip,width=0.15\textwidth,trim={0 0 0 0}]{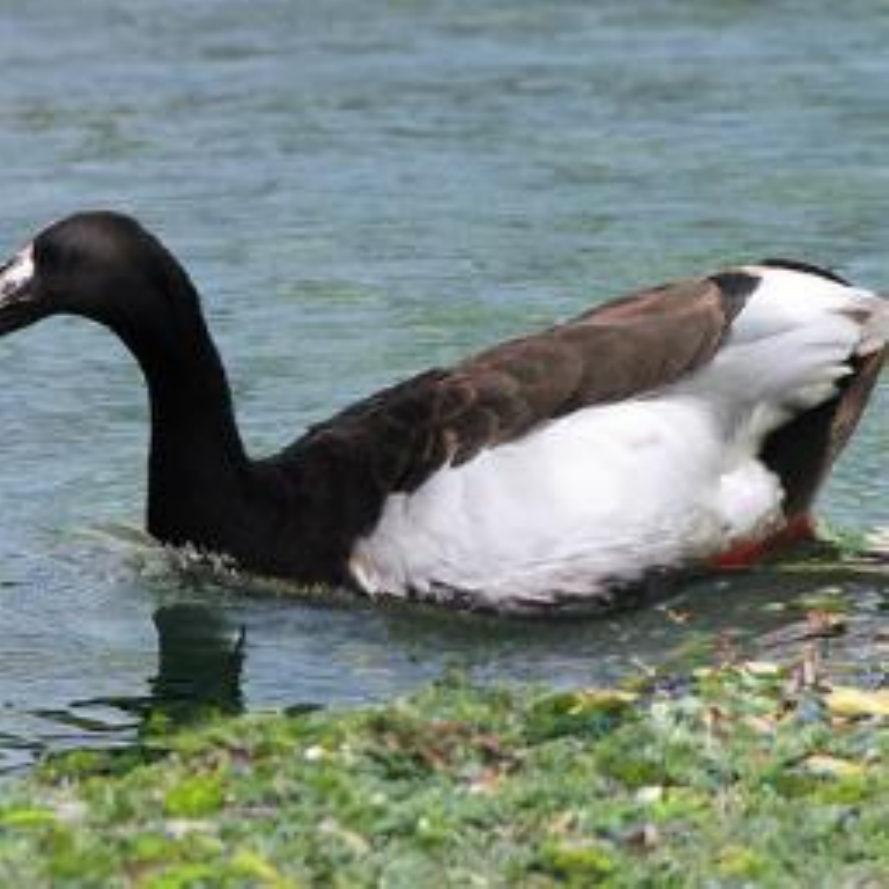} &
        \adjincludegraphics[clip,width=0.15\textwidth,trim={0 0 0 0}]{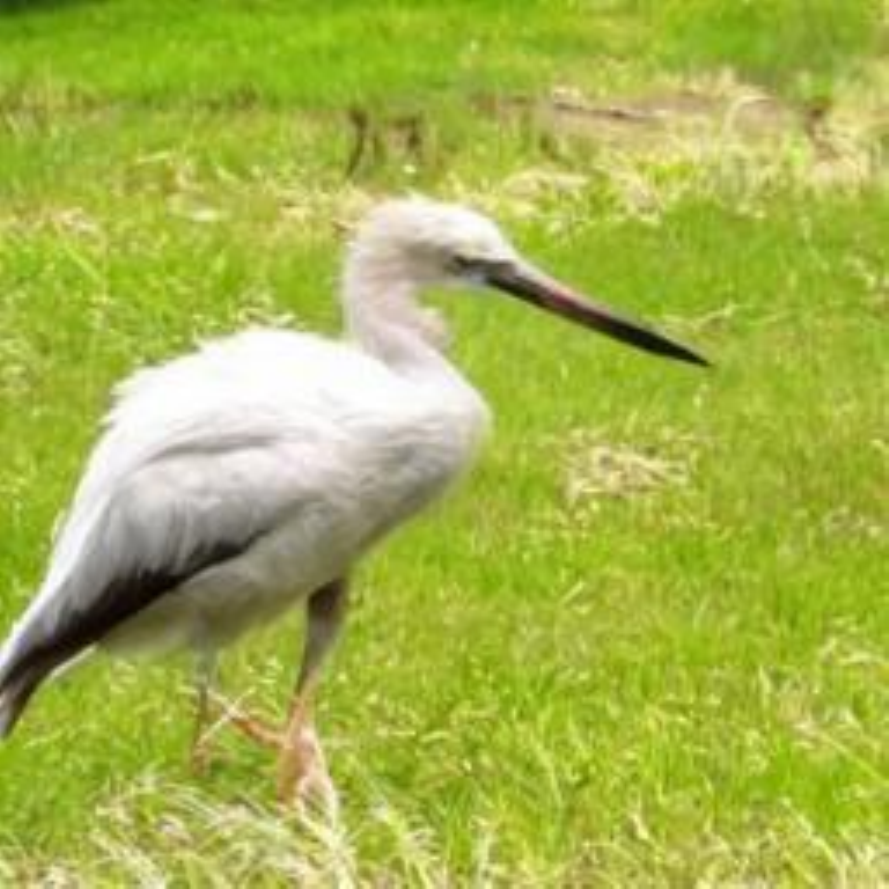} &
        \adjincludegraphics[clip,width=0.15\textwidth,trim={0 0 0 0}]{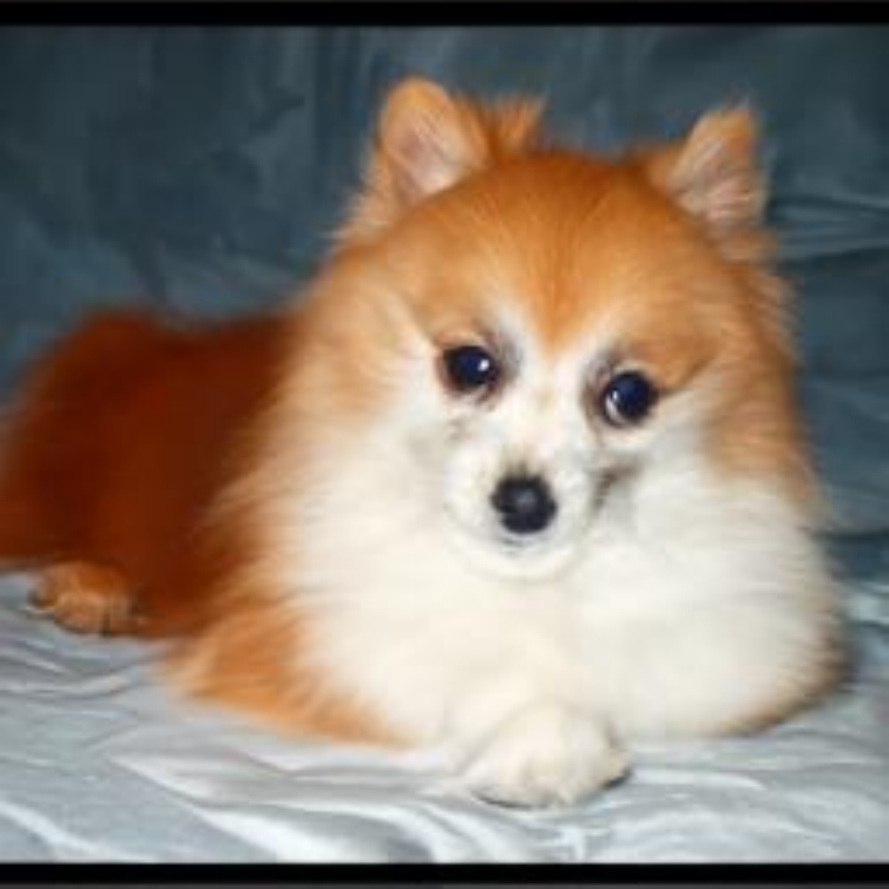} &
        \adjincludegraphics[clip,width=0.15\textwidth,trim={0 0 0 0}]{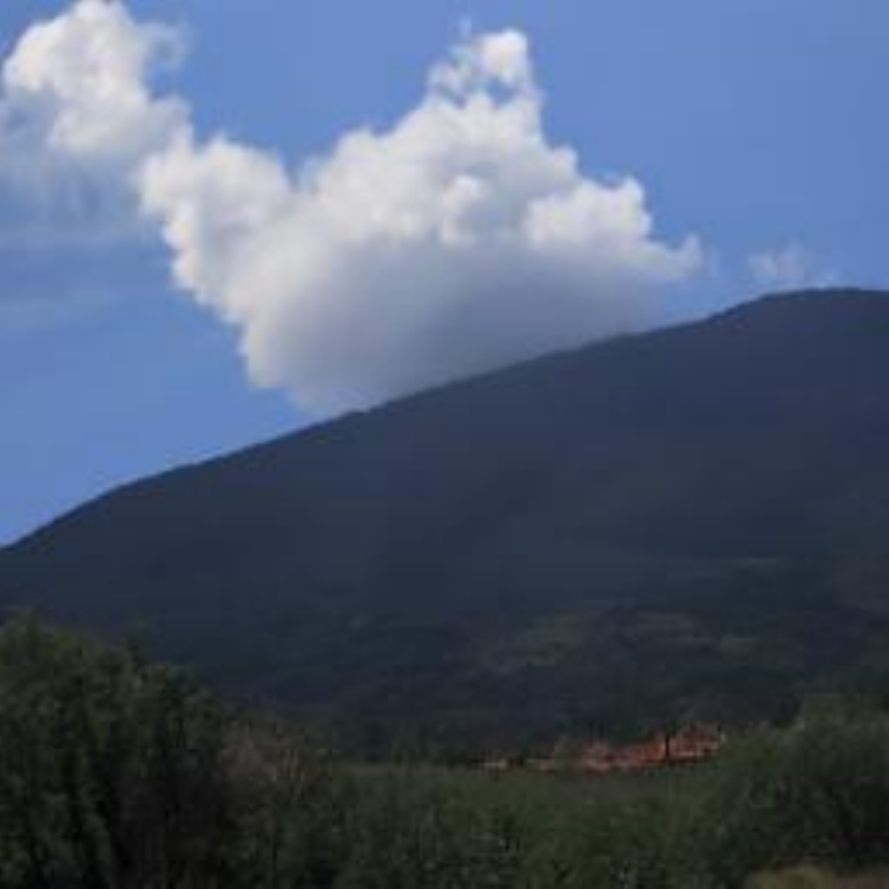} \\

        \raisebox{0.3\height}{\rotatebox{90}{\scriptsize PPAP ($N=10$)}} 
        \adjincludegraphics[clip,width=0.15\textwidth,trim={0 0 0 0}]{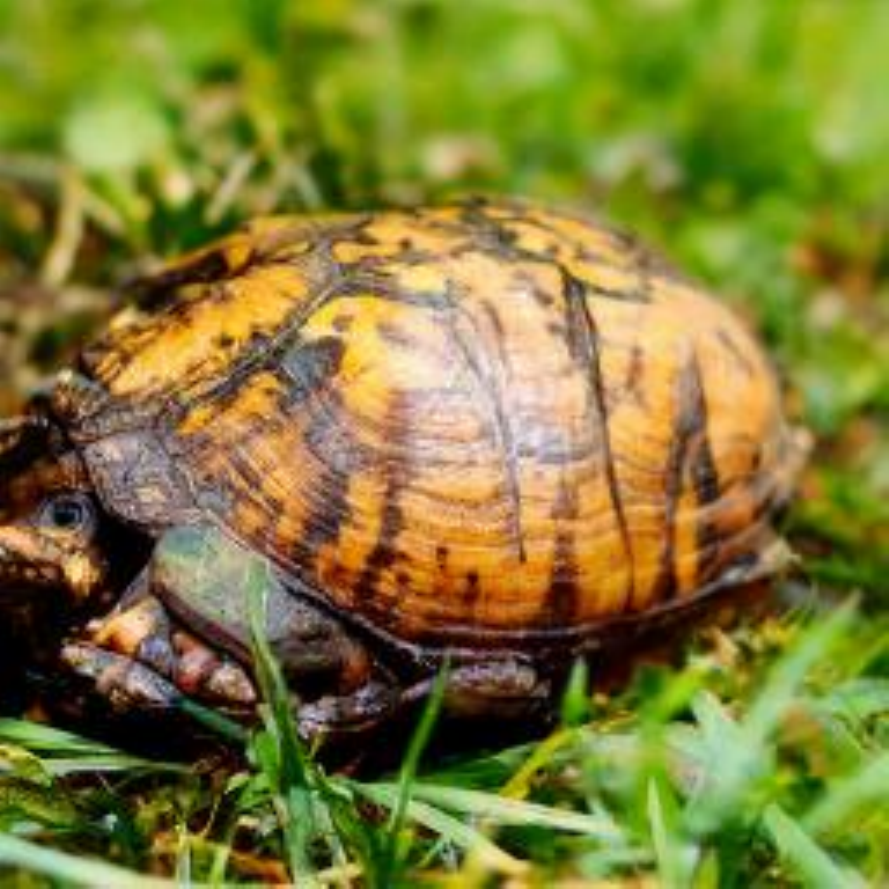} &
        \adjincludegraphics[clip,width=0.15\textwidth,trim={0 0 0 0}]{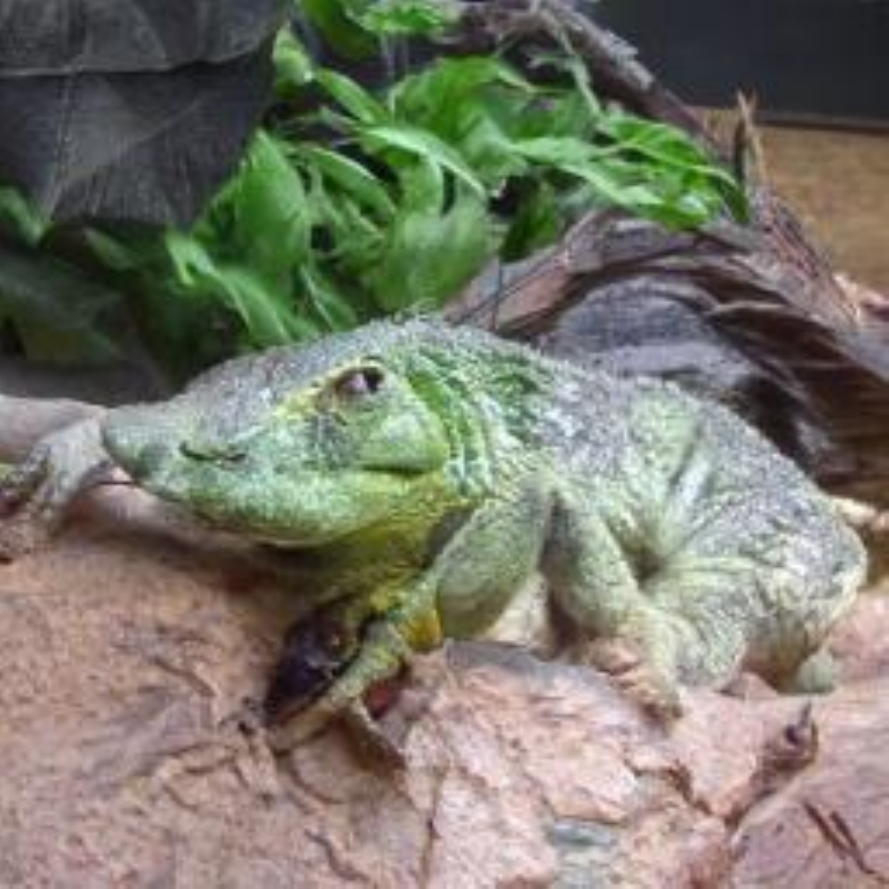} &
        \adjincludegraphics[clip,width=0.15\textwidth,trim={0 0 0 0}]{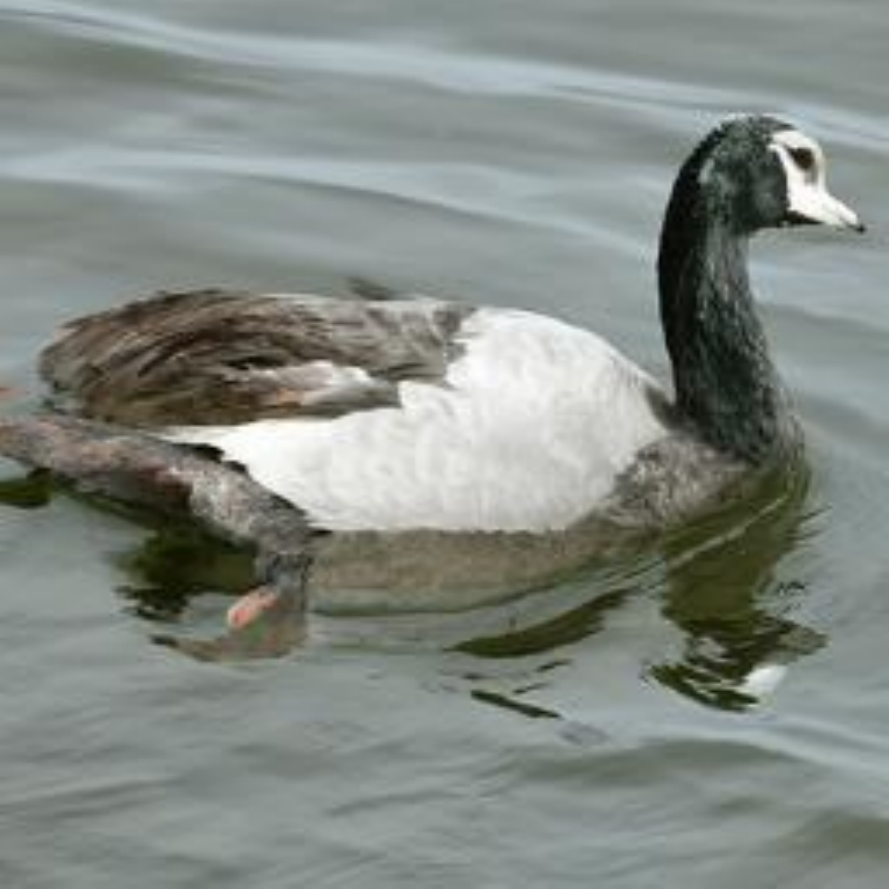} &
        \adjincludegraphics[clip,width=0.15\textwidth,trim={0 0 0 0}]{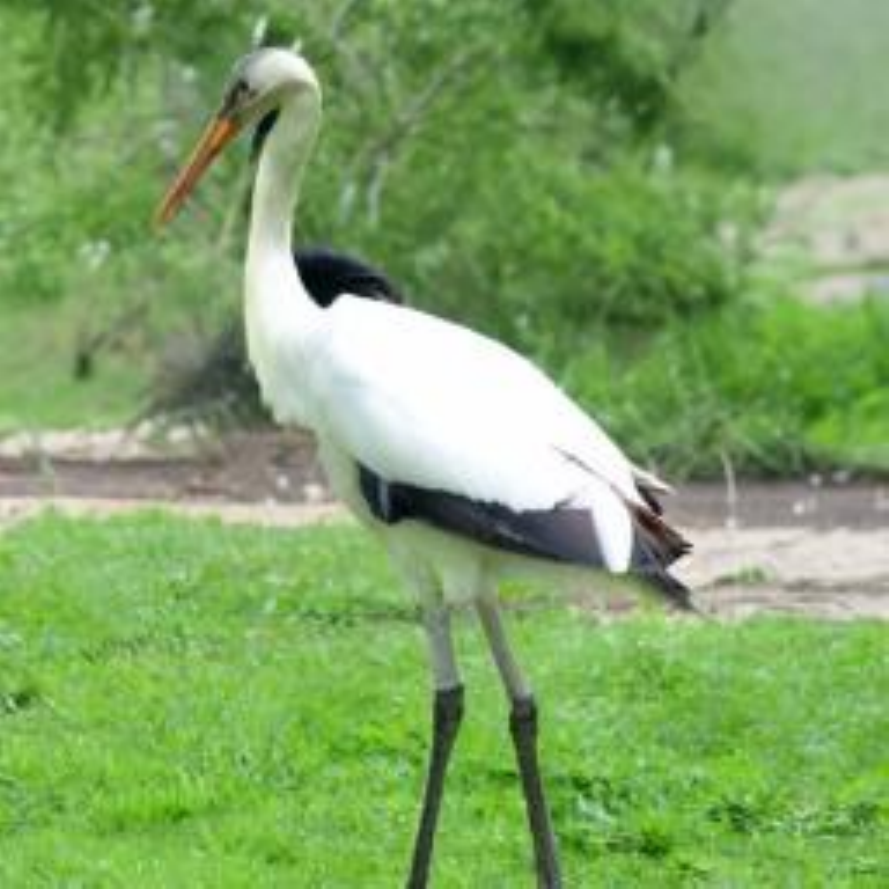} &
        \adjincludegraphics[clip,width=0.15\textwidth,trim={0 0 0 0}]{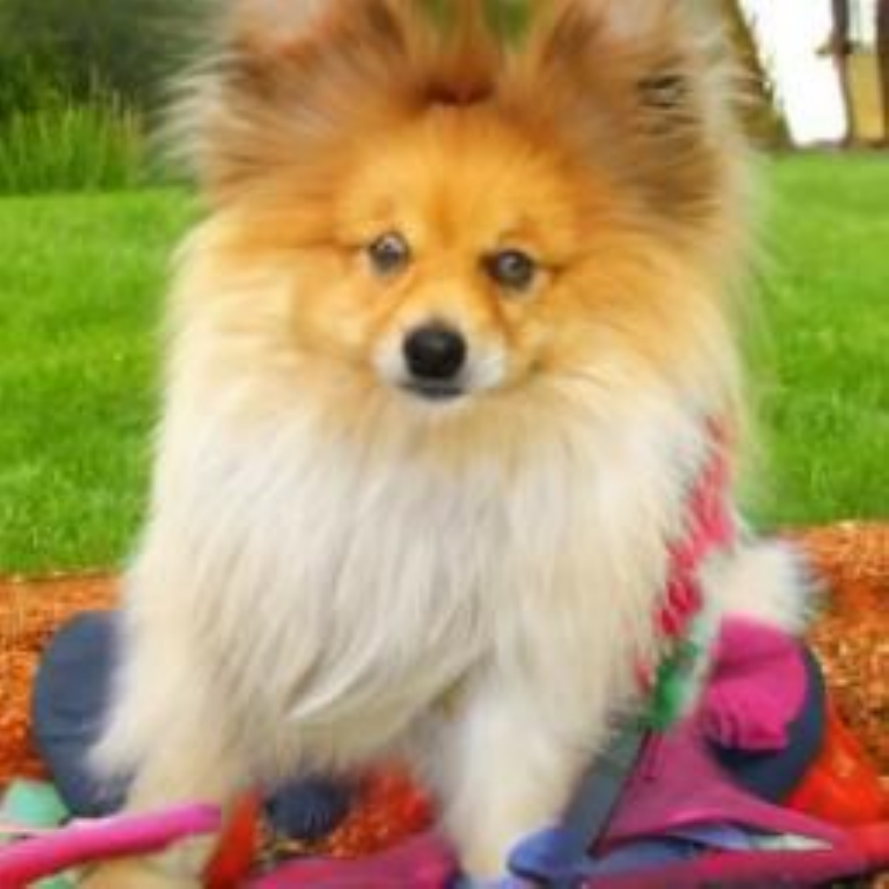} &
        \adjincludegraphics[clip,width=0.15\textwidth,trim={0 0 0 0}]{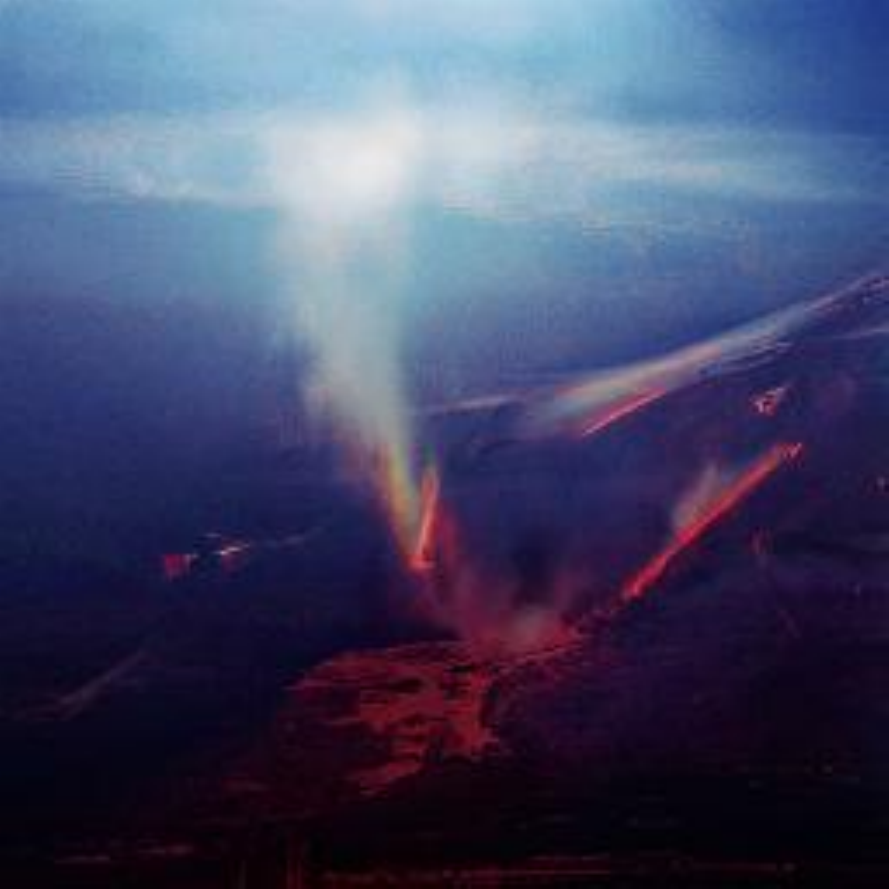} \\
        
        \footnotesize ~~~~~~Box turtle & \footnotesize  Iguana & \footnotesize  Goose & \footnotesize  Crane & \footnotesize Pomeranian & \footnotesize Volcano \\
    \end{tabular}
    \caption{
    Qualitative results on ImageNet class conditional generation with DDPM 250 steps by guiding unconditional ADM with ResNet50.
    }
    \label{apx:fig_qualitative_resnet_ddpm}

\end{figure*}

\begin{figure*}[h]

    \centering
    \begin{tabular}{@{\hspace{2mm}}c@{\hspace{2mm}}c@{\hspace{2mm}}c@{\hspace{2mm}}c@{\hspace{2mm}}c@{\hspace{2mm}}c@{\hspace{2mm}}c@{\hspace{2mm}}c}
    
        \raisebox{0.2\height}{\rotatebox{90}{\scriptsize Na\"ive Off-the-shelf}} 
        \adjincludegraphics[clip,width=0.15\textwidth,trim={0 0 0 0}]{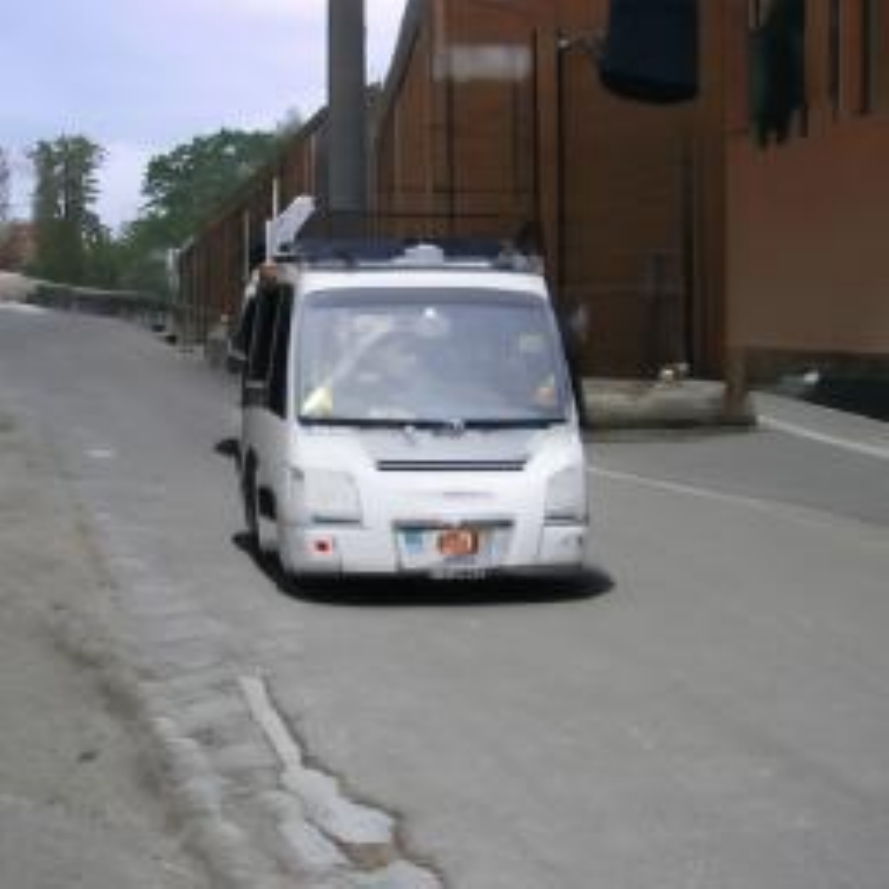} &
        \adjincludegraphics[clip,width=0.15\textwidth,trim={0 0 0 0}]{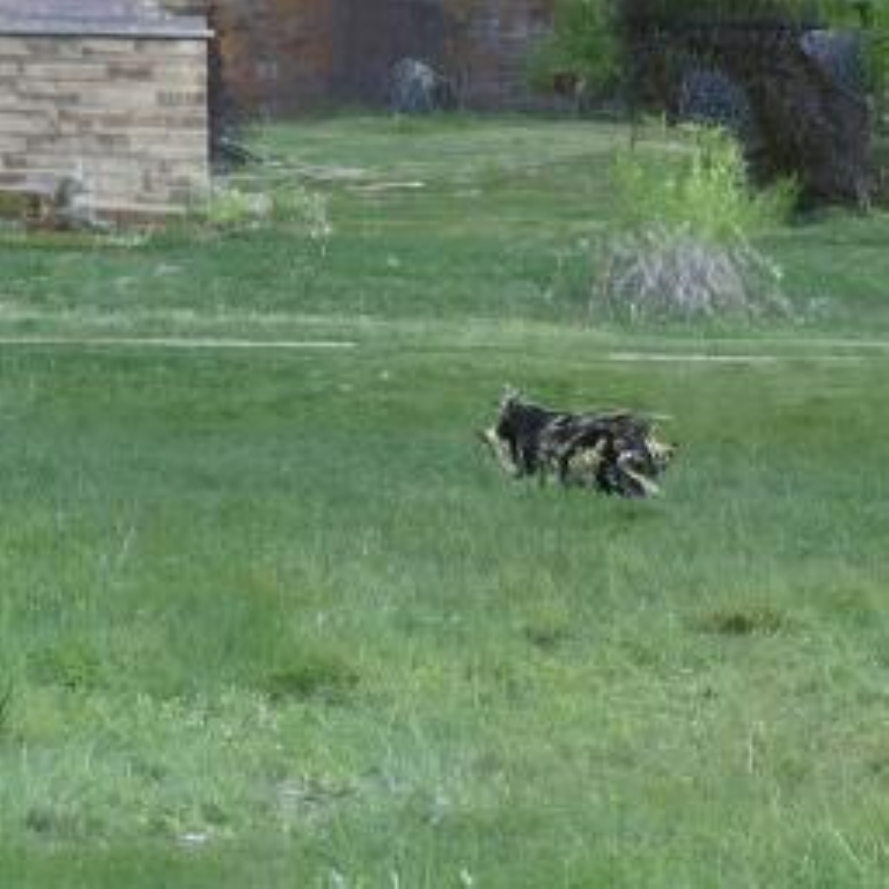} &
        \adjincludegraphics[clip,width=0.15\textwidth,trim={0 0 0 0}]{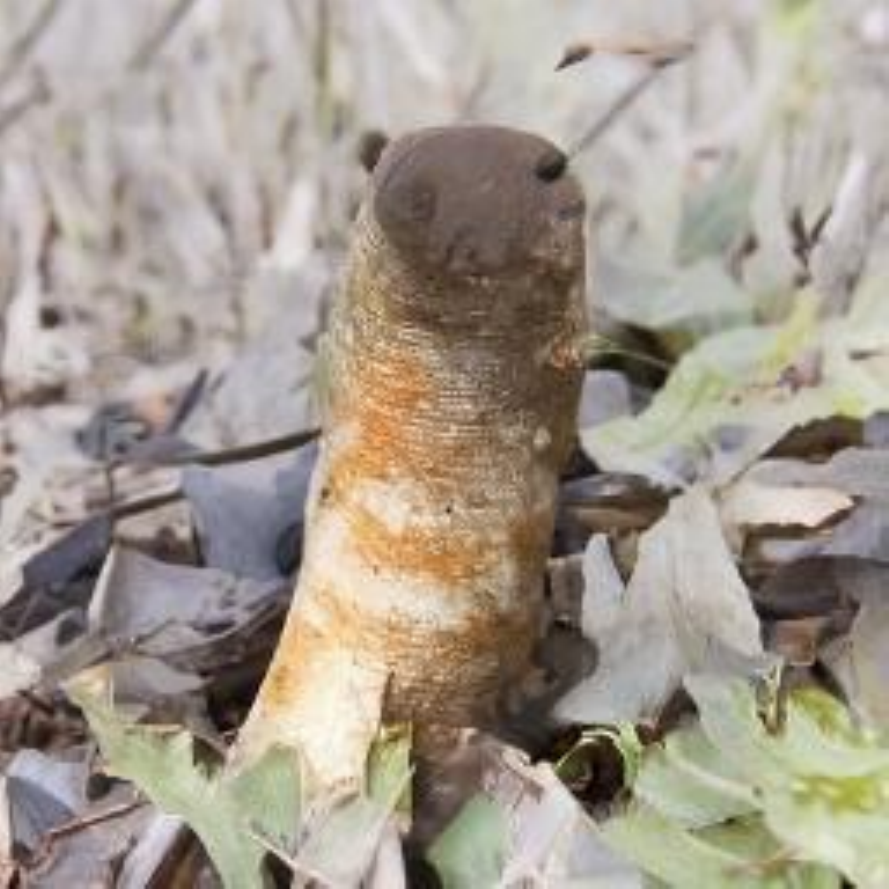} &
        \adjincludegraphics[clip,width=0.15\textwidth,trim={0 0 0 0}]{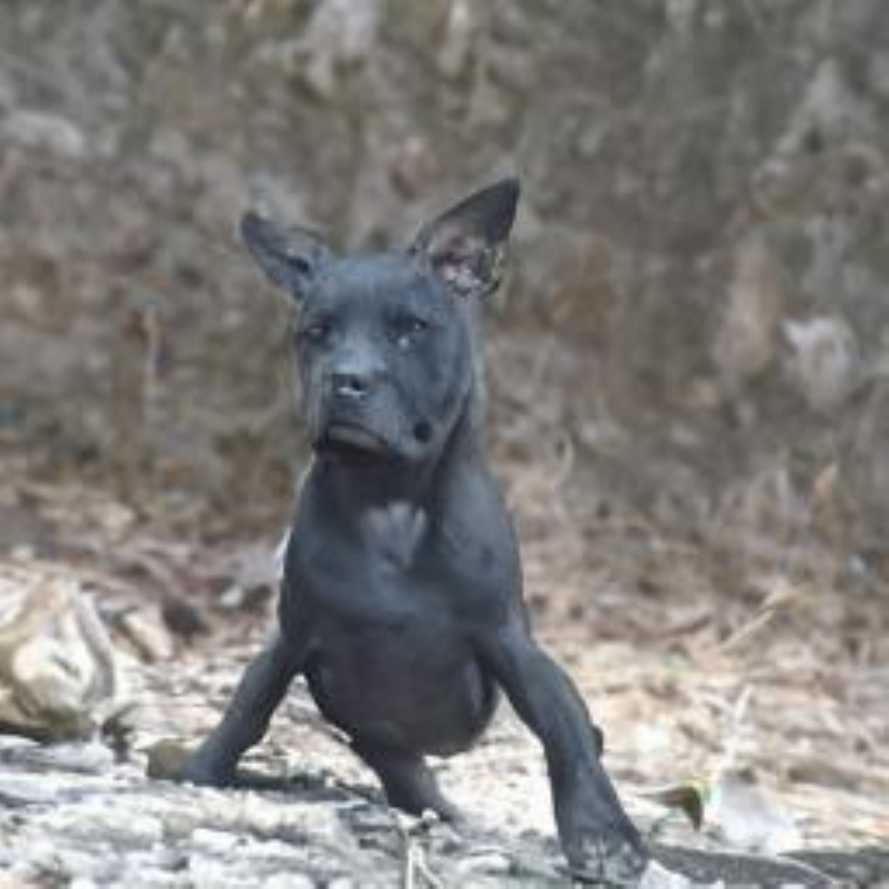} &
        \adjincludegraphics[clip,width=0.15\textwidth,trim={0 0 0 0}]{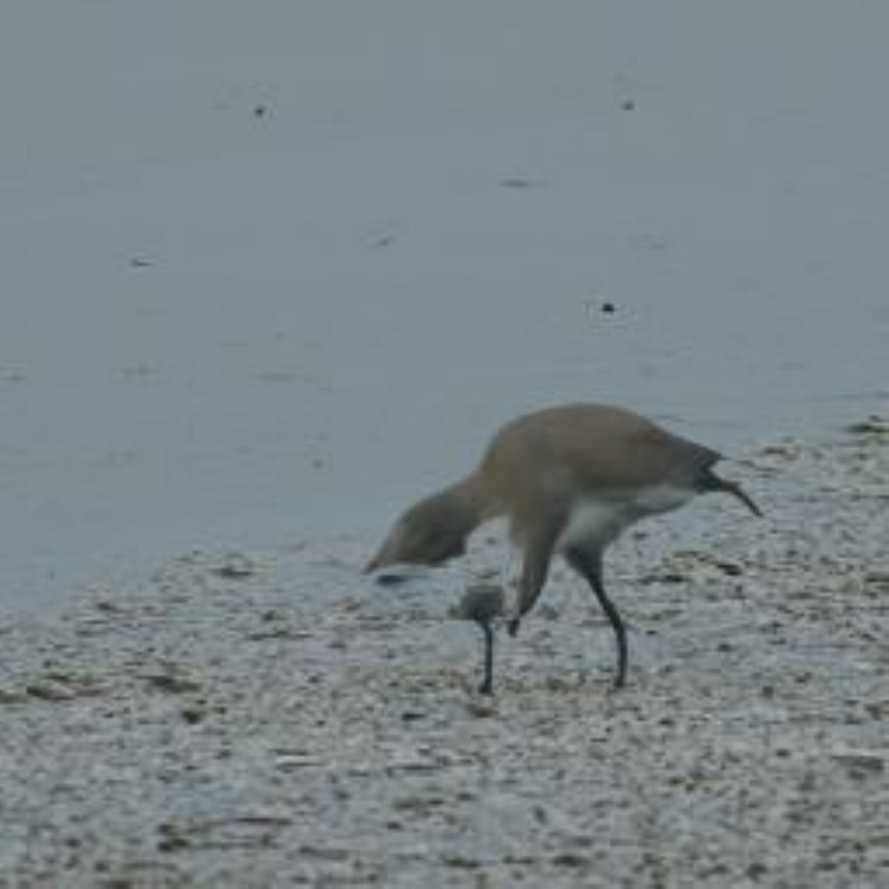} &
        \adjincludegraphics[clip,width=0.15\textwidth,trim={0 0 0 0}]{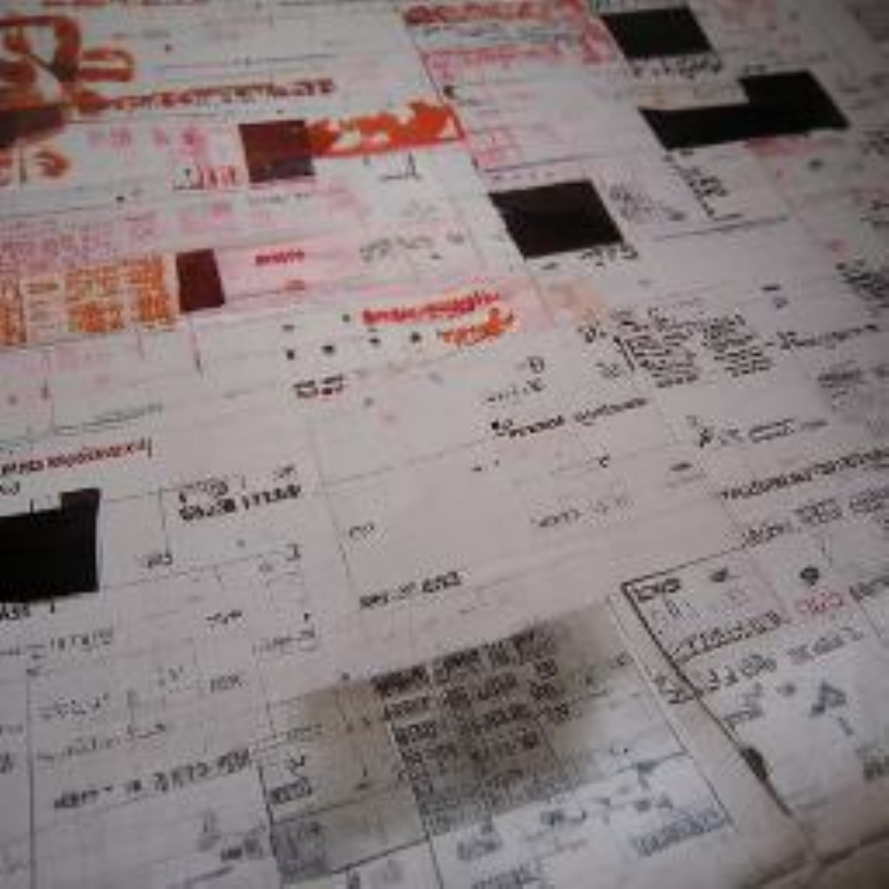} \\

            \raisebox{0.2\height}{\rotatebox{90}{\scriptsize Single Noise-aware}} 
        \adjincludegraphics[clip,width=0.15\textwidth,trim={0 0 0 0}]{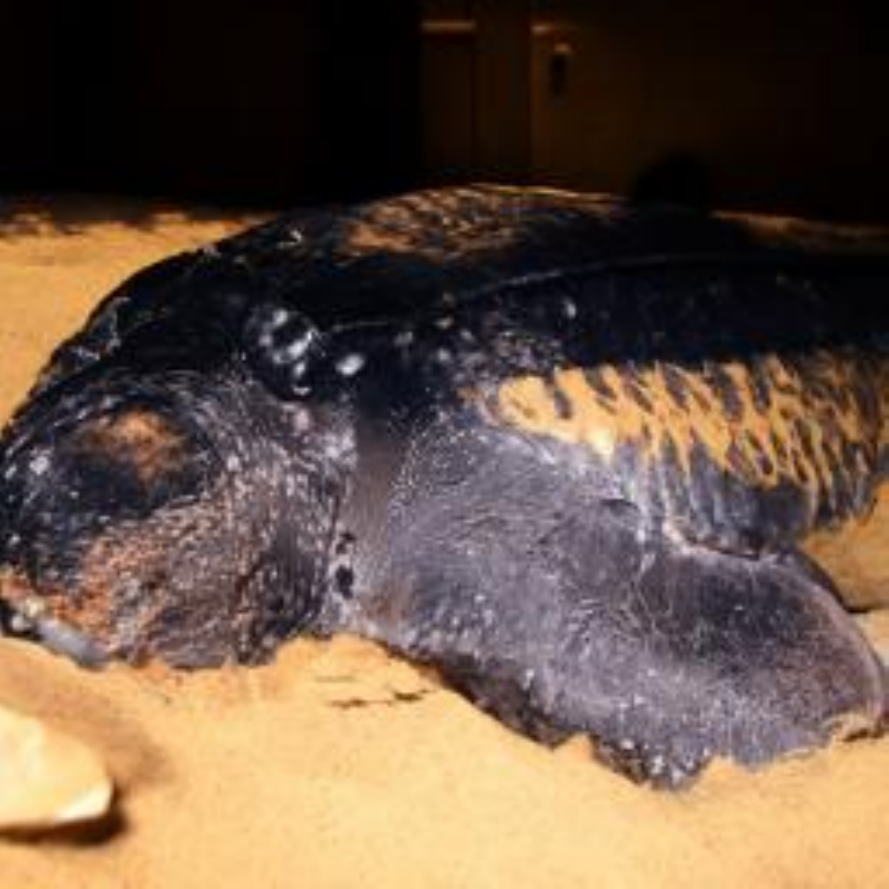} &
        \adjincludegraphics[clip,width=0.15\textwidth,trim={0 0 0 0}]{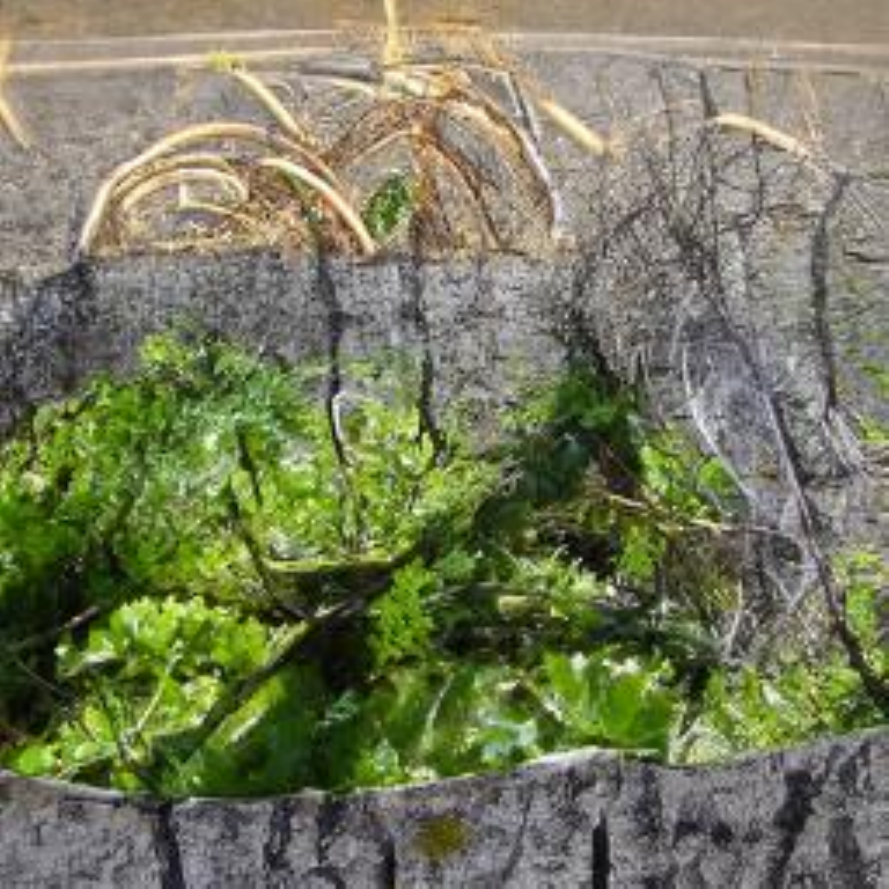} &
        \adjincludegraphics[clip,width=0.15\textwidth,trim={0 0 0 0}]{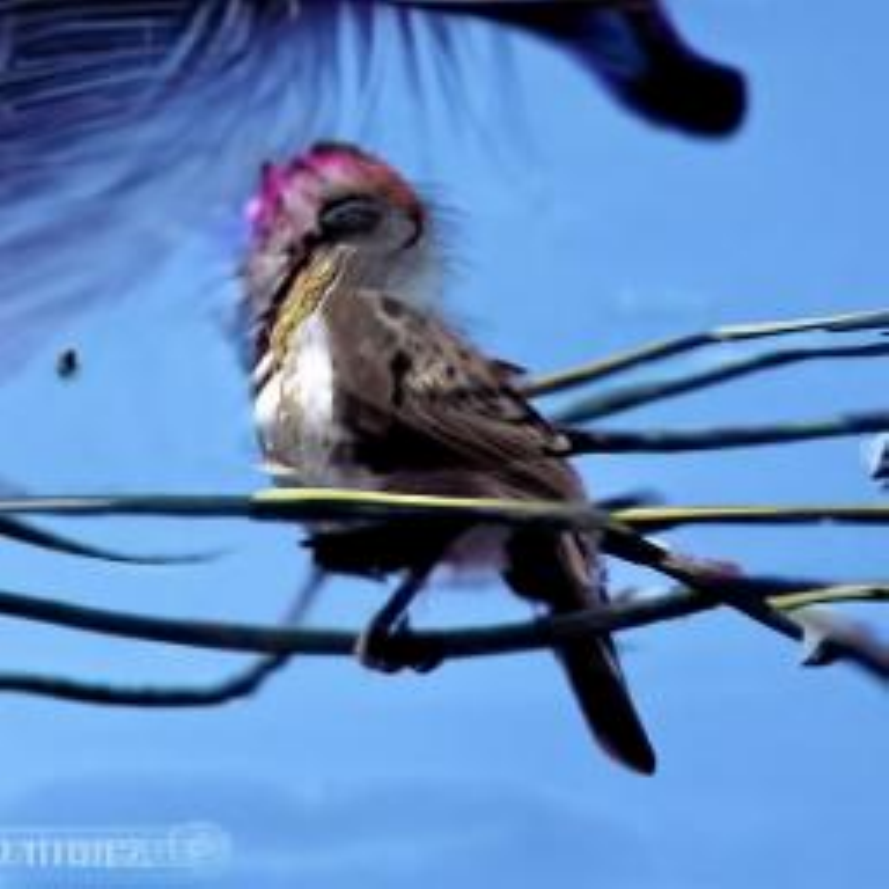} &
        \adjincludegraphics[clip,width=0.15\textwidth,trim={0 0 0 0}]{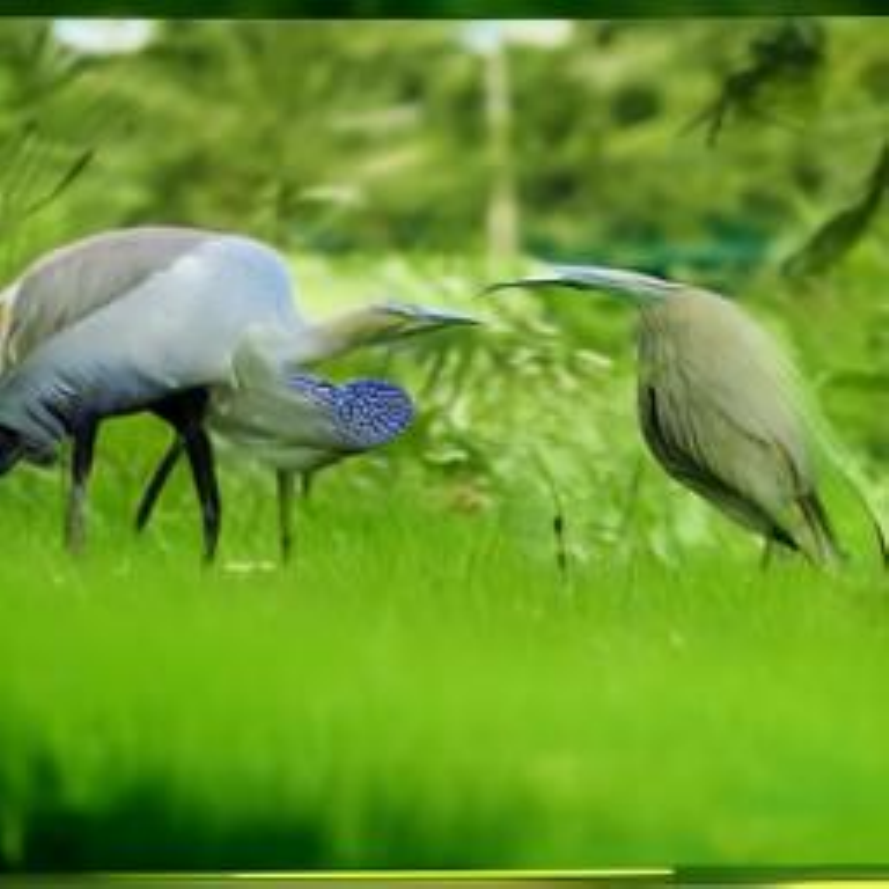} &
        \adjincludegraphics[clip,width=0.15\textwidth,trim={0 0 0 0}]{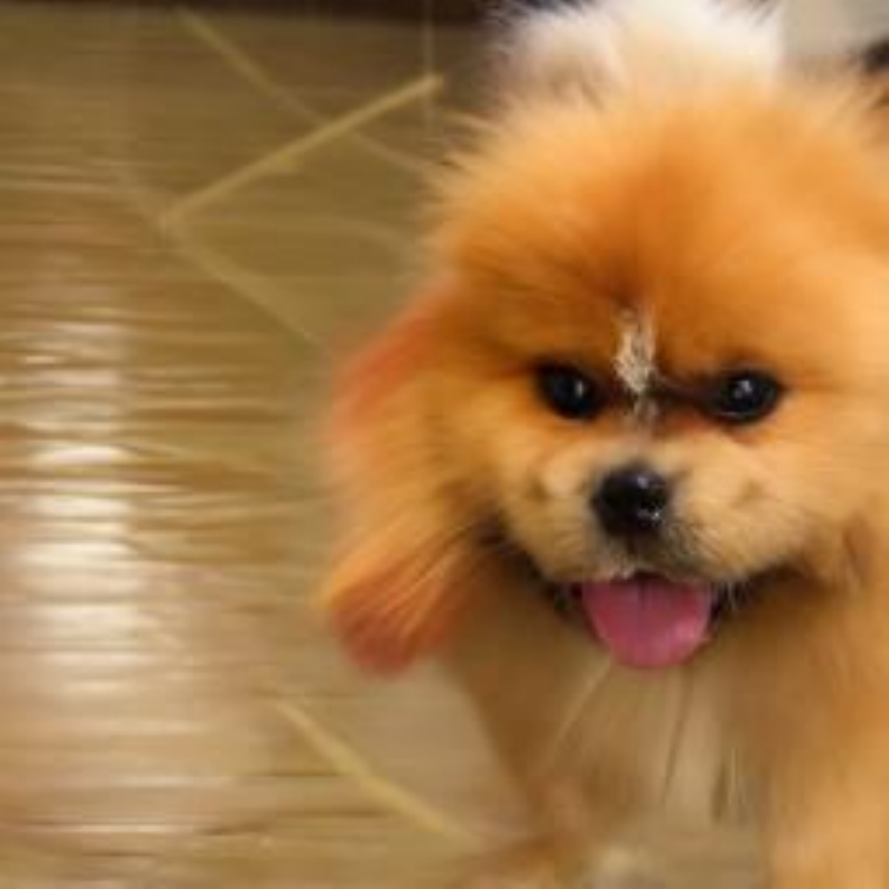} &
        \adjincludegraphics[clip,width=0.15\textwidth,trim={0 0 0 0}]{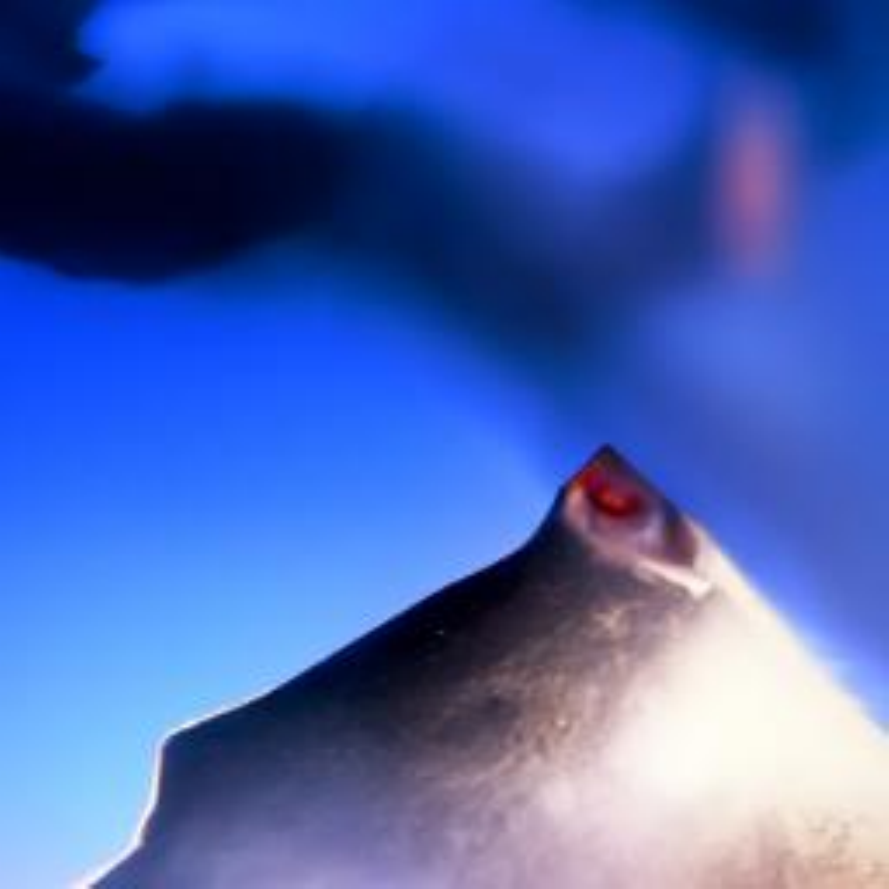} \\

        \raisebox{0\height}{\rotatebox{90}{\scriptsize Supervised Multi-Experts}} 
        \adjincludegraphics[clip,width=0.15\textwidth,trim={0 0 0 0}]{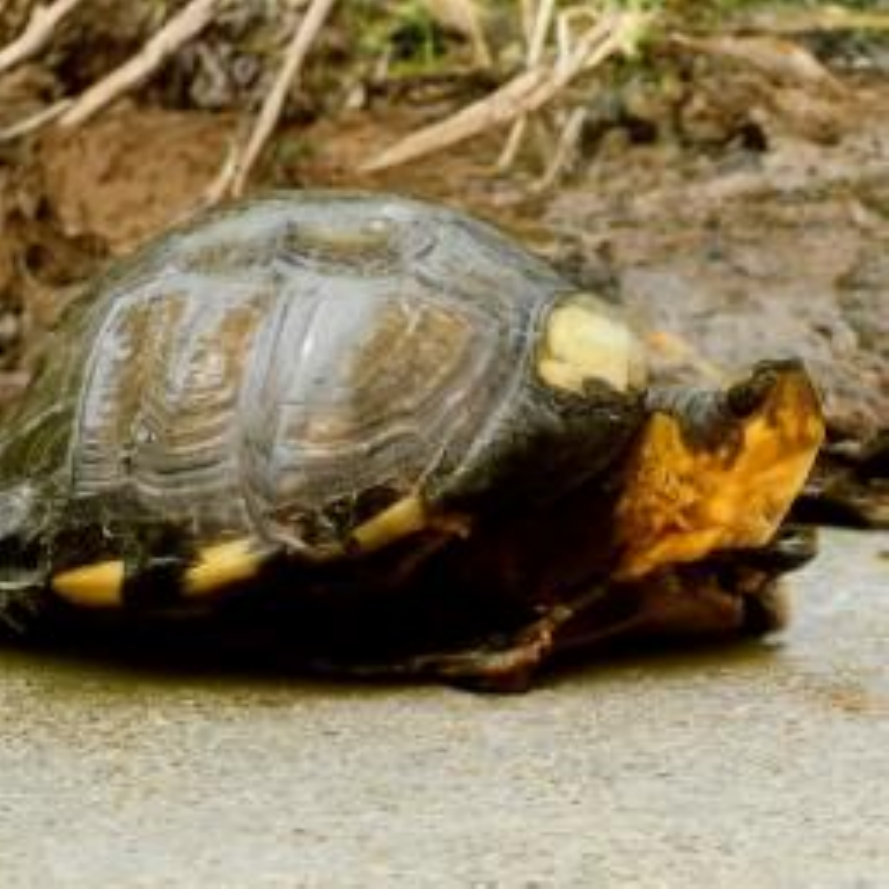} &
        \adjincludegraphics[clip,width=0.15\textwidth,trim={0 0 0 0}]{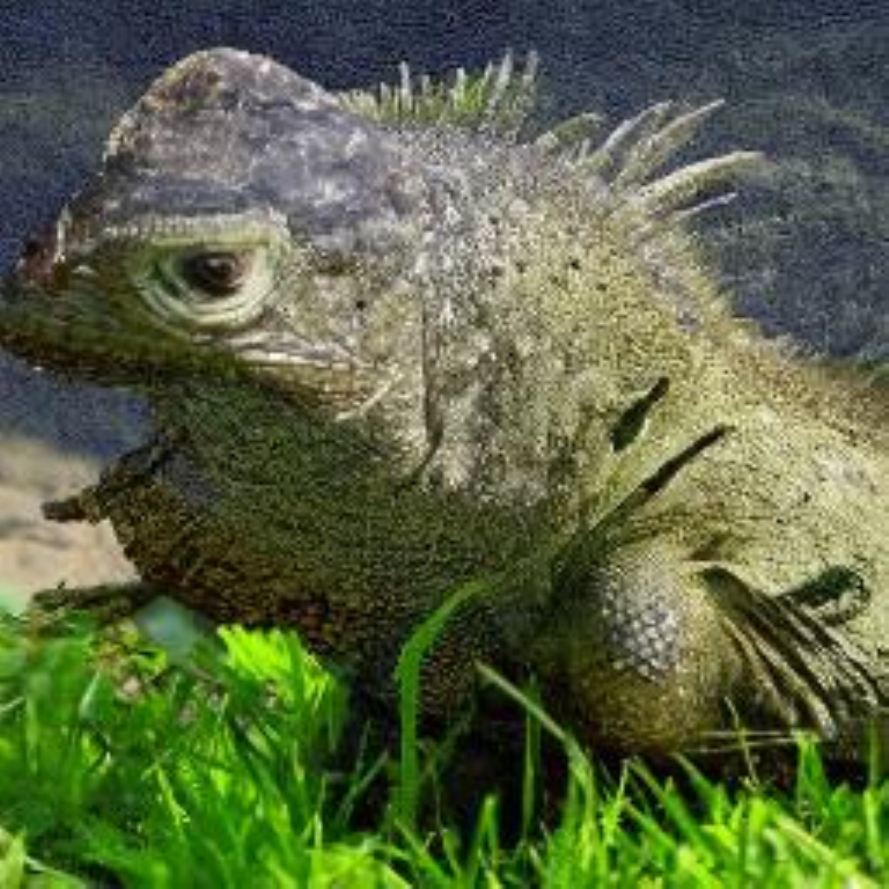} &
        \adjincludegraphics[clip,width=0.15\textwidth,trim={0 0 0 0}]{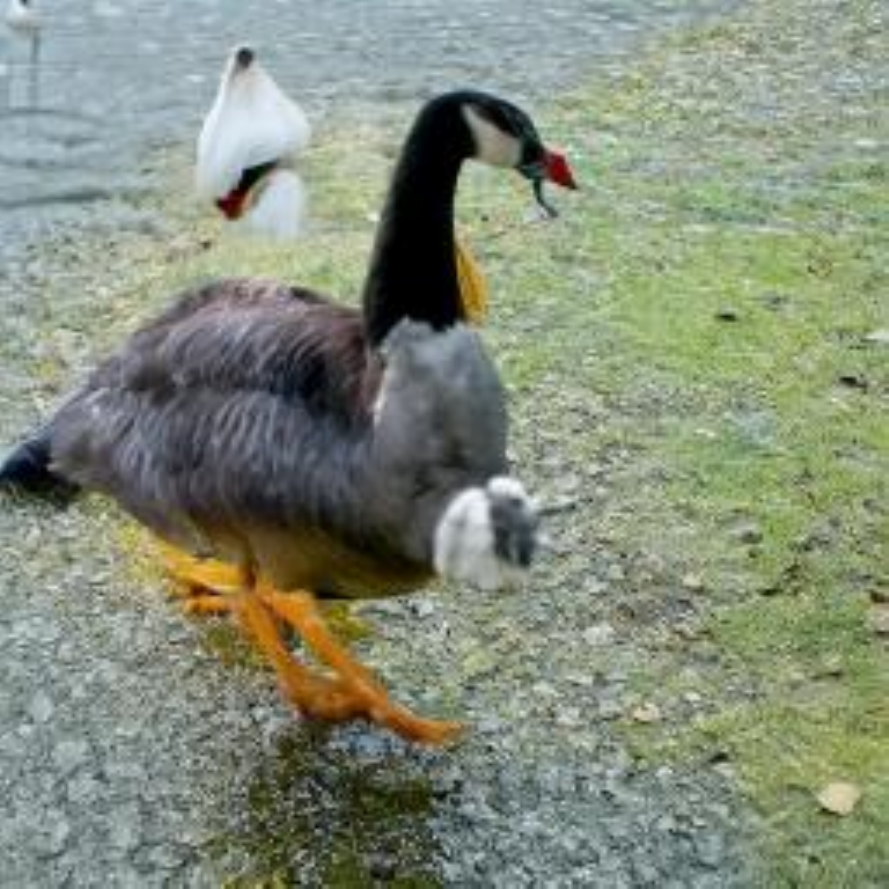} &
        \adjincludegraphics[clip,width=0.15\textwidth,trim={0 0 0 0}]{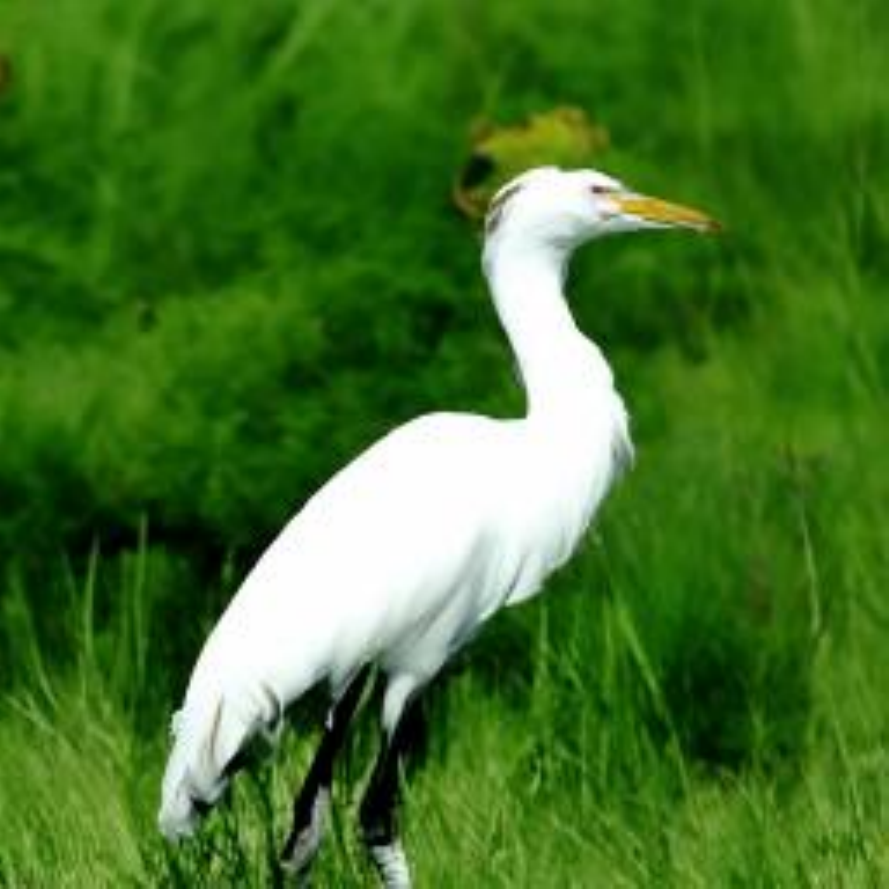} &
        \adjincludegraphics[clip,width=0.15\textwidth,trim={0 0 0 0}]{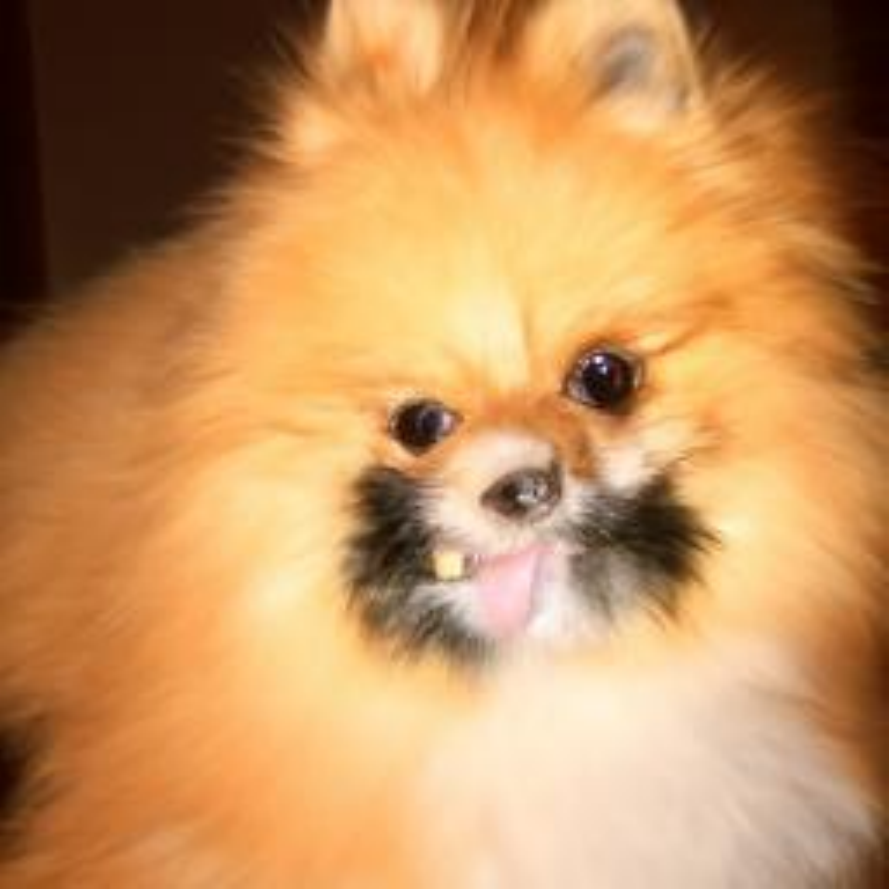} &
        \adjincludegraphics[clip,width=0.15\textwidth,trim={0 0 0 0}]{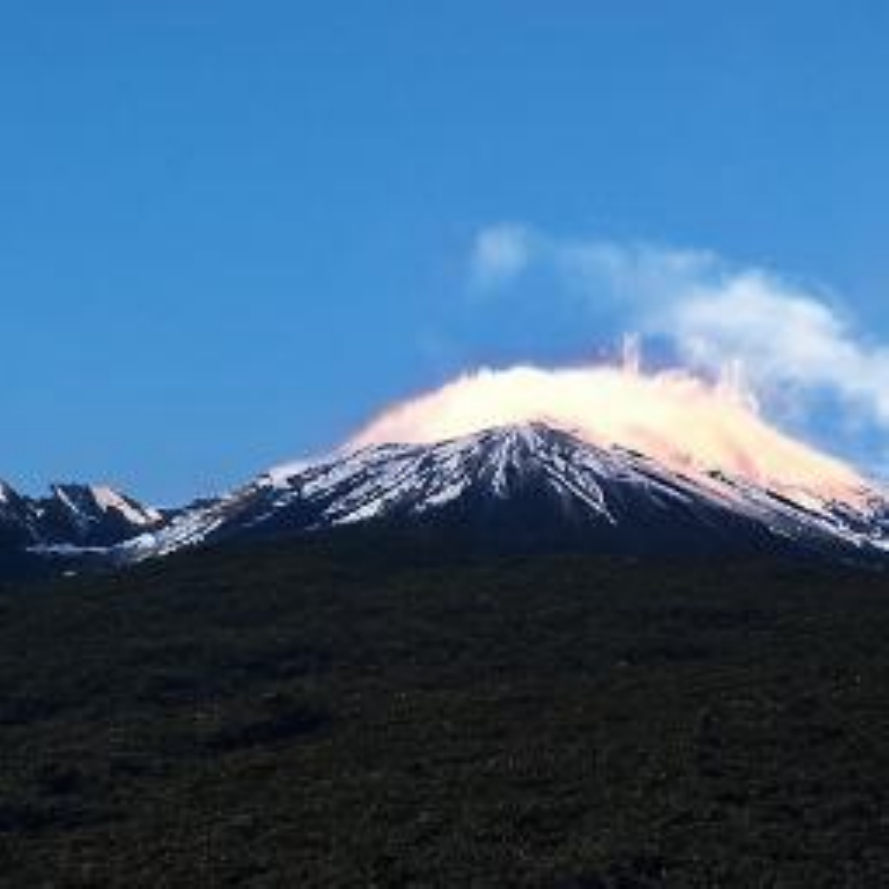} \\

        \raisebox{0.35\height}{\rotatebox{90}{\scriptsize PPAP ($N=5$)}} 
        \adjincludegraphics[clip,width=0.15\textwidth,trim={0 0 0 0}]{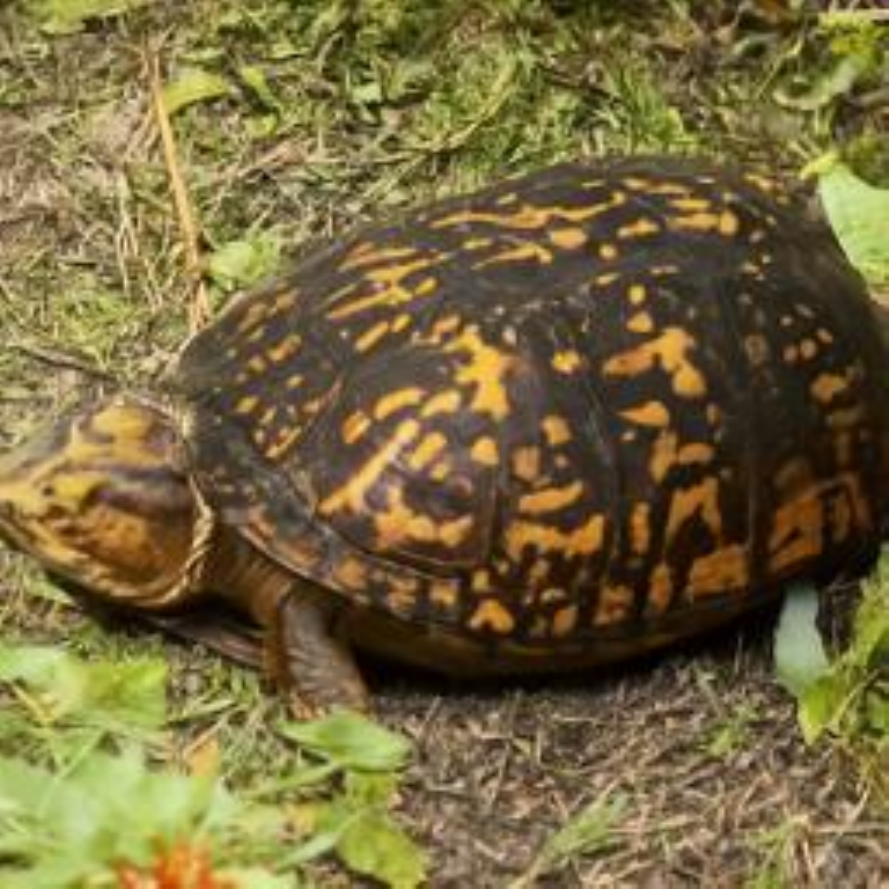} &
        \adjincludegraphics[clip,width=0.15\textwidth,trim={0 0 0 0}]{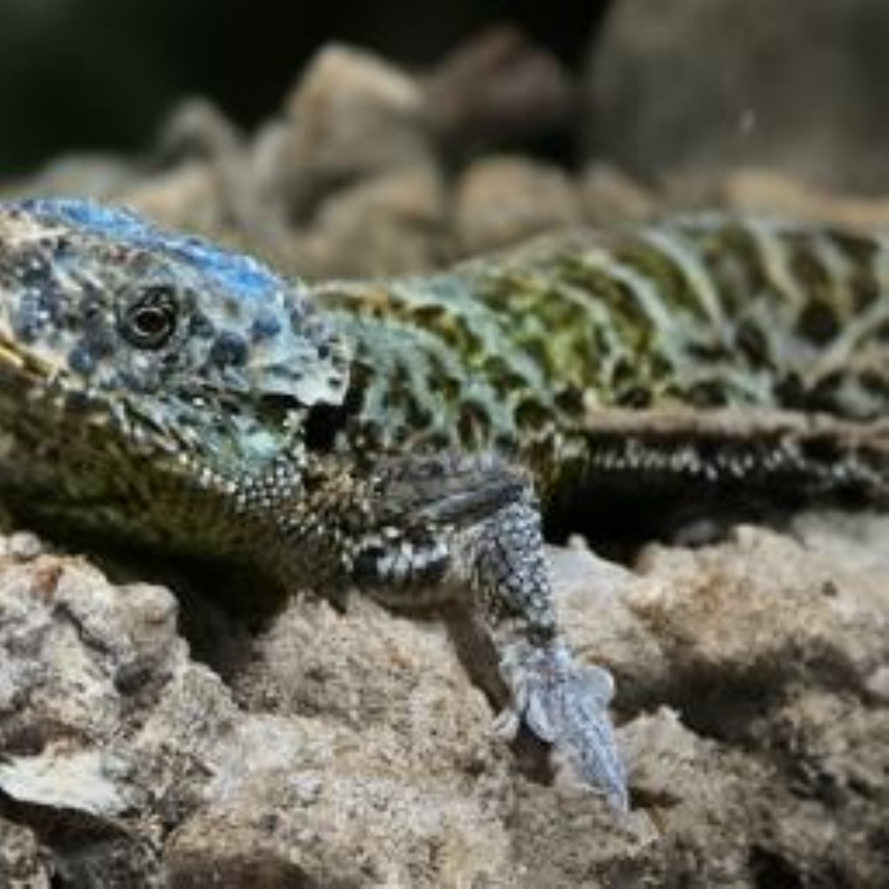} &
        \adjincludegraphics[clip,width=0.15\textwidth,trim={0 0 0 0}]{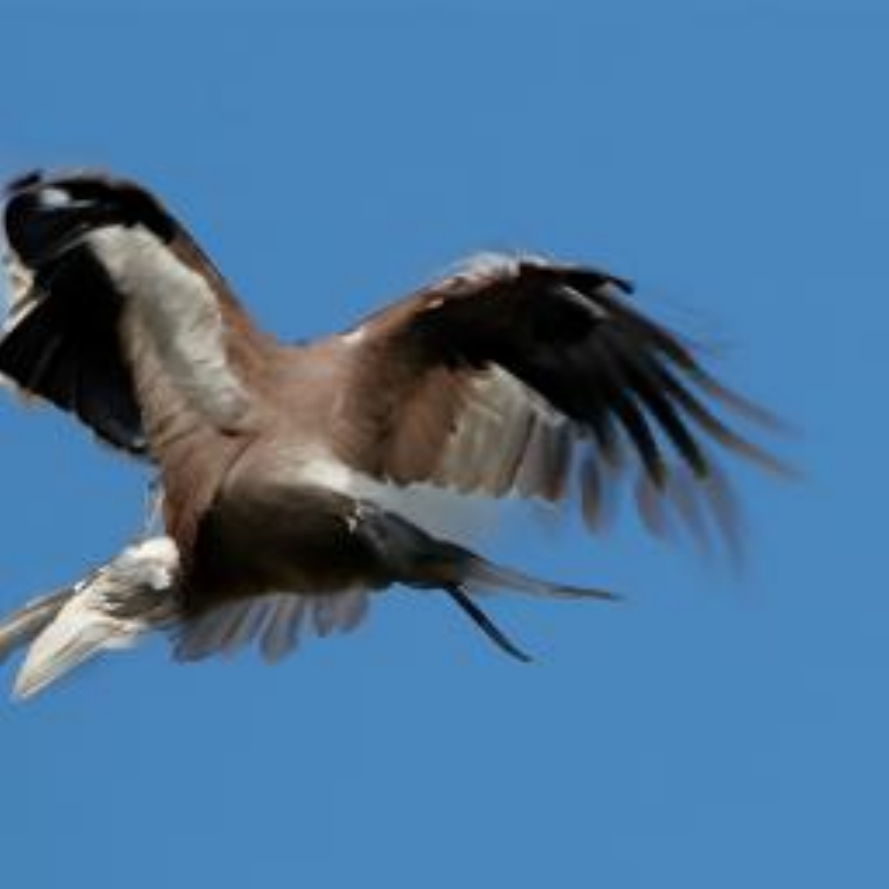} &
        \adjincludegraphics[clip,width=0.15\textwidth,trim={0 0 0 0}]{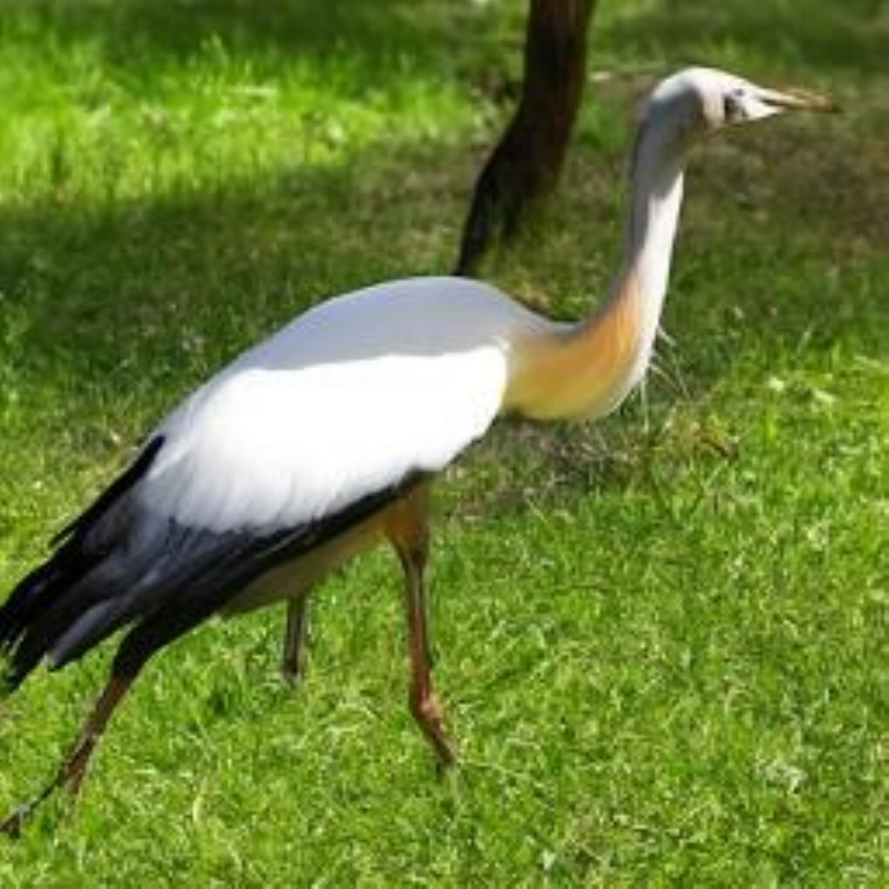} &
        \adjincludegraphics[clip,width=0.15\textwidth,trim={0 0 0 0}]{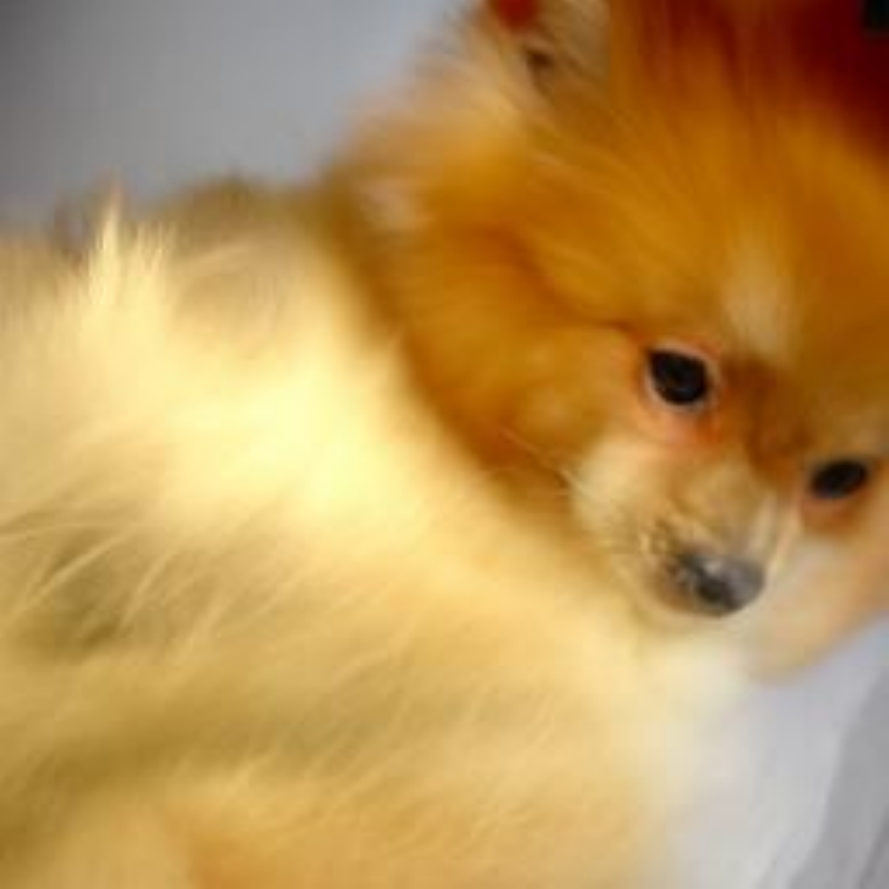} &
        \adjincludegraphics[clip,width=0.15\textwidth,trim={0 0 0 0}]{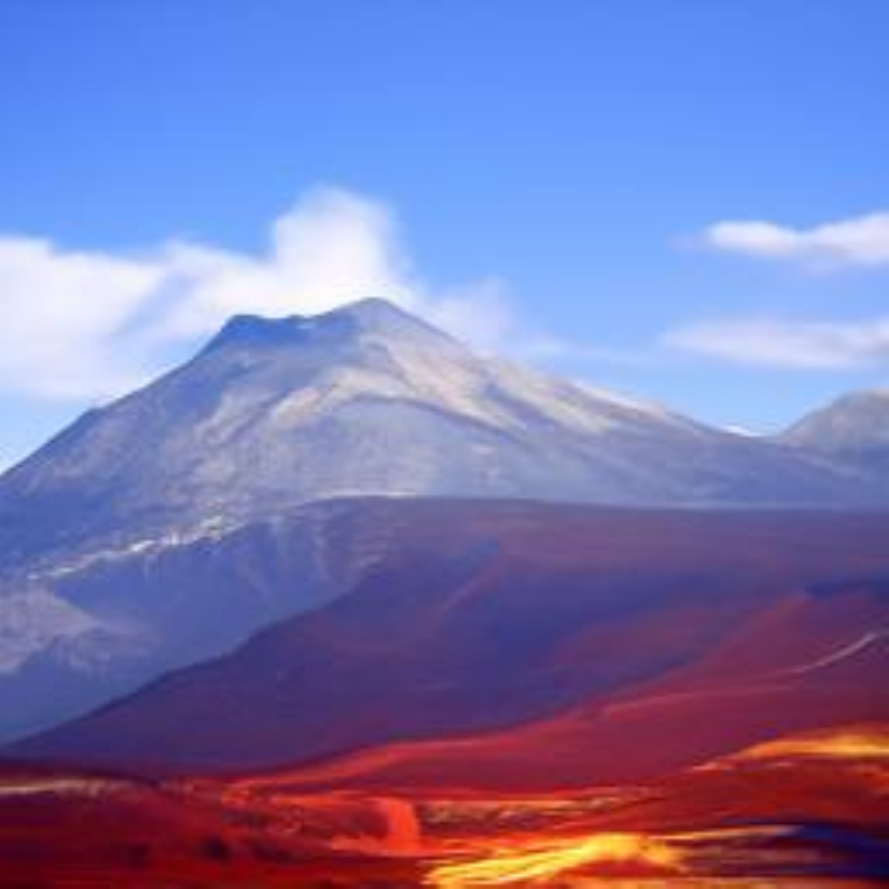} \\

        \raisebox{0.3\height}{\rotatebox{90}{\scriptsize PPAP ($N=10$)}} 
        \adjincludegraphics[clip,width=0.15\textwidth,trim={0 0 0 0}]{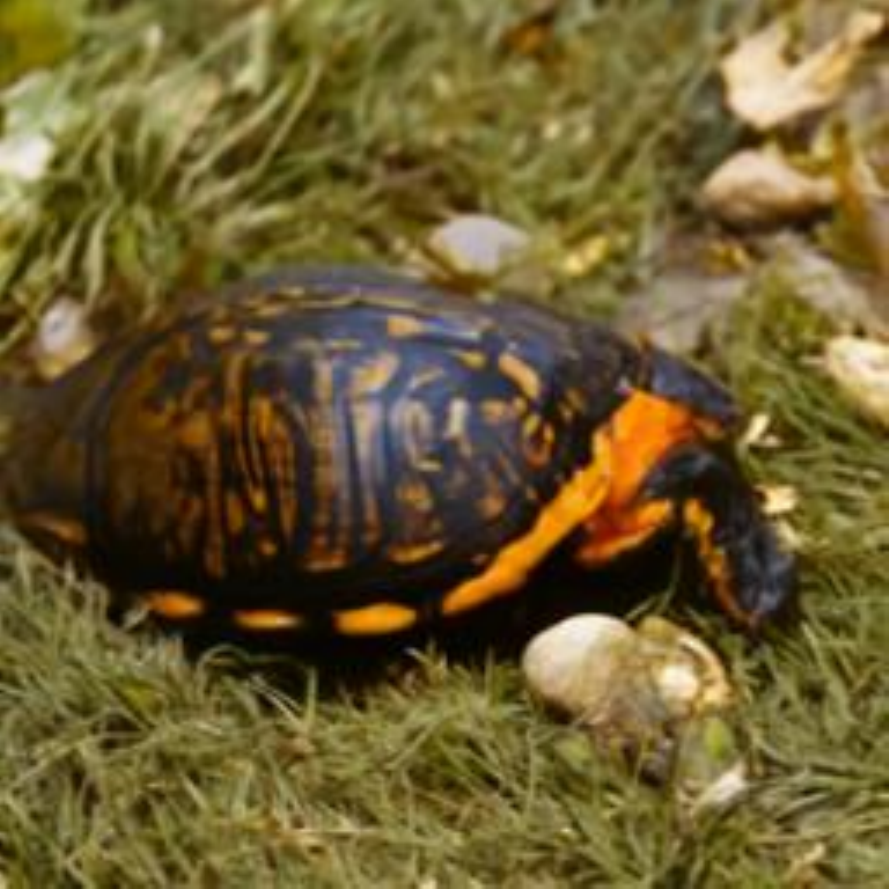} &
        \adjincludegraphics[clip,width=0.15\textwidth,trim={0 0 0 0}]{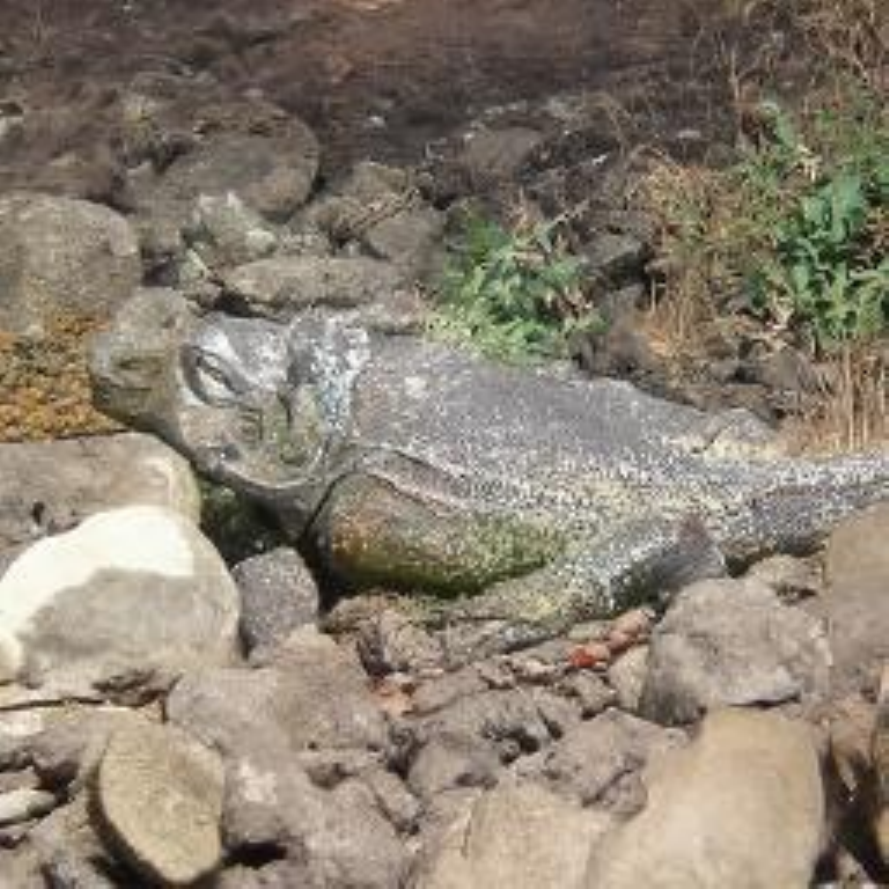} &
        \adjincludegraphics[clip,width=0.15\textwidth,trim={0 0 0 0}]{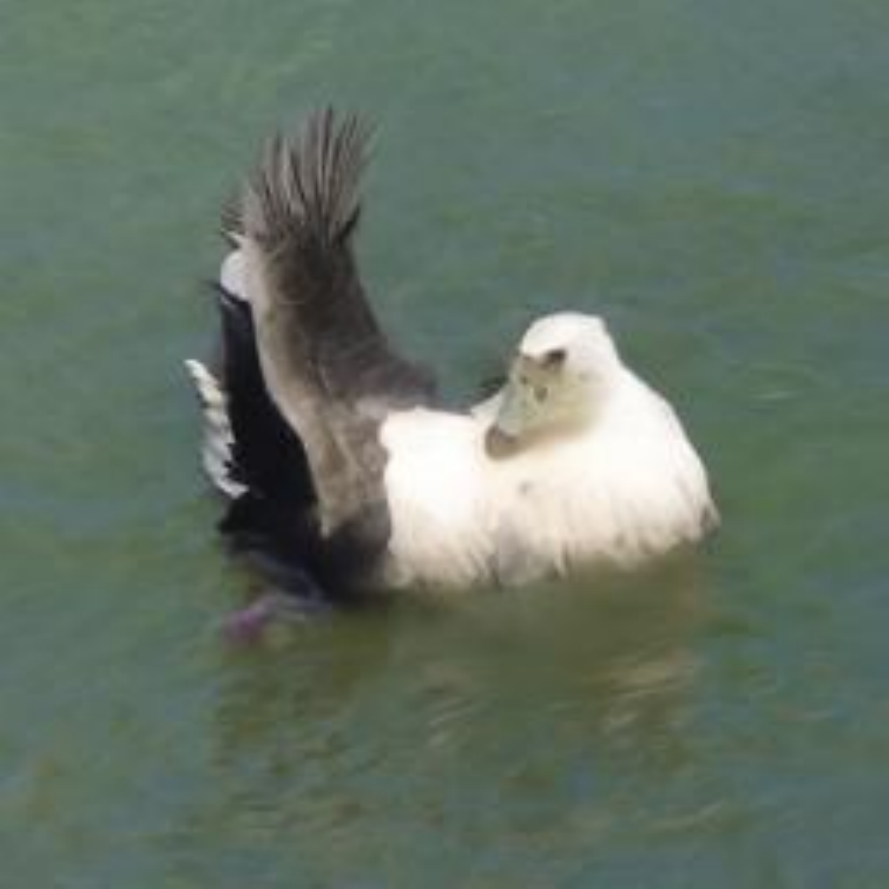} &
        \adjincludegraphics[clip,width=0.15\textwidth,trim={0 0 0 0}]{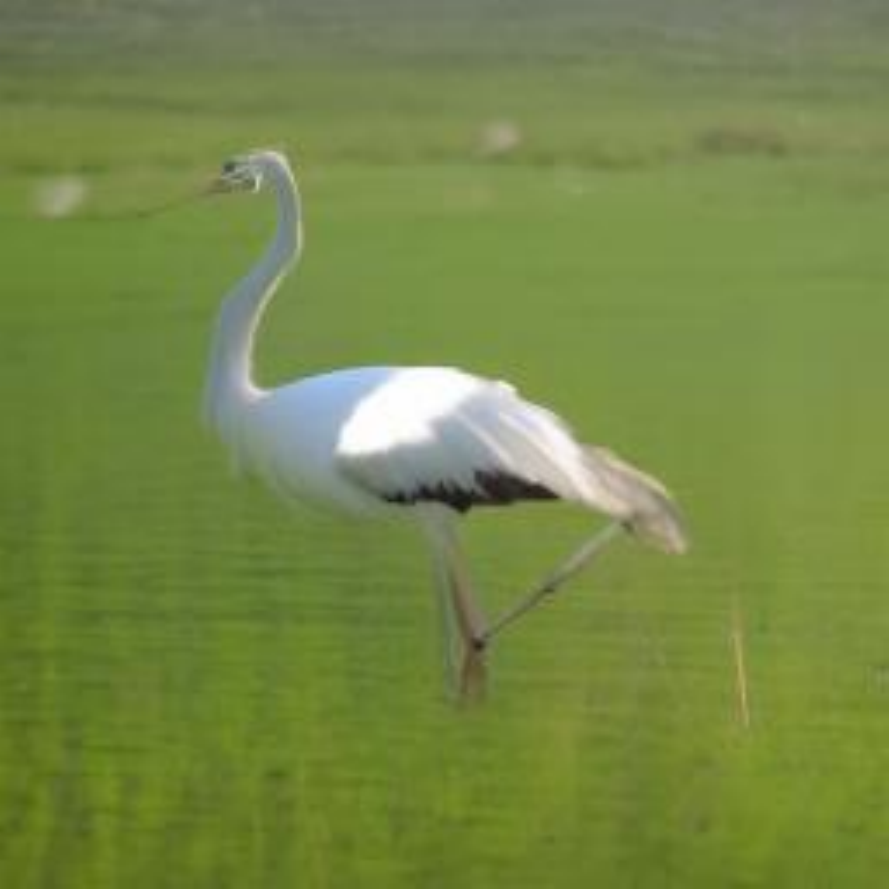} &
        \adjincludegraphics[clip,width=0.15\textwidth,trim={0 0 0 0}]{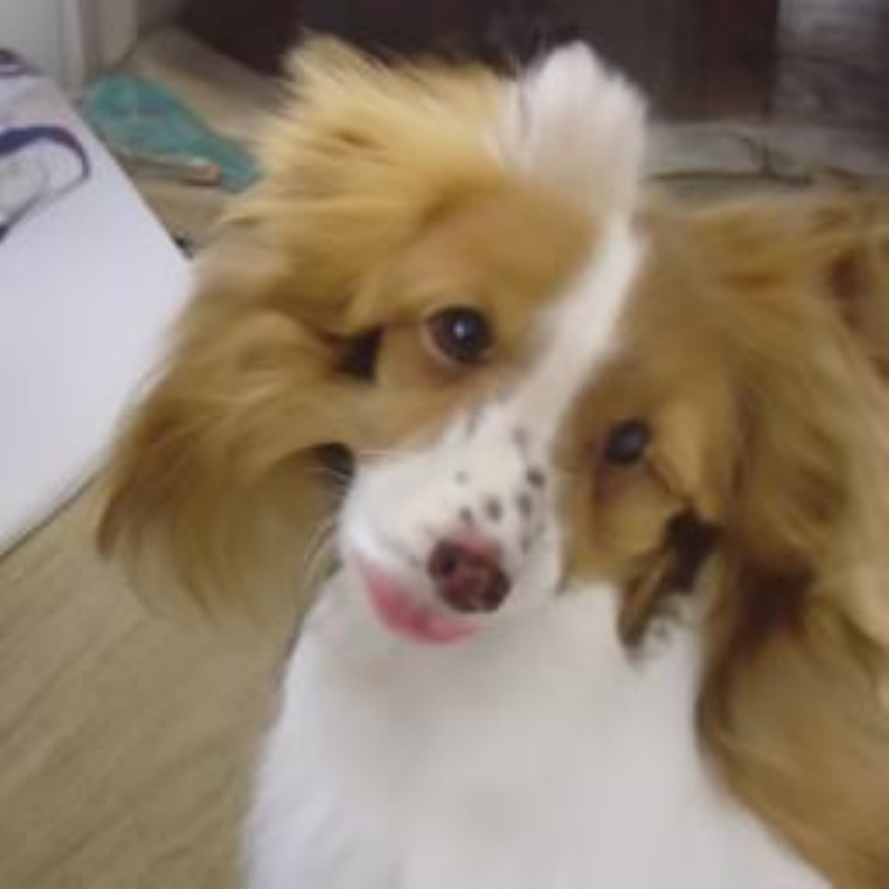} &
        \adjincludegraphics[clip,width=0.15\textwidth,trim={0 0 0 0}]{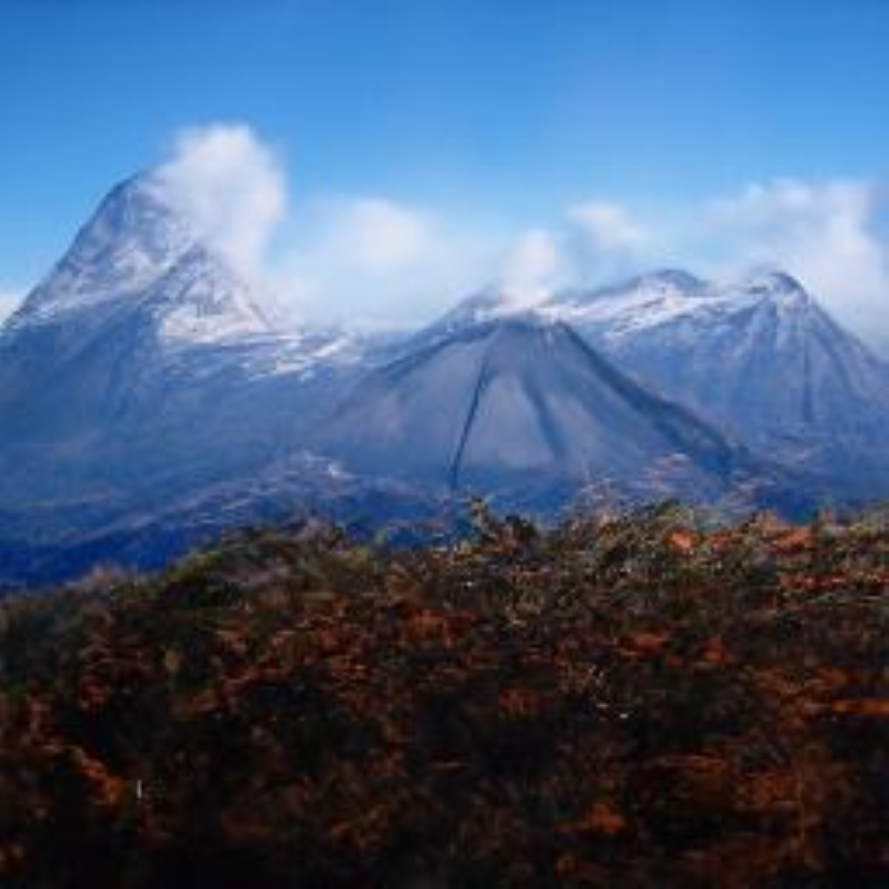} \\
        
        \footnotesize ~~~~~~Box turtle & \footnotesize  Iguana & \footnotesize  Goose & \footnotesize  Crane & \footnotesize Pomeranian & \footnotesize Volcano \\
    \end{tabular}
    \caption{
    Qualitative results on ImageNet class conditional generation with DDIM 25 steps by guiding unconditional ADM with ResNet50.
    }
    \label{apx:fig_qualitative_resnet_ddim}

\end{figure*}

Due to limited spaces, we only show qualitative results with DDPM 250 steps in Section~\textcolor{red}{5.1}.
We illustrate more qualitative results with DDPM 250 steps and DDIM 25 steps in Fig.~\ref{apx:fig_qualitative_resnet_ddpm} and Fig.~\ref{apx:fig_qualitative_resnet_ddim}, respectively.

\subsection{Guided ADM with DeiT-S}
\label{apx:subsec:adm+deit}

\begin{figure*}[t]

    \centering
    \begin{tabular}{@{\hspace{2mm}}c@{\hspace{2mm}}c@{\hspace{2mm}}c@{\hspace{2mm}}c@{\hspace{2mm}}c@{\hspace{2mm}}c@{\hspace{2mm}}c@{\hspace{2mm}}c}
    
        \raisebox{0.2\height}{\rotatebox{90}{\scriptsize Na\"ive Off-the-shelf}} 
        \adjincludegraphics[clip,width=0.15\textwidth,trim={0 0 0 0}]{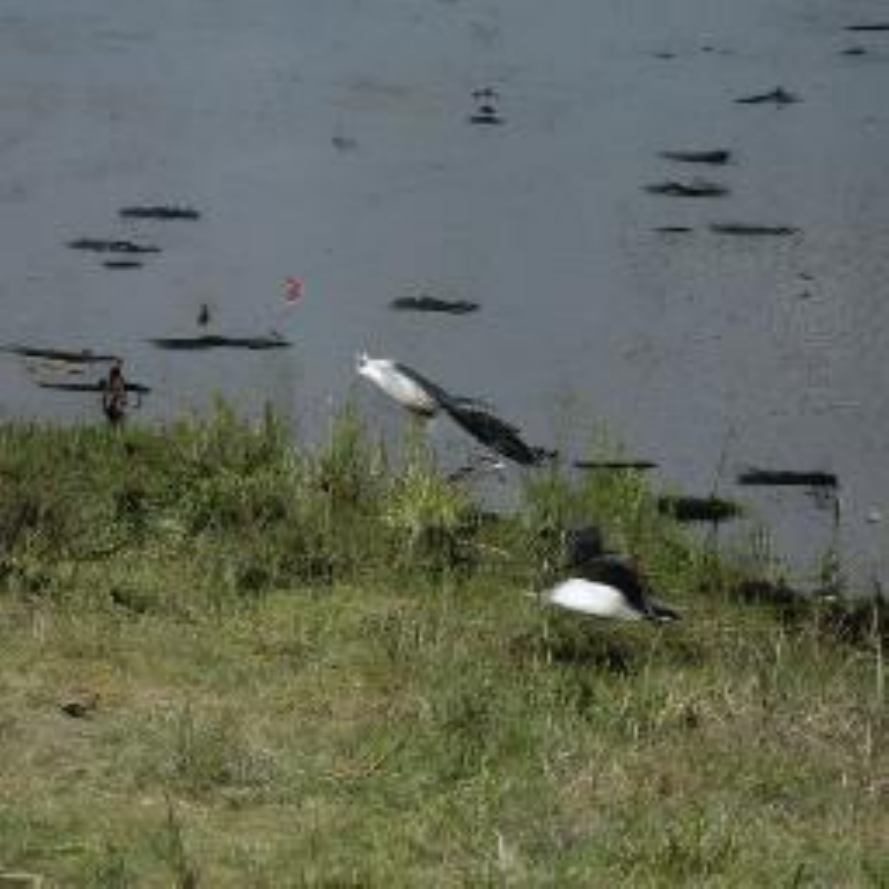} &
        \adjincludegraphics[clip,width=0.15\textwidth,trim={0 0 0 0}]{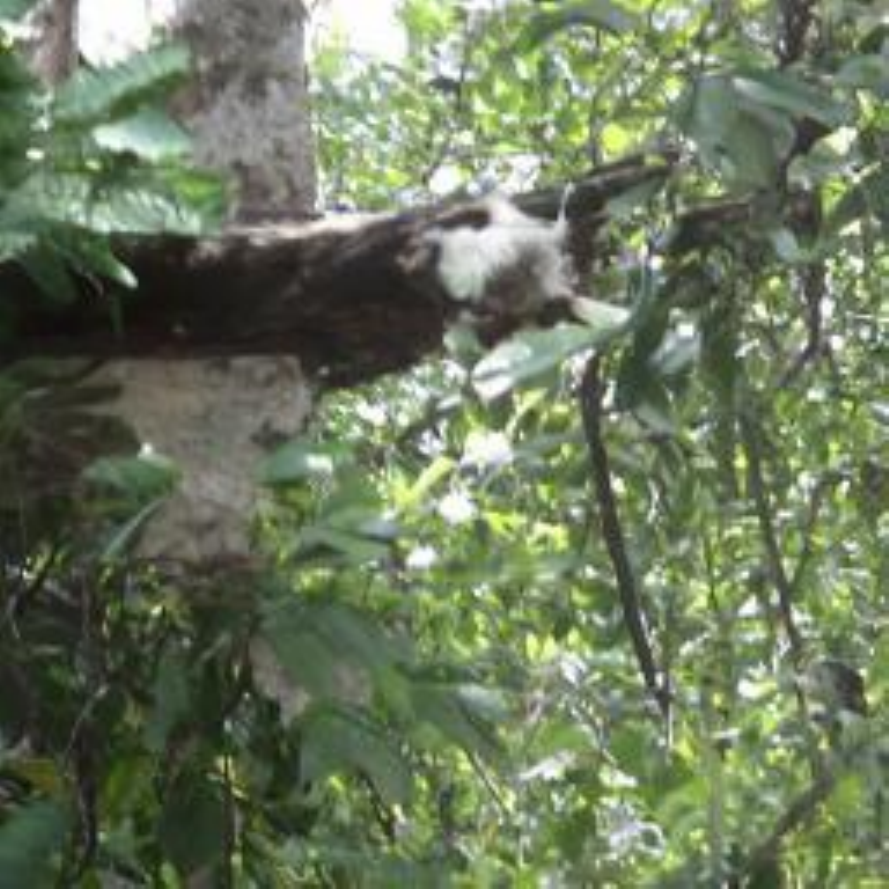} &
        \adjincludegraphics[clip,width=0.15\textwidth,trim={0 0 0 0}]{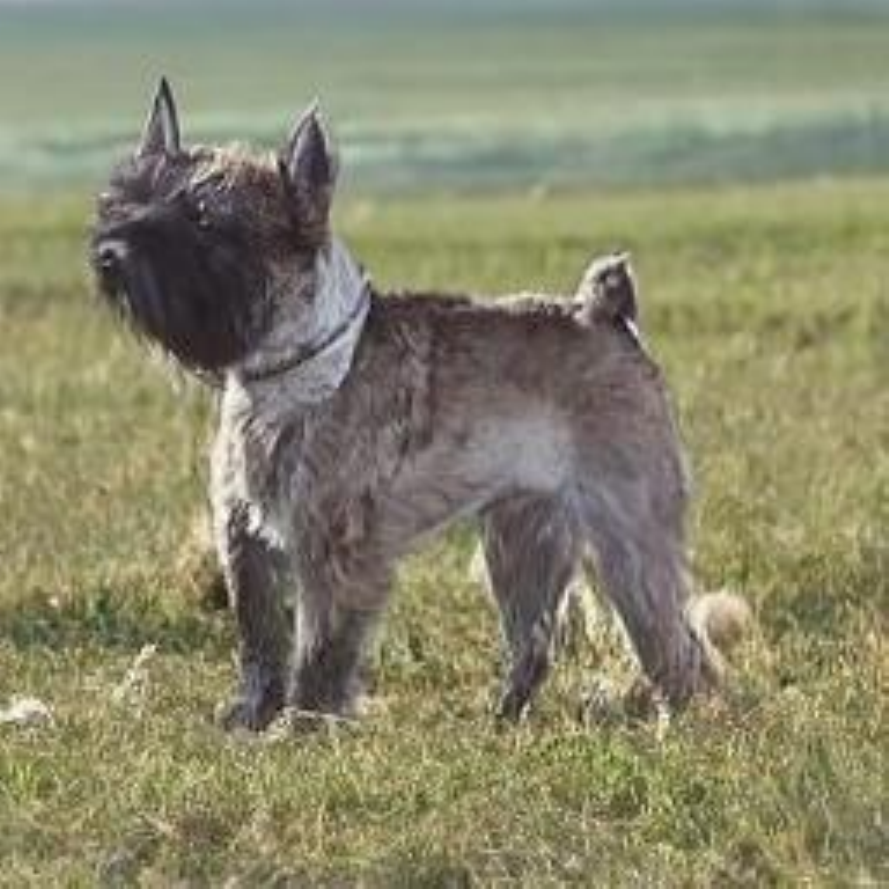} &
        \adjincludegraphics[clip,width=0.15\textwidth,trim={0 0 0 0}]{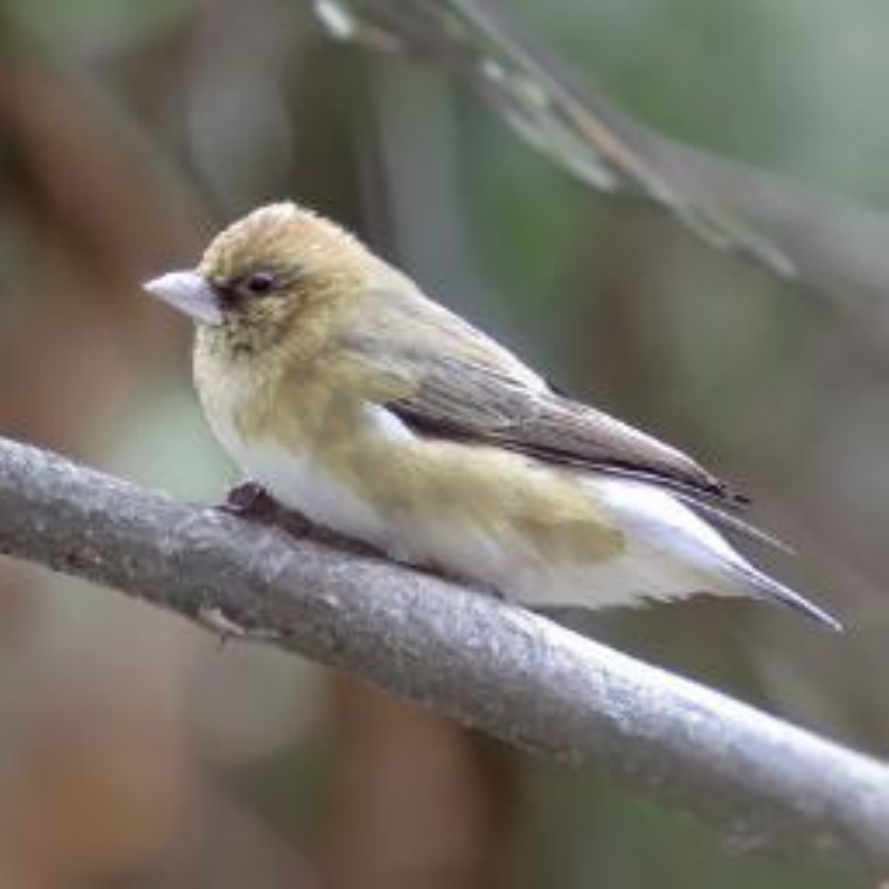} &
        \adjincludegraphics[clip,width=0.15\textwidth,trim={0 0 0 0}]{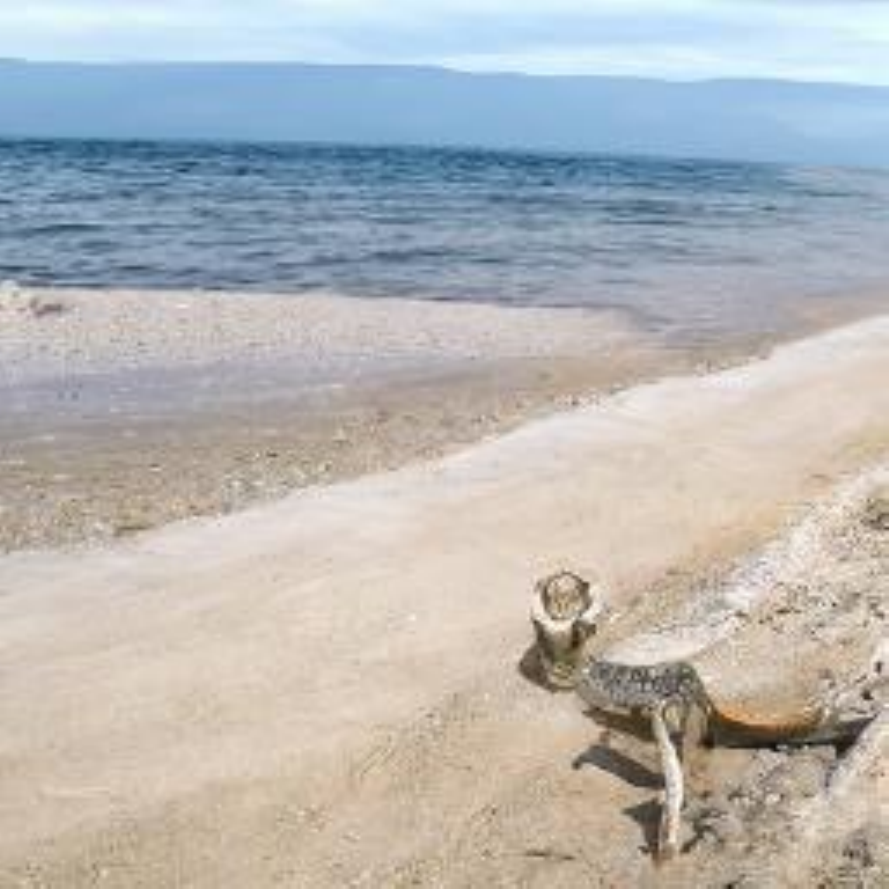} &
        \adjincludegraphics[clip,width=0.15\textwidth,trim={0 0 0 0}]{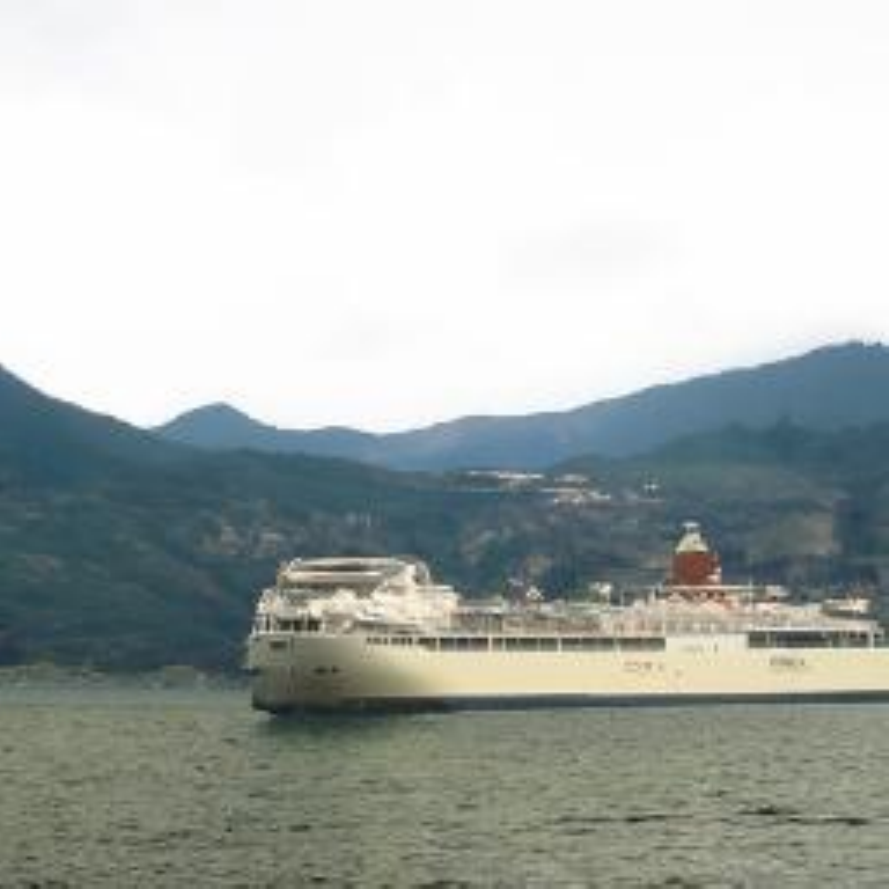} \\

            \raisebox{0.2\height}{\rotatebox{90}{\scriptsize Single Noise-aware}} 
        \adjincludegraphics[clip,width=0.15\textwidth,trim={0 0 0 0}]{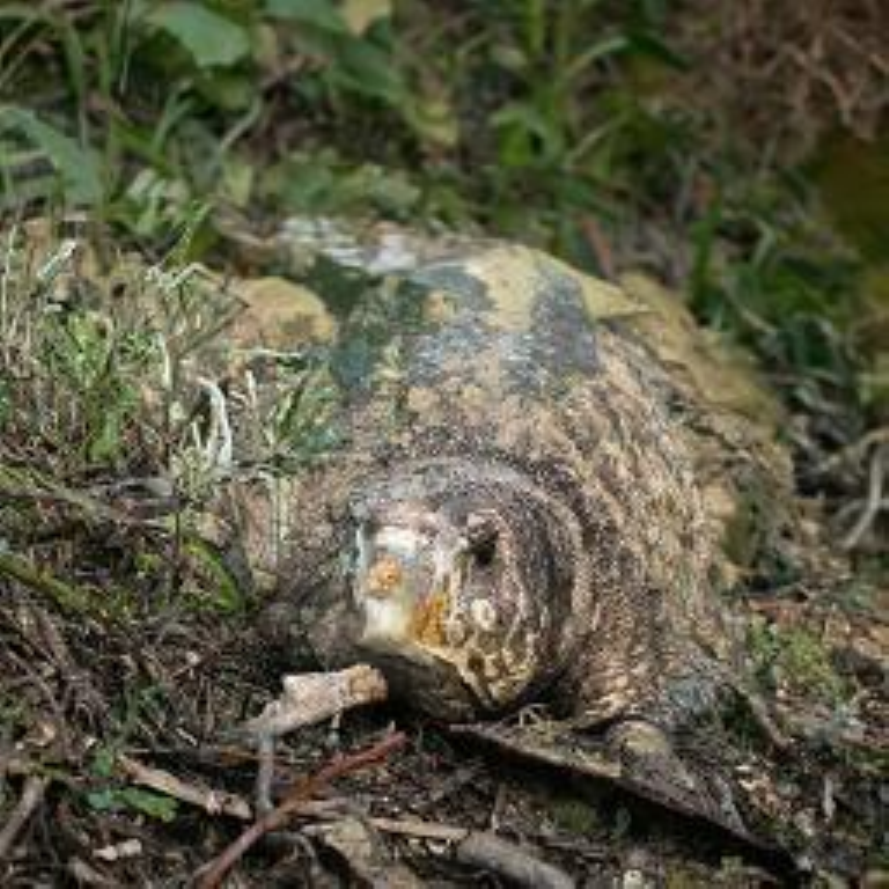} &
        \adjincludegraphics[clip,width=0.15\textwidth,trim={0 0 0 0}]{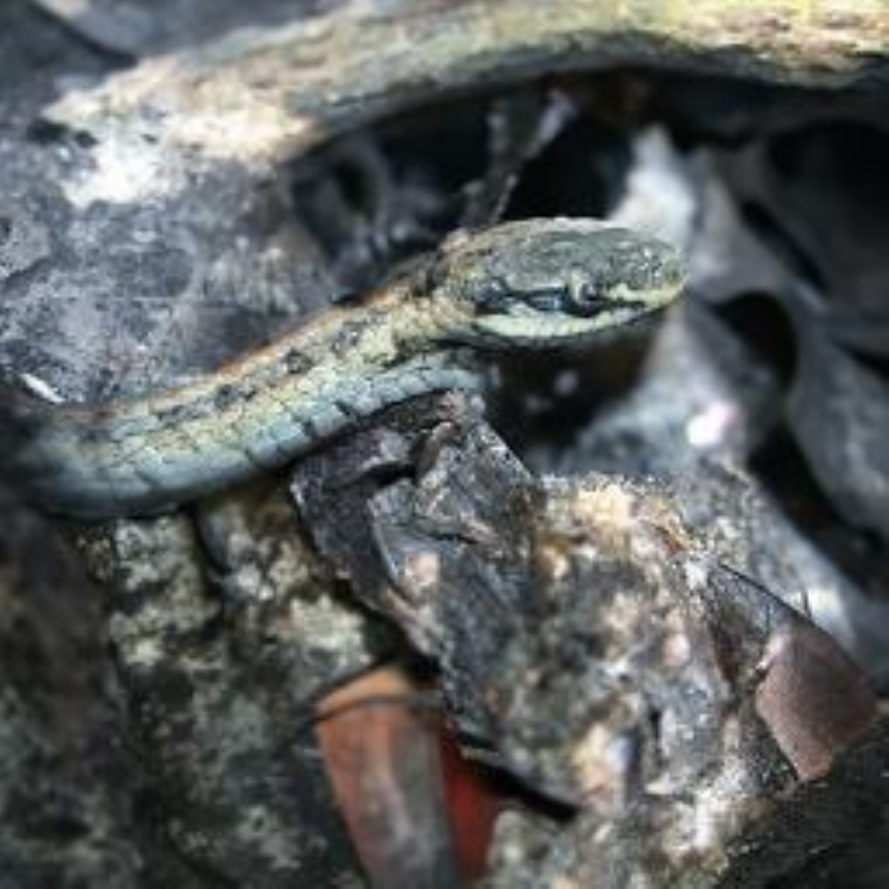} &
        \adjincludegraphics[clip,width=0.15\textwidth,trim={0 0 0 0}]{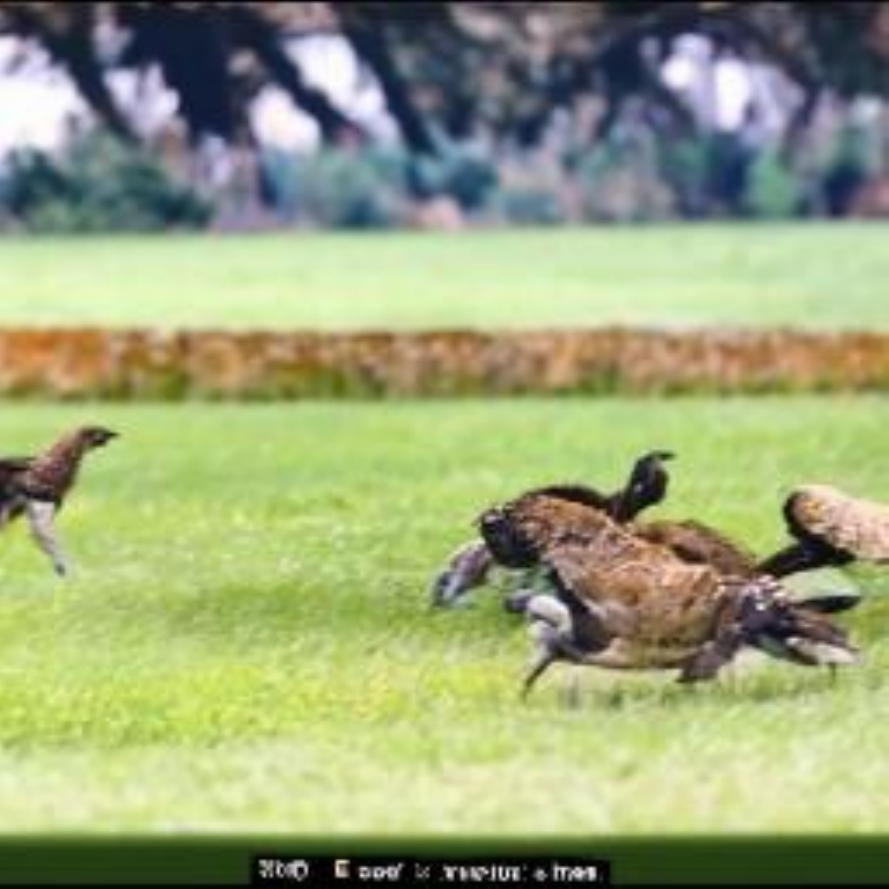} &
        \adjincludegraphics[clip,width=0.15\textwidth,trim={0 0 0 0}]{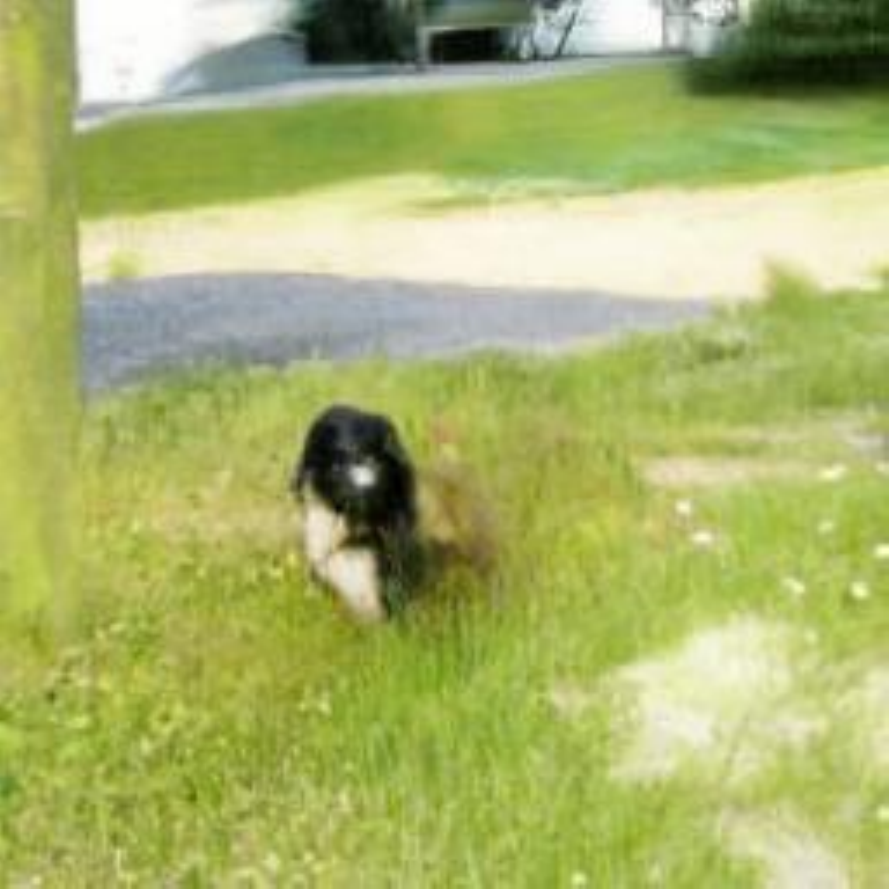} &
        \adjincludegraphics[clip,width=0.15\textwidth,trim={0 0 0 0}]{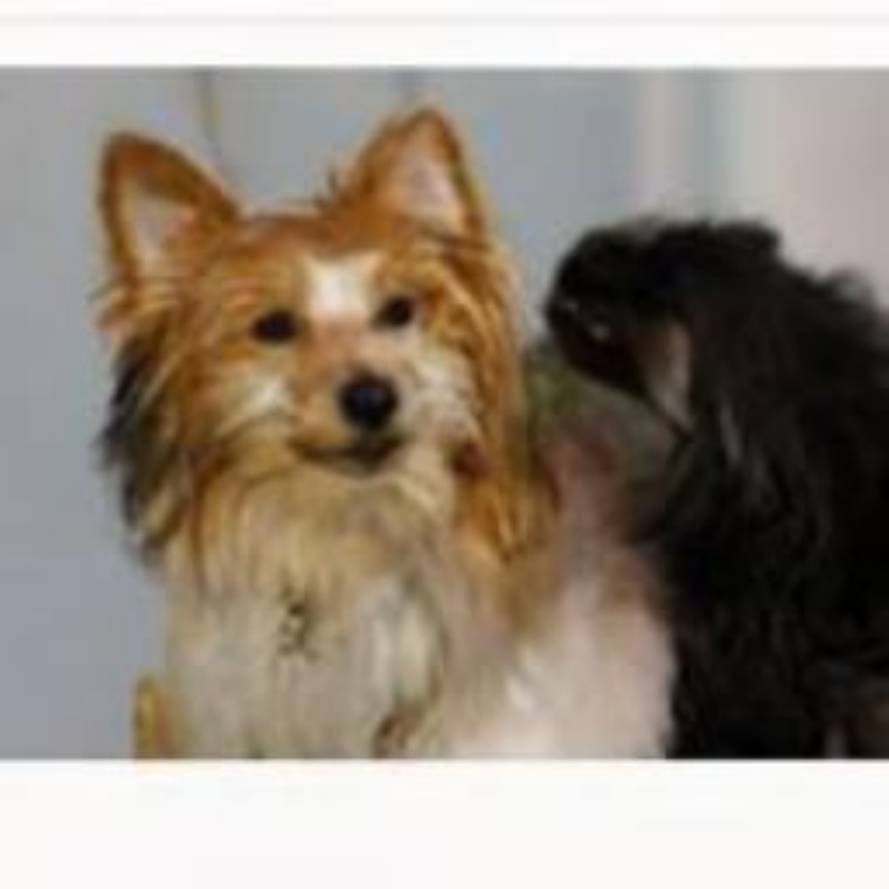} &
        \adjincludegraphics[clip,width=0.15\textwidth,trim={0 0 0 0}]{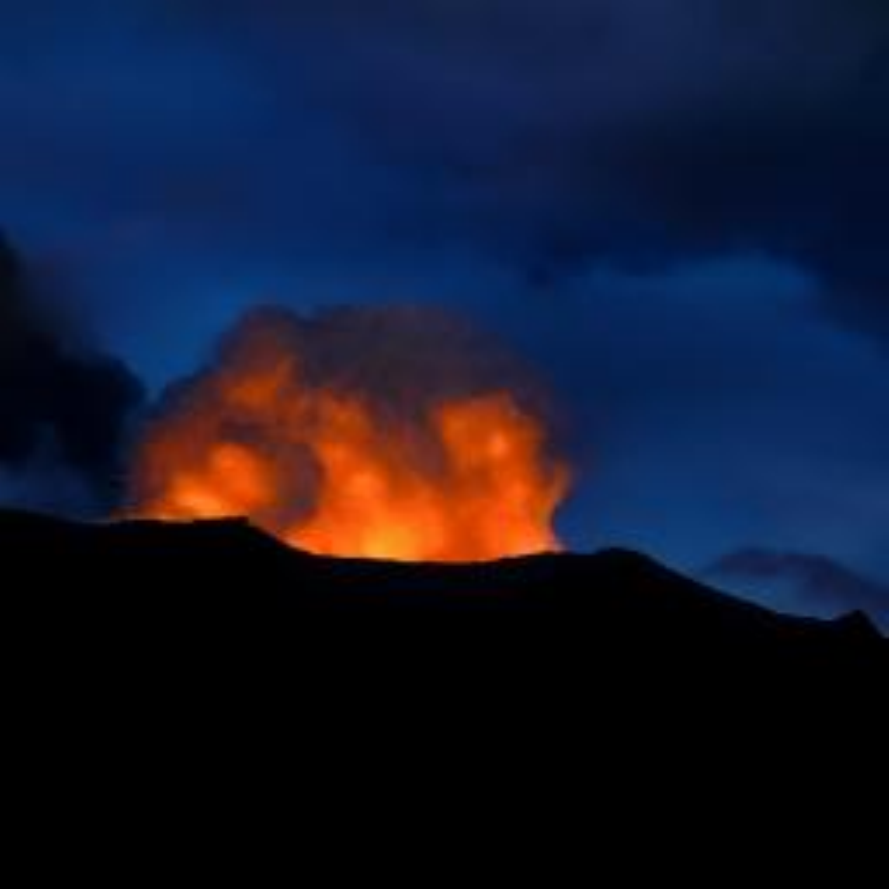} \\

        \raisebox{0\height}{\rotatebox{90}{\scriptsize Supervised Multi-Experts}} 
        \adjincludegraphics[clip,width=0.15\textwidth,trim={0 0 0 0}]{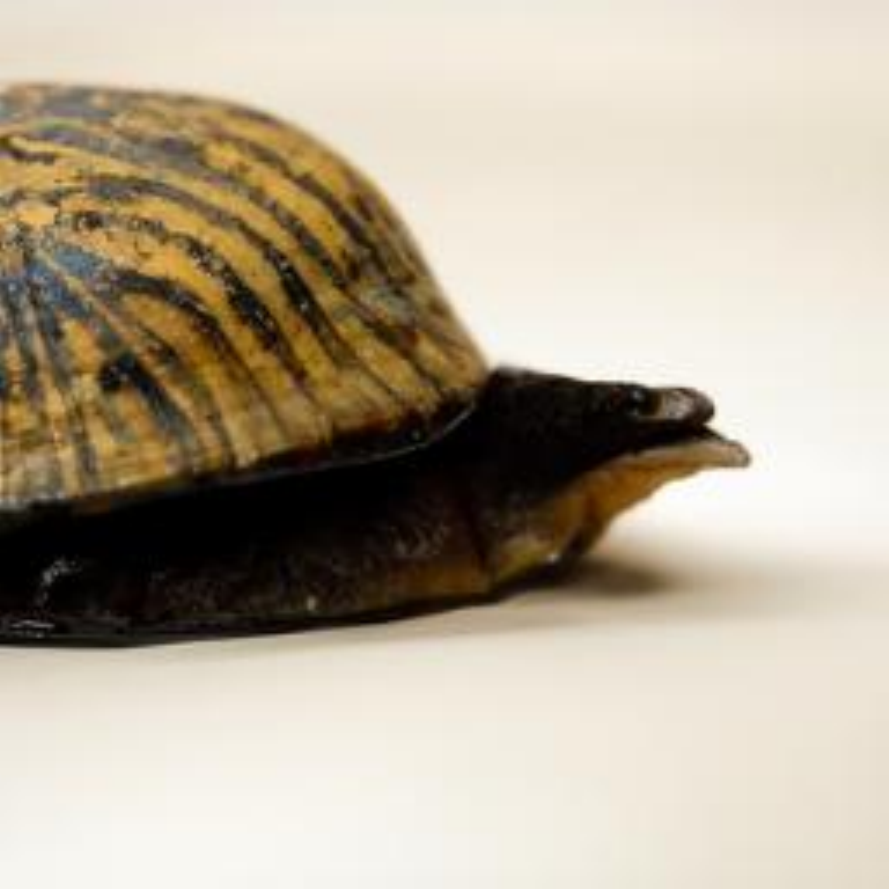} &
        \adjincludegraphics[clip,width=0.15\textwidth,trim={0 0 0 0}]{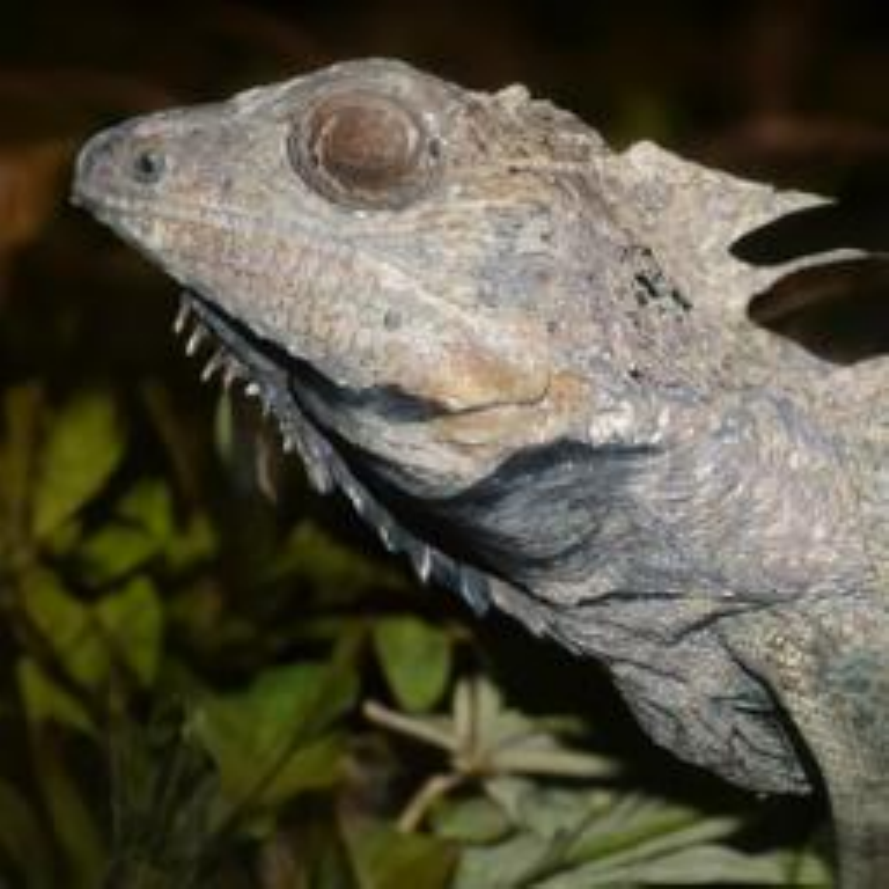} &
        \adjincludegraphics[clip,width=0.15\textwidth,trim={0 0 0 0}]{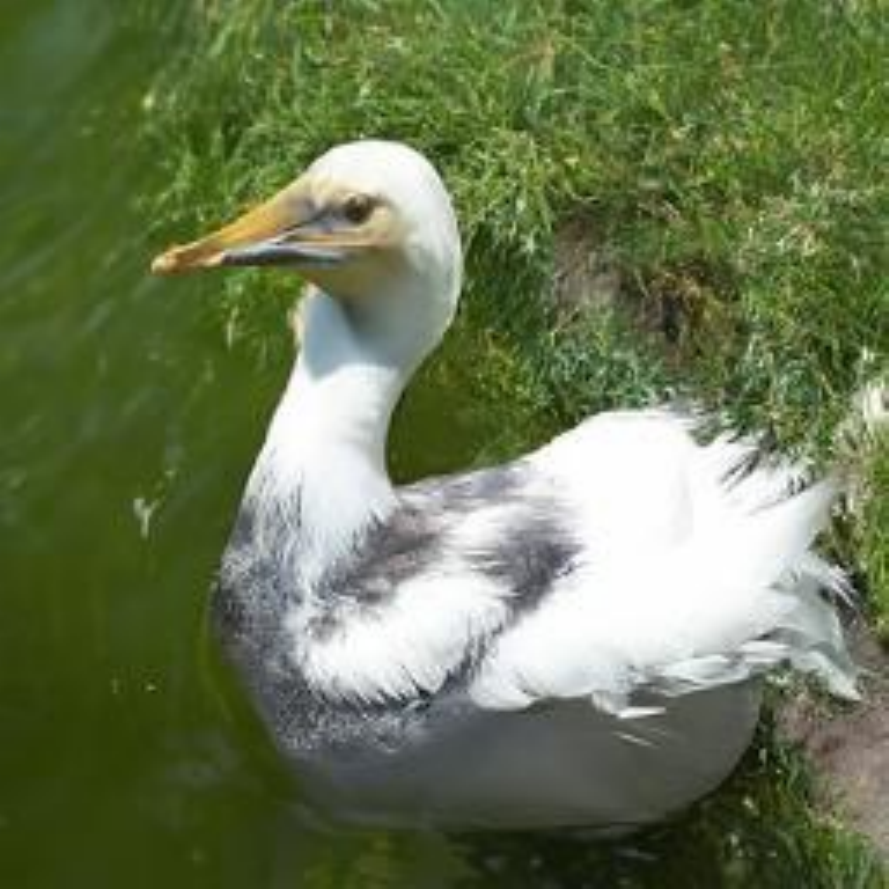} &
        \adjincludegraphics[clip,width=0.15\textwidth,trim={0 0 0 0}]{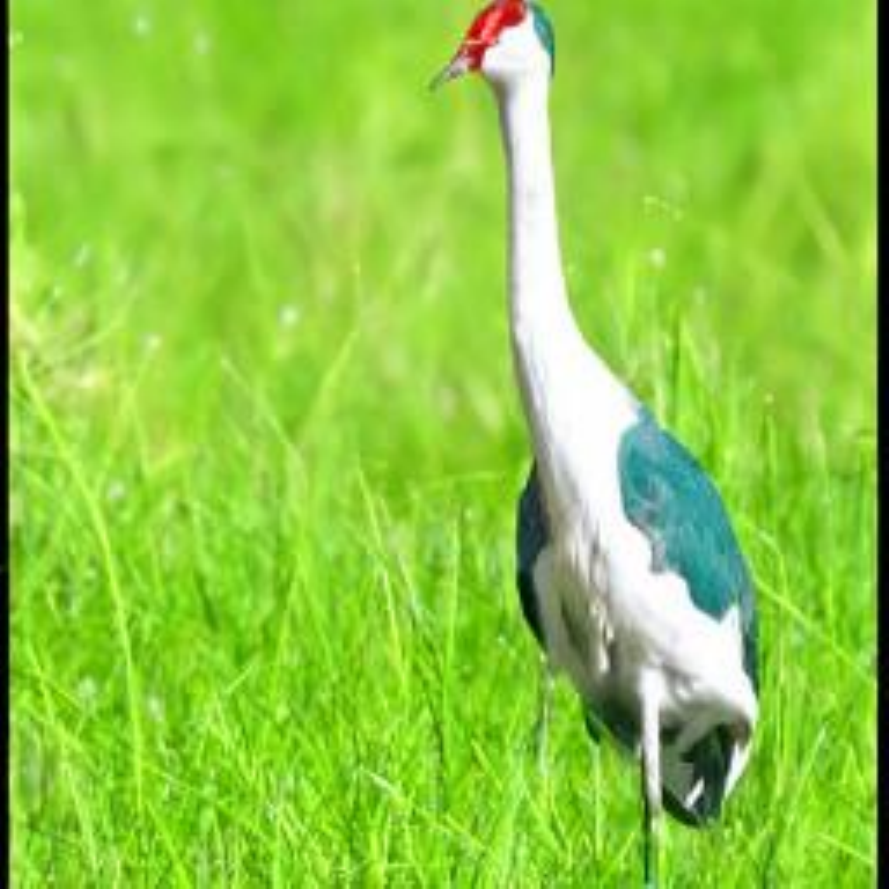} &
        \adjincludegraphics[clip,width=0.15\textwidth,trim={0 0 0 0}]{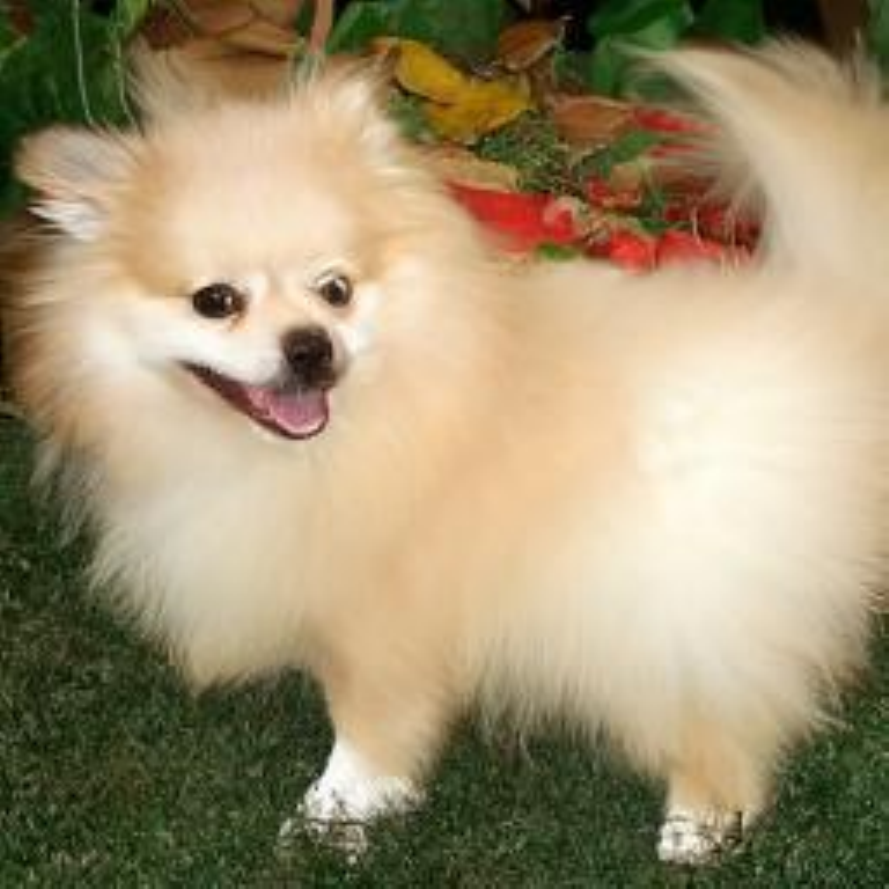} &
        \adjincludegraphics[clip,width=0.15\textwidth,trim={0 0 0 0}]{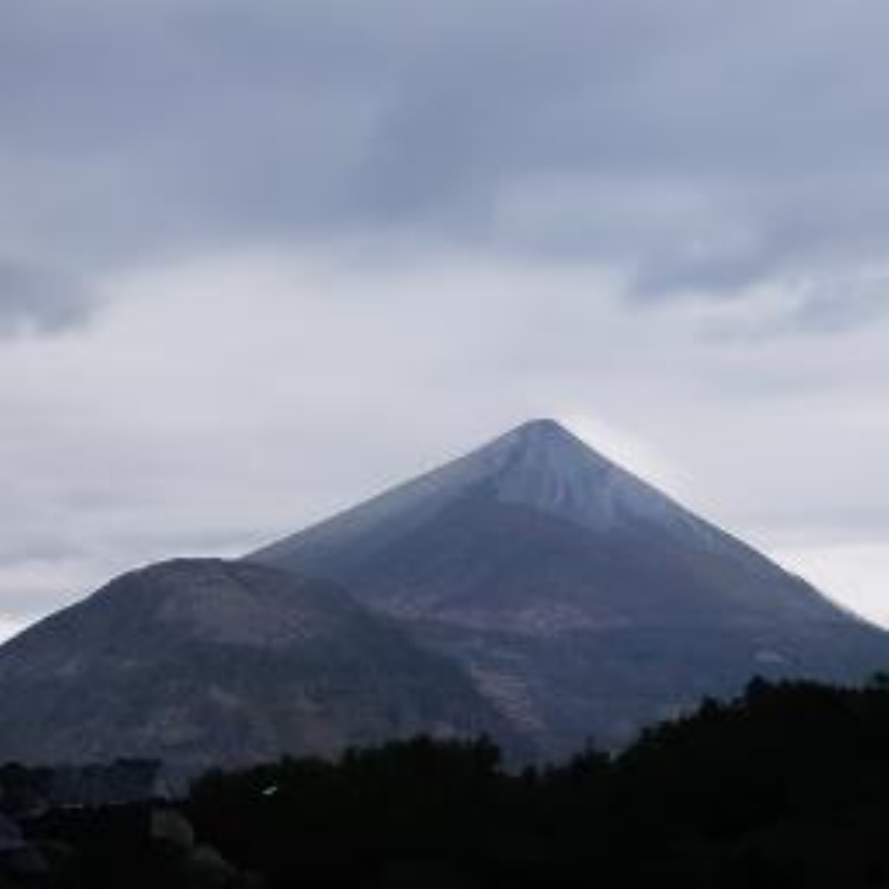} \\

        \raisebox{0.35\height}{\rotatebox{90}{\scriptsize PPAP ($N=5$)}} 
        \adjincludegraphics[clip,width=0.15\textwidth,trim={0 0 0 0}]{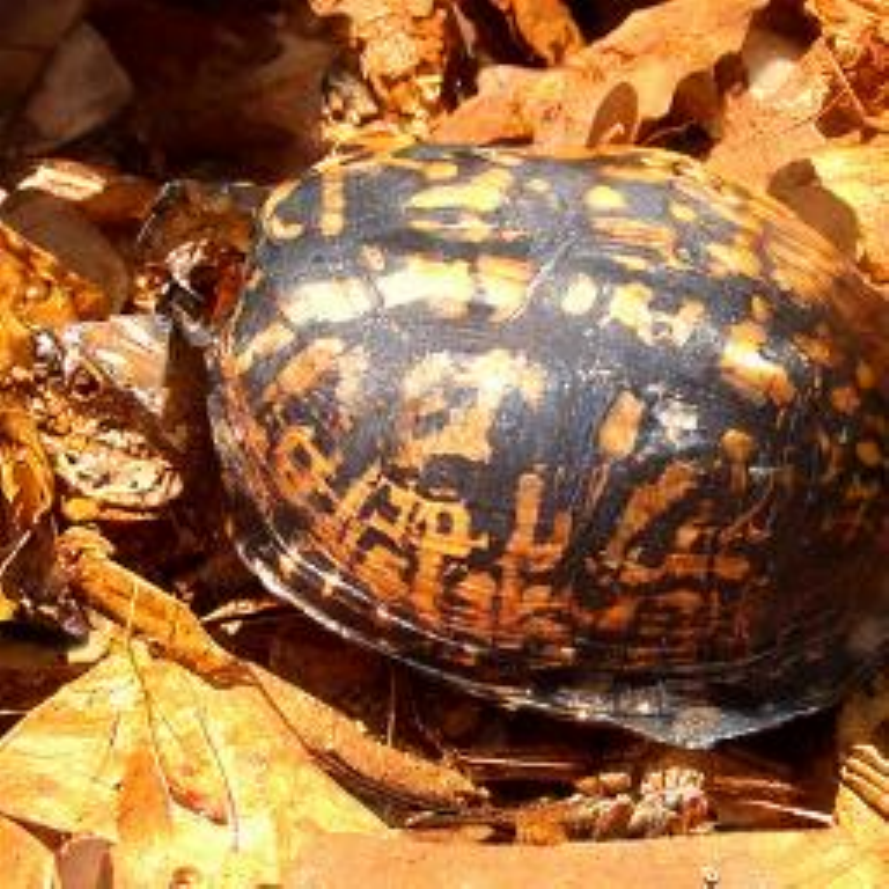} &
        \adjincludegraphics[clip,width=0.15\textwidth,trim={0 0 0 0}]{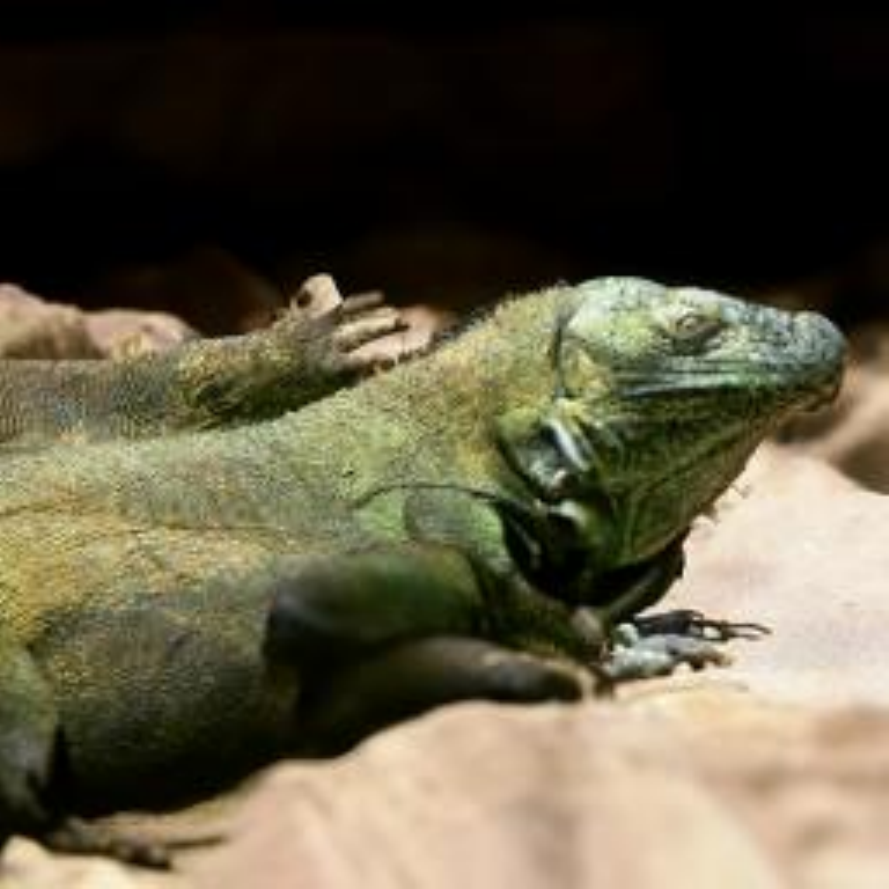} &
        \adjincludegraphics[clip,width=0.15\textwidth,trim={0 0 0 0}]{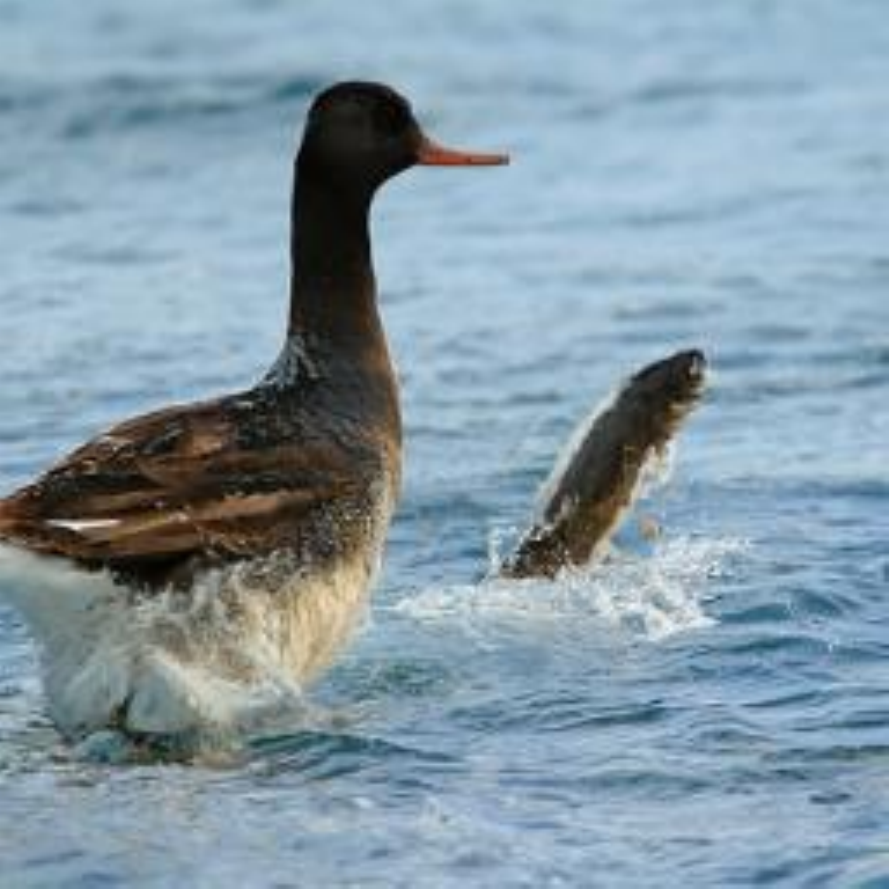} &
        \adjincludegraphics[clip,width=0.15\textwidth,trim={0 0 0 0}]{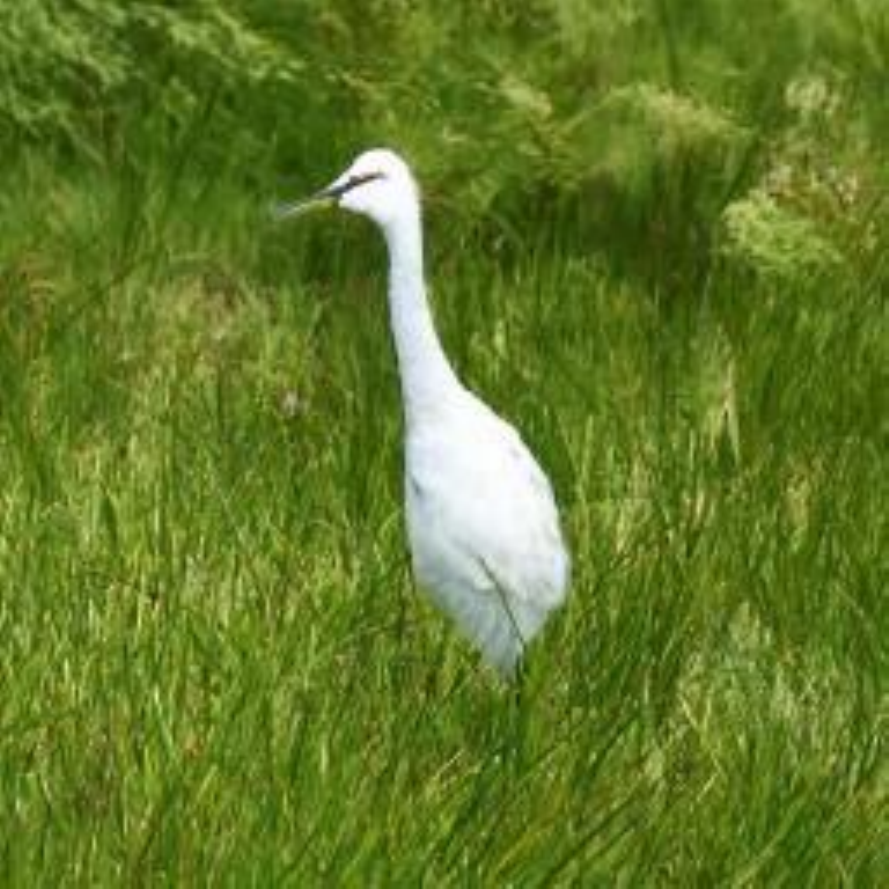} &
        \adjincludegraphics[clip,width=0.15\textwidth,trim={0 0 0 0}]{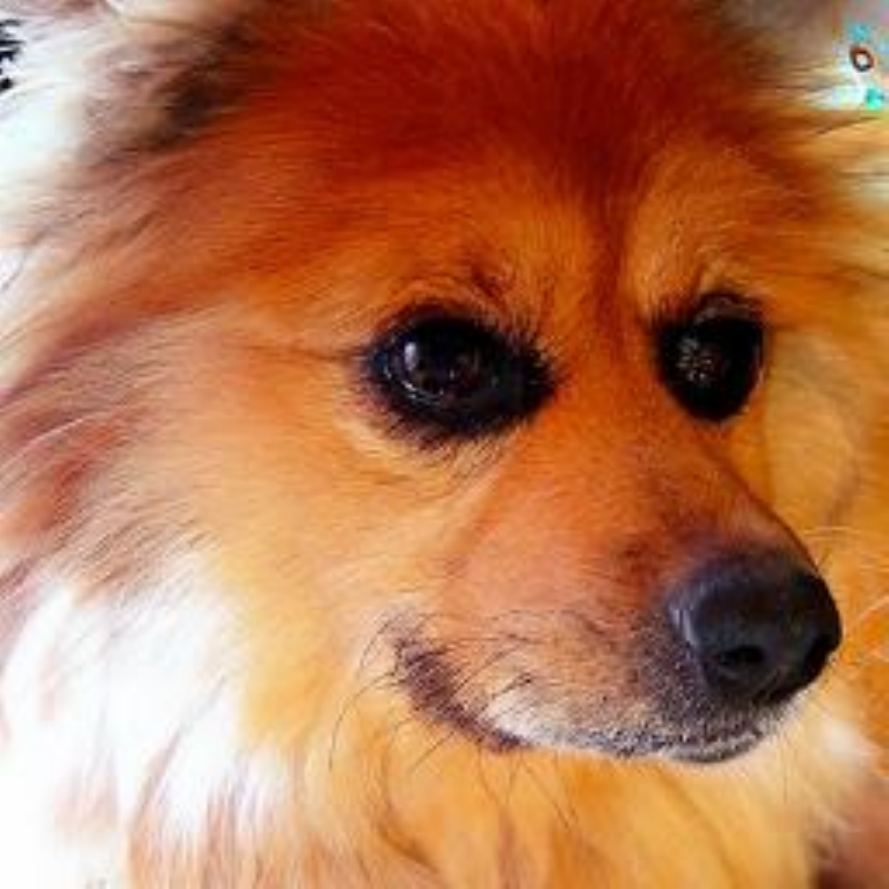} &
        \adjincludegraphics[clip,width=0.15\textwidth,trim={0 0 0 0}]{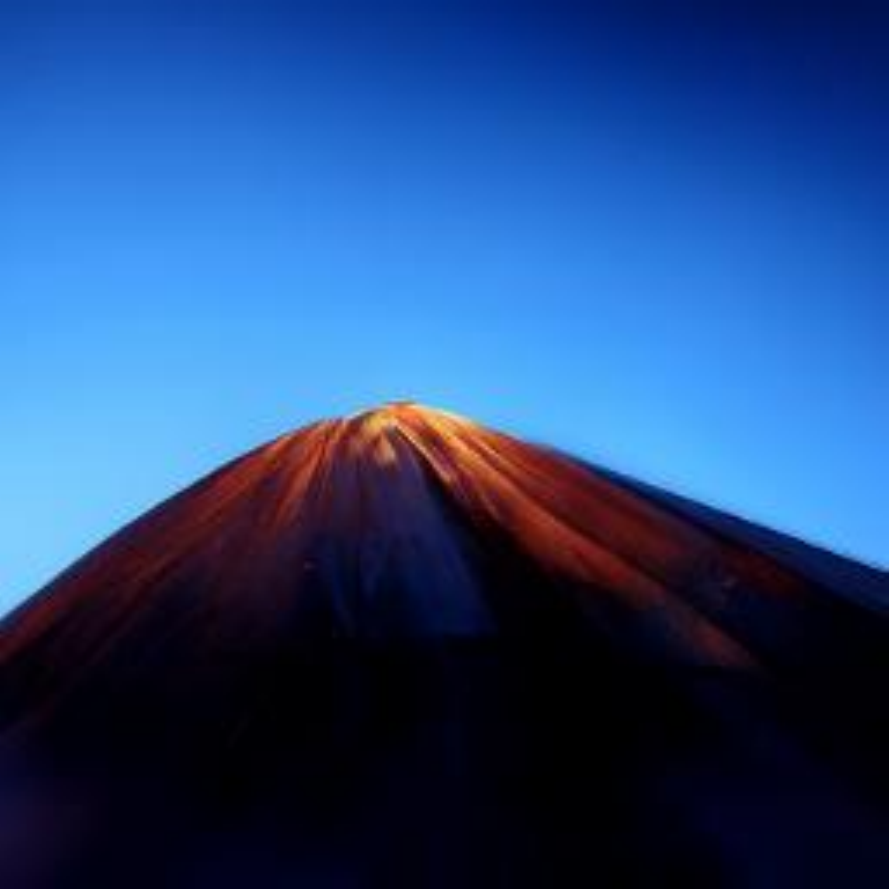} \\

        \raisebox{0.3\height}{\rotatebox{90}{\scriptsize PPAP ($N=10$)}} 
        \adjincludegraphics[clip,width=0.15\textwidth,trim={0 0 0 0}]{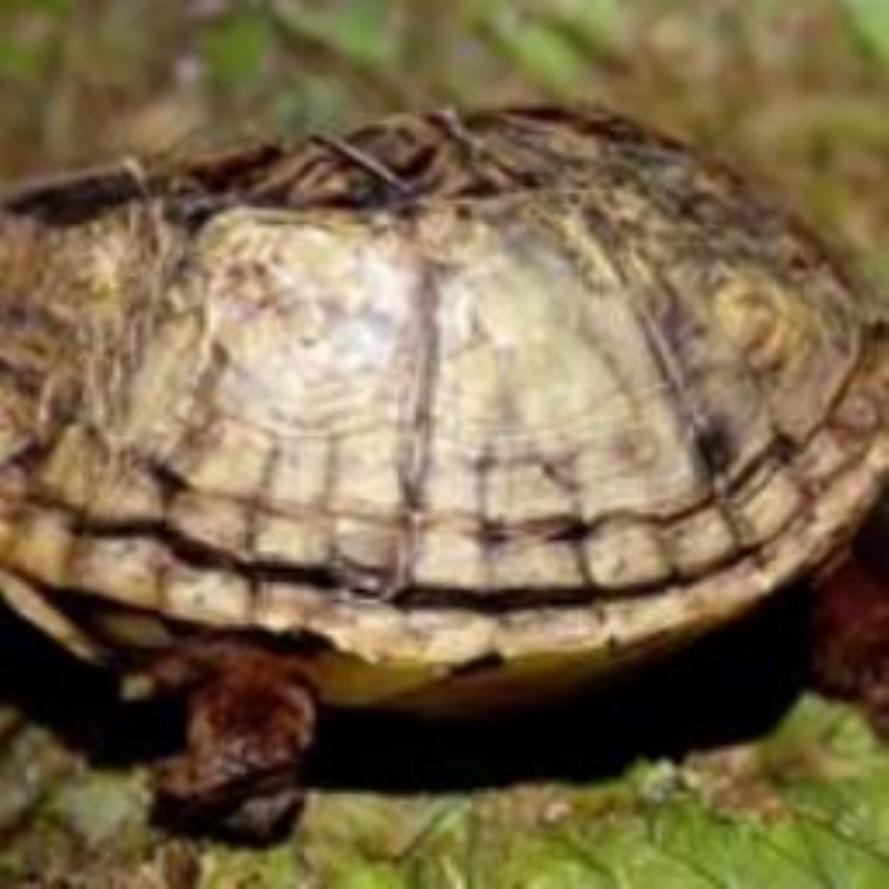} &
        \adjincludegraphics[clip,width=0.15\textwidth,trim={0 0 0 0}]{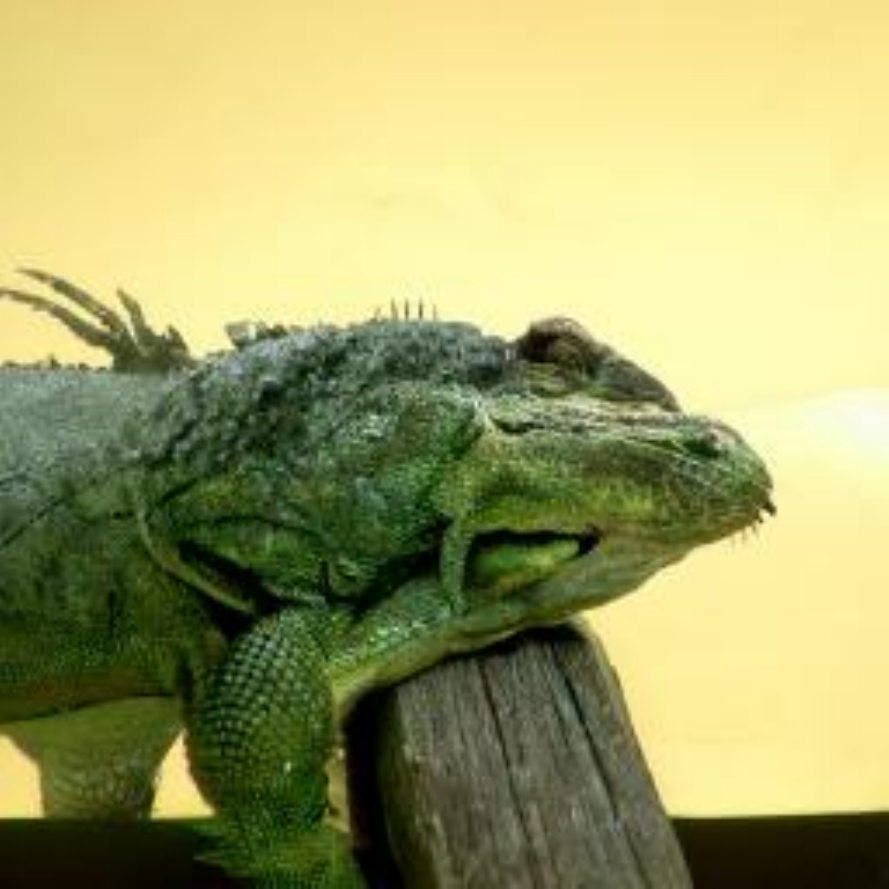} &
        \adjincludegraphics[clip,width=0.15\textwidth,trim={0 0 0 0}]{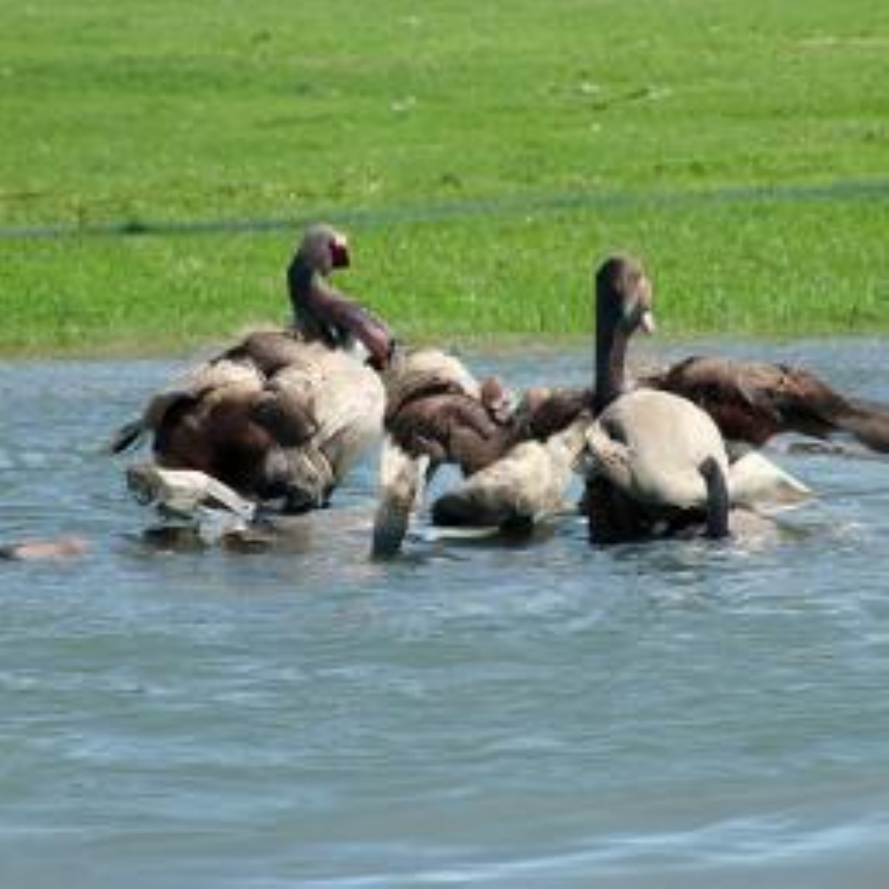} &
        \adjincludegraphics[clip,width=0.15\textwidth,trim={0 0 0 0}]{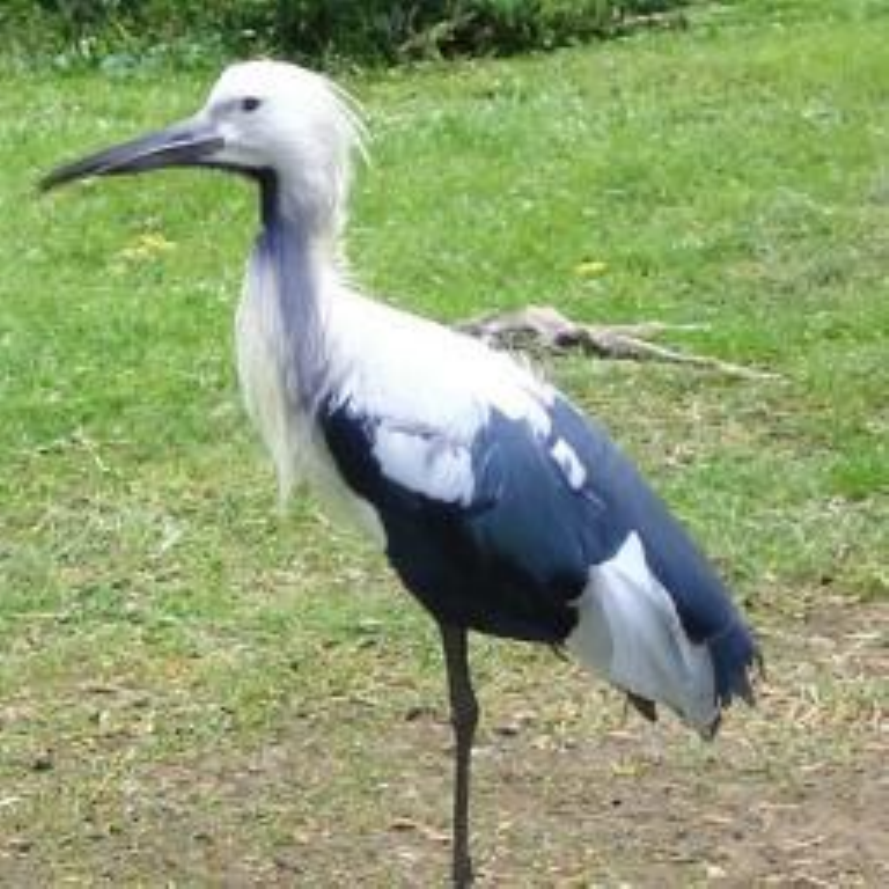} &
        \adjincludegraphics[clip,width=0.15\textwidth,trim={0 0 0 0}]{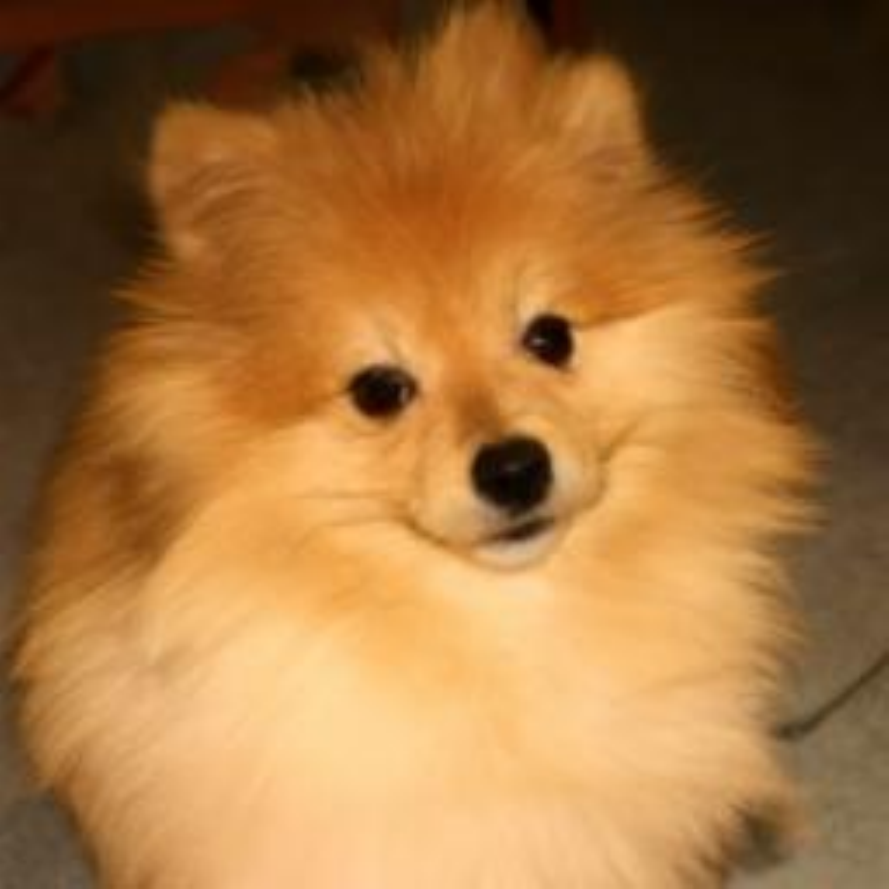} &
        \adjincludegraphics[clip,width=0.15\textwidth,trim={0 0 0 0}]{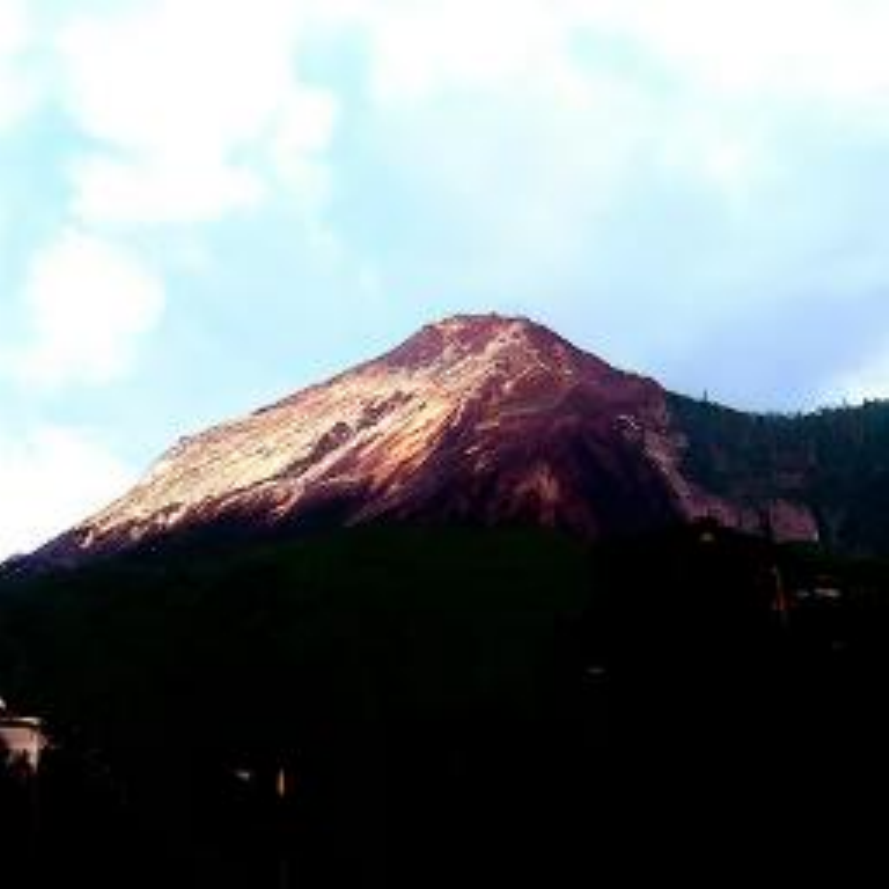} \\
        
        \footnotesize ~~~~~~Box turtle & \footnotesize  Iguana & \footnotesize  Goose & \footnotesize  Crane & \footnotesize Pomeranian & \footnotesize Volcano \\
    \end{tabular}
    \caption{
    Qualitative results on ImageNet class conditional generation with DDPM 250 steps by guiding unconditional ADM with DeiT-S.
    }
    \label{apx:fig_qualitative_deit_ddpm}

\end{figure*}

\begin{figure*}[t]

    \centering
    \begin{tabular}{@{\hspace{2mm}}c@{\hspace{2mm}}c@{\hspace{2mm}}c@{\hspace{2mm}}c@{\hspace{2mm}}c@{\hspace{2mm}}c@{\hspace{2mm}}c@{\hspace{2mm}}c}
    
        \raisebox{0.2\height}{\rotatebox{90}{\scriptsize Na\"ive Off-the-shelf}} 
        \adjincludegraphics[clip,width=0.15\textwidth,trim={0 0 0 0}]{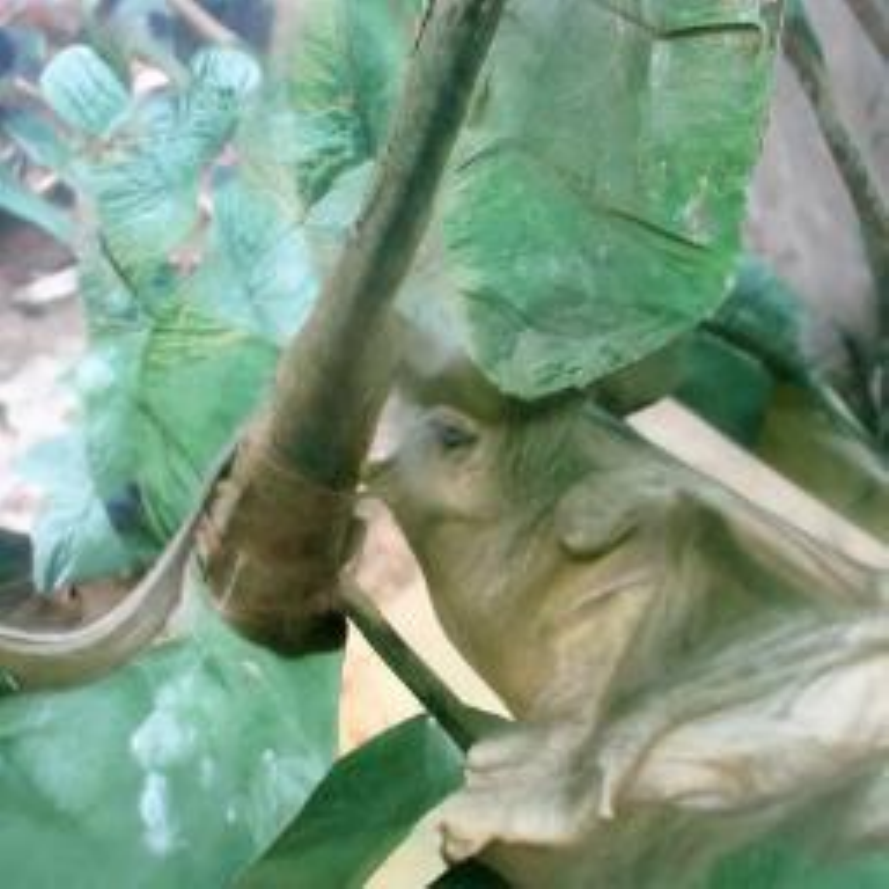} &
        \adjincludegraphics[clip,width=0.15\textwidth,trim={0 0 0 0}]{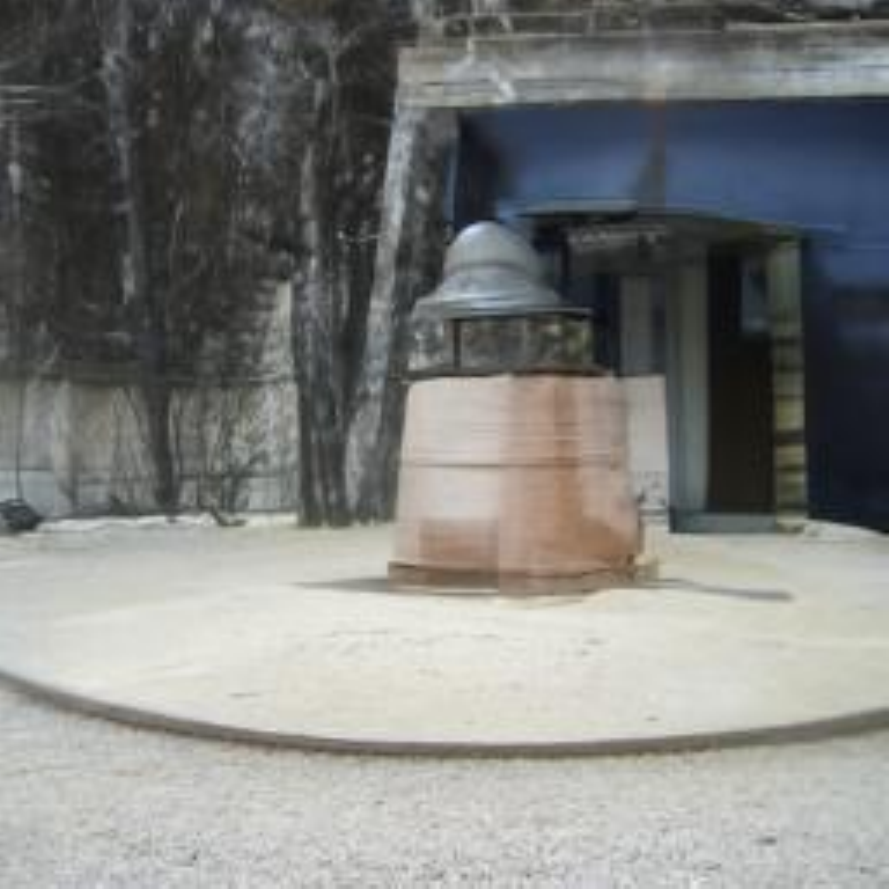} &
        \adjincludegraphics[clip,width=0.15\textwidth,trim={0 0 0 0}]{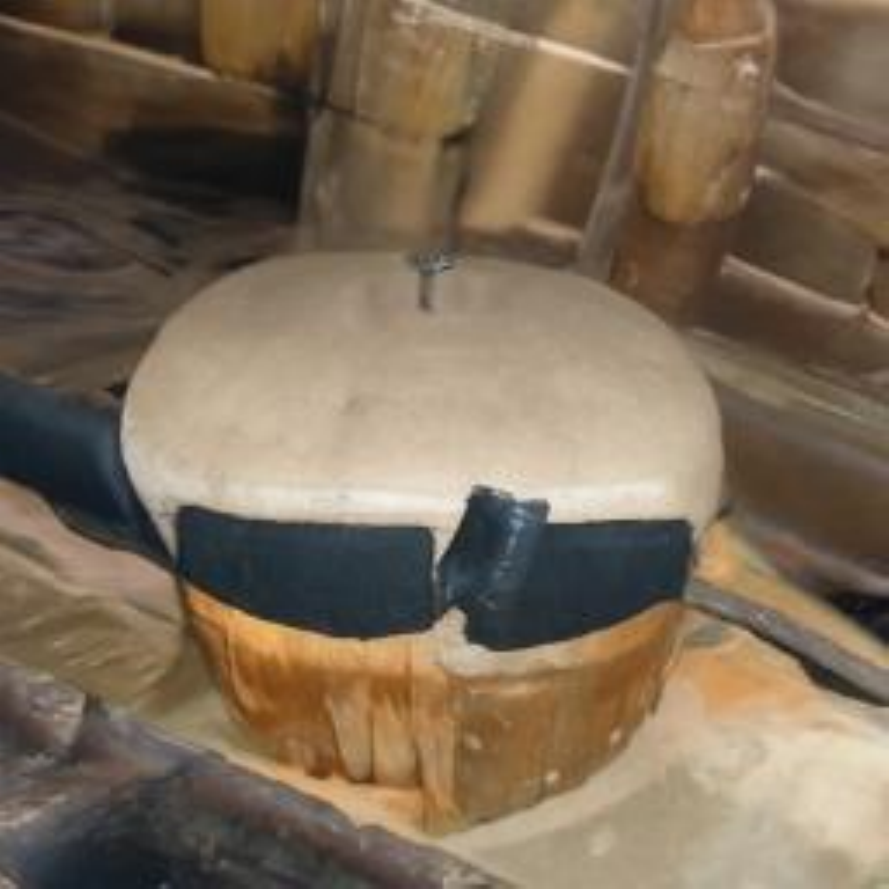} &
        \adjincludegraphics[clip,width=0.15\textwidth,trim={0 0 0 0}]{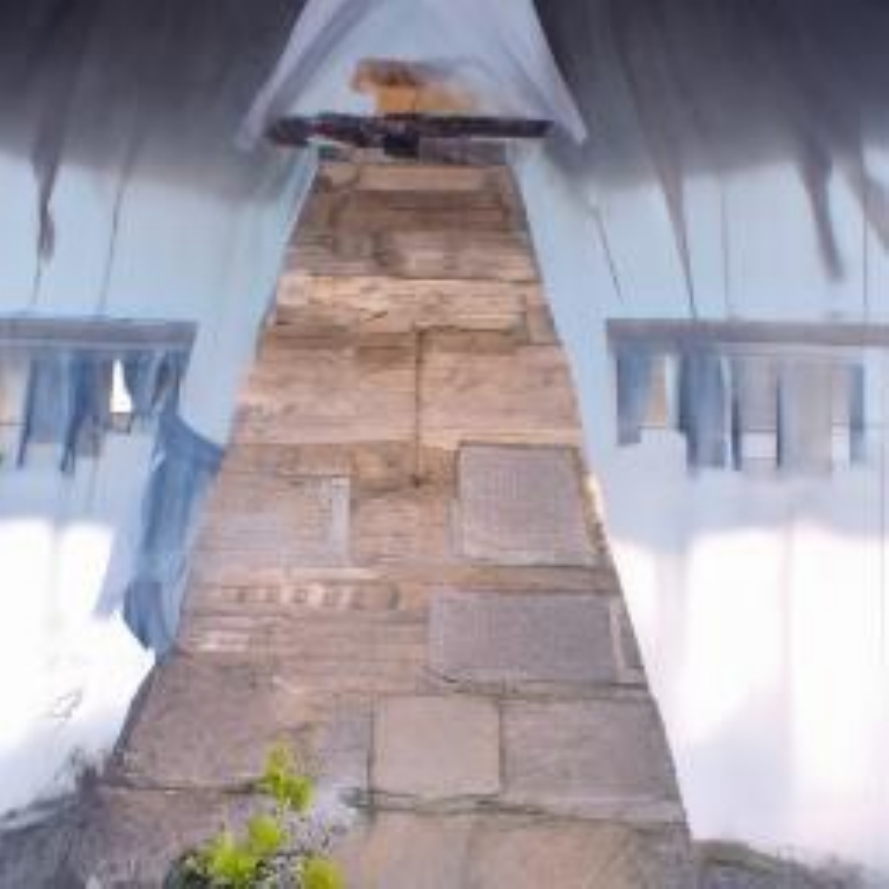} &
        \adjincludegraphics[clip,width=0.15\textwidth,trim={0 0 0 0}]{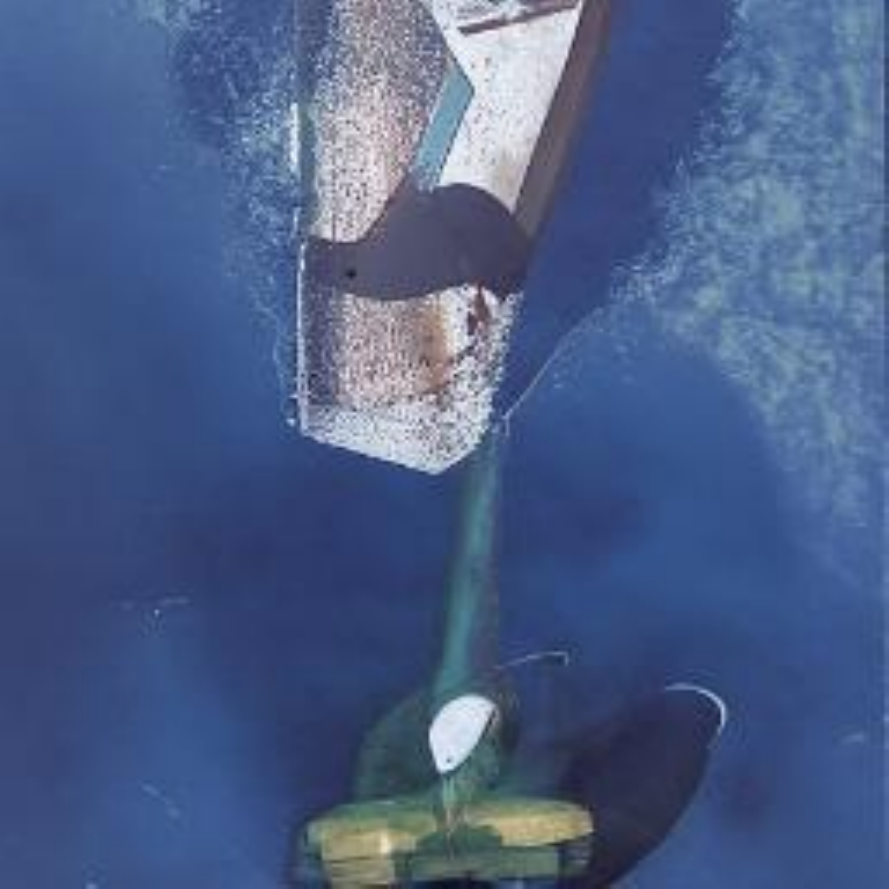} &
        \adjincludegraphics[clip,width=0.15\textwidth,trim={0 0 0 0}]{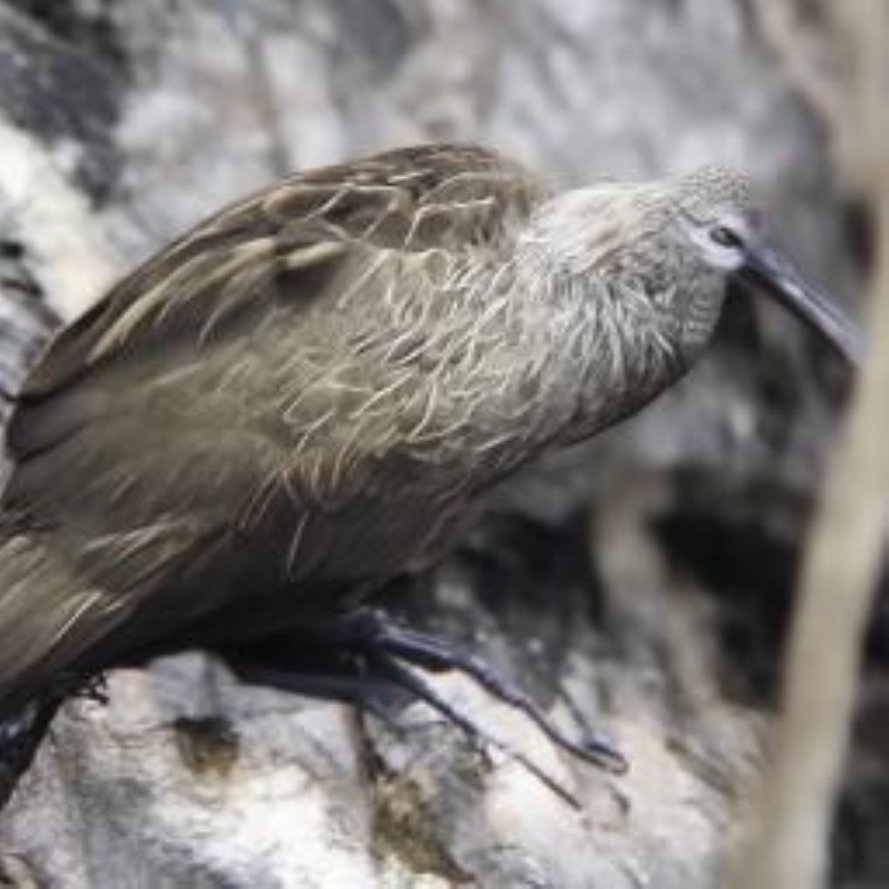} \\

            \raisebox{0.2\height}{\rotatebox{90}{\scriptsize Single Noise-aware}} 
        \adjincludegraphics[clip,width=0.15\textwidth,trim={0 0 0 0}]{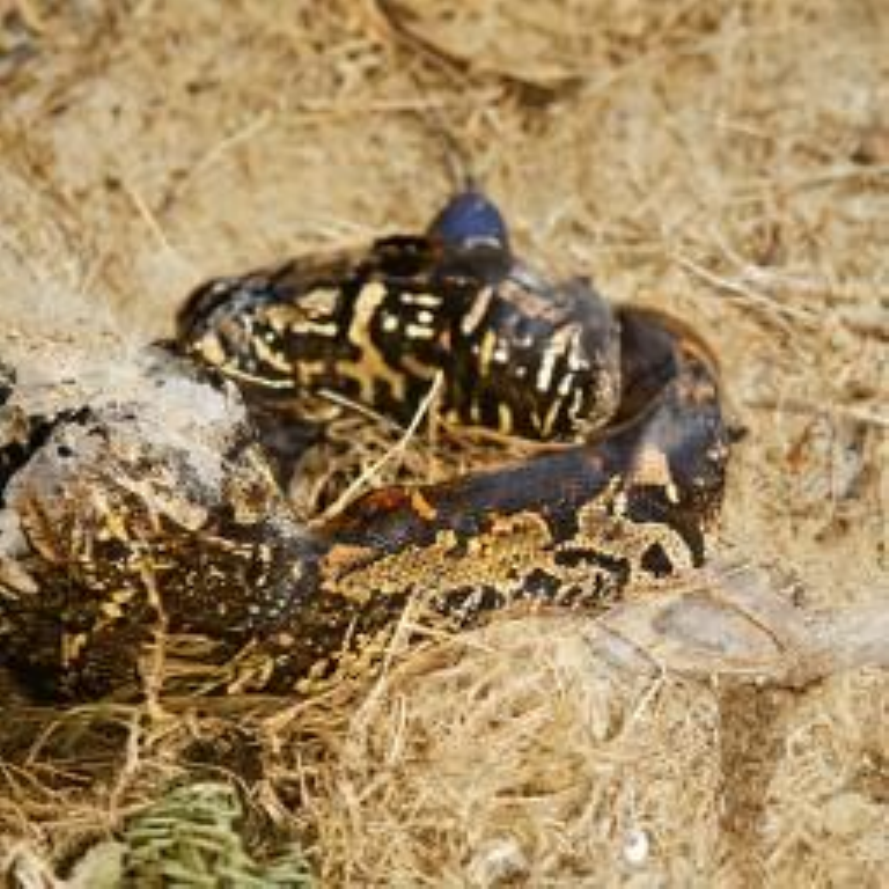} &
        \adjincludegraphics[clip,width=0.15\textwidth,trim={0 0 0 0}]{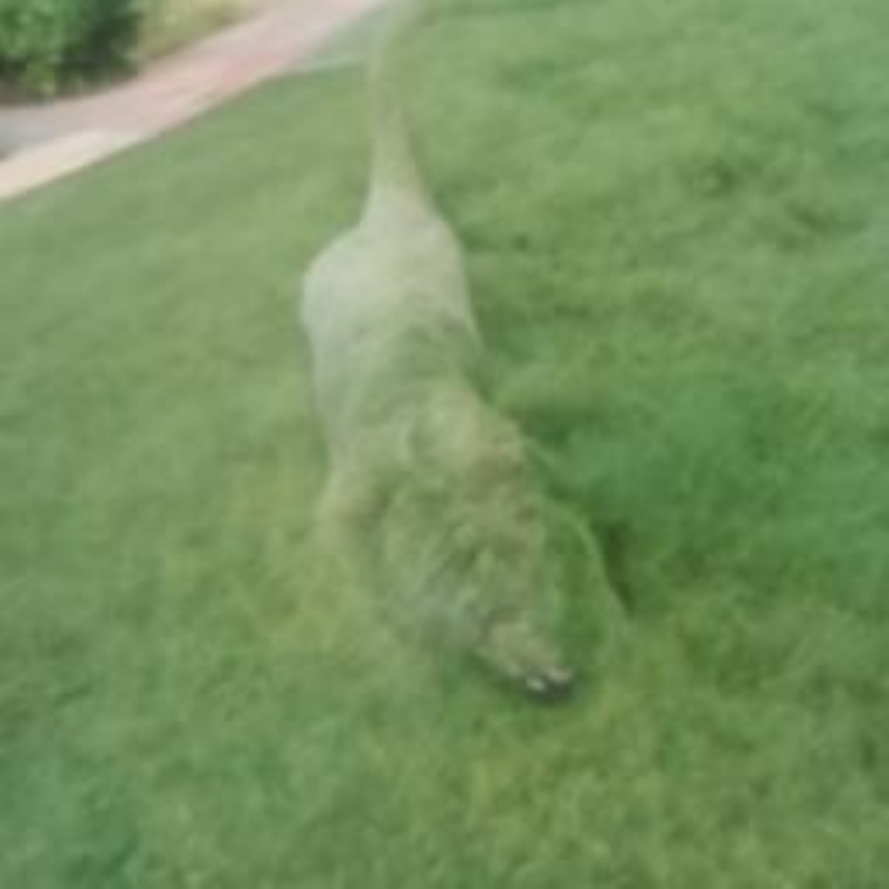} &
        \adjincludegraphics[clip,width=0.15\textwidth,trim={0 0 0 0}]{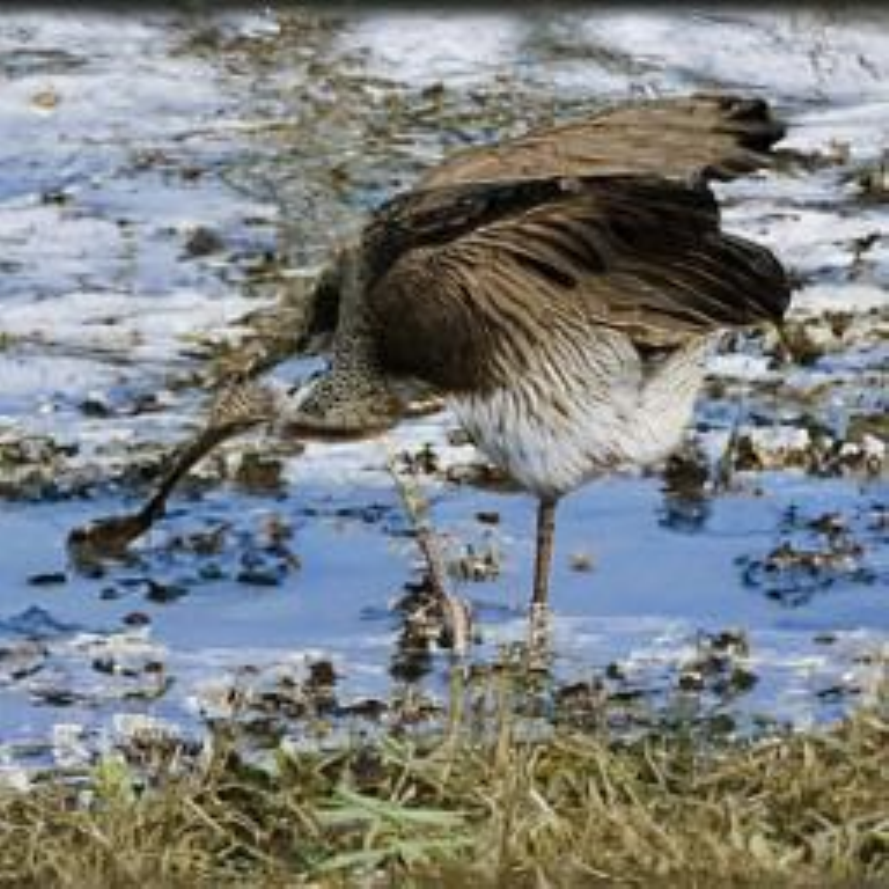} &
        \adjincludegraphics[clip,width=0.15\textwidth,trim={0 0 0 0}]{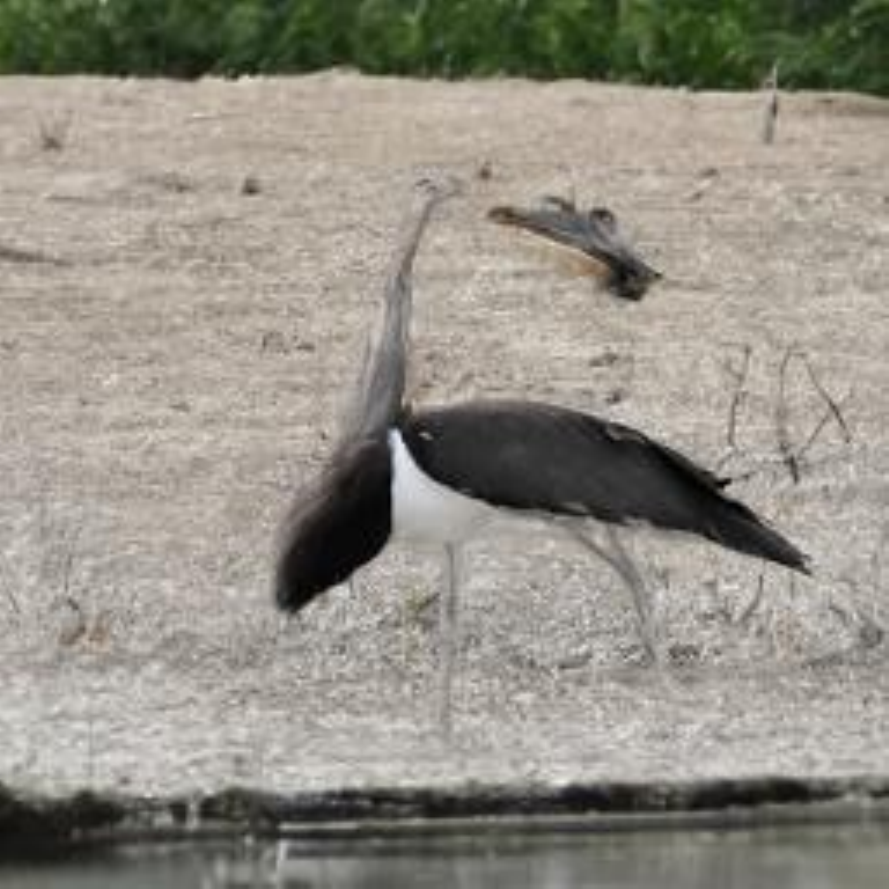} &
        \adjincludegraphics[clip,width=0.15\textwidth,trim={0 0 0 0}]{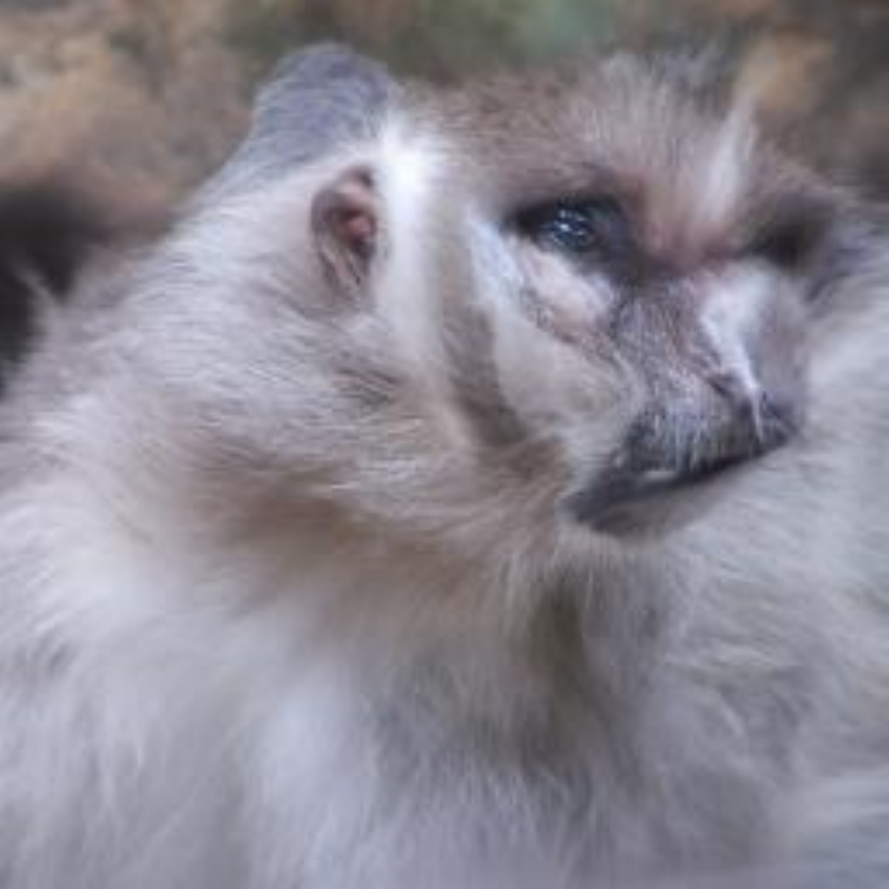} &
        \adjincludegraphics[clip,width=0.15\textwidth,trim={0 0 0 0}]{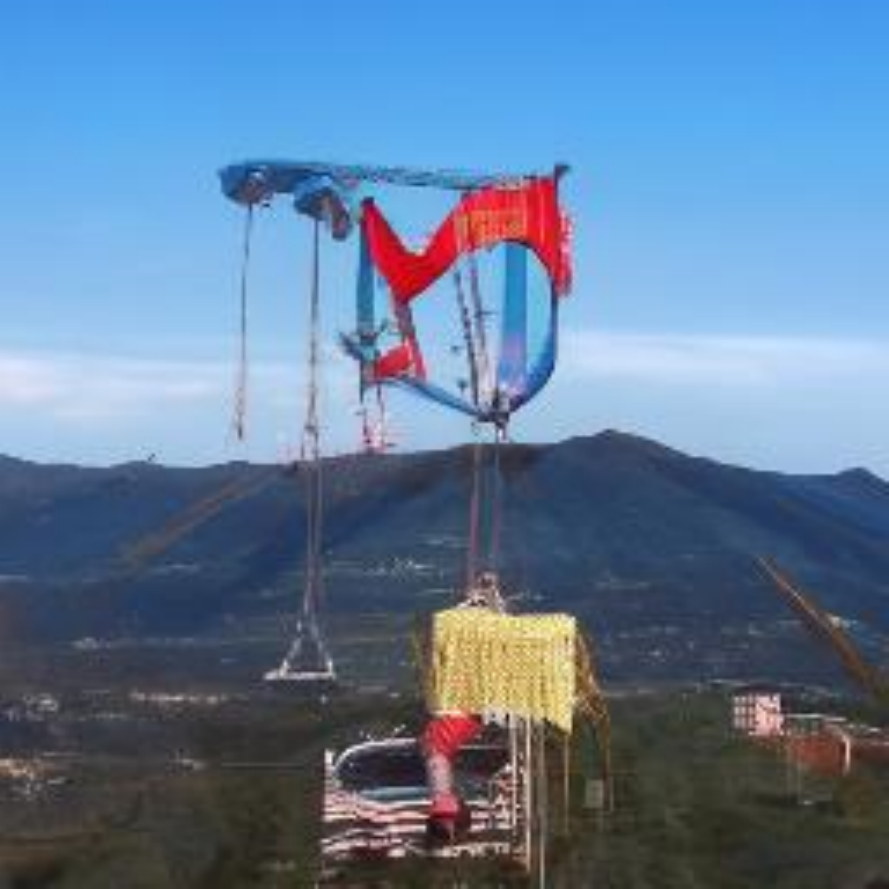} \\

        \raisebox{0\height}{\rotatebox{90}{\scriptsize Supervised Multi-Experts}} 
        \adjincludegraphics[clip,width=0.15\textwidth,trim={0 0 0 0}]{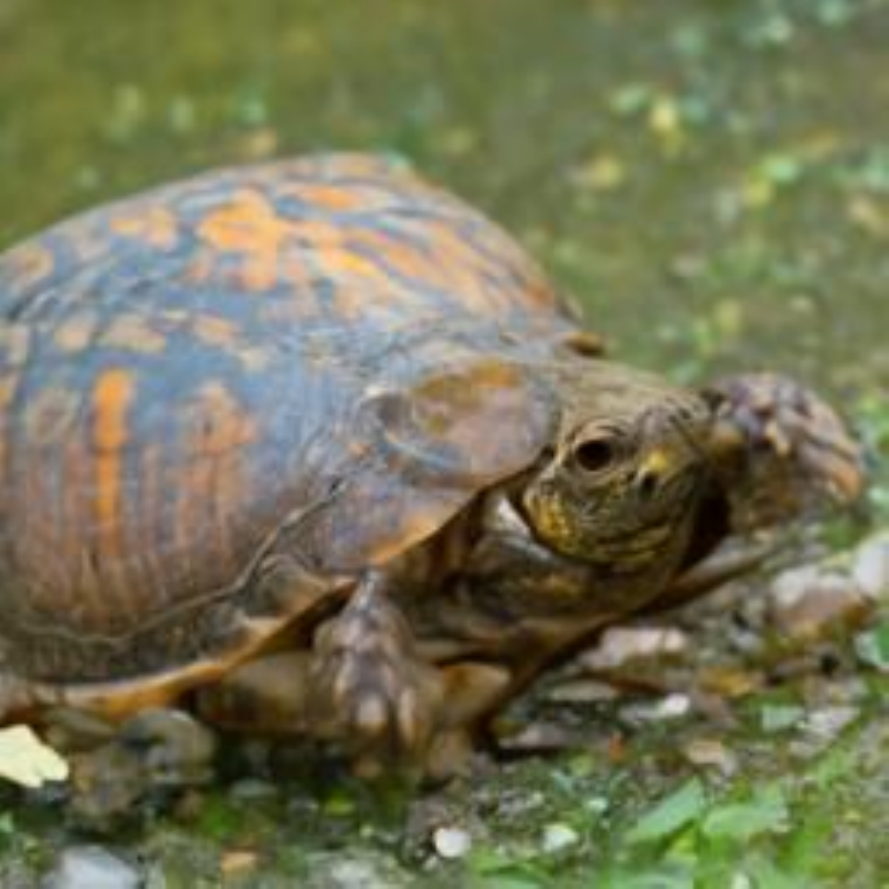} &
        \adjincludegraphics[clip,width=0.15\textwidth,trim={0 0 0 0}]{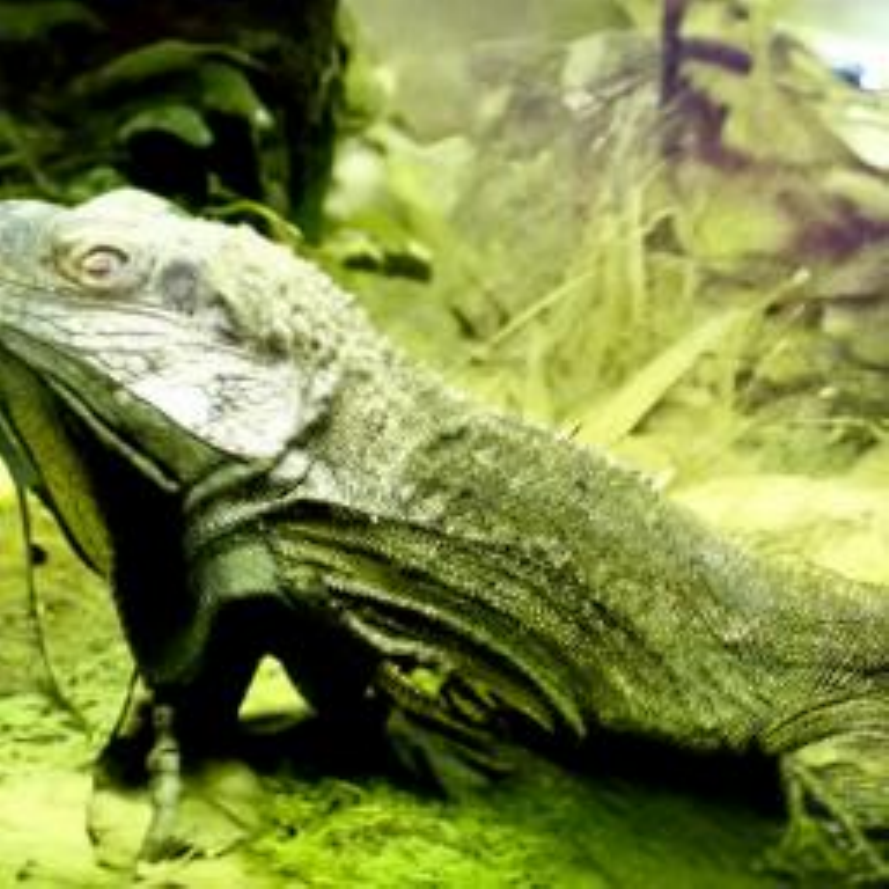} &
        \adjincludegraphics[clip,width=0.15\textwidth,trim={0 0 0 0}]{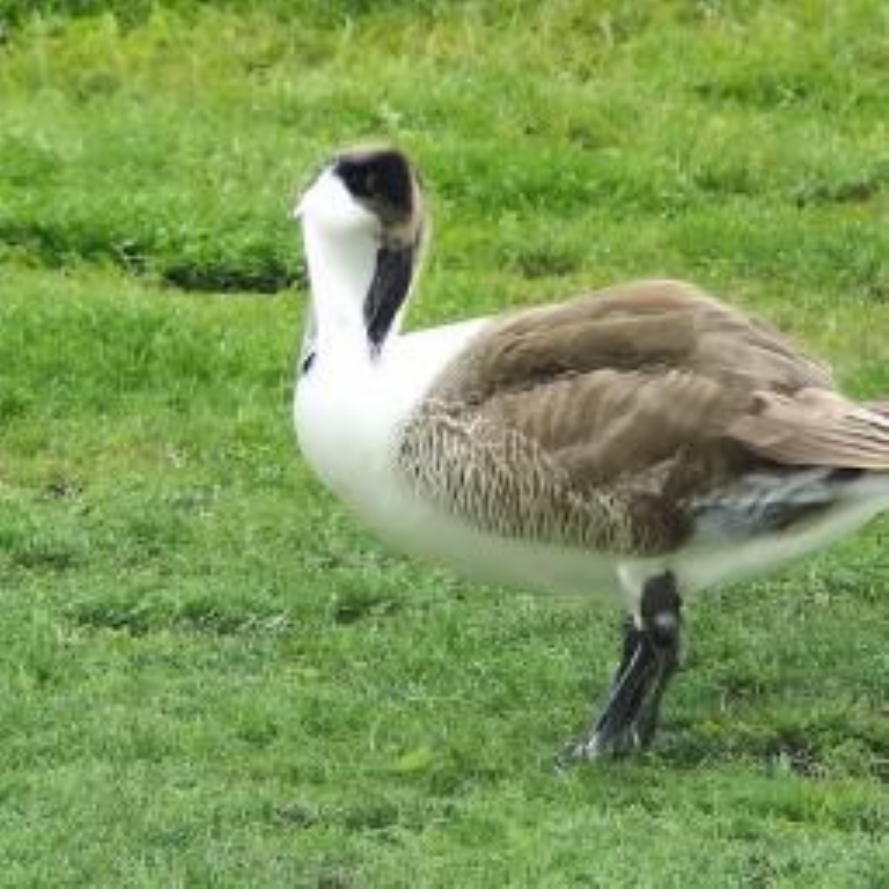} &
        \adjincludegraphics[clip,width=0.15\textwidth,trim={0 0 0 0}]{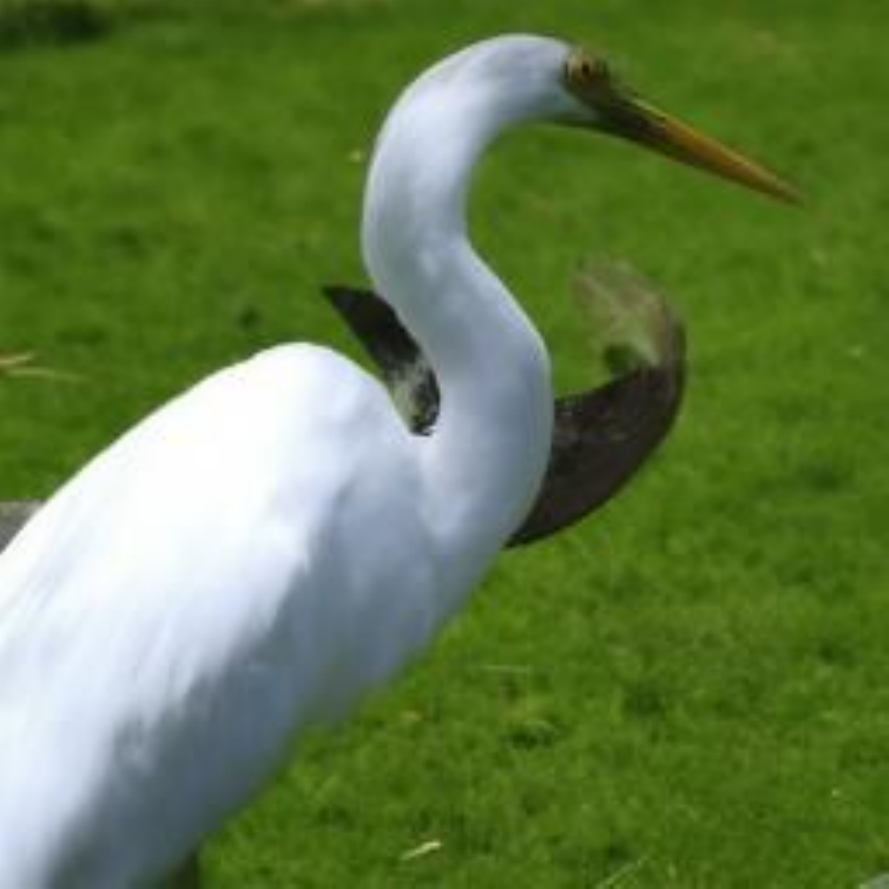} &
        \adjincludegraphics[clip,width=0.15\textwidth,trim={0 0 0 0}]{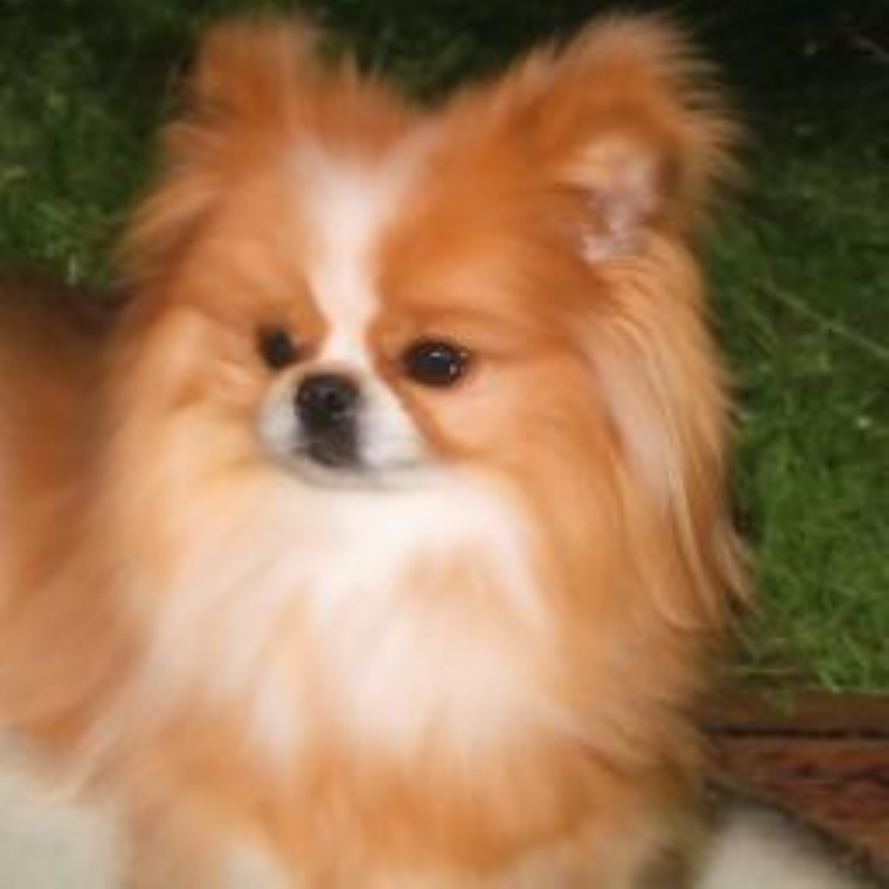} &
        \adjincludegraphics[clip,width=0.15\textwidth,trim={0 0 0 0}]{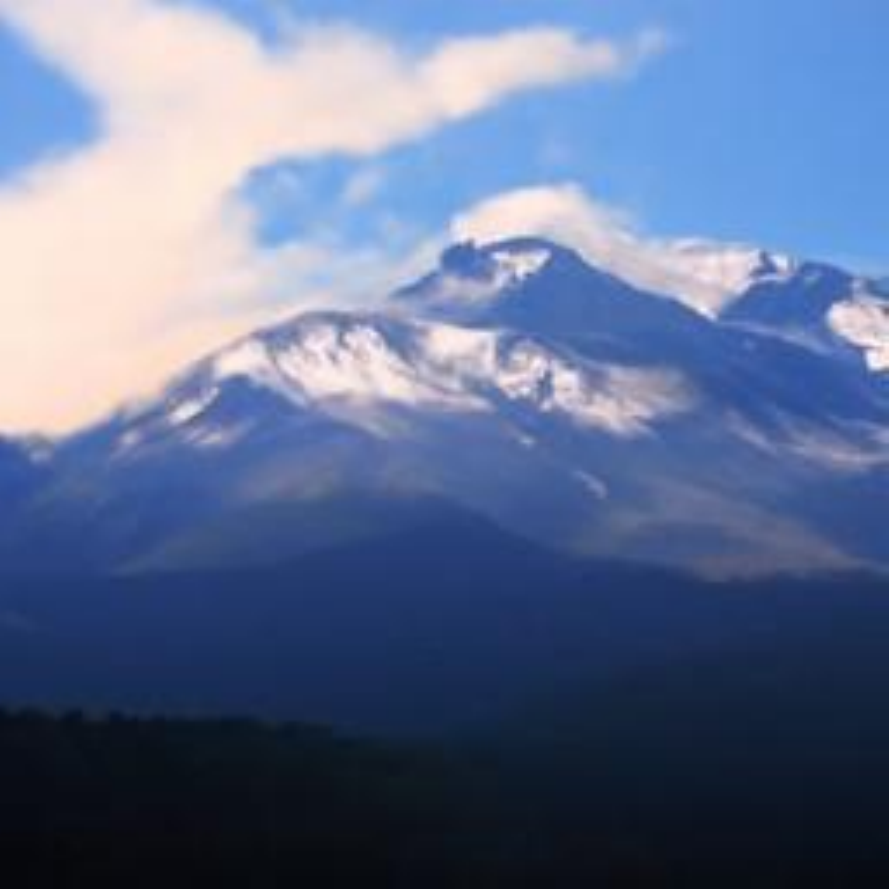} \\

        \raisebox{0.35\height}{\rotatebox{90}{\scriptsize PPAP ($N=5$)}} 
        \adjincludegraphics[clip,width=0.15\textwidth,trim={0 0 0 0}]{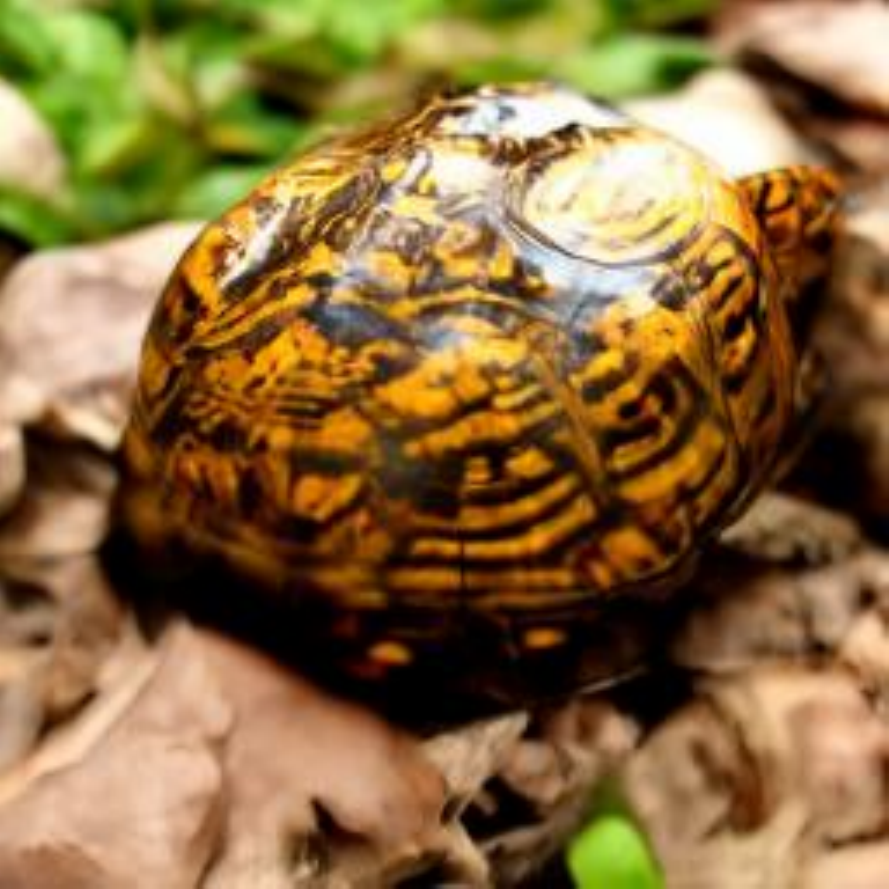} &
        \adjincludegraphics[clip,width=0.15\textwidth,trim={0 0 0 0}]{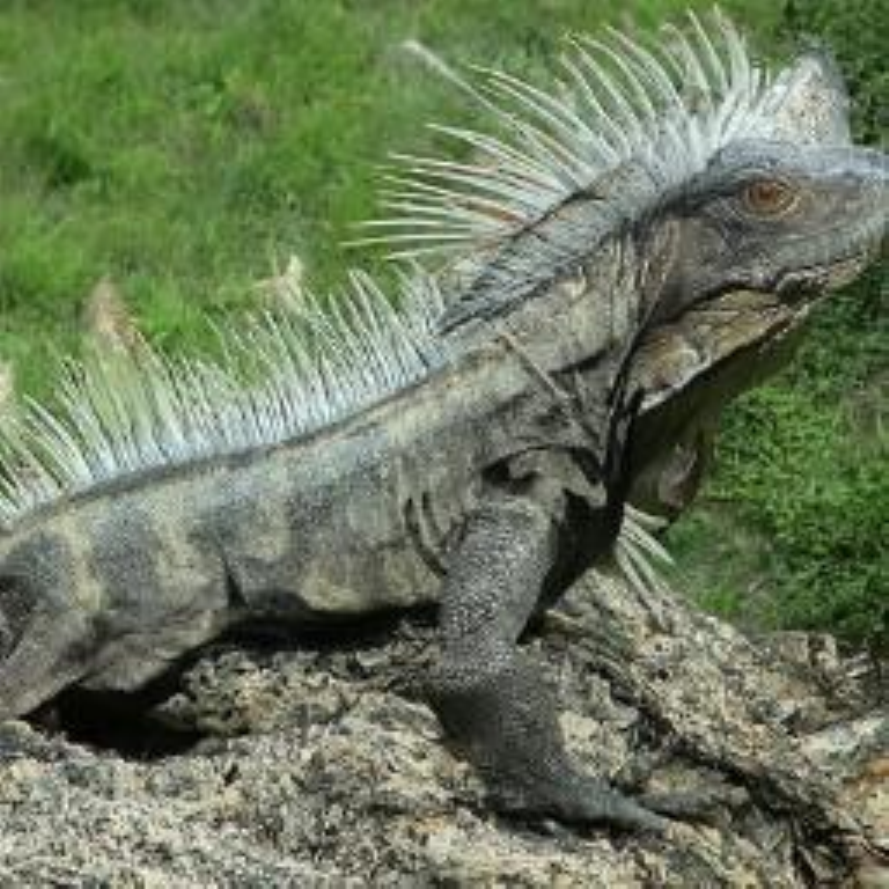} &
        \adjincludegraphics[clip,width=0.15\textwidth,trim={0 0 0 0}]{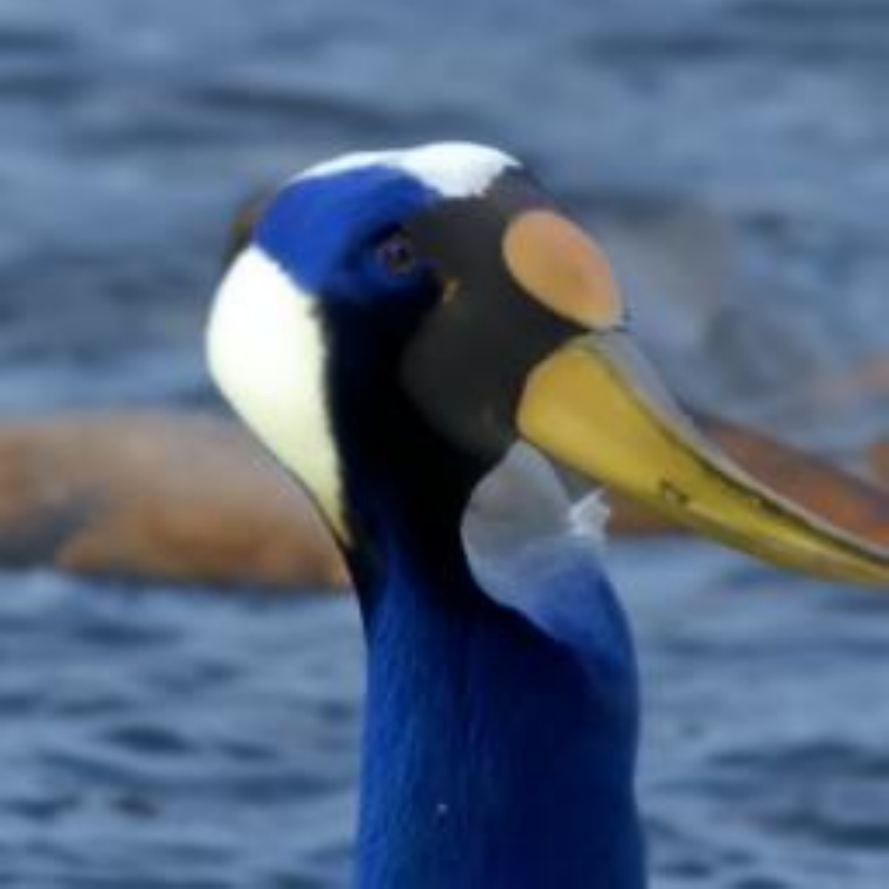} &
        \adjincludegraphics[clip,width=0.15\textwidth,trim={0 0 0 0}]{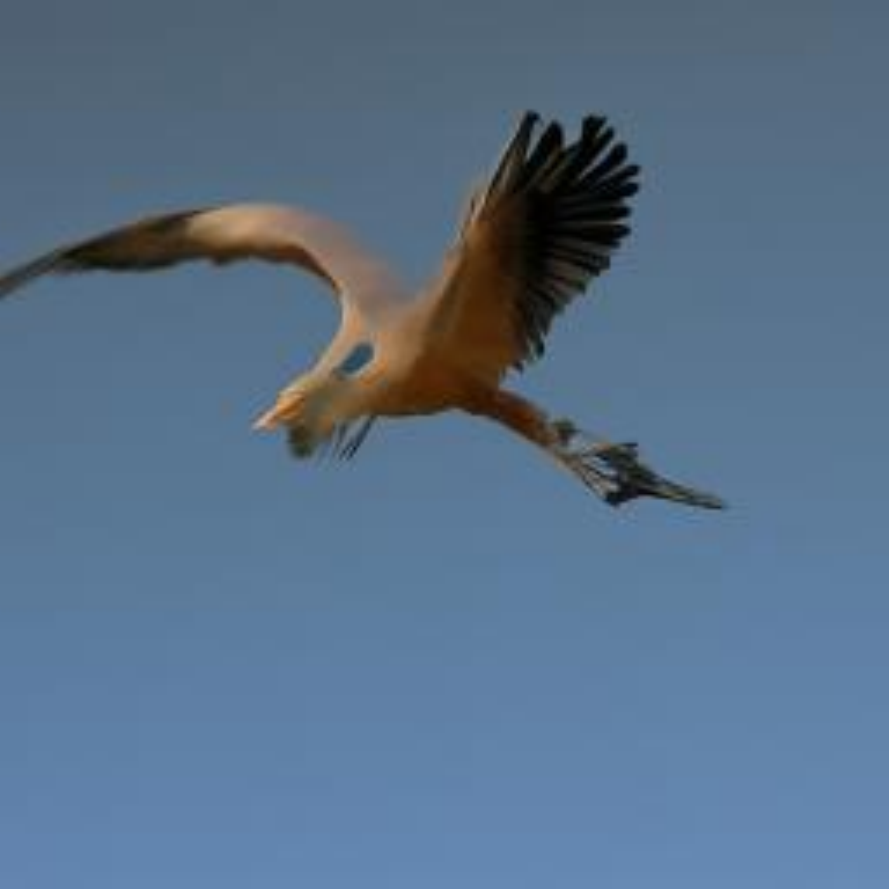} &
        \adjincludegraphics[clip,width=0.15\textwidth,trim={0 0 0 0}]{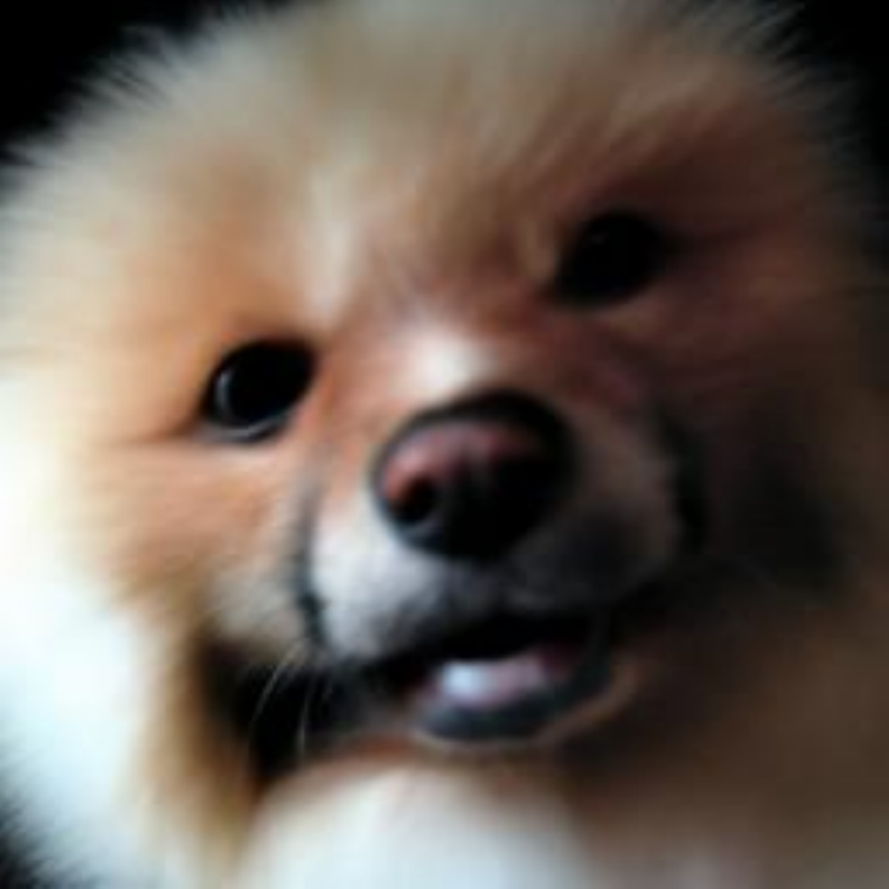} &
        \adjincludegraphics[clip,width=0.15\textwidth,trim={0 0 0 0}]{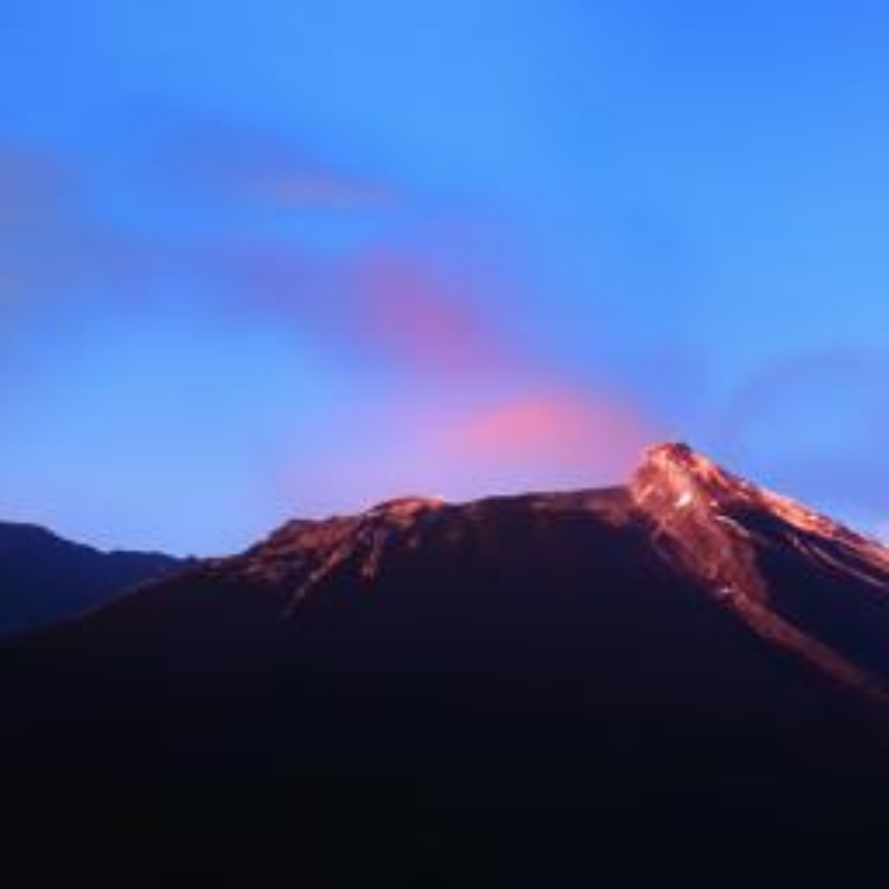} \\

        \raisebox{0.3\height}{\rotatebox{90}{\scriptsize PPAP ($N=10$)}} 
        \adjincludegraphics[clip,width=0.15\textwidth,trim={0 0 0 0}]{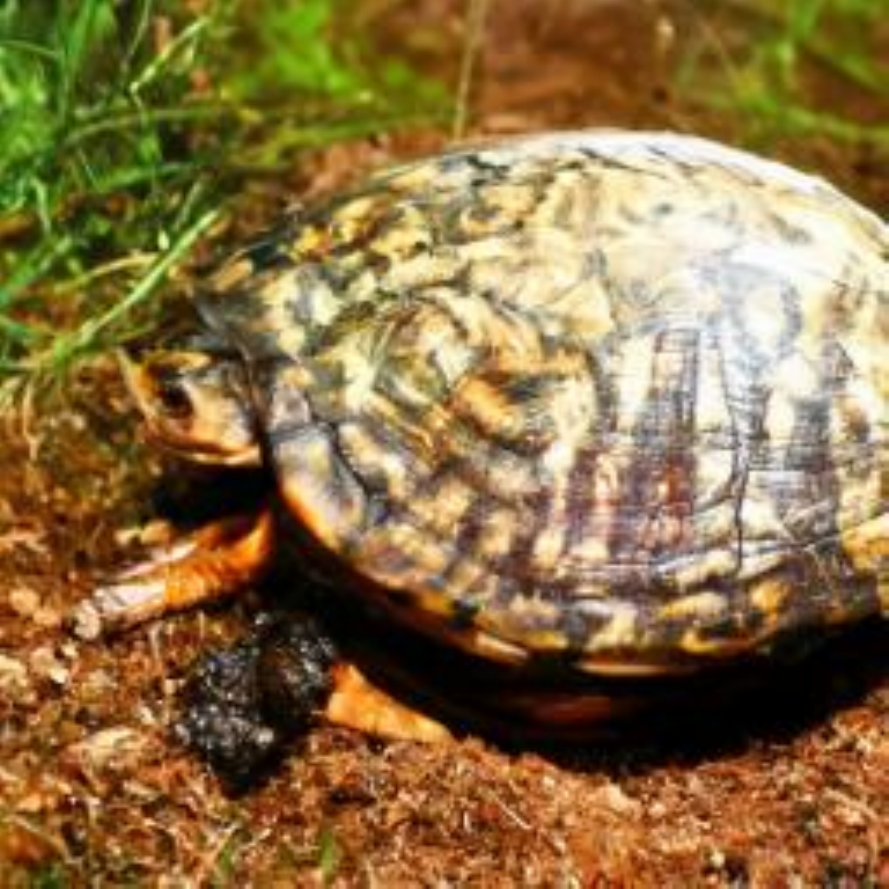} &
        \adjincludegraphics[clip,width=0.15\textwidth,trim={0 0 0 0}]{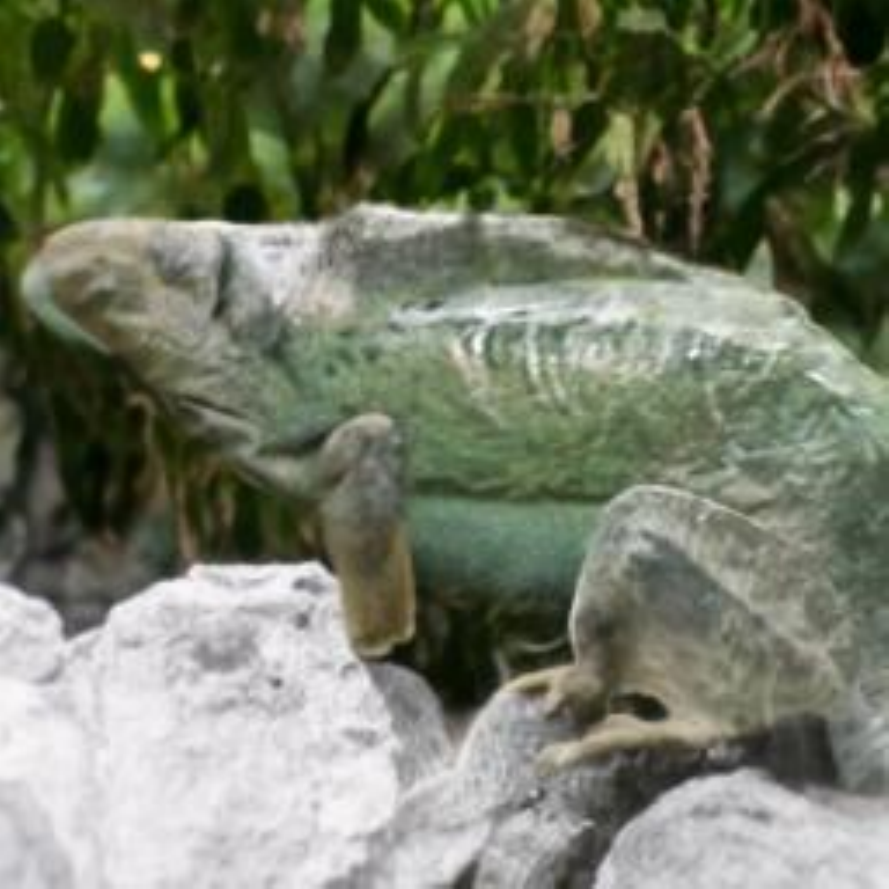} &
        \adjincludegraphics[clip,width=0.15\textwidth,trim={0 0 0 0}]{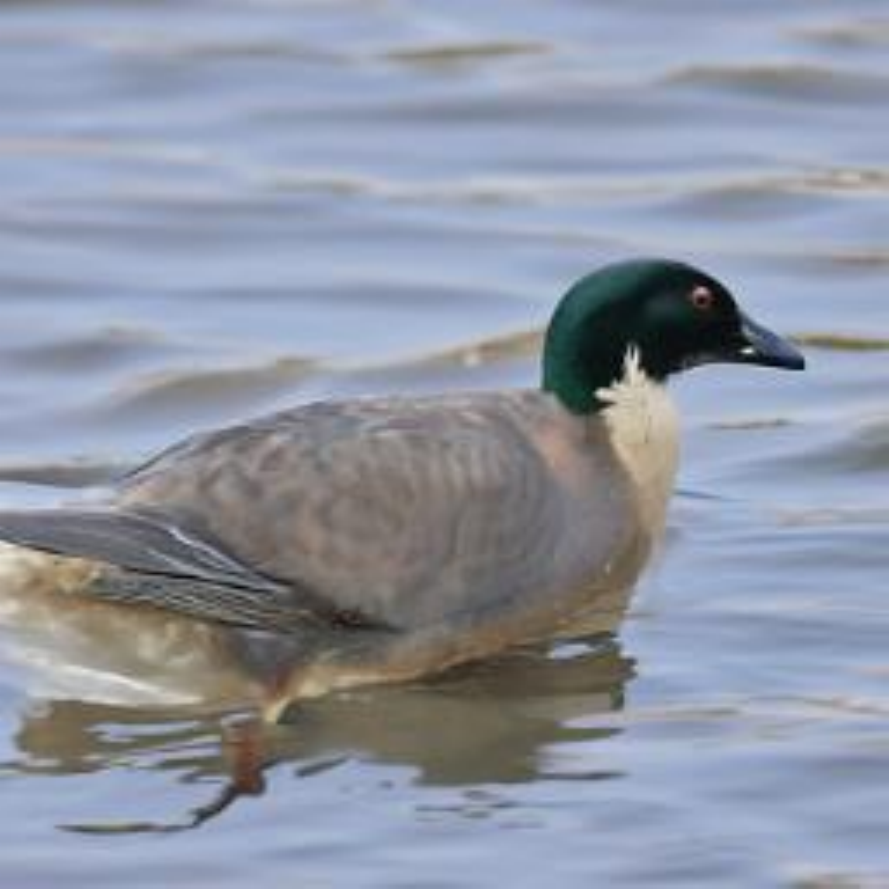} &
        \adjincludegraphics[clip,width=0.15\textwidth,trim={0 0 0 0}]{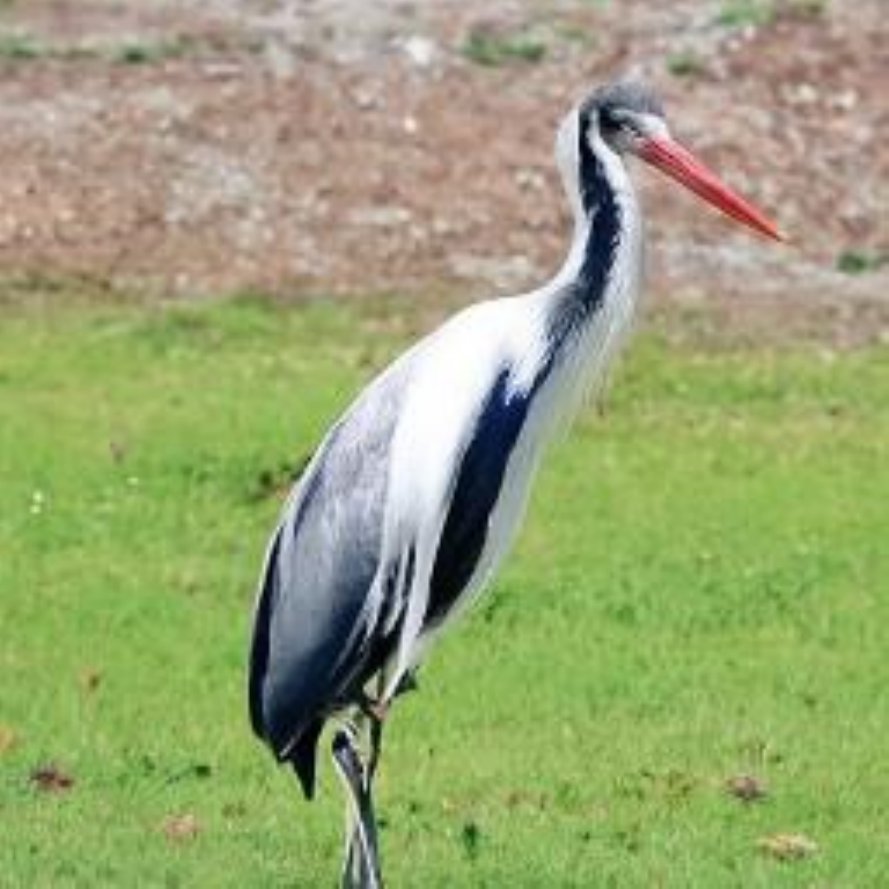} &
        \adjincludegraphics[clip,width=0.15\textwidth,trim={0 0 0 0}]{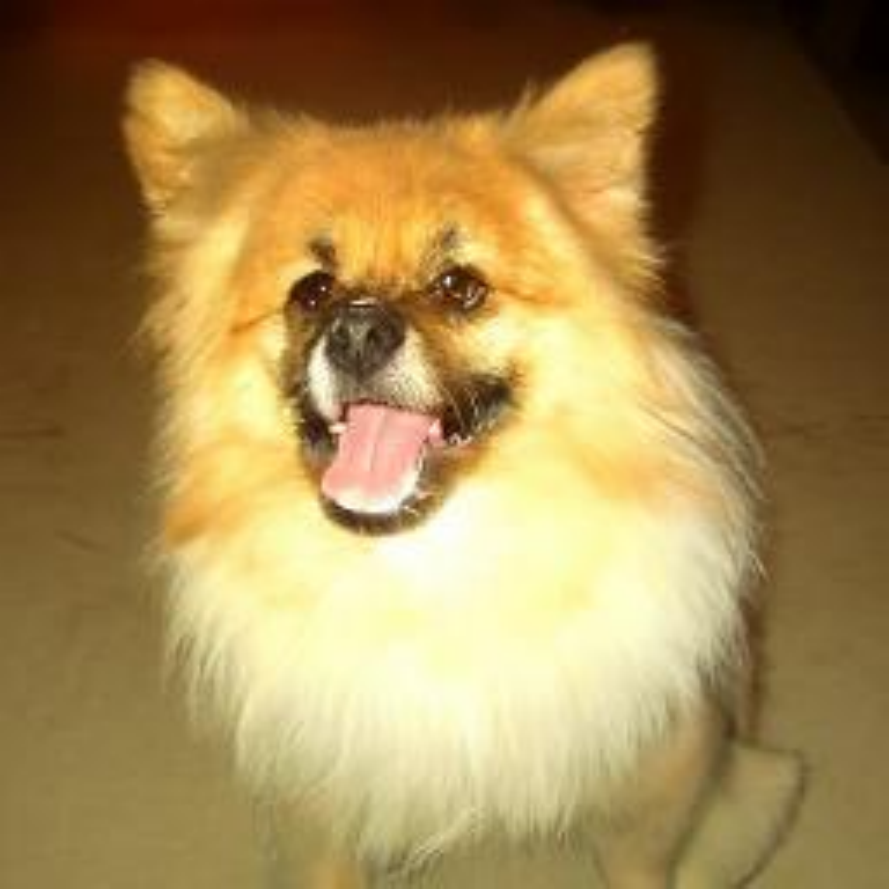} &
        \adjincludegraphics[clip,width=0.15\textwidth,trim={0 0 0 0}]{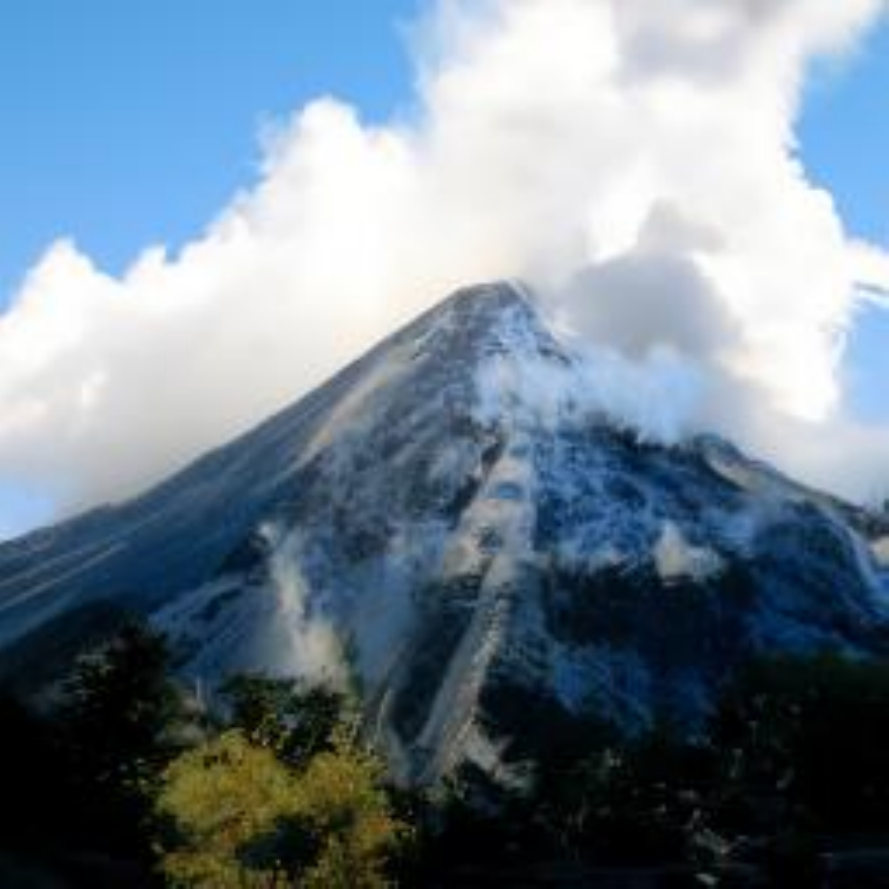} \\
        
        \footnotesize ~~~~~~Box turtle & \footnotesize  Iguana & \footnotesize  Goose & \footnotesize  Crane & \footnotesize Pomeranian & \footnotesize Volcano \\
    \end{tabular}
    \caption{
    Qualitative results on ImageNet class conditional generation with DDIM 25 steps by guiding unconditional ADM with DeiT-S.
    }
    \label{apx:fig_qualitative_deit_ddim}

\end{figure*}

Only qualitative results with DDPM 250 steps are presented in Section~\textcolor{red}{5.1}, because of limited spaces.
Fig.~\ref{apx:fig_qualitative_deit_ddpm} and Fig.~\ref{apx:fig_qualitative_deit_ddim} show qualitative results with DDIM 250 steps and DDIM 25 steps, respectively.

\section{More Qualitative Results on GLIDE}
\label{apx:sec:glide-more-qual}
We note that the reference batch for measuring quantitative results such as FID, IS, Precision, and Recall is not valid since data for training GLIDE is not publicly available.
Instead, we present various qualitative results to show that PPAP can guide GLIDE.

\subsection{GLIDE + ResNet-50}
\label{apx:subsec:glide+res}
\begin{figure*}[t]
    \centering
    \begin{tabular}{@{}c@{\hspace{0.5mm}}c@{\hspace{0.5mm}}c@{\hspace{0.5mm}}c@{\hspace{0.5mm}}c@{\hspace{0.5mm}}c@{\hspace{0.5mm}}c@{\hspace{0.5mm}}c}
        \raisebox{0.4\height}{\rotatebox{90}{\scriptsize PPAP (Ours)}} 
        \adjincludegraphics[clip,width=0.115\textwidth,trim={0 0 0 0}]{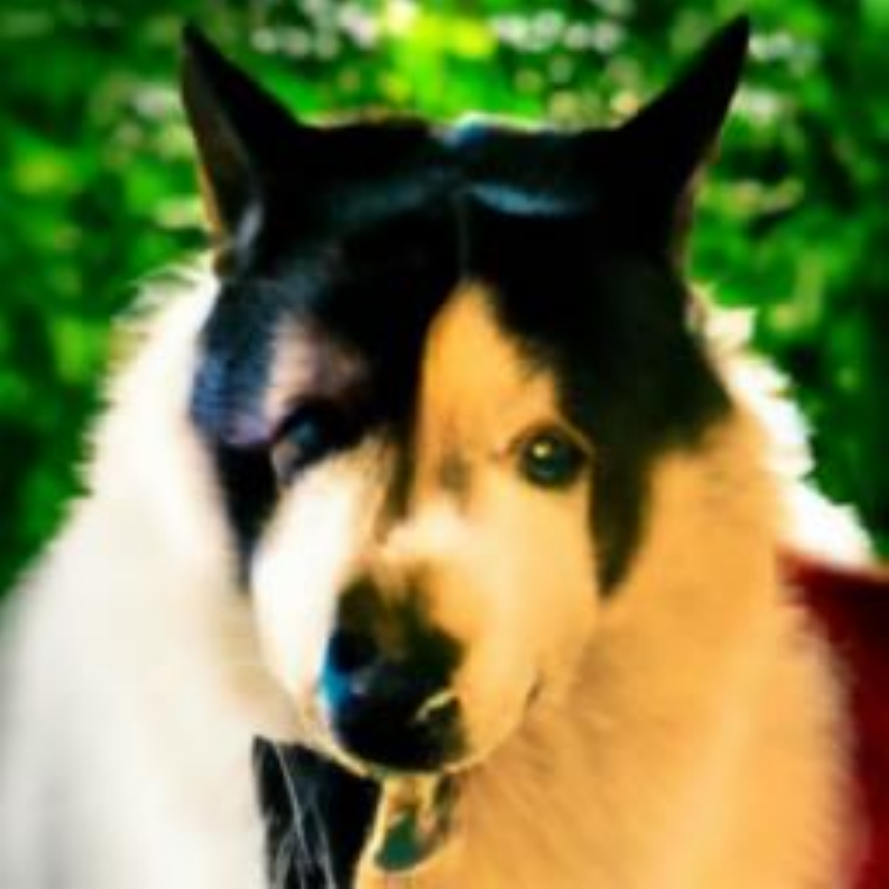} &
        \adjincludegraphics[clip,width=0.115\textwidth,trim={0 0 0 0}]{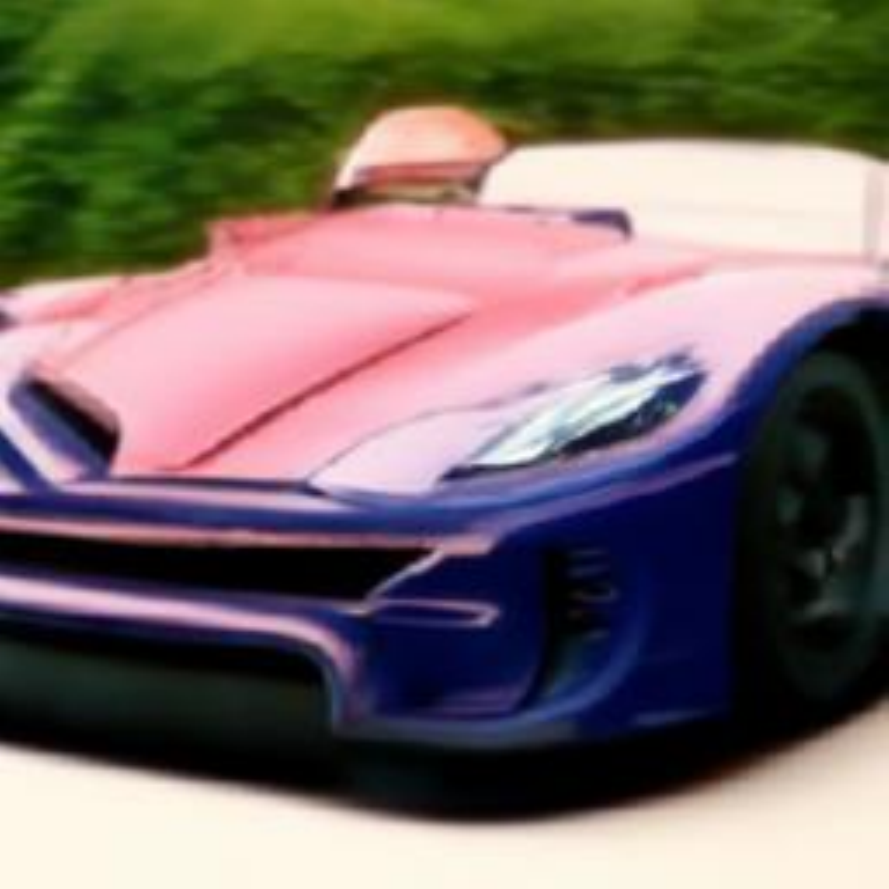} &
        \adjincludegraphics[clip,width=0.115\textwidth,trim={0 0 0 0}]{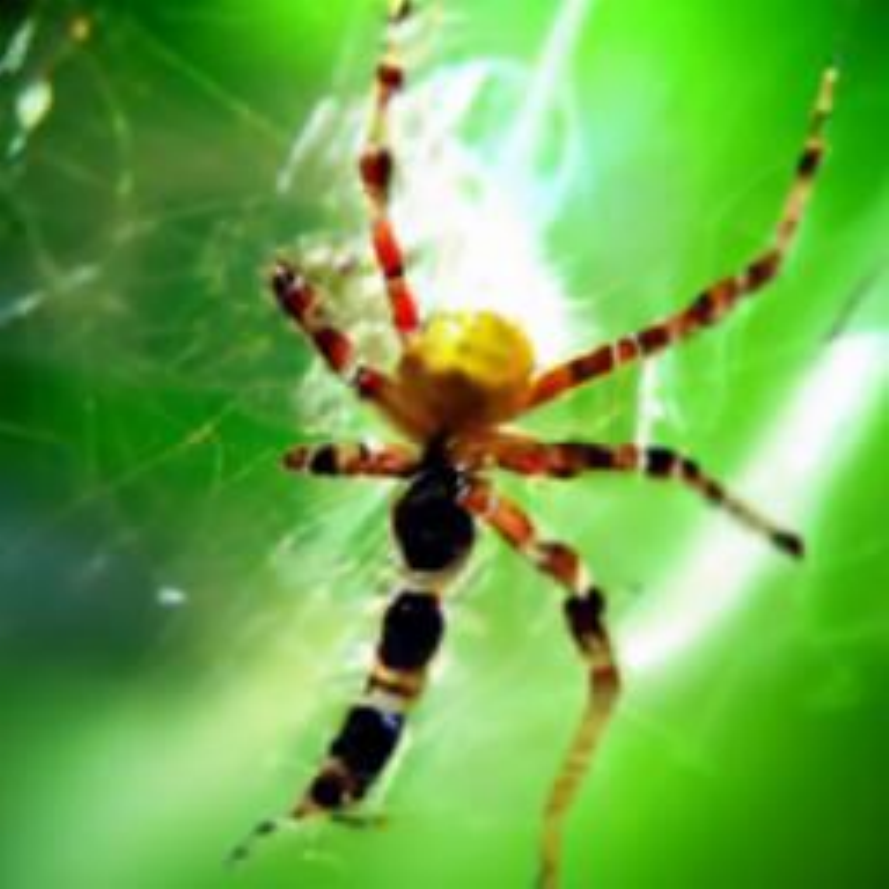} &
        \adjincludegraphics[clip,width=0.115\textwidth,trim={0 0 0 0}]{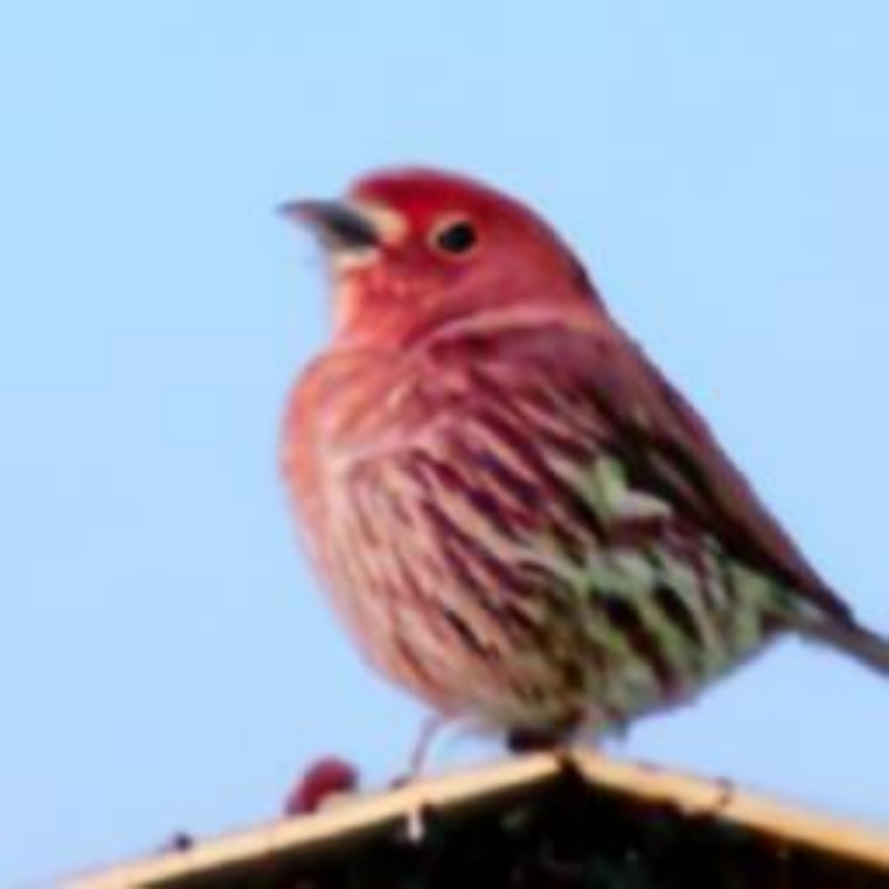} &
        \adjincludegraphics[clip,width=0.115\textwidth,trim={0 0 0 0}]{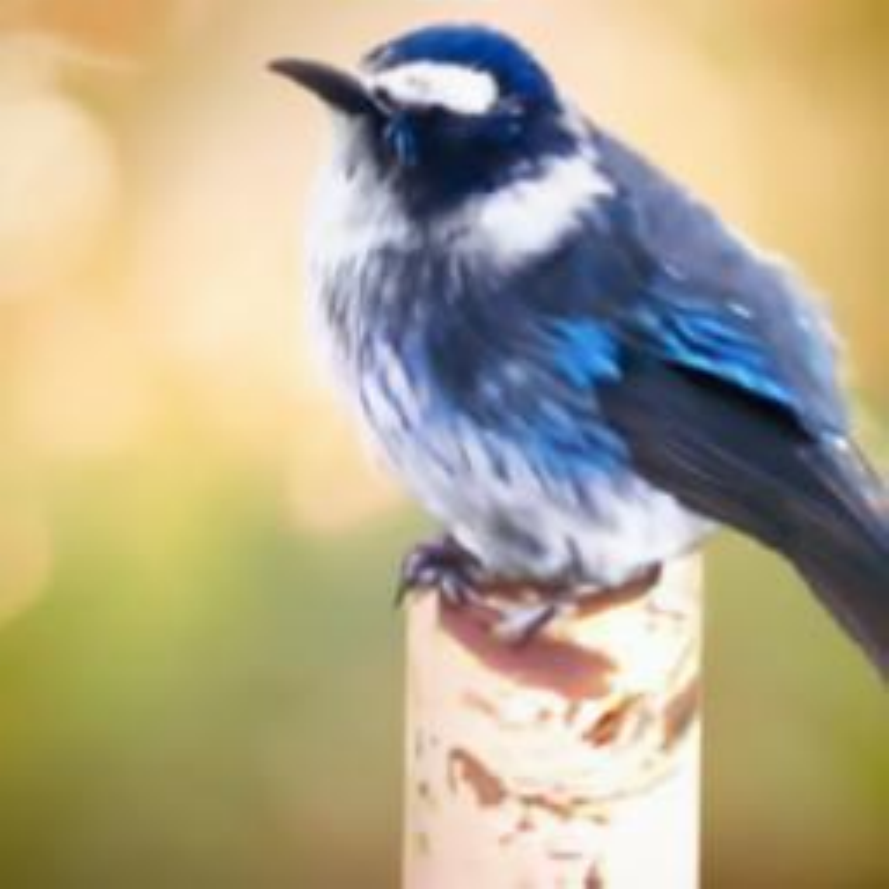} &
        \adjincludegraphics[clip,width=0.115\textwidth,trim={0 0 0 0}]{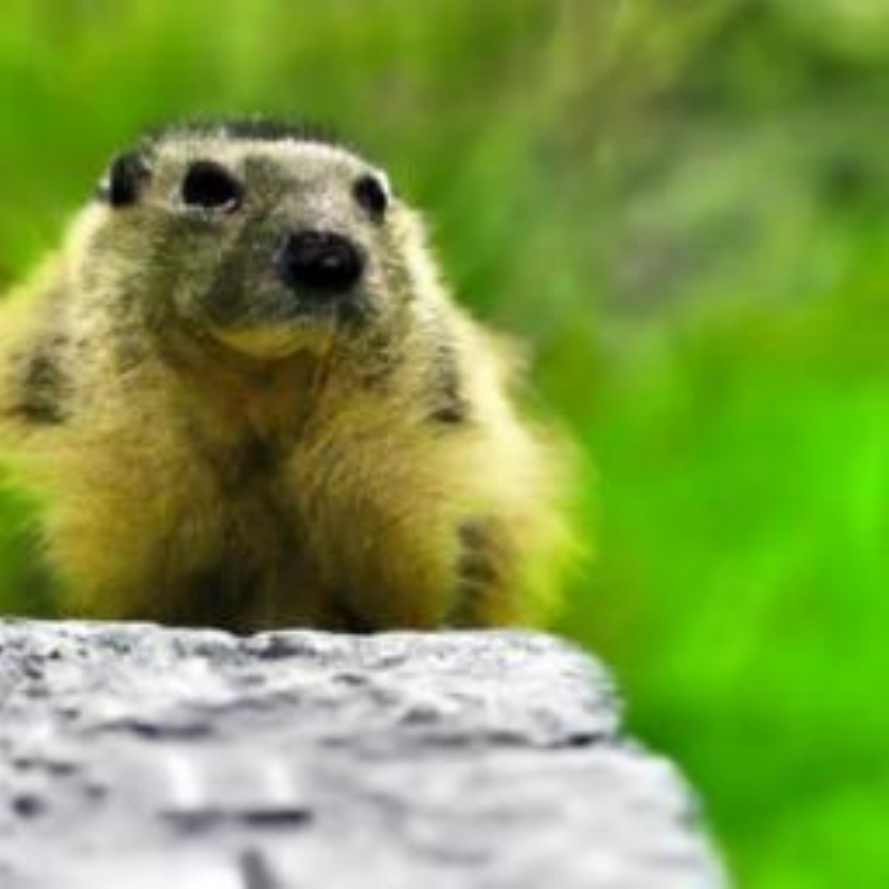} &
        \adjincludegraphics[clip,width=0.115\textwidth,trim={0 0 0 0}]{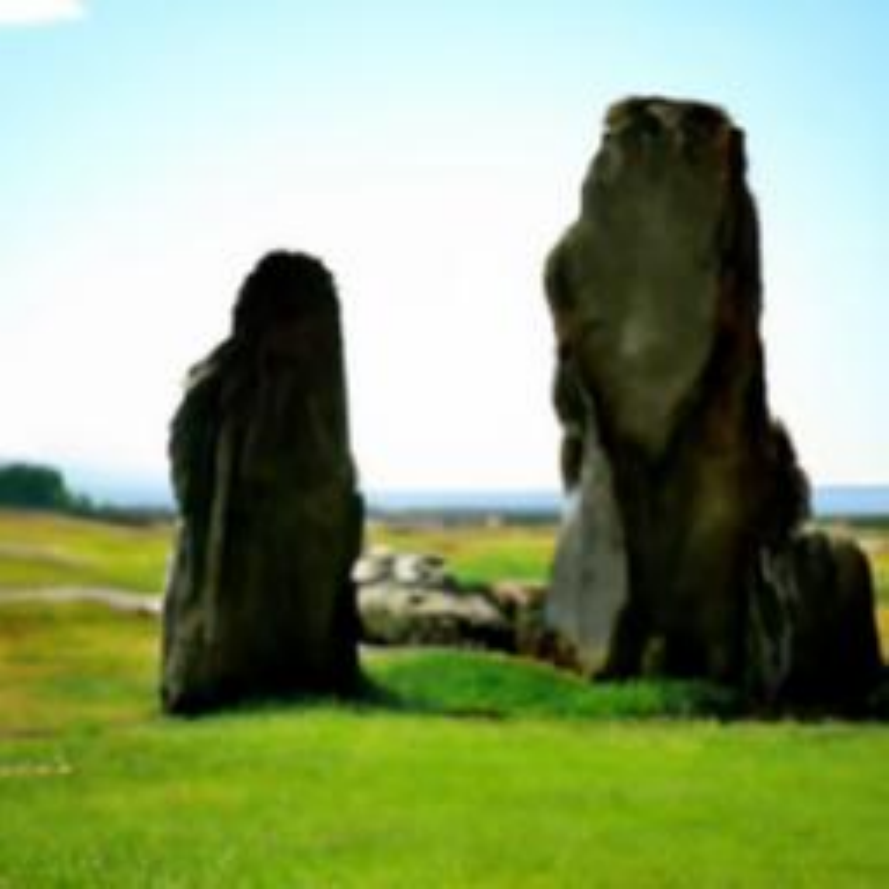} &
        \adjincludegraphics[clip,width=0.115\textwidth,trim={0 0 0 0}]{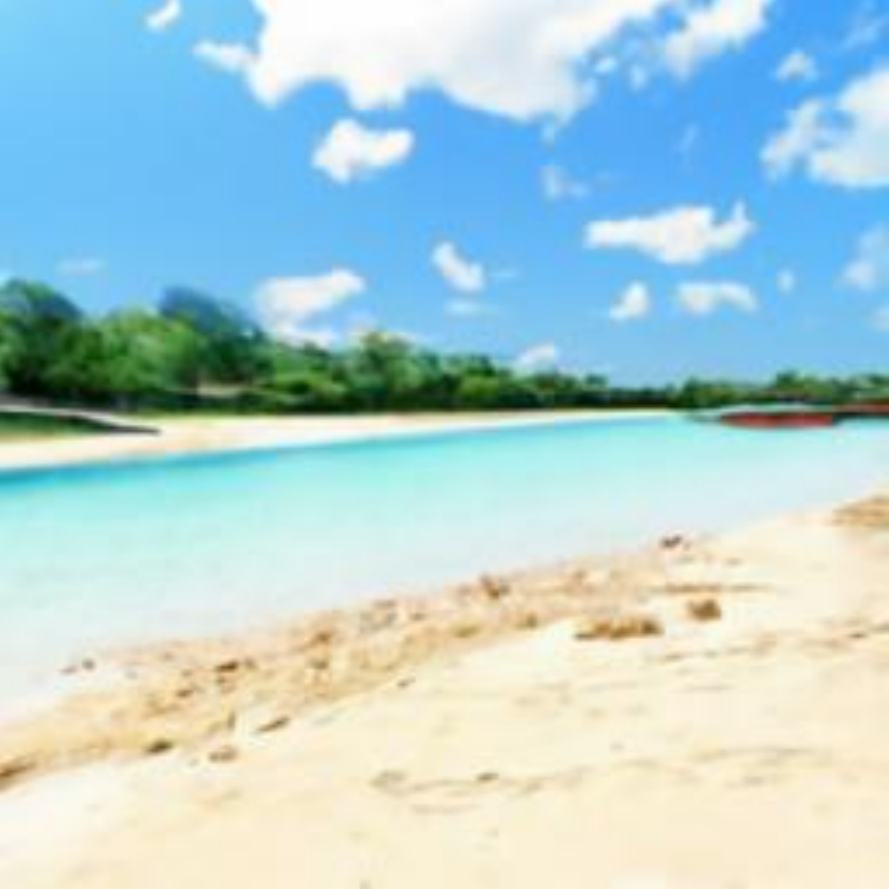} \\
        
        \raisebox{1.3\height}{\rotatebox{90}{\scriptsize Na\"ive}} 
        \adjincludegraphics[clip,width=0.115\textwidth,trim={0 0 0 0}]{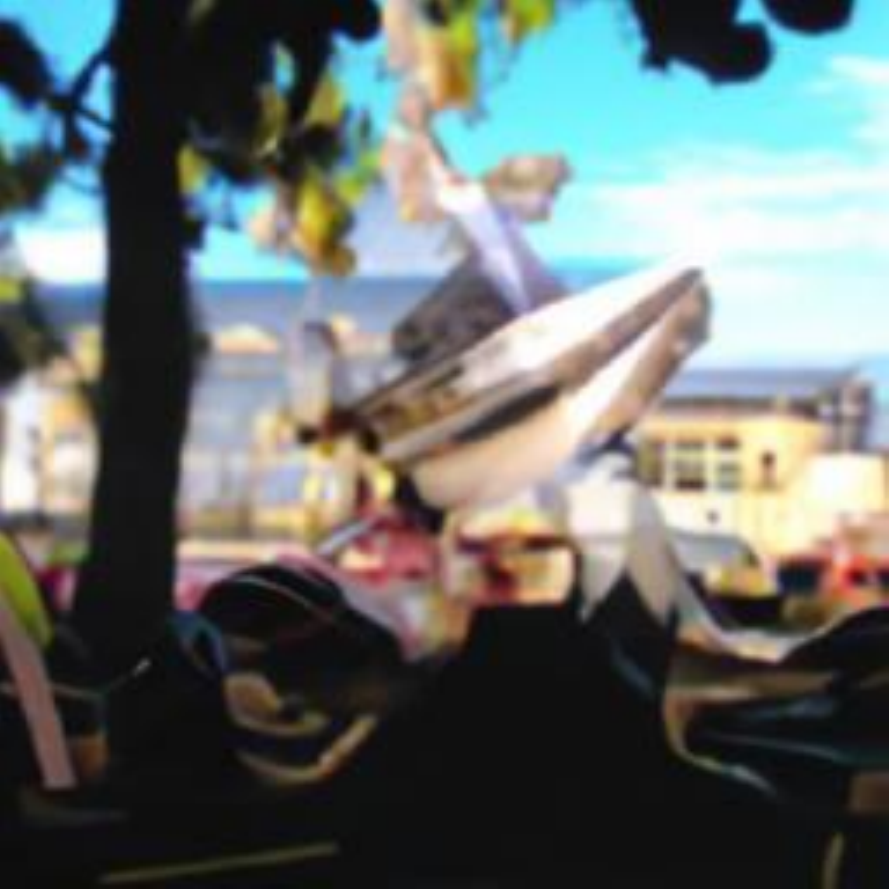} &
        \adjincludegraphics[clip,width=0.115\textwidth,trim={0 0 0 0}]{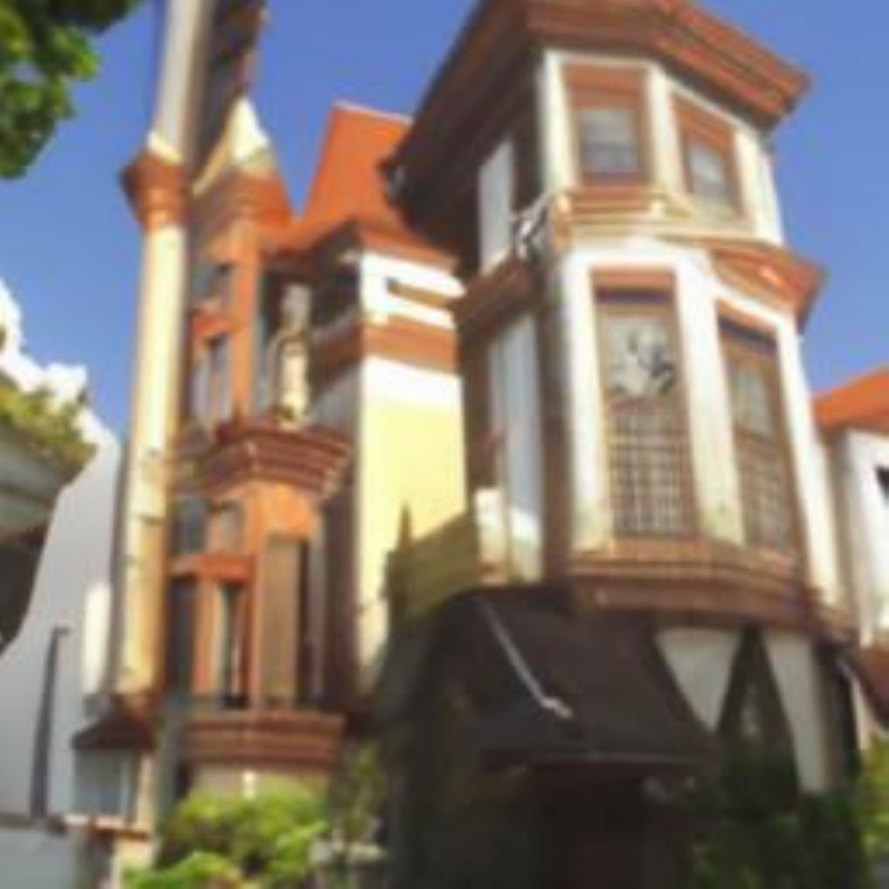} &
        \adjincludegraphics[clip,width=0.115\textwidth,trim={0 0 0 0}]{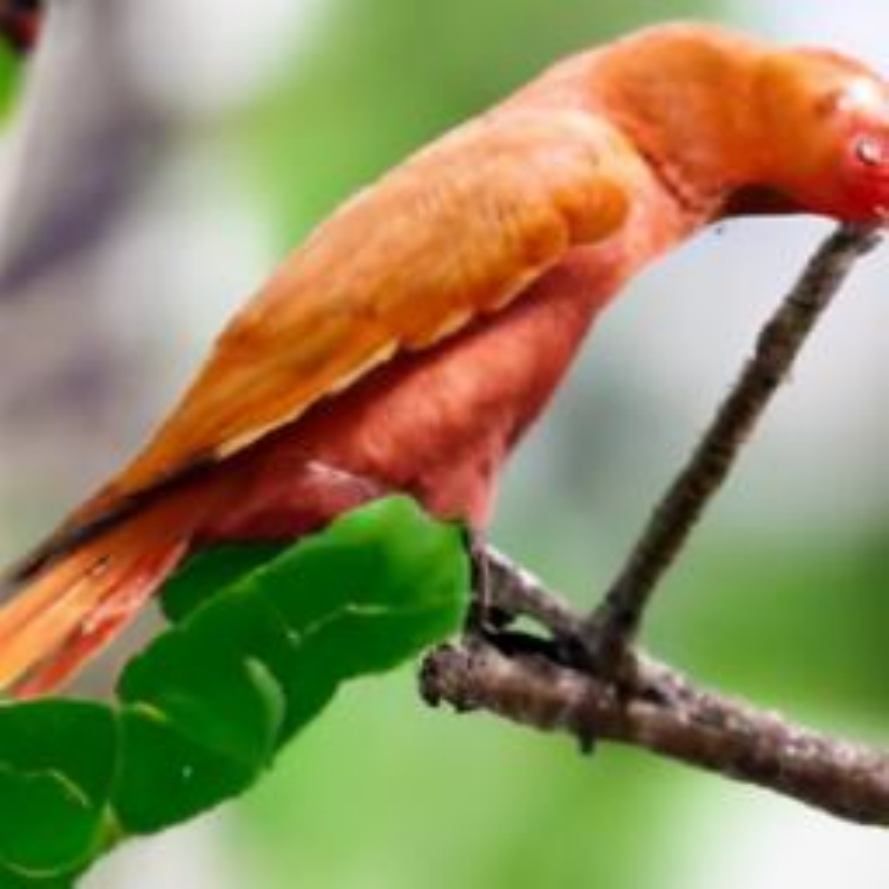} &
        \adjincludegraphics[clip,width=0.115\textwidth,trim={0 0 0 0}]{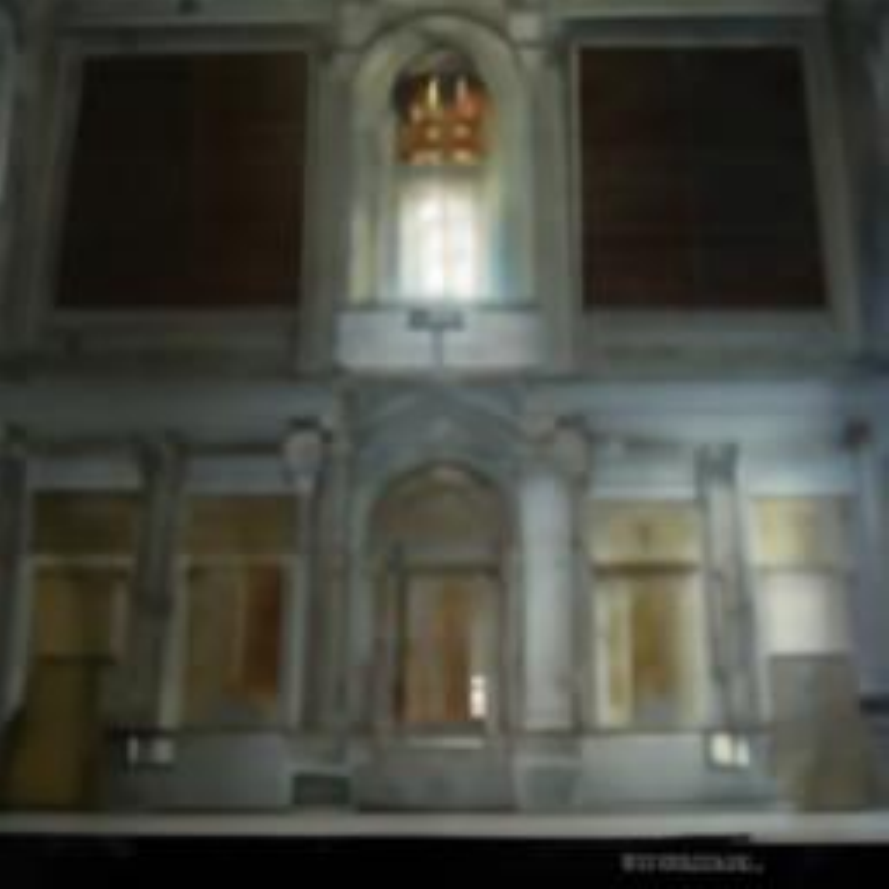} &
        \adjincludegraphics[clip,width=0.115\textwidth,trim={0 0 0 0}]{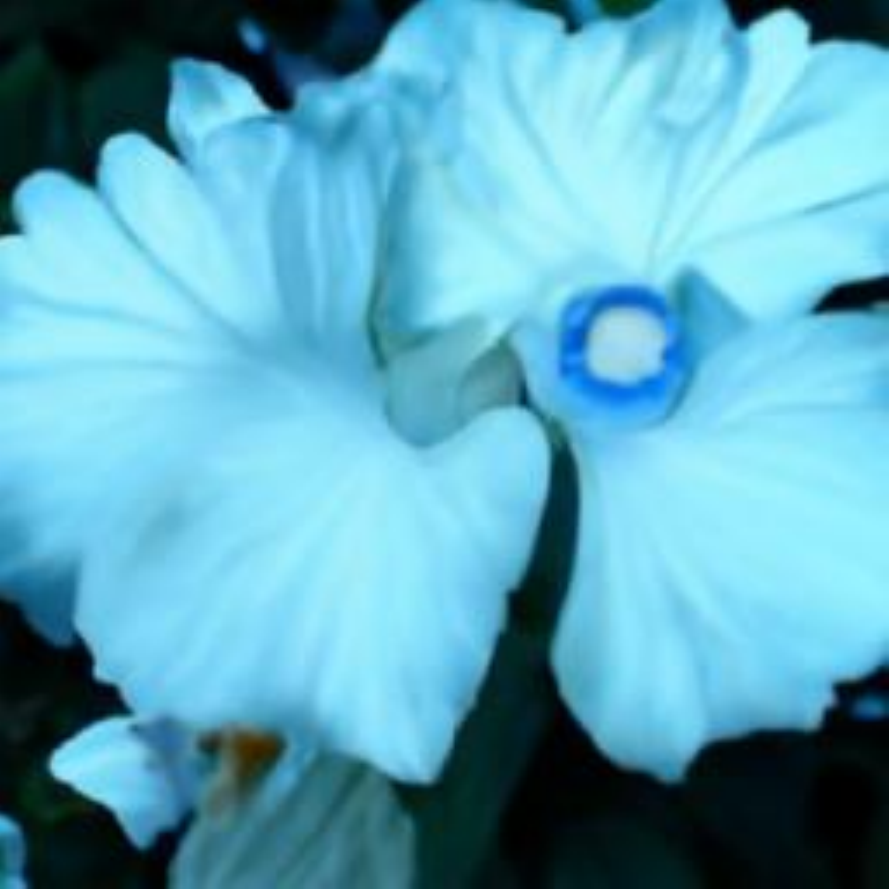} &
        \adjincludegraphics[clip,width=0.115\textwidth,trim={0 0 0 0}]{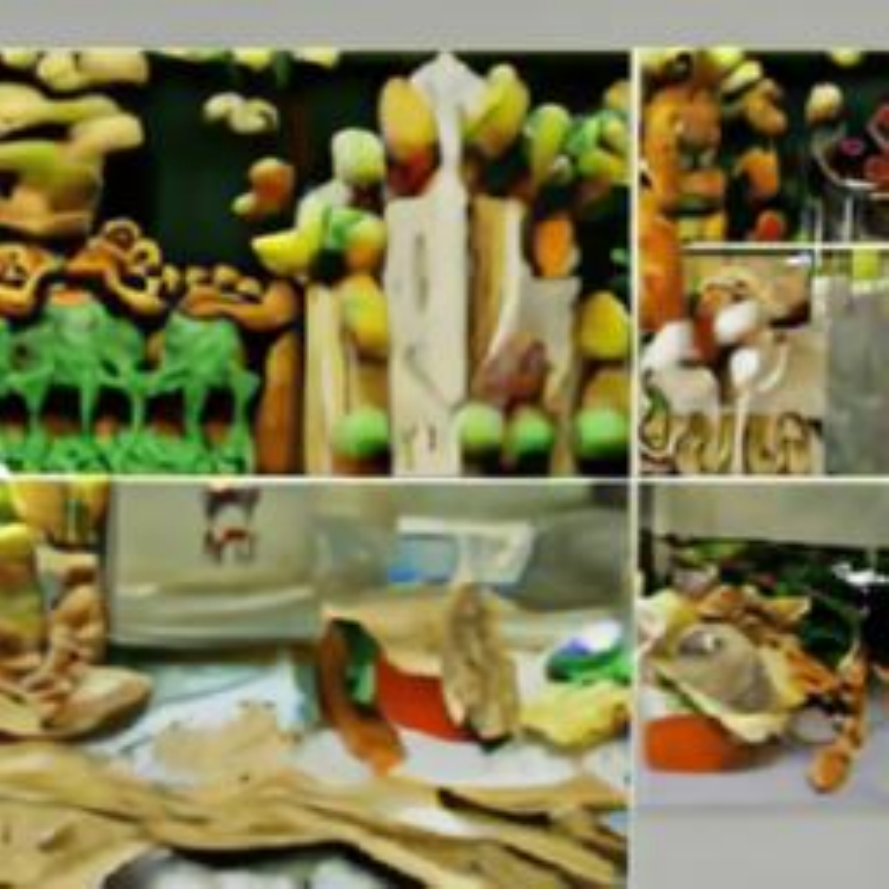} &
        \adjincludegraphics[clip,width=0.115\textwidth,trim={0 0 0 0}]{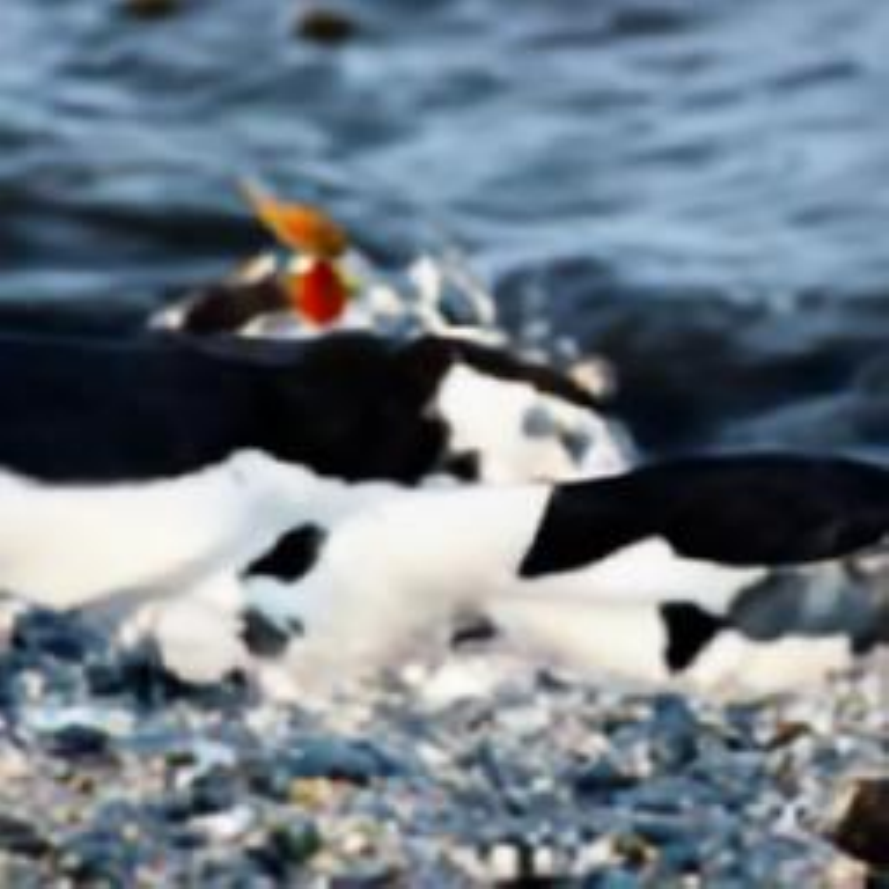} &
        \adjincludegraphics[clip,width=0.115\textwidth,trim={0 0 0 0}]{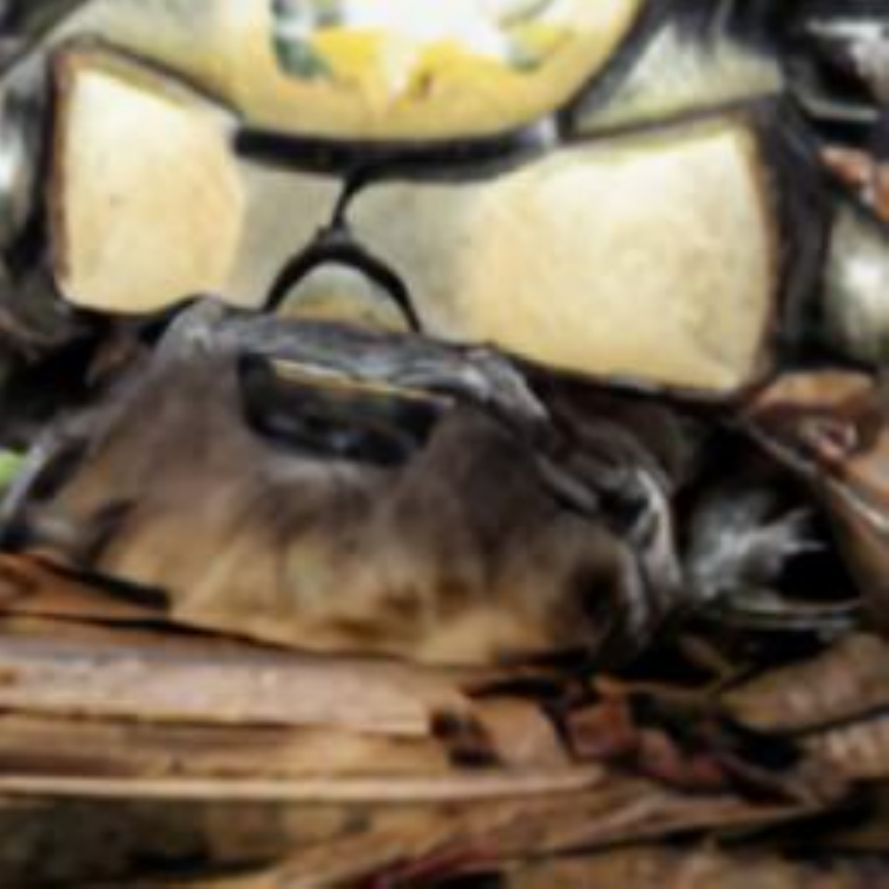} \\
        \footnotesize ~~~~Collie & \footnotesize  Convertible & \footnotesize  Garden spider & \footnotesize  House finch & \footnotesize Jay & \footnotesize Marmot & \footnotesize Megalith & \footnotesize Sandbar
    \end{tabular}
    \caption{Generated images by guiding GLIDE with ResNet50 classifier. Our framework PPAP-5 succeeds in the guidance of diffusion, but the na\"ive off-the-shelf model fails.}
    \vspace{-1mm}
    \label{fig:ap_glide_resnet_more}
\end{figure*}

\begin{figure*}[t]

    \centering
    \begin{tabular}{@{\hspace{0.5mm}}c@{\hspace{0.5mm}}c@{\hspace{0.5mm}}c@{\hspace{0.5mm}}c|c@{\hspace{0.5mm}}c@{\hspace{0.5mm}}c@{\hspace{0.5mm}}c}
        \raisebox{0.55\height}{\rotatebox{90}{\scriptsize Coral reef}} 
        \adjincludegraphics[clip,width=0.112\textwidth,trim={0 0 0 0}]{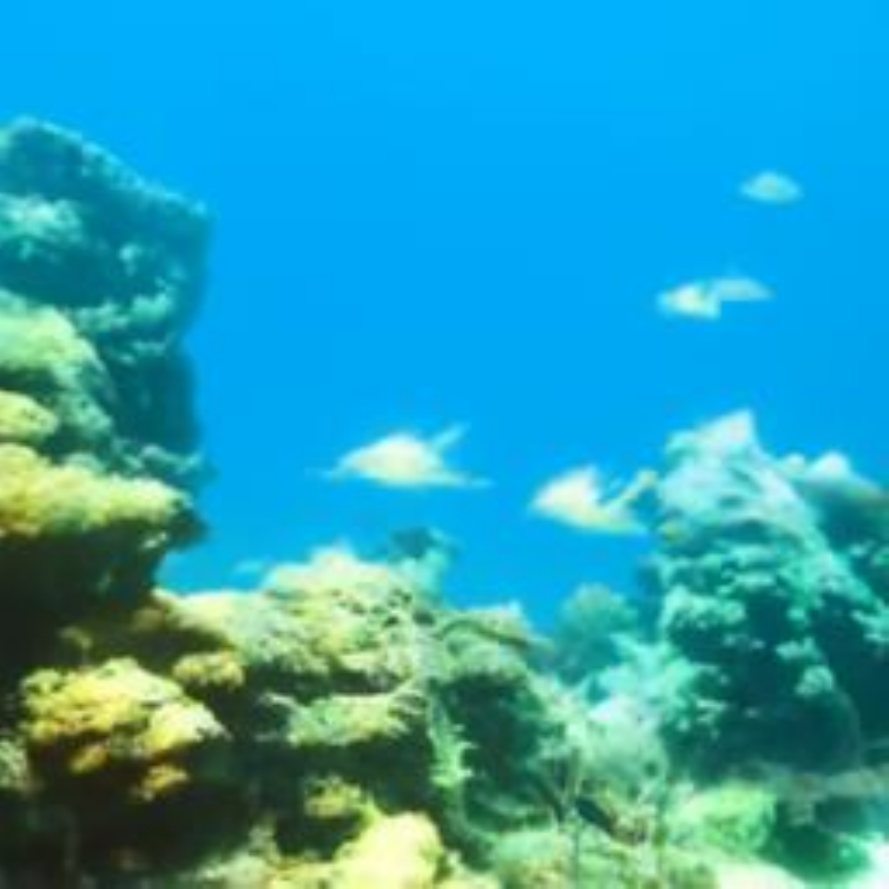} &
        \adjincludegraphics[clip,width=0.112\textwidth,trim={0 0 0 0}]{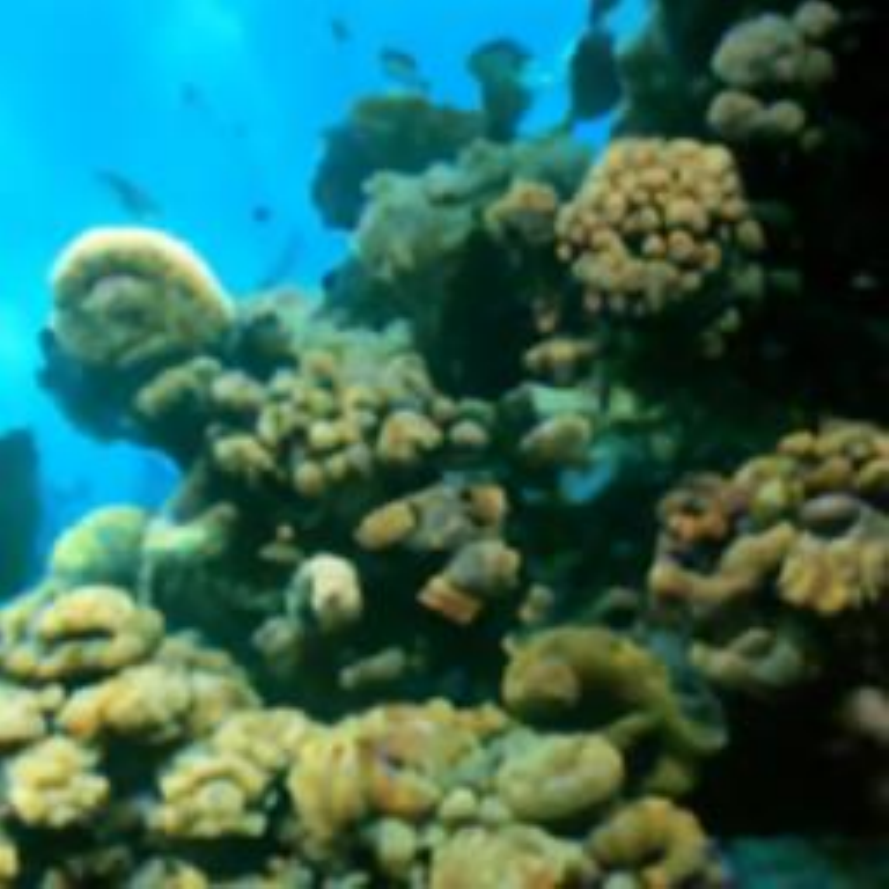} &
        \adjincludegraphics[clip,width=0.112\textwidth,trim={0 0 0 0}]{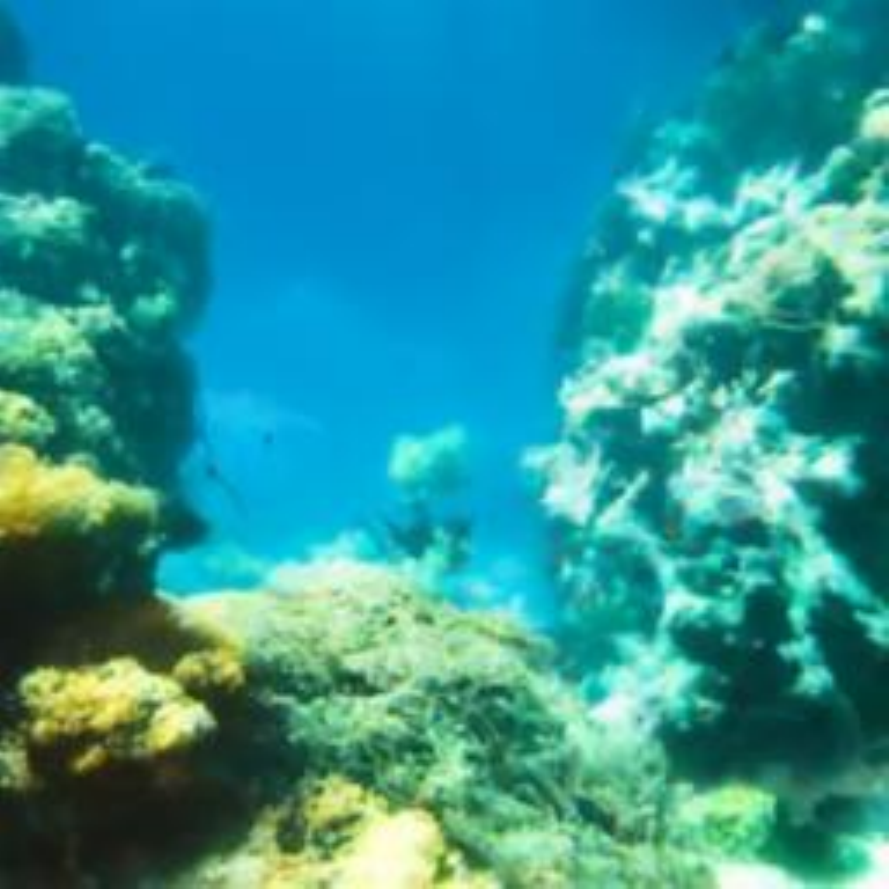} & 
        \adjincludegraphics[clip,width=0.112\textwidth,trim={0 0 0 0}]{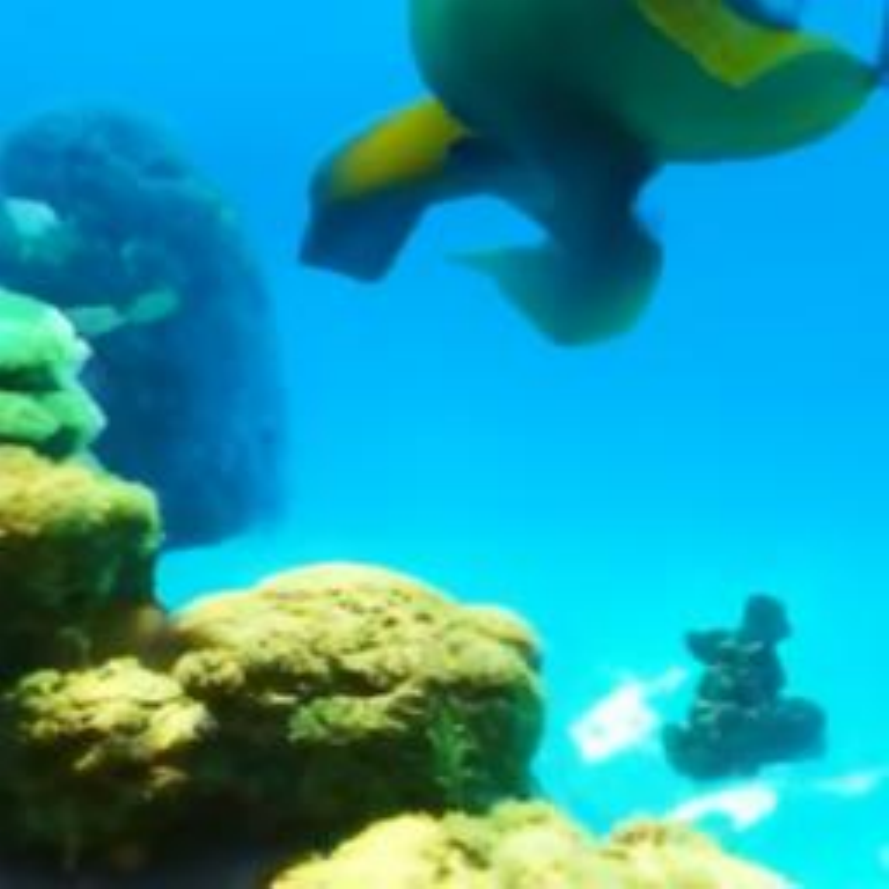} &
        \adjincludegraphics[clip,width=0.112\textwidth,trim={0 0 0 0}]{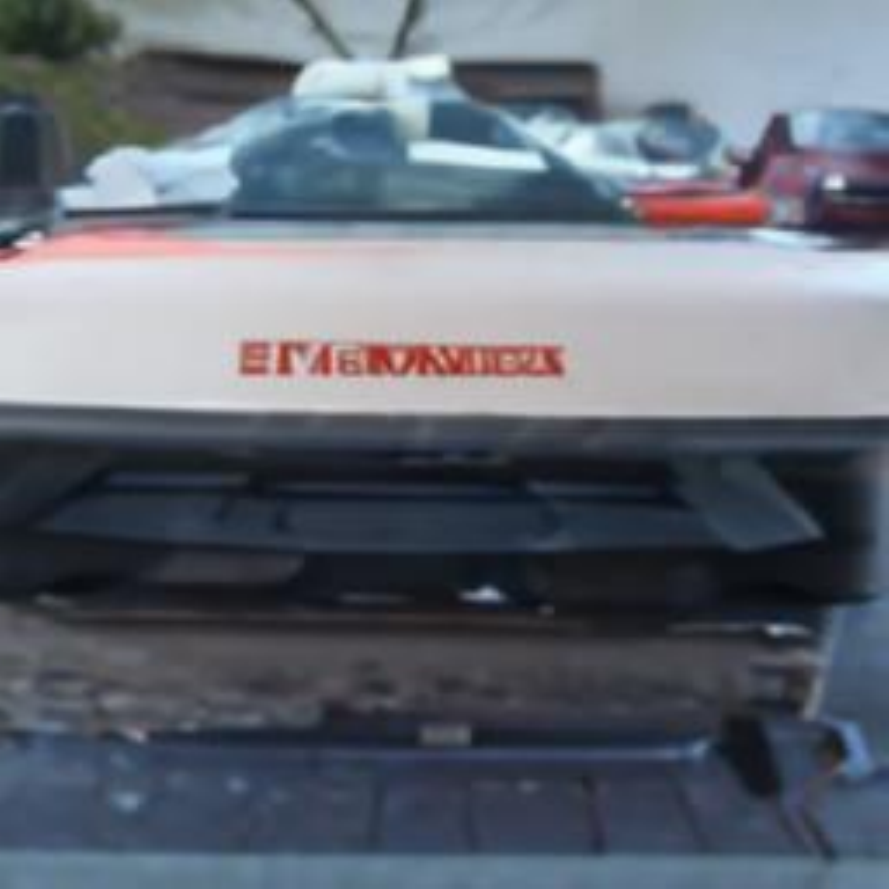} &
        \adjincludegraphics[clip,width=0.112\textwidth,trim={0 0 0 0}]{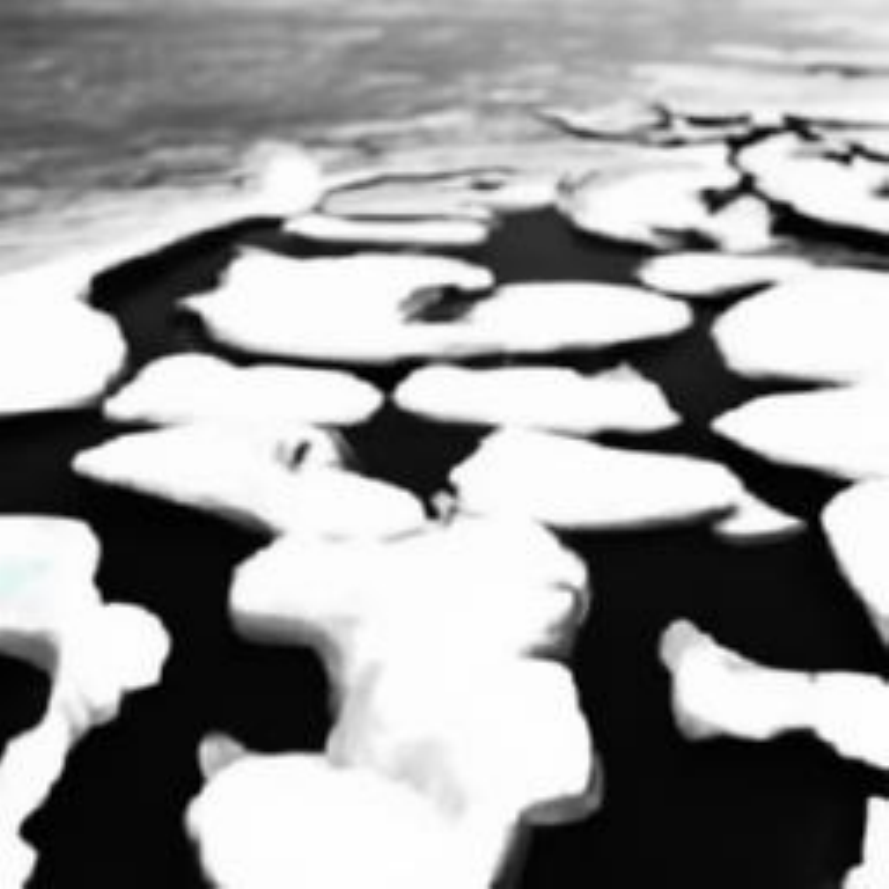} &
        \adjincludegraphics[clip,width=0.112\textwidth,trim={0 0 0 0}]{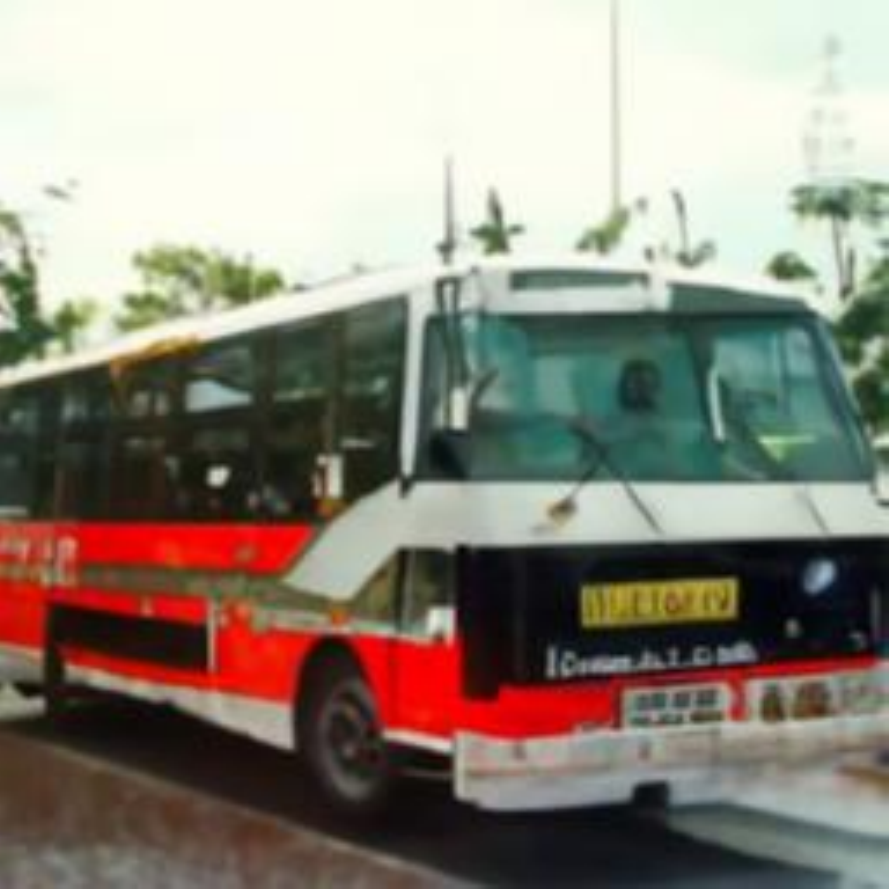} &
        \adjincludegraphics[clip,width=0.112\textwidth,trim={0 0 0 0}]{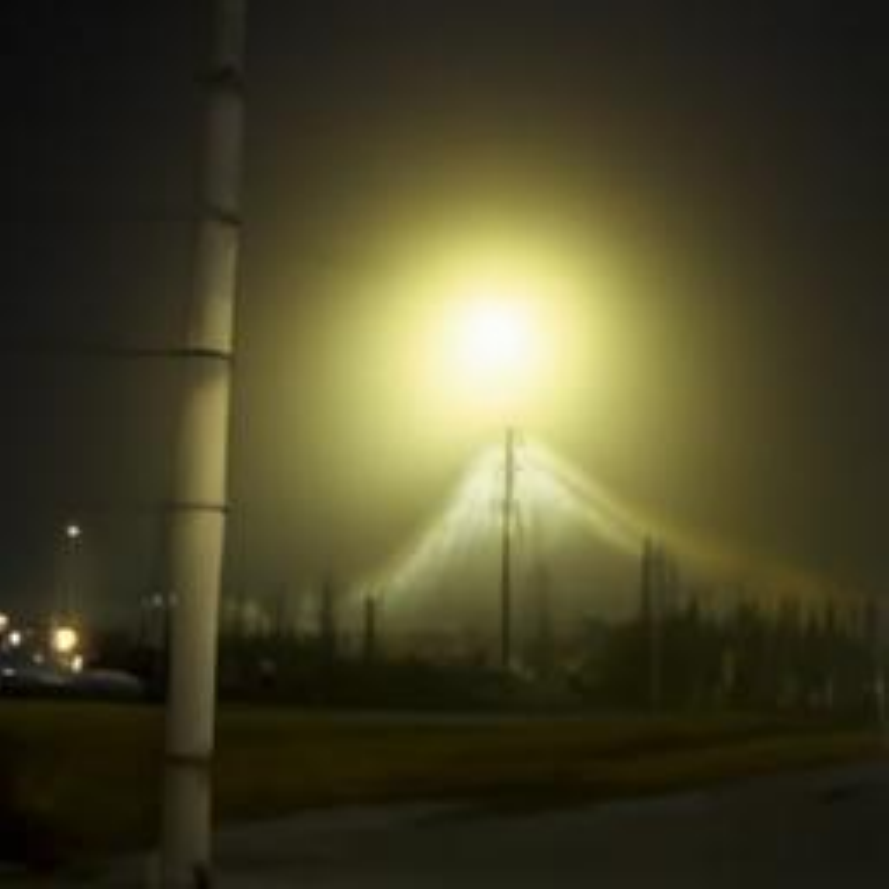} \\

        \raisebox{0.2\height}{\rotatebox{90}{\scriptsize Golden retriever}} 
        \adjincludegraphics[clip,width=0.112\textwidth,trim={0 0 0 0}]{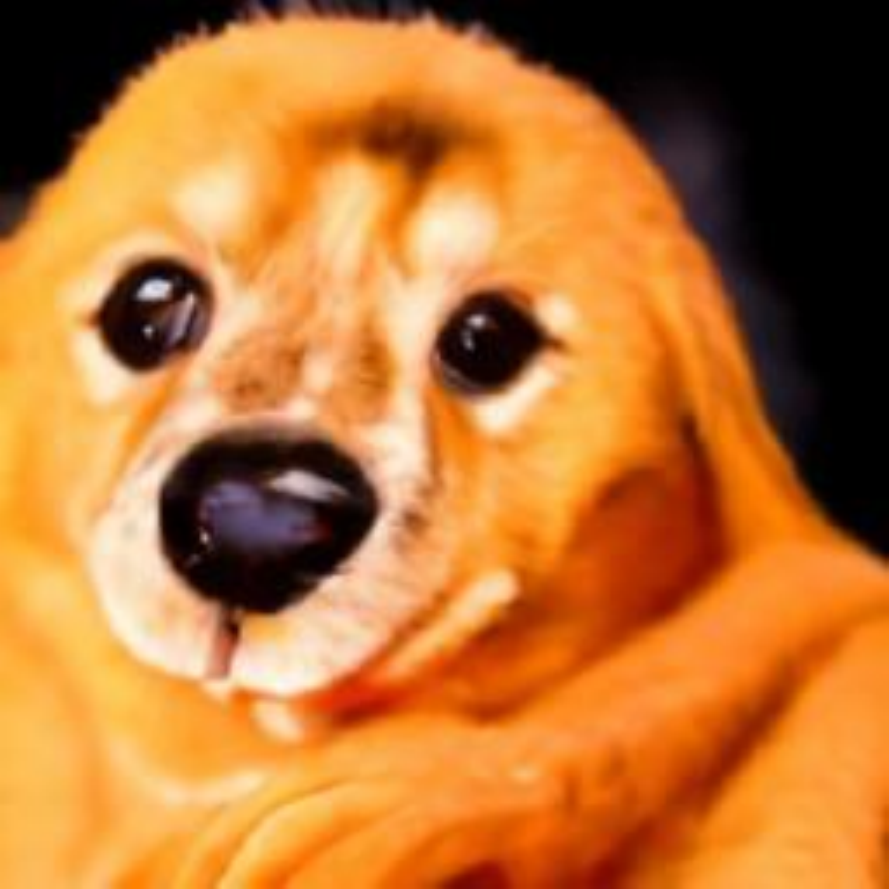} &
        \adjincludegraphics[clip,width=0.112\textwidth,trim={0 0 0 0}]{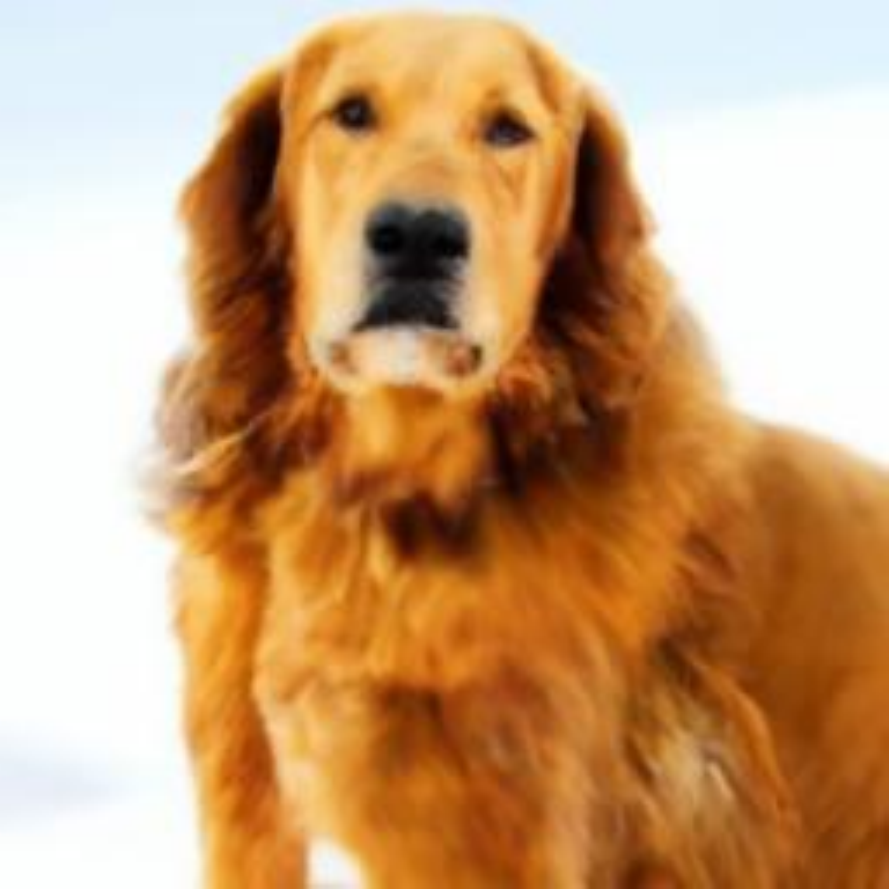} &
        \adjincludegraphics[clip,width=0.112\textwidth,trim={0 0 0 0}]{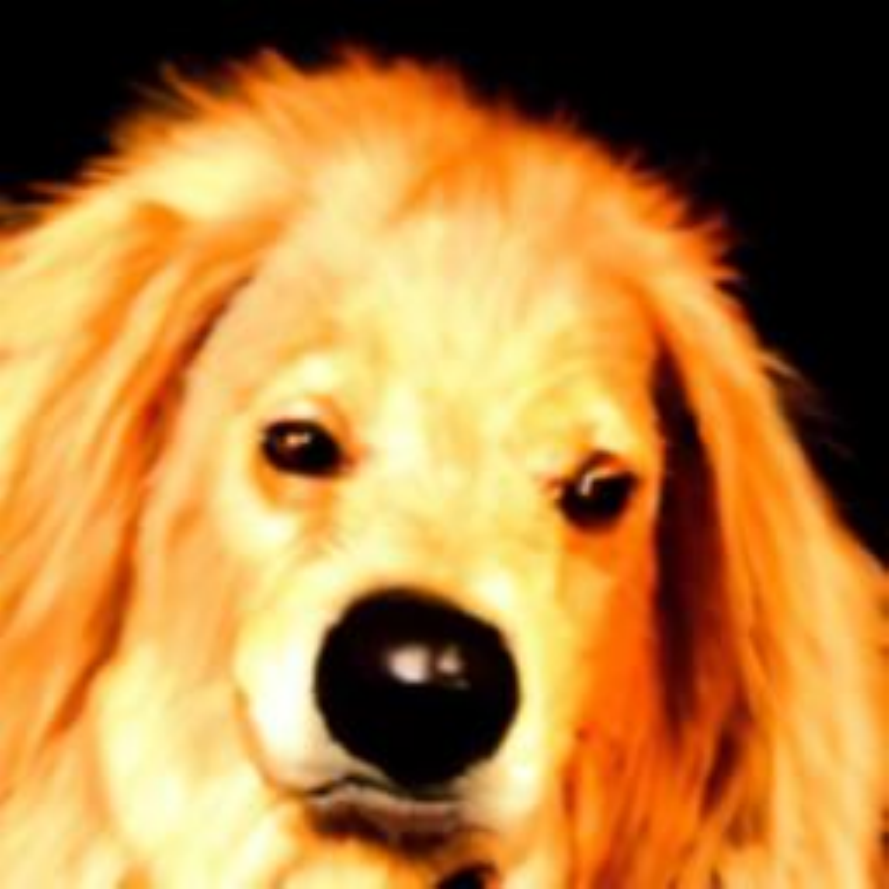} & 
        \adjincludegraphics[clip,width=0.112\textwidth,trim={0 0 0 0}]{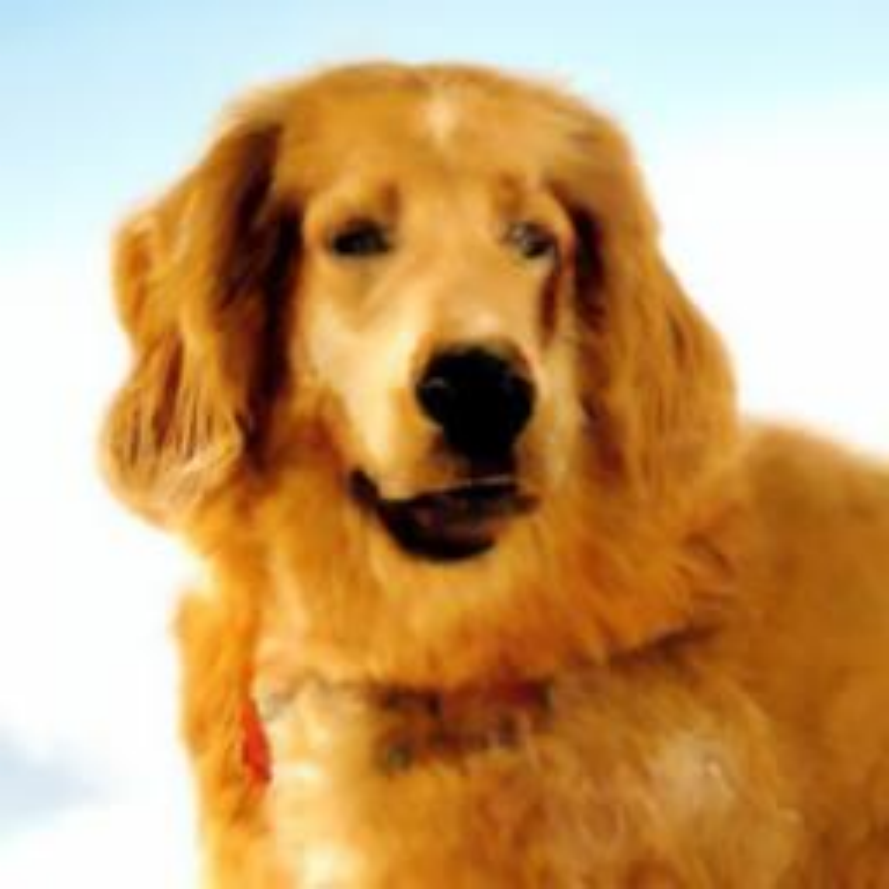} &
        \adjincludegraphics[clip,width=0.112\textwidth,trim={0 0 0 0}]{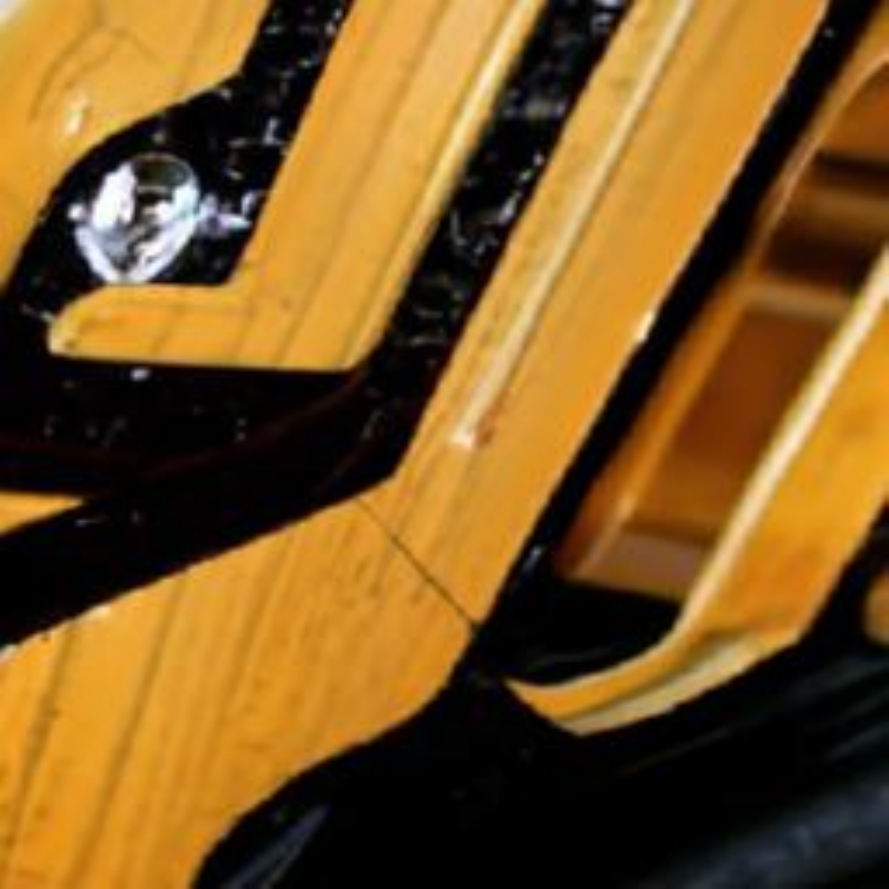} &
        \adjincludegraphics[clip,width=0.112\textwidth,trim={0 0 0 0}]{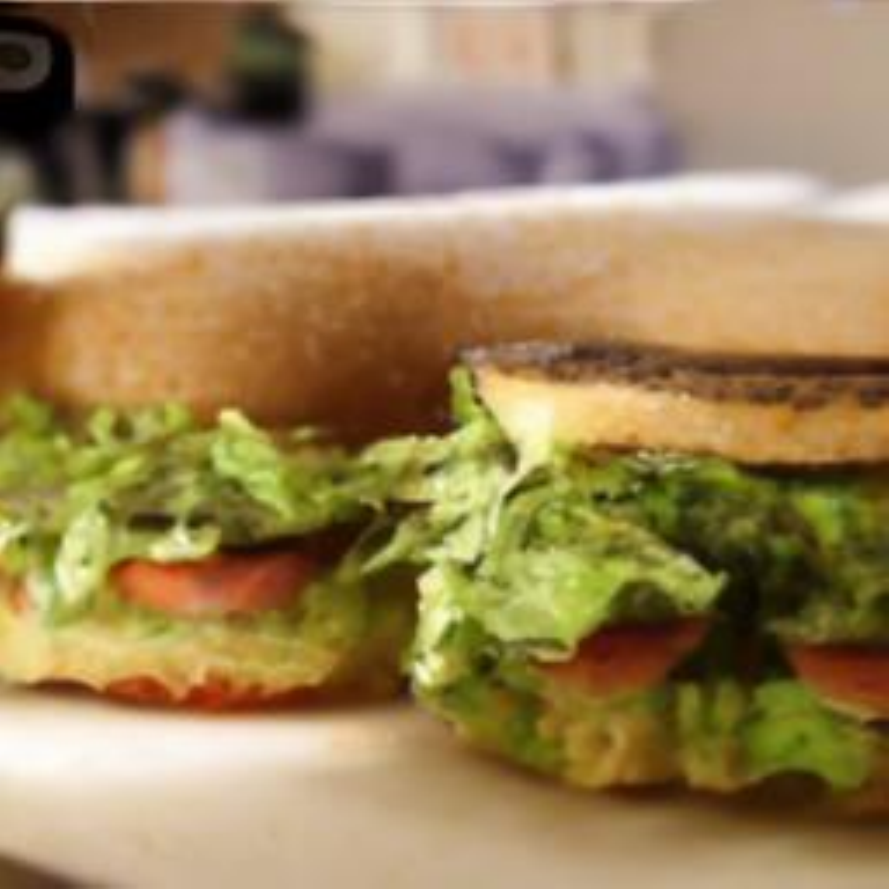} &
        \adjincludegraphics[clip,width=0.112\textwidth,trim={0 0 0 0}]{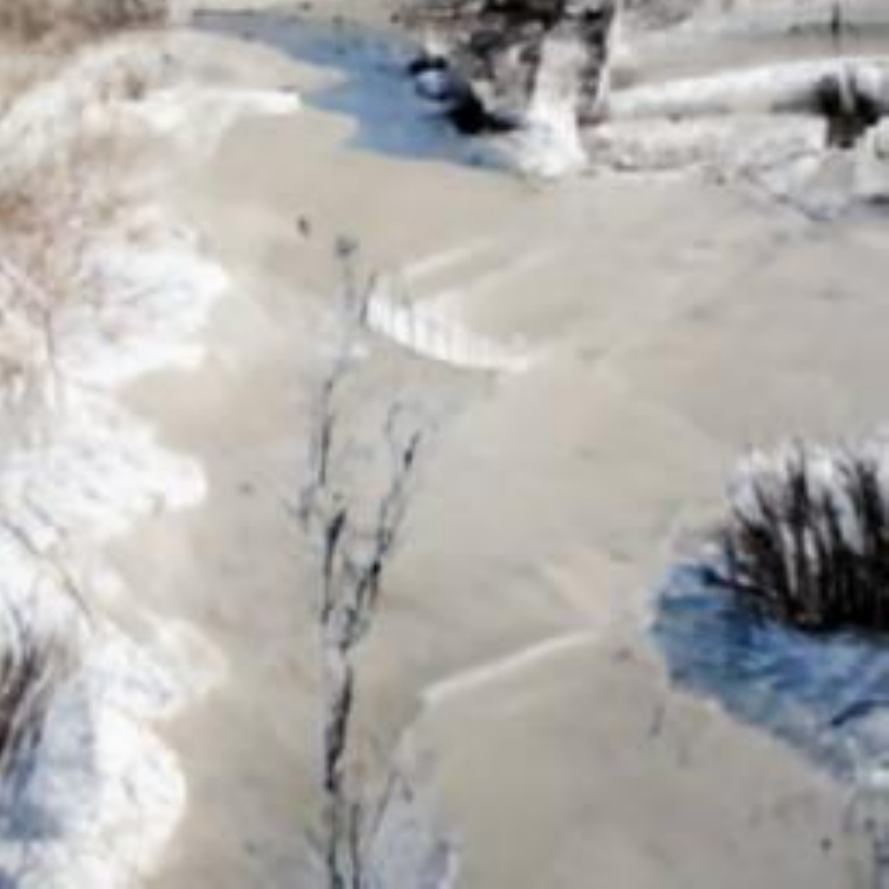} &
        \adjincludegraphics[clip,width=0.112\textwidth,trim={0 0 0 0}]{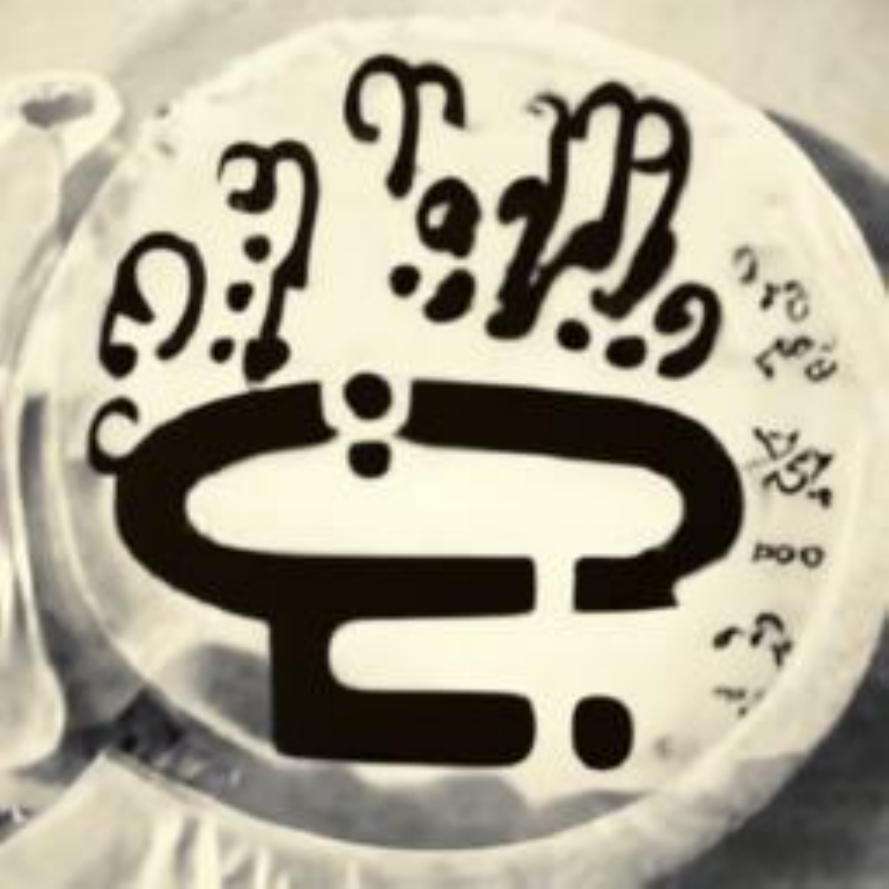} \\

        \raisebox{0.8\height}{\rotatebox{90}{\scriptsize Obelisk}} 
        \adjincludegraphics[clip,width=0.112\textwidth,trim={0 0 0 0}]{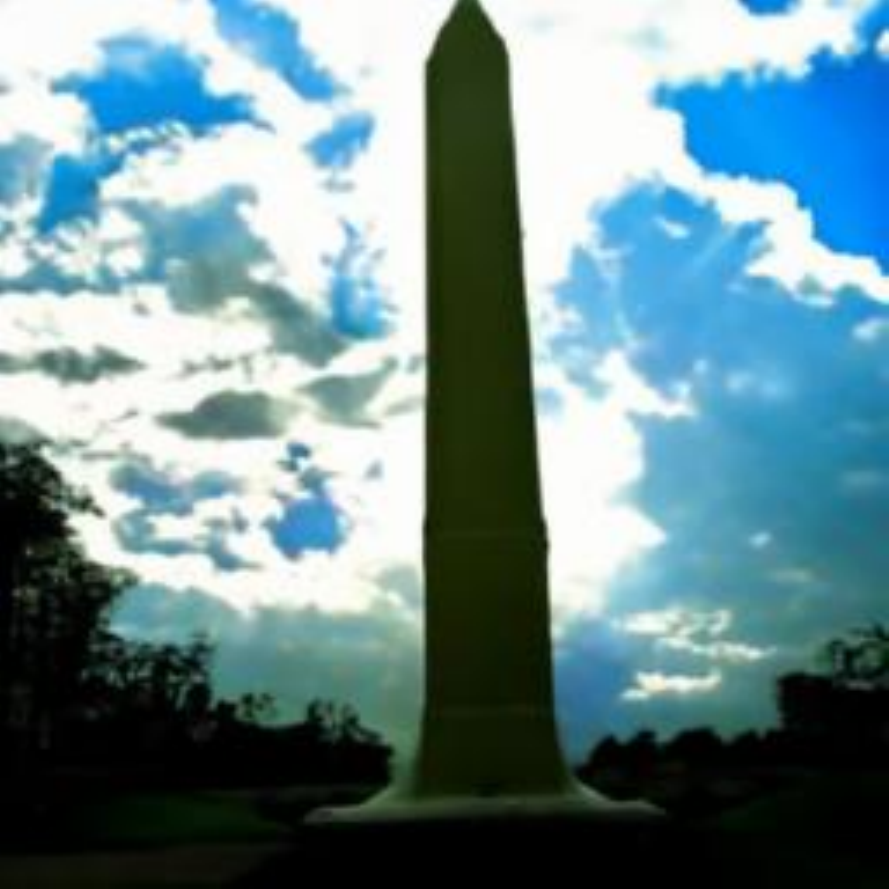} &
        \adjincludegraphics[clip,width=0.112\textwidth,trim={0 0 0 0}]{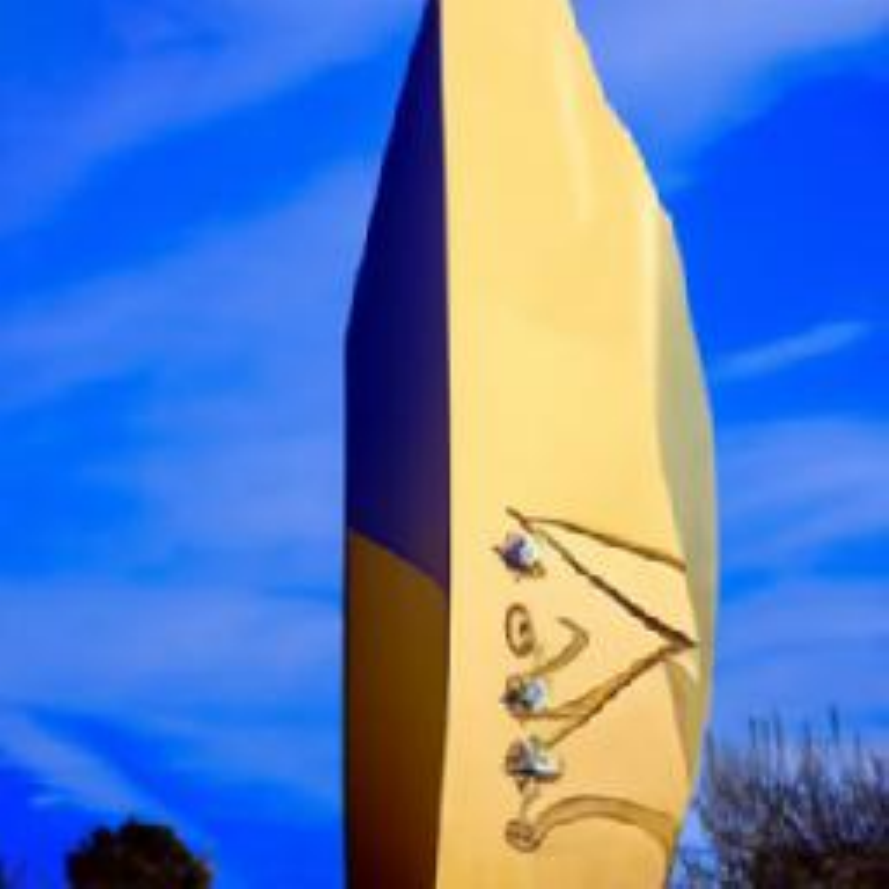} &
        \adjincludegraphics[clip,width=0.112\textwidth,trim={0 0 0 0}]{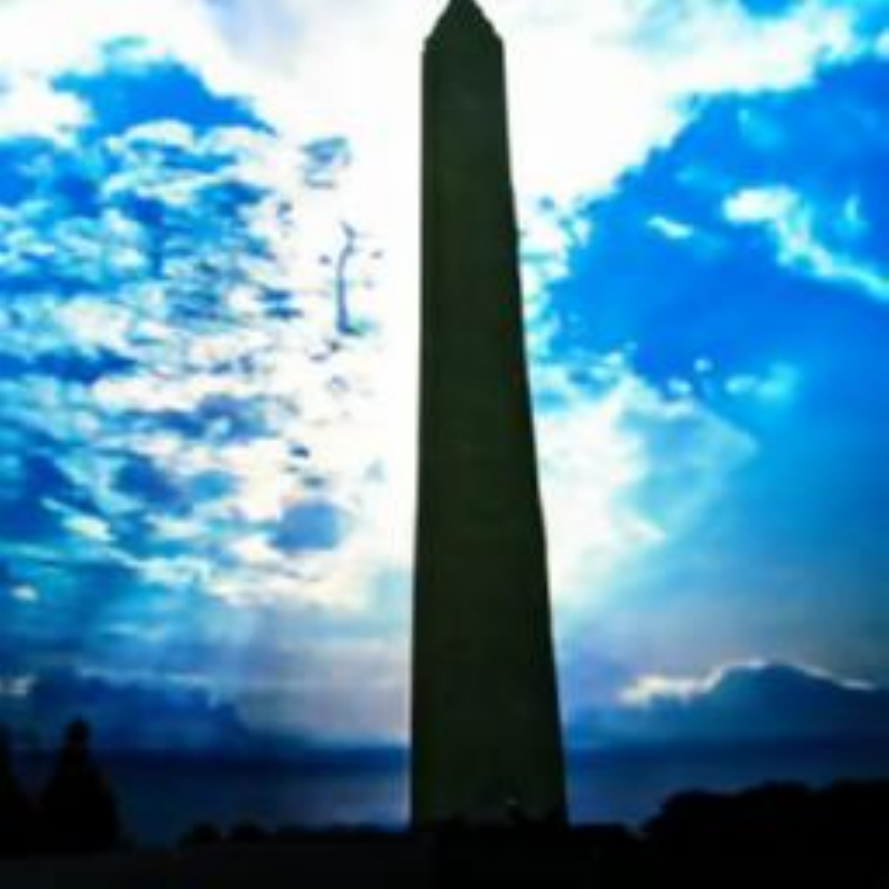} & 
        \adjincludegraphics[clip,width=0.112\textwidth,trim={0 0 0 0}]{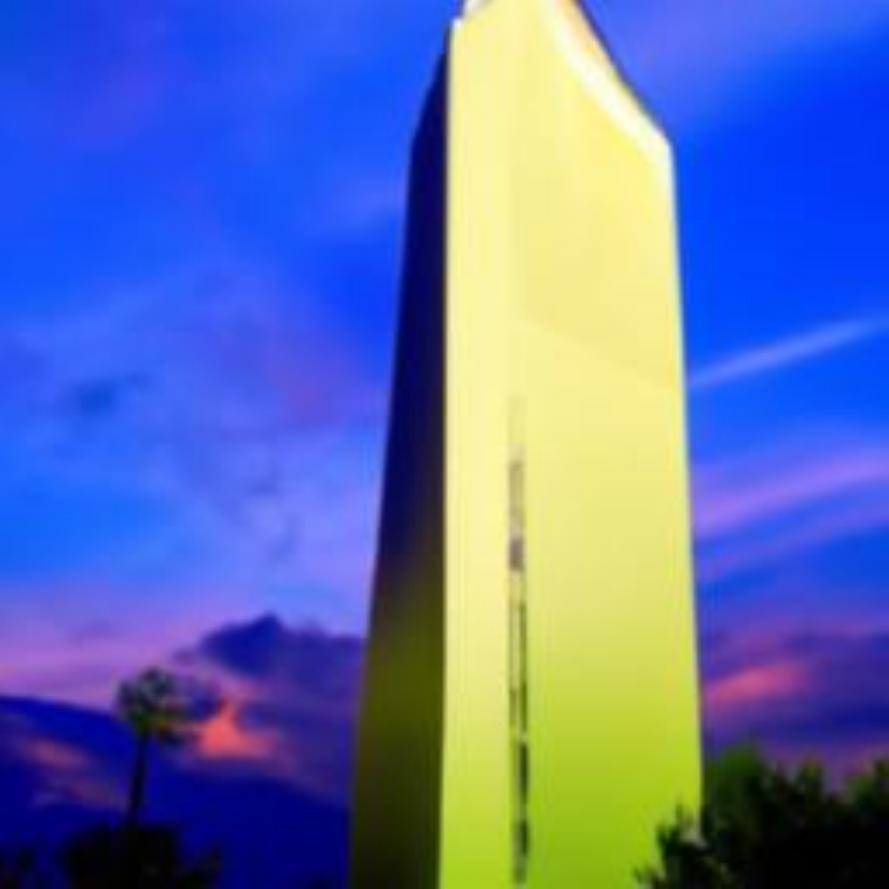} &
        \adjincludegraphics[clip,width=0.112\textwidth,trim={0 0 0 0}]{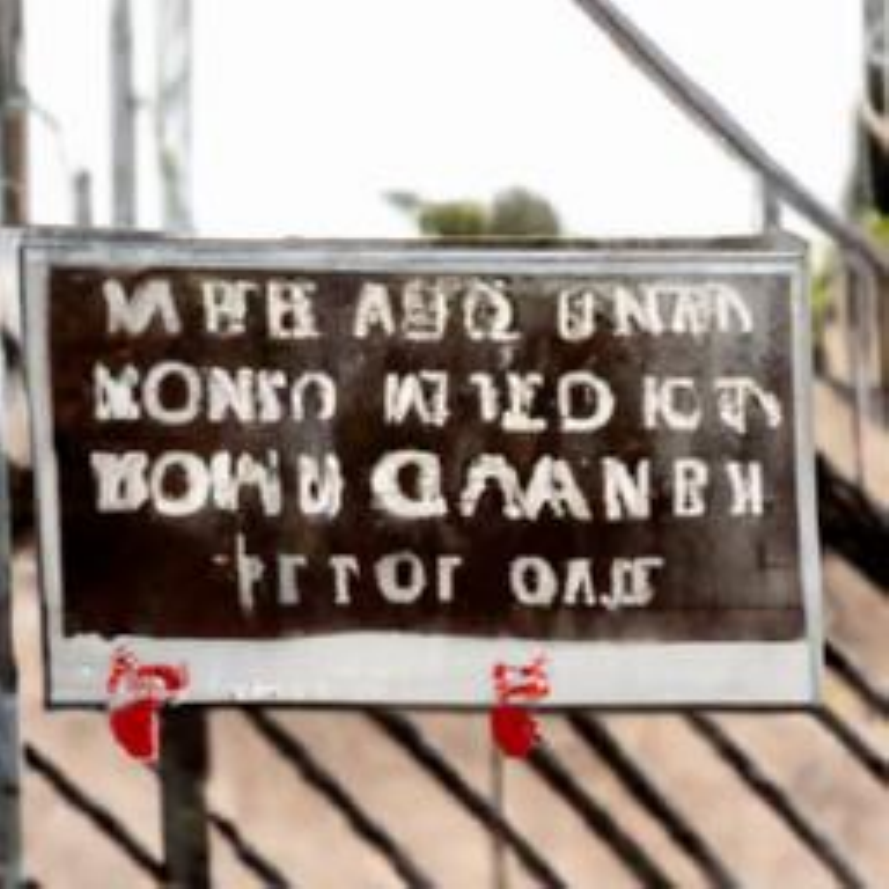} &
        \adjincludegraphics[clip,width=0.112\textwidth,trim={0 0 0 0}]{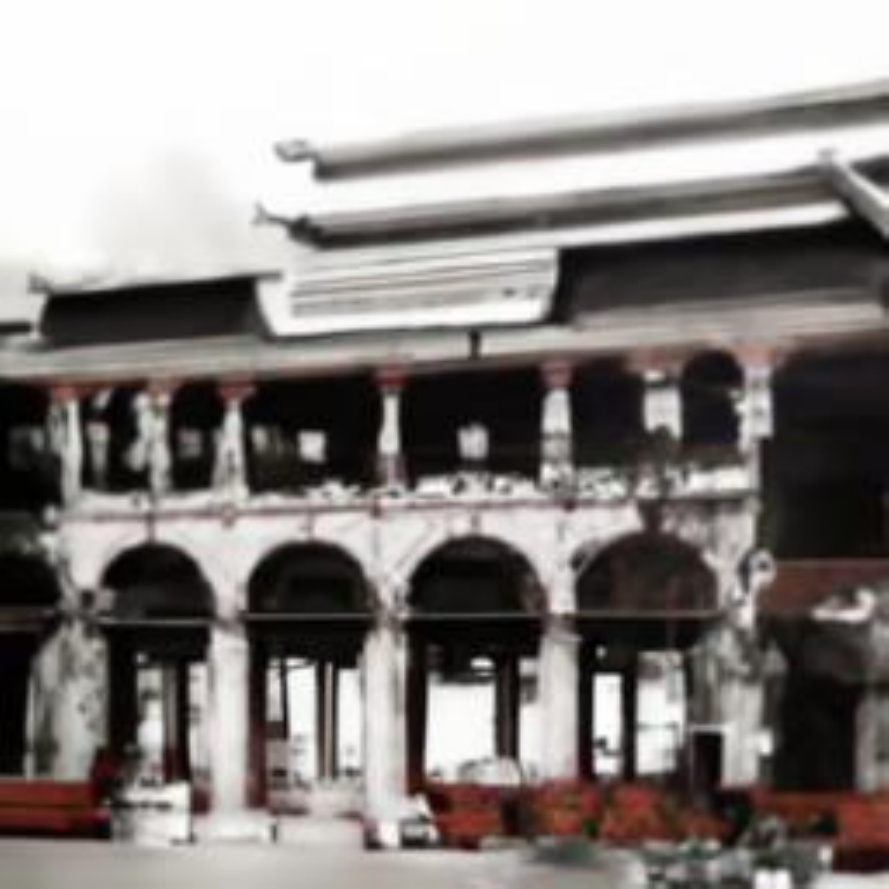} &
        \adjincludegraphics[clip,width=0.112\textwidth,trim={0 0 0 0}]{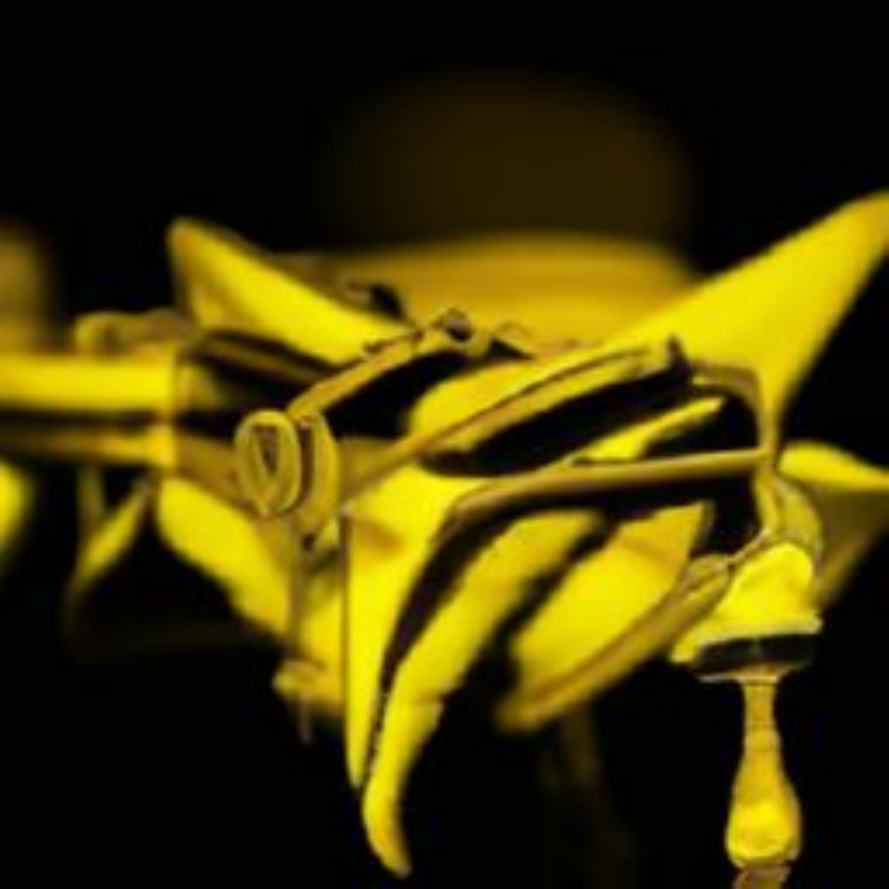} &
        \adjincludegraphics[clip,width=0.112\textwidth,trim={0 0 0 0}]{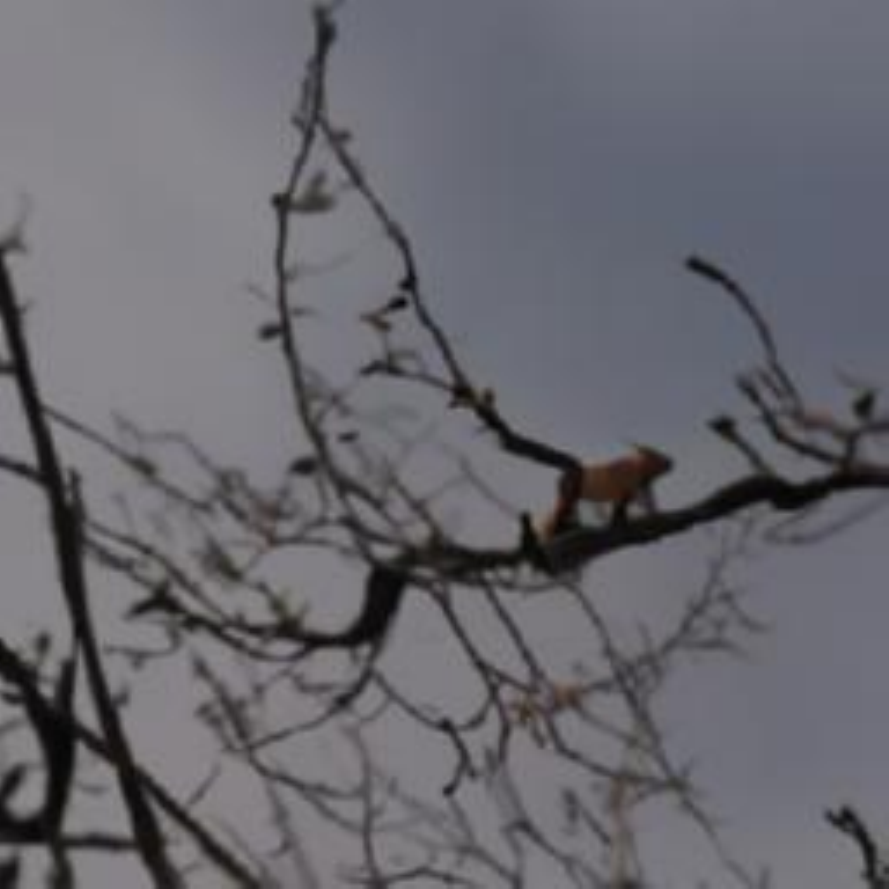} \\

        \raisebox{1.6\height}{\rotatebox{90}{\scriptsize Tick}} 
        \adjincludegraphics[clip,width=0.112\textwidth,trim={0 0 0 0}]{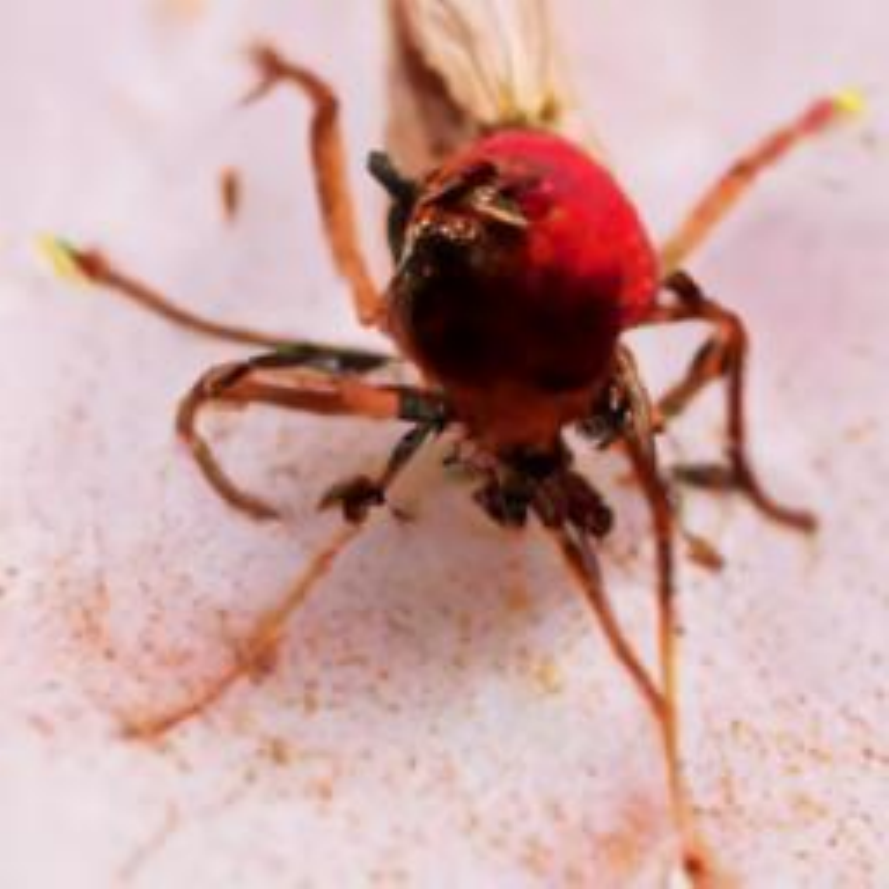} &
        \adjincludegraphics[clip,width=0.112\textwidth,trim={0 0 0 0}]{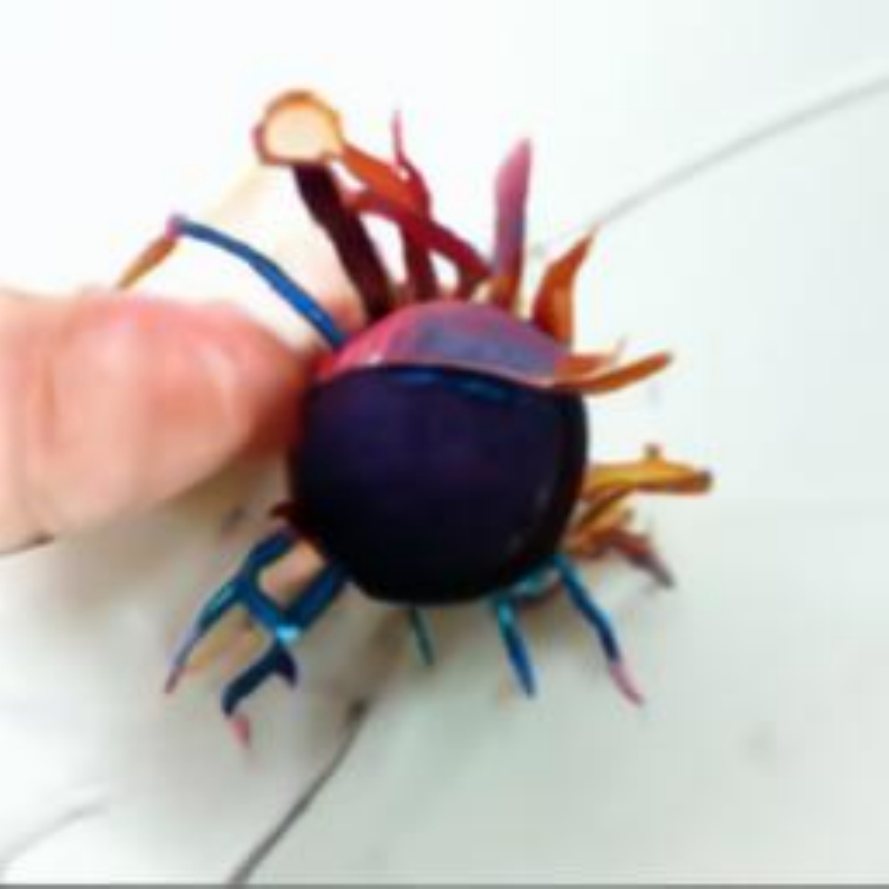} &
        \adjincludegraphics[clip,width=0.112\textwidth,trim={0 0 0 0}]{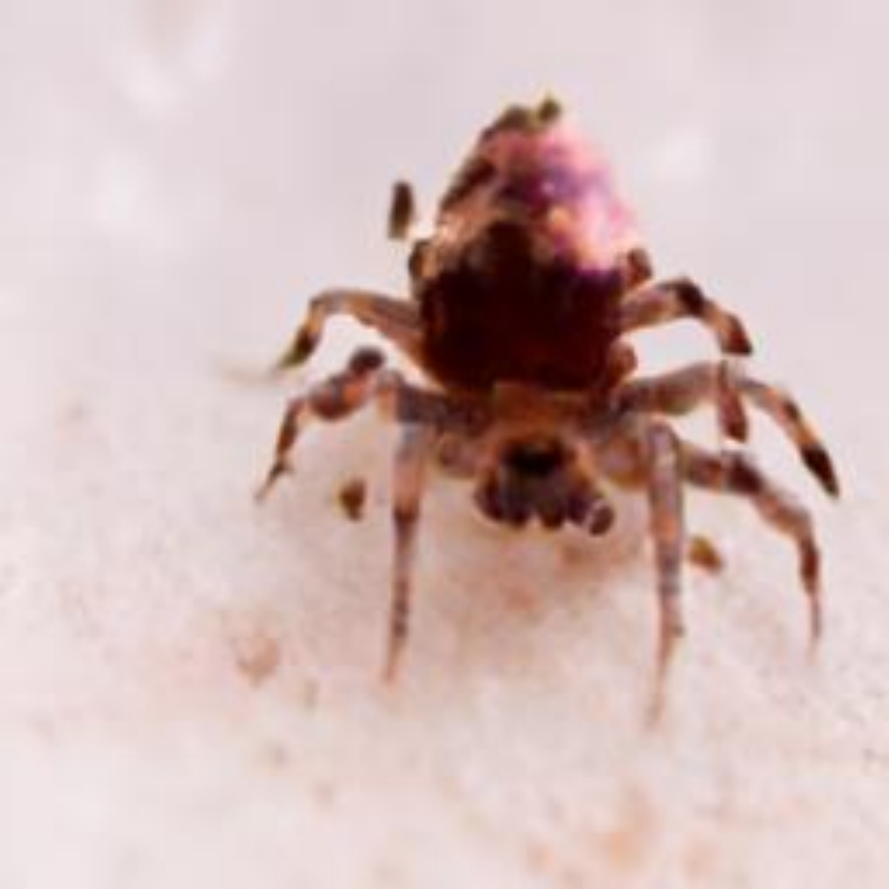} & 
        \adjincludegraphics[clip,width=0.112\textwidth,trim={0 0 0 0}]{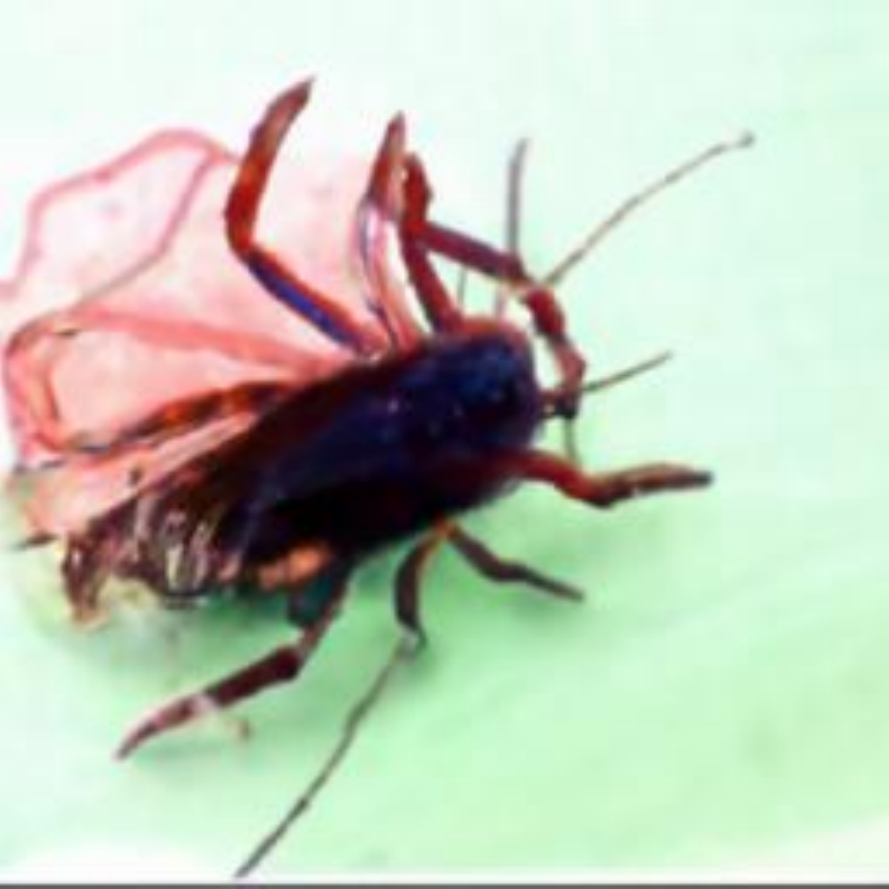} &
        \adjincludegraphics[clip,width=0.112\textwidth,trim={0 0 0 0}]{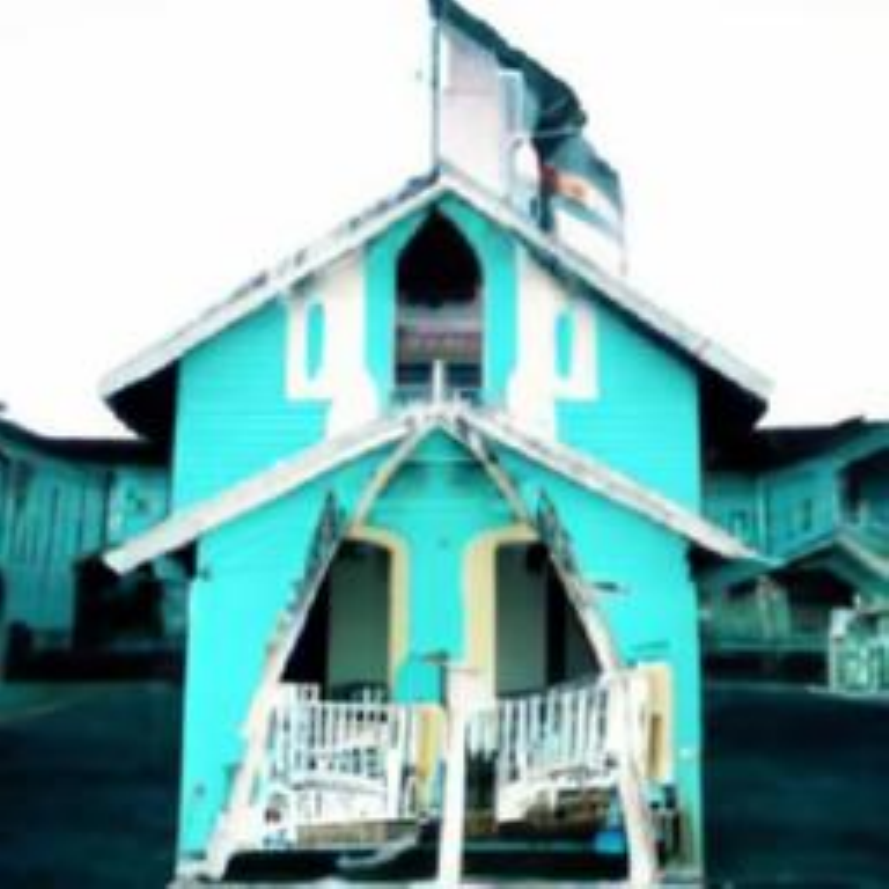} &
        \adjincludegraphics[clip,width=0.112\textwidth,trim={0 0 0 0}]{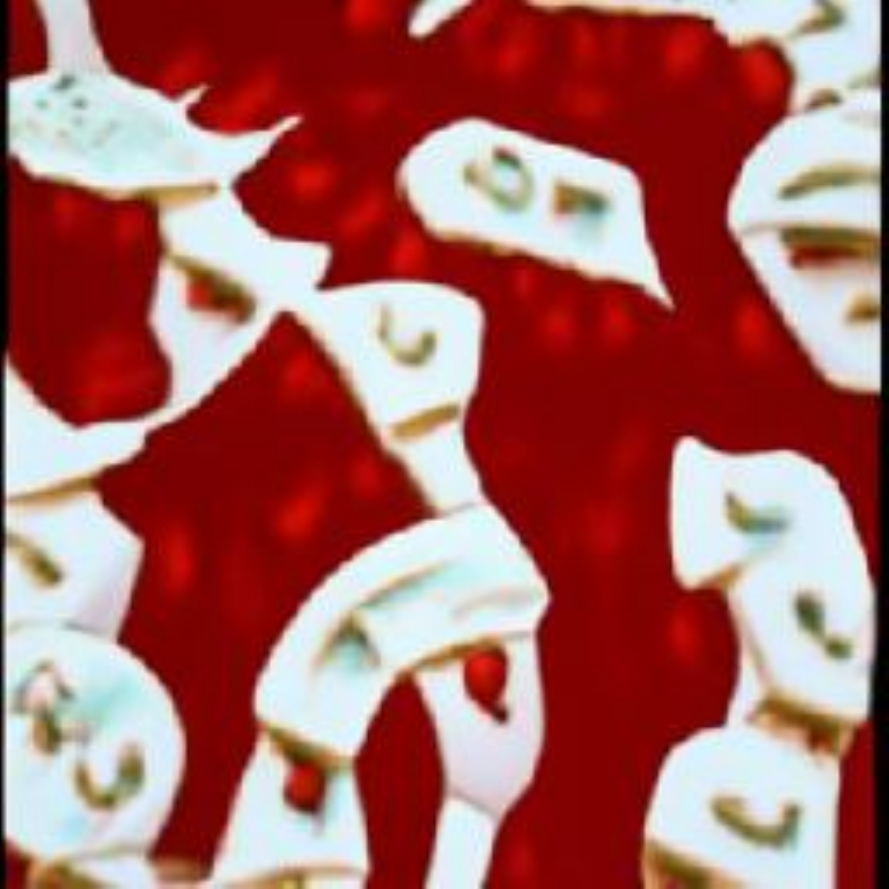} &
        \adjincludegraphics[clip,width=0.112\textwidth,trim={0 0 0 0}]{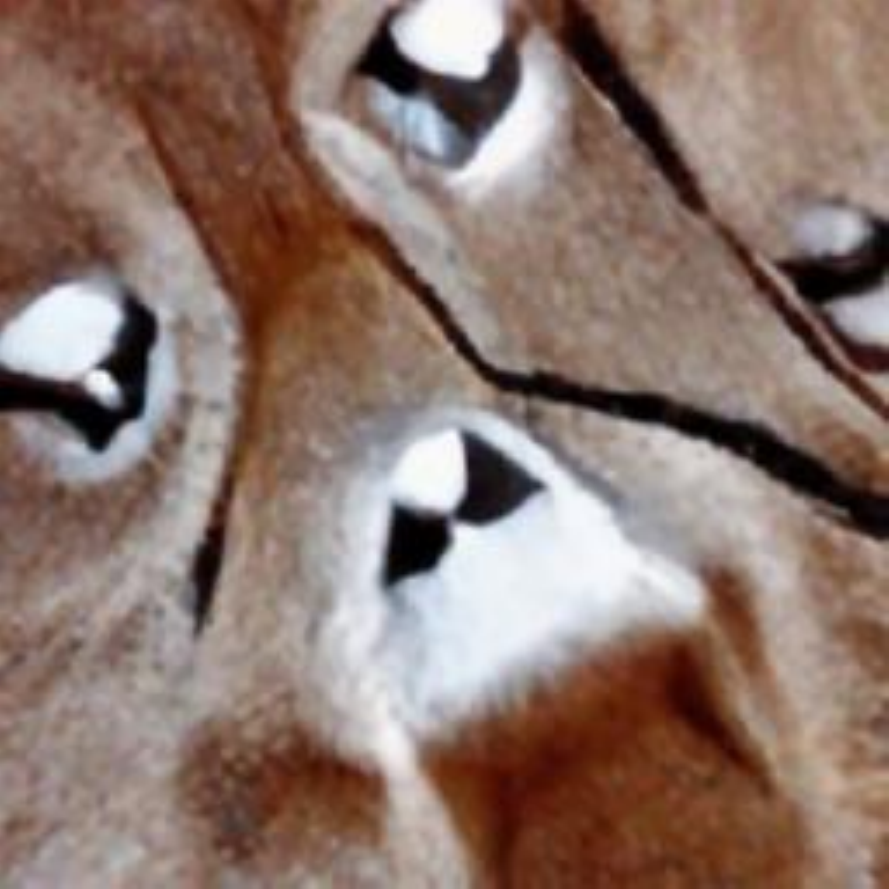} &
        \adjincludegraphics[clip,width=0.112\textwidth,trim={0 0 0 0}]{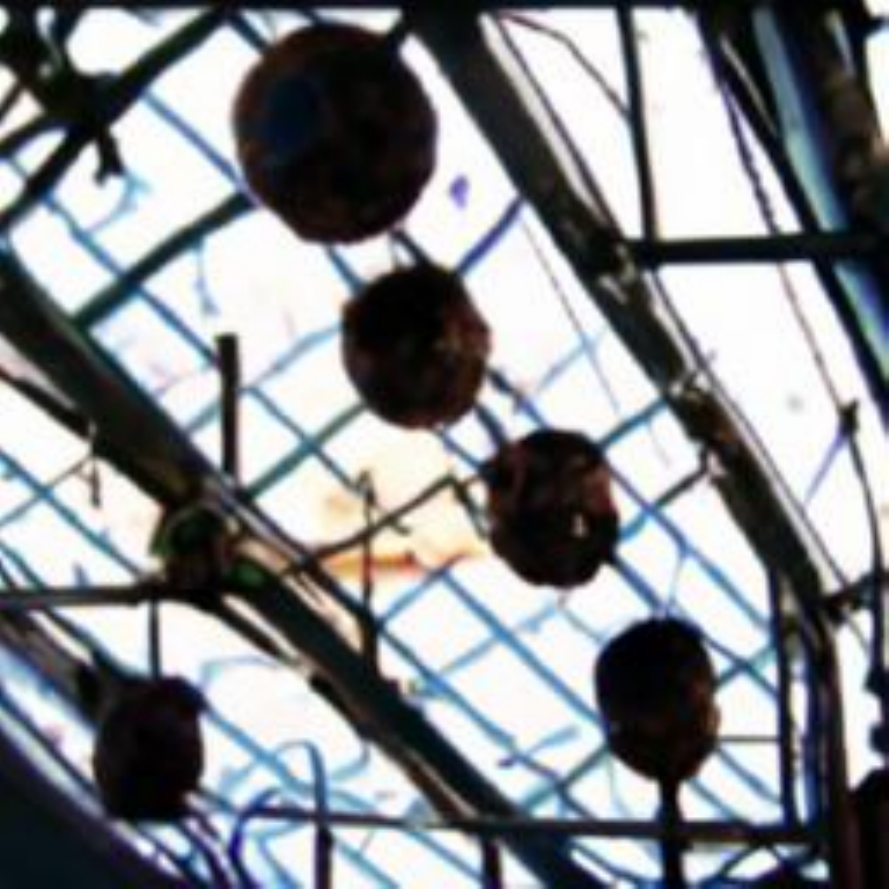} \\

        \raisebox{1.3\height}{\rotatebox{90}{\scriptsize Dock}} 
        \adjincludegraphics[clip,width=0.112\textwidth,trim={0 0 0 0}]{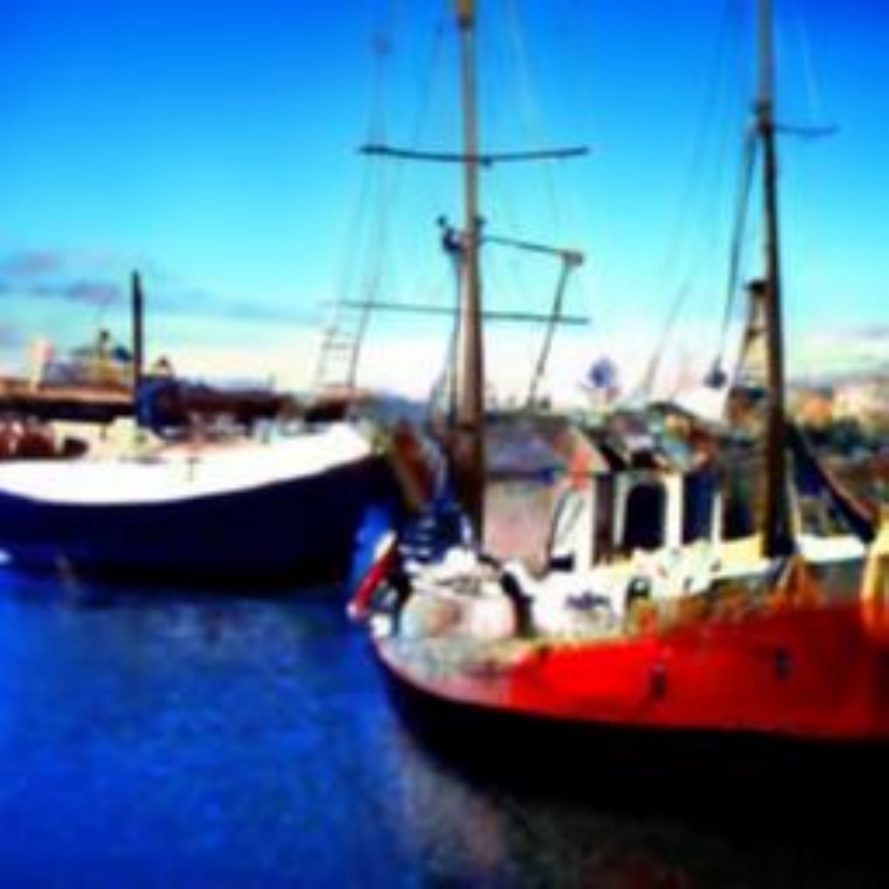} &
        \adjincludegraphics[clip,width=0.112\textwidth,trim={0 0 0 0}]{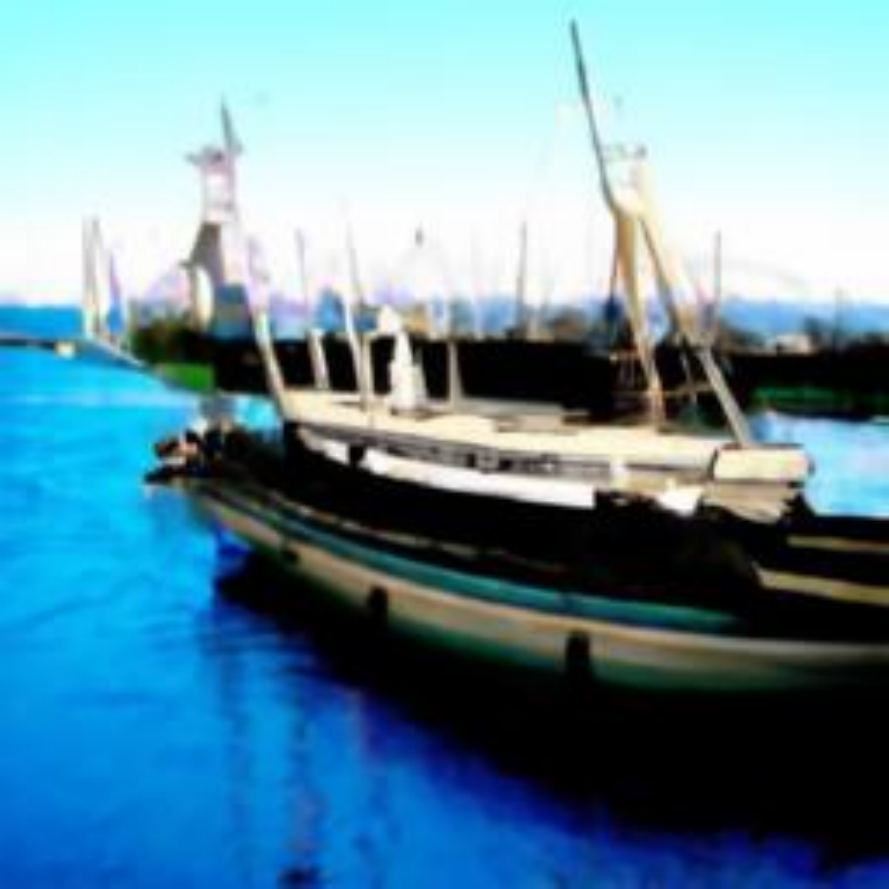} &
        \adjincludegraphics[clip,width=0.112\textwidth,trim={0 0 0 0}]{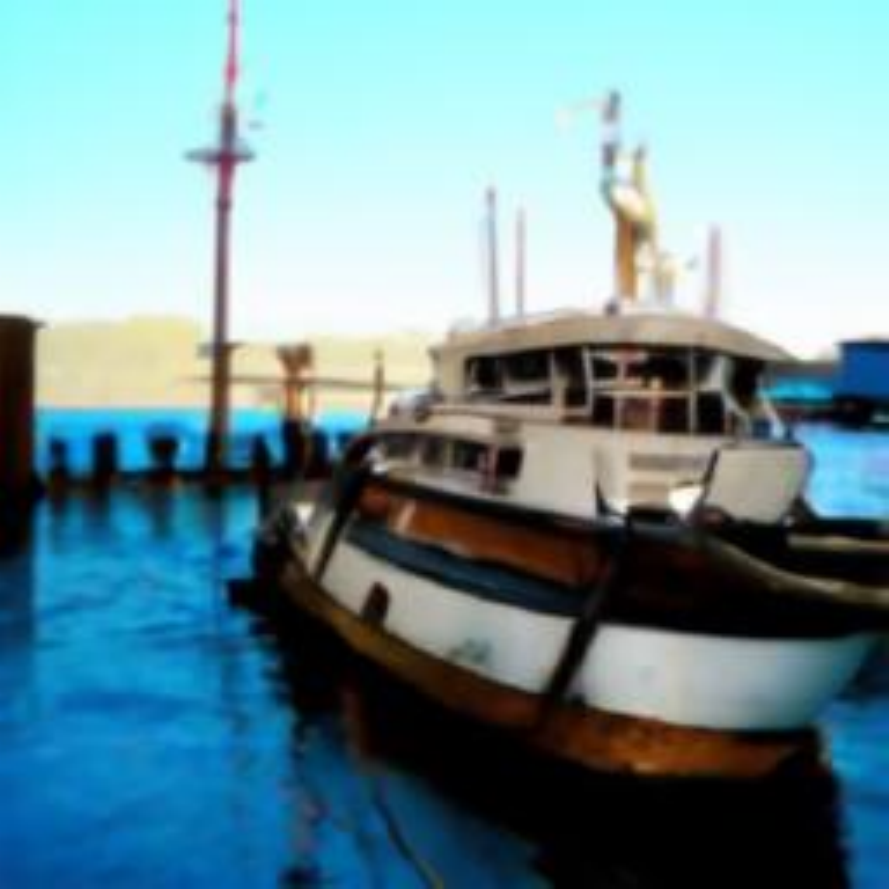} & 
        \adjincludegraphics[clip,width=0.112\textwidth,trim={0 0 0 0}]{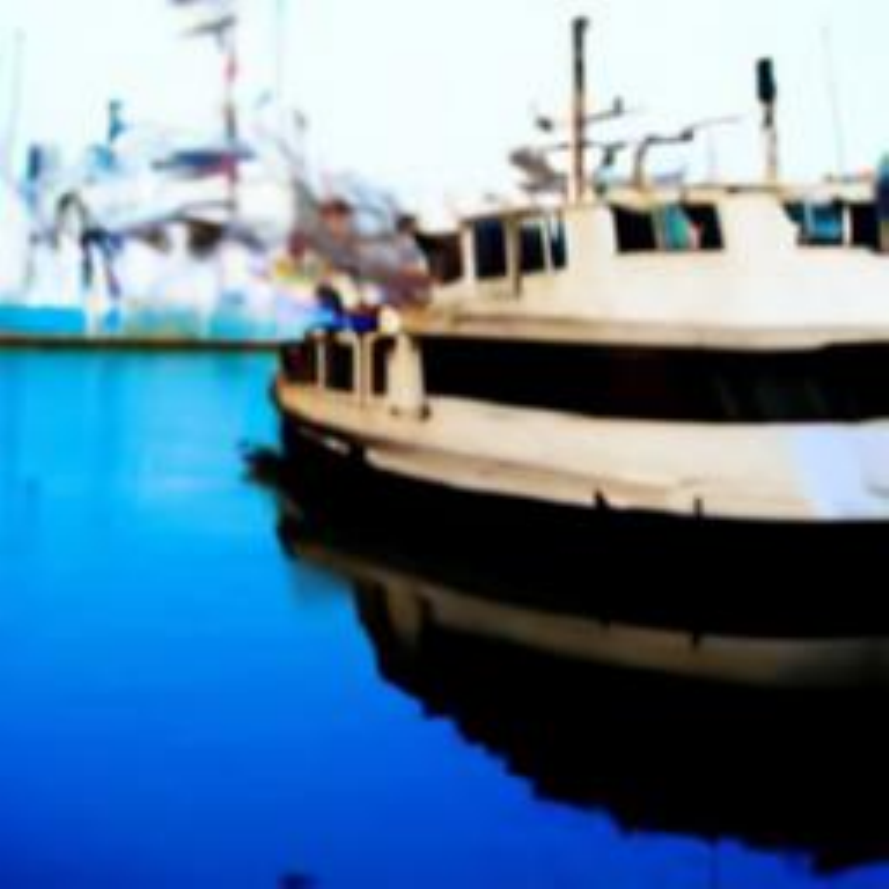} &
        \adjincludegraphics[clip,width=0.112\textwidth,trim={0 0 0 0}]{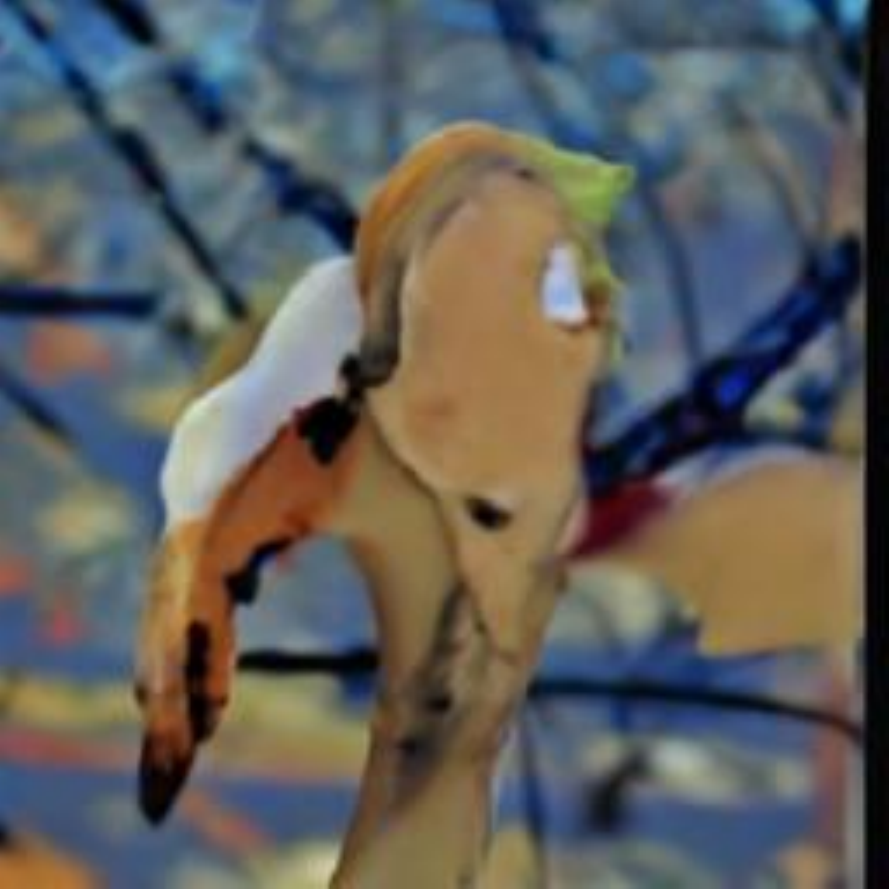} &
        \adjincludegraphics[clip,width=0.112\textwidth,trim={0 0 0 0}]{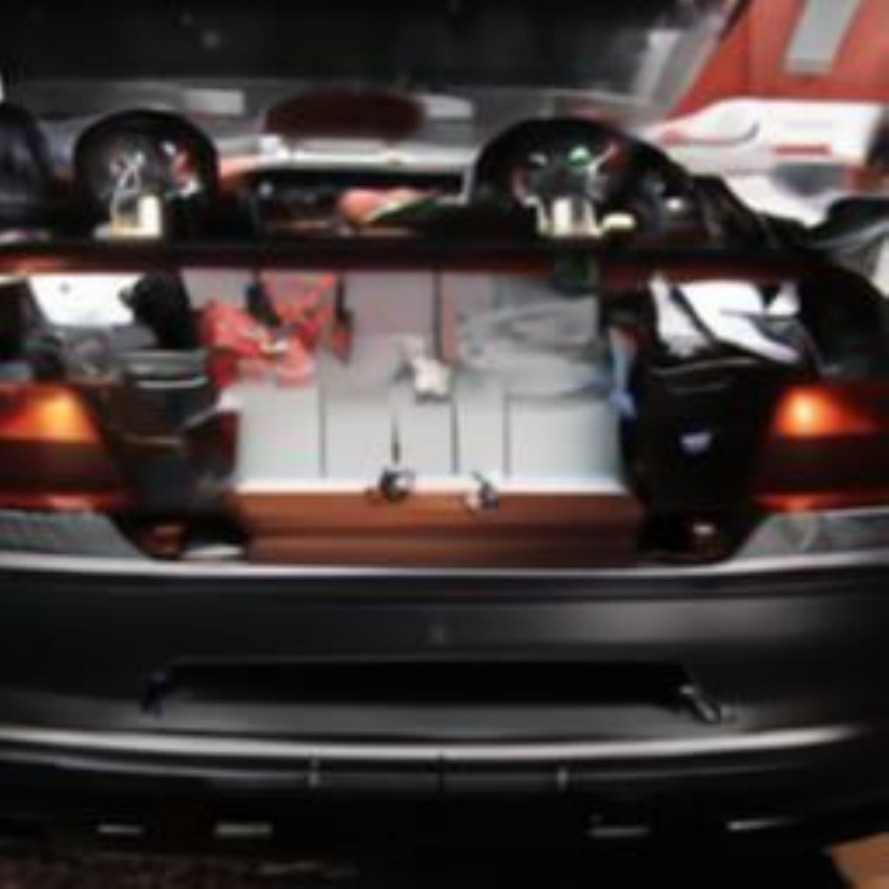} &
        \adjincludegraphics[clip,width=0.112\textwidth,trim={0 0 0 0}]{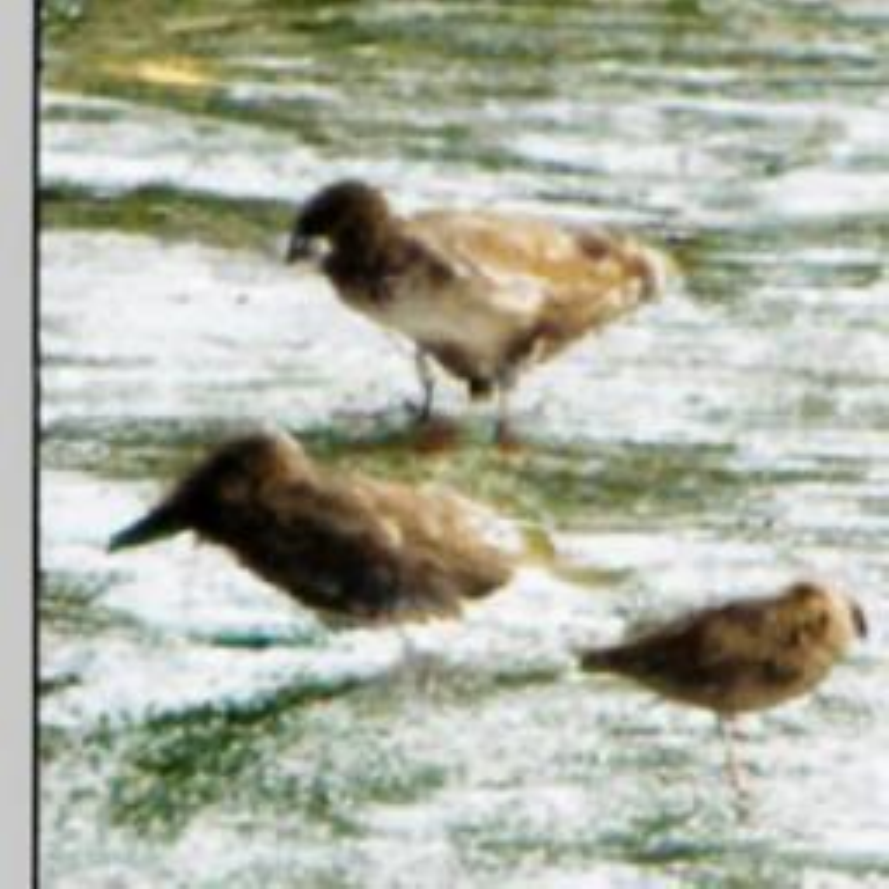} &
        \adjincludegraphics[clip,width=0.112\textwidth,trim={0 0 0 0}]{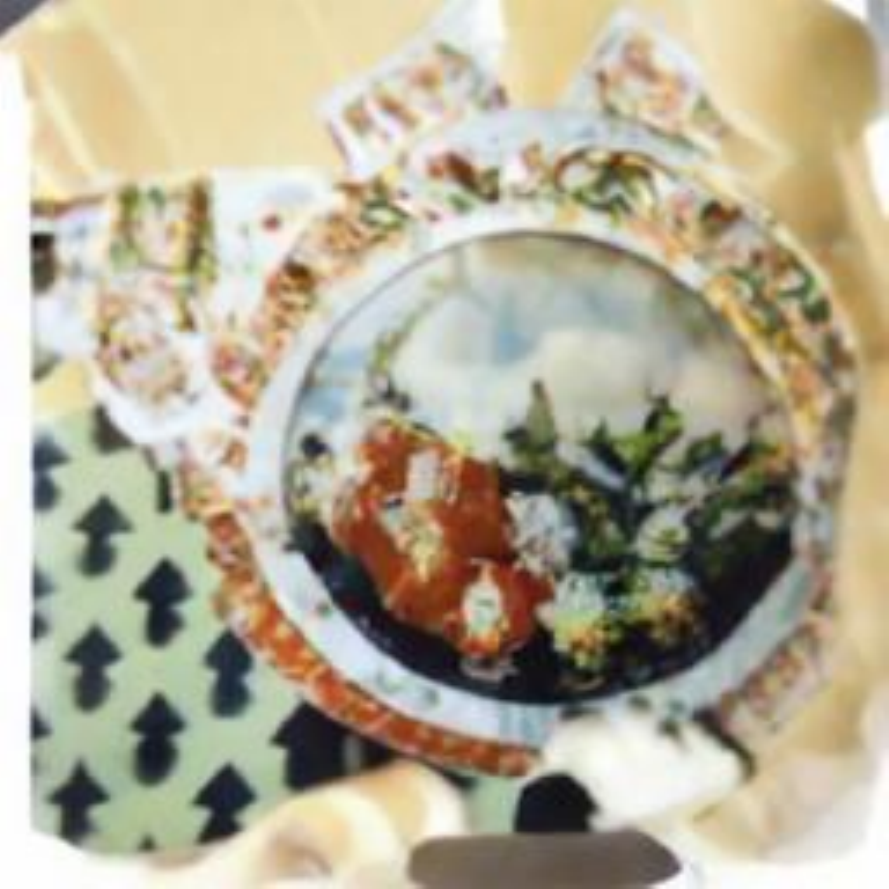} \\
        
        \multicolumn{4}{c}{PPAP} &
        \multicolumn{4}{c}{Na\"ive off-the-shelf}
    \end{tabular}
    
    \caption{
    Qualitative results of in-class variations by guiding GLIDE with ResNet50 classifier. Our PPAP framework can generate proper images corresponding to the given class, but the na\"ive off-the-shelf fails.
    }
    \label{apx:fig_glide_resnet_class}

\end{figure*}

To show the effectiveness of PPAP, we illustrate more qualitative results in Fig.~\ref{fig:ap_glide_resnet_more}.
Furthermore, we also show multiple generated images per class in Fig.~\ref{apx:fig_glide_resnet_class}.

\subsection{GLIDE + Depth}
\label{aps:subsec:glide+depth}

\begin{figure*}[t]

    \centering
    \begin{tabular}{@{\hspace{2mm}}c@{\hspace{2mm}}c@{\hspace{2mm}}c@{\hspace{2mm}}c@{\hspace{2mm}}c@{\hspace{2mm}}c@{\hspace{2mm}}c@{\hspace{2mm}}c@{\hspace{2mm}}c}
        \raisebox{0.7\height}{\rotatebox{90}{\scriptsize Original}} 
        \adjincludegraphics[clip,width=0.11\textwidth,trim={0 0 0 0}]{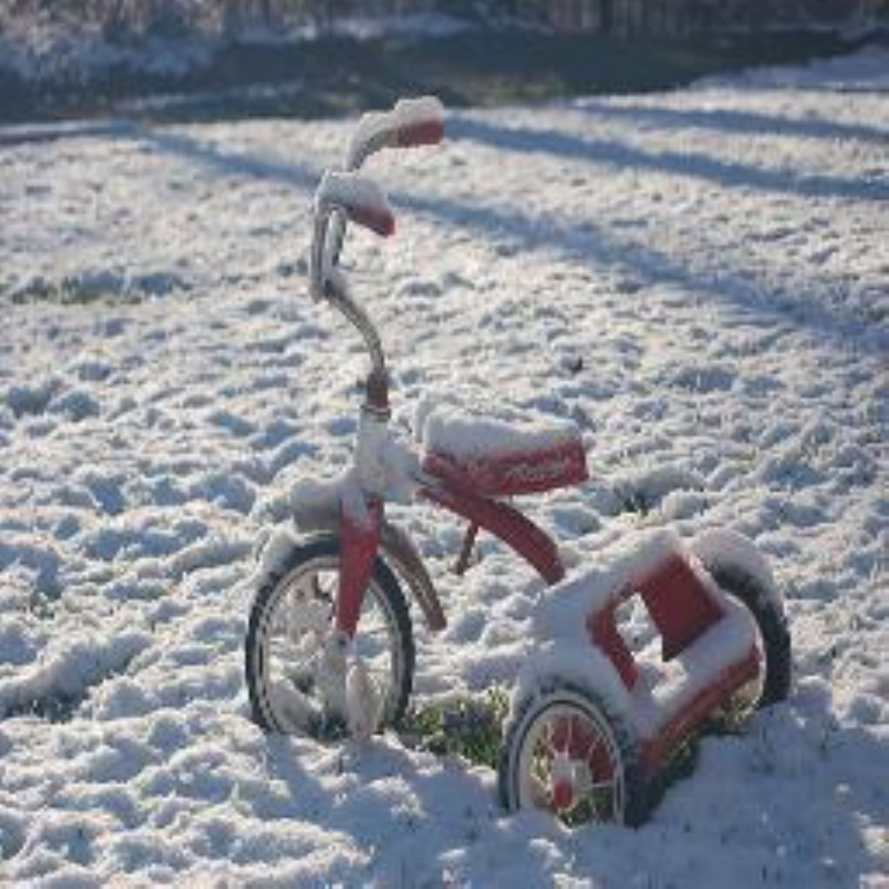} &
        \adjincludegraphics[clip,width=0.11\textwidth,trim={0 0 0 0}]{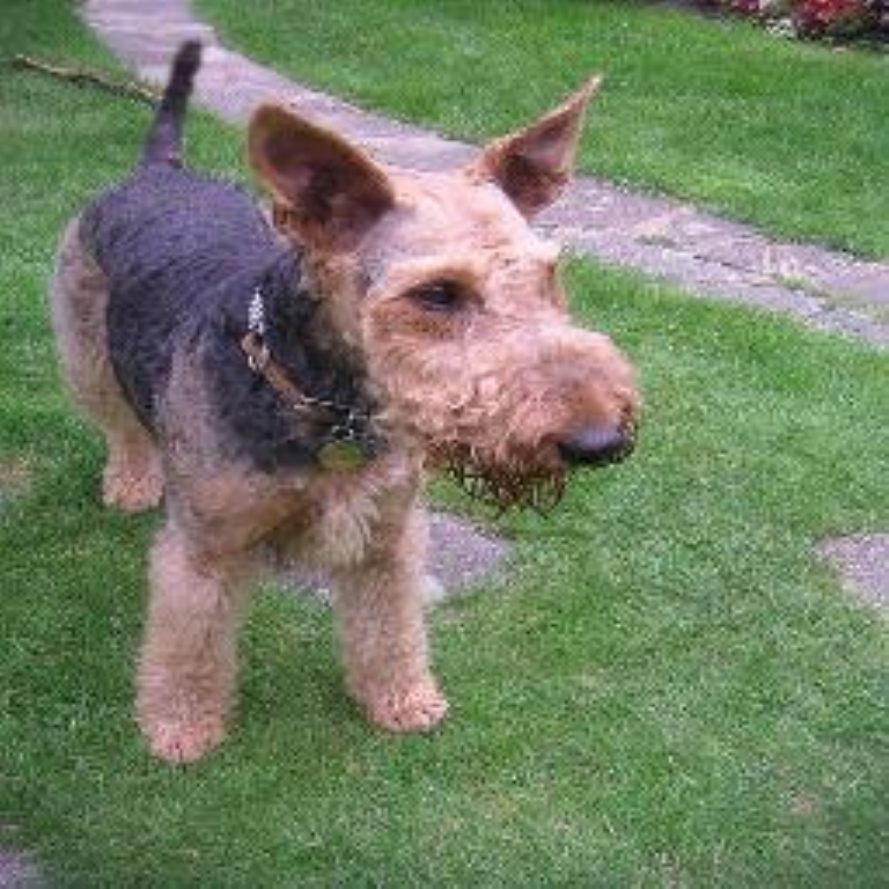} &
        \adjincludegraphics[clip,width=0.11\textwidth,trim={0 0 0 0}]{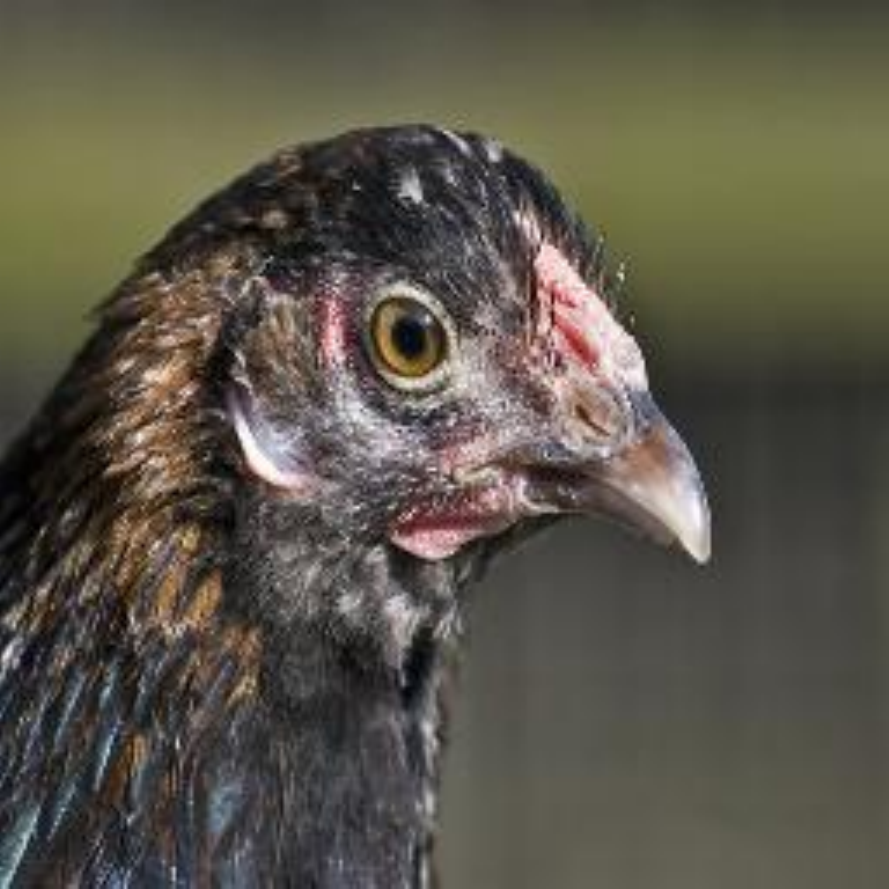} &
        \adjincludegraphics[clip,width=0.11\textwidth,trim={0 0 0 0}]{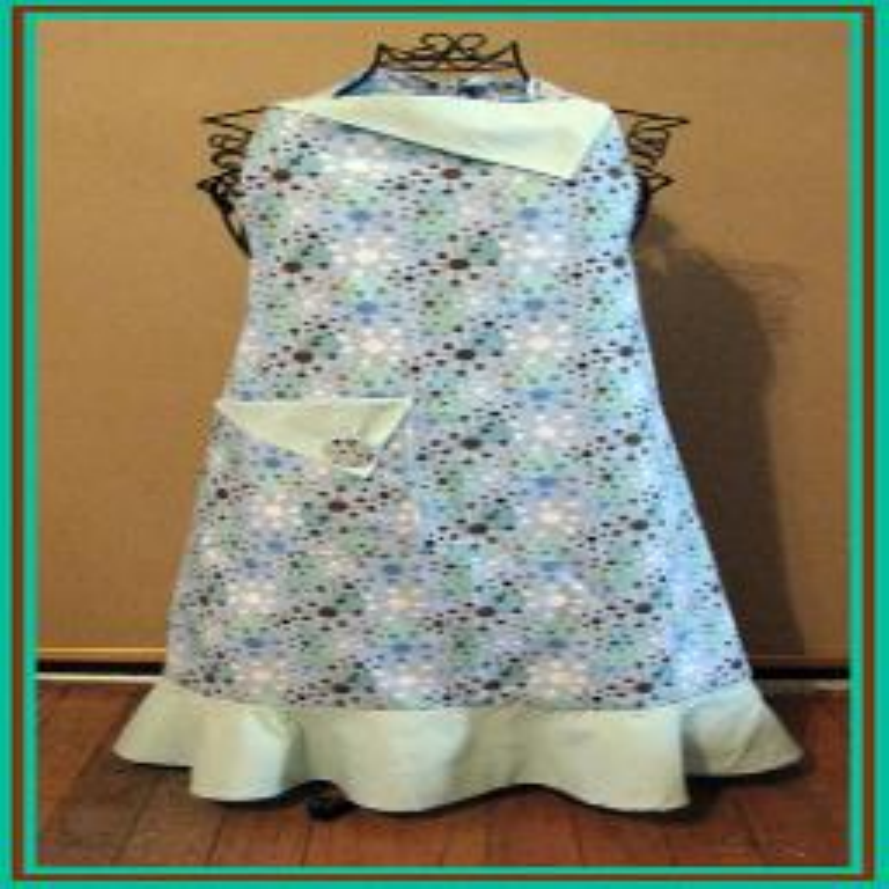} &
        \adjincludegraphics[clip,width=0.11\textwidth,trim={0 0 0 0}]{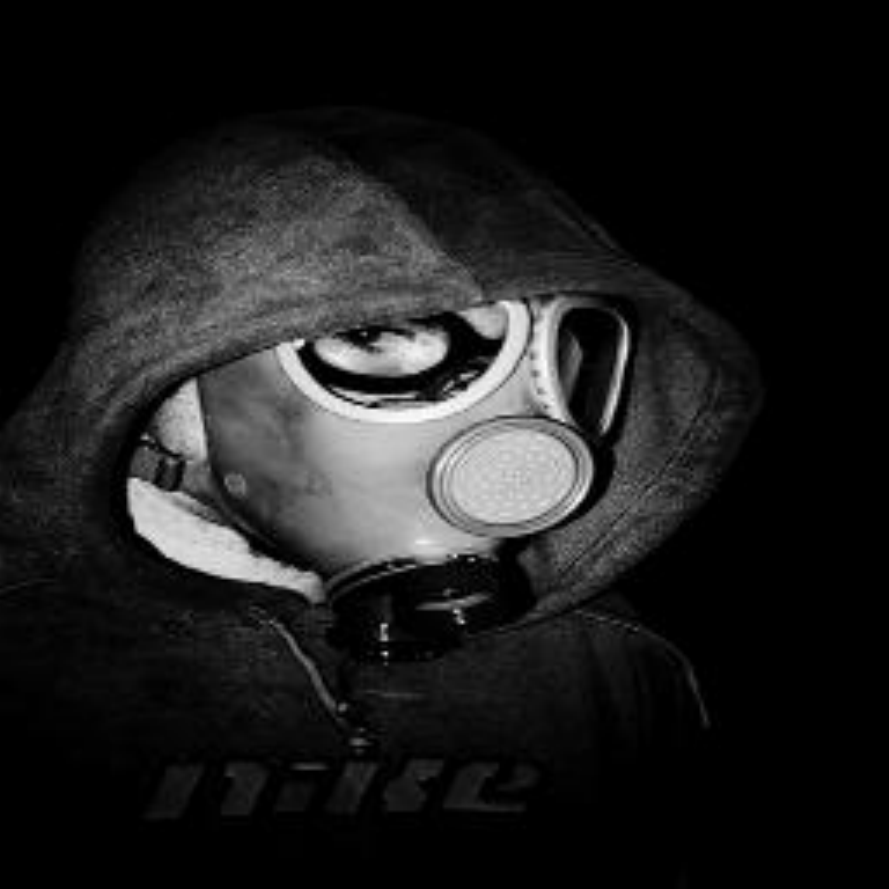} &
        \adjincludegraphics[clip,width=0.11\textwidth,trim={0 0 0 0}]{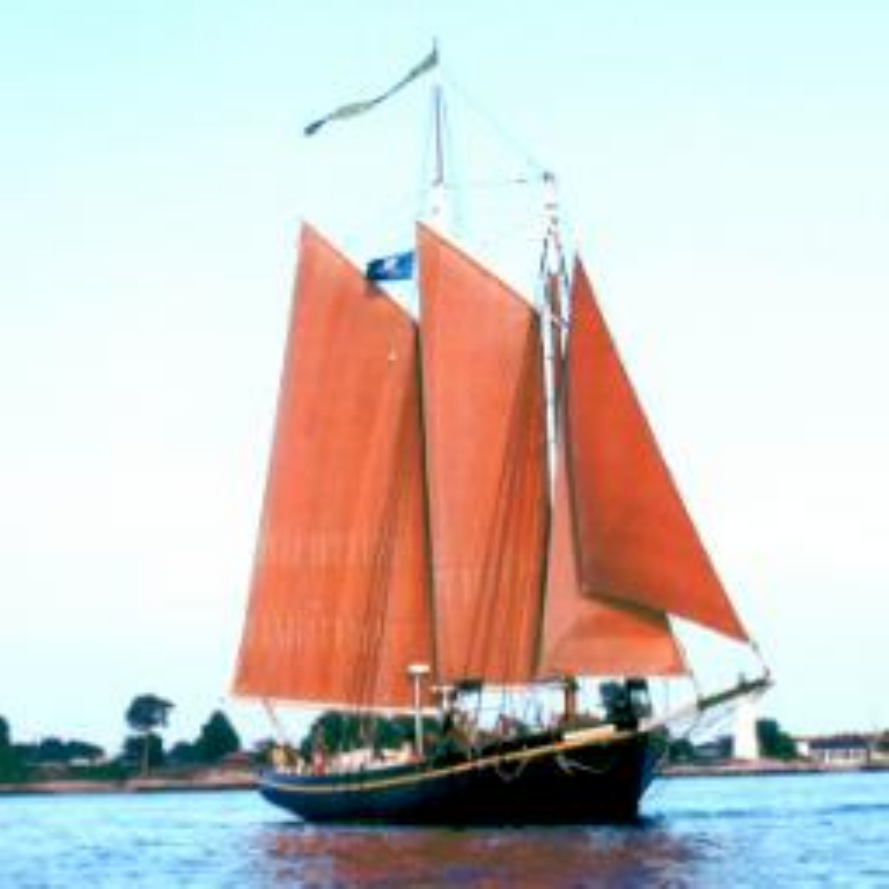} &
        \adjincludegraphics[clip,width=0.11\textwidth,trim={0 0 0 0}]{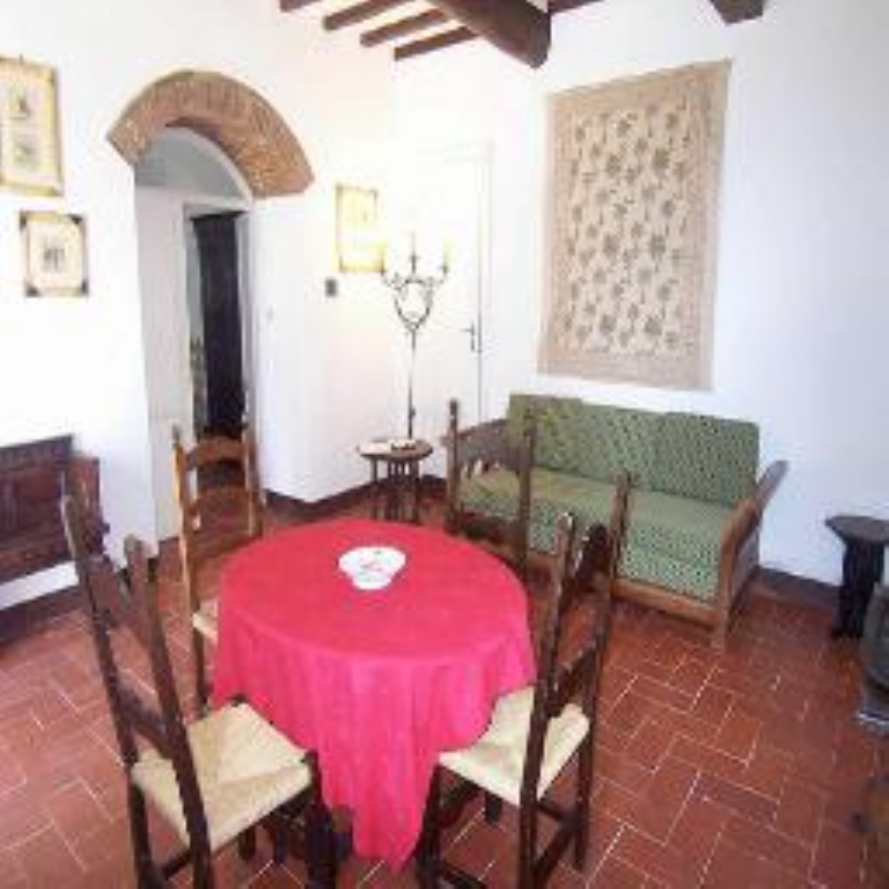} &
        \adjincludegraphics[clip,width=0.11\textwidth,trim={0 0 0 0}]{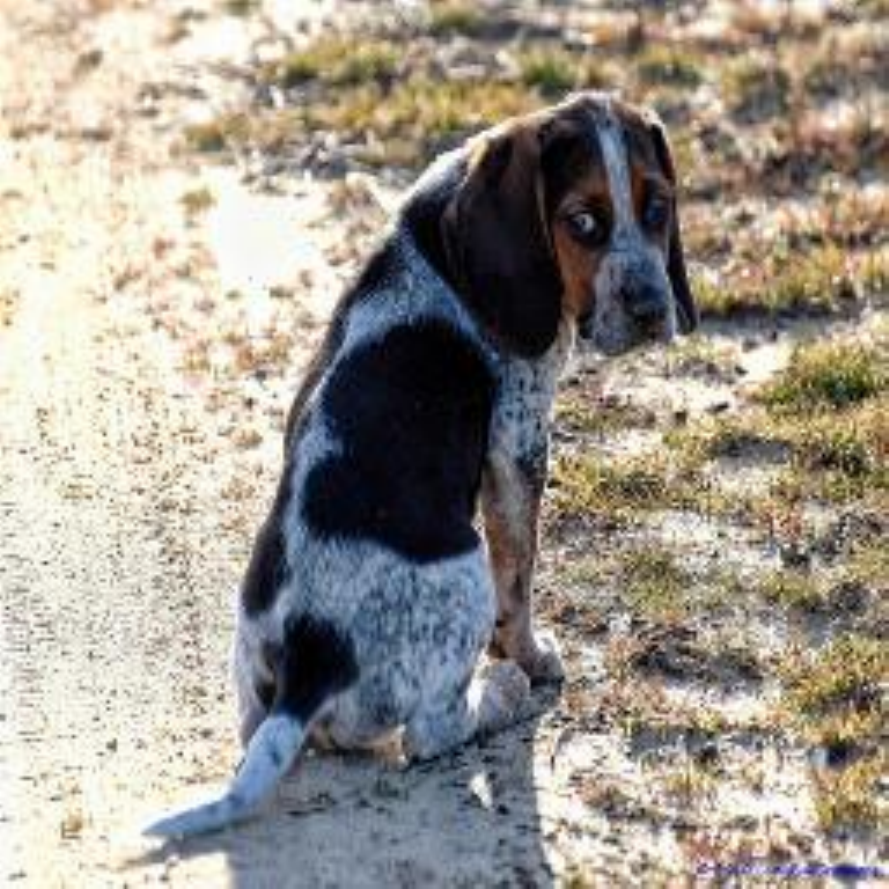} \\

        \raisebox{0.5\height}{\rotatebox{90}{\scriptsize Depth map}} 
        \adjincludegraphics[clip,width=0.11\textwidth,trim={0 0 0 0}]{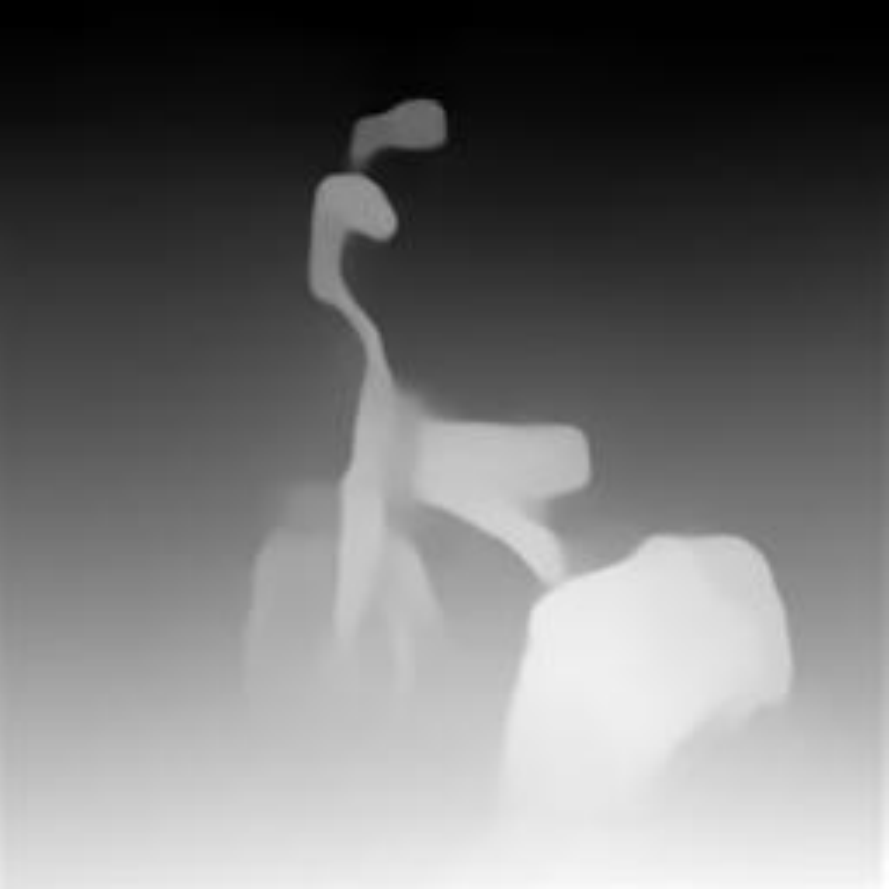} &
        \adjincludegraphics[clip,width=0.11\textwidth,trim={0 0 0 0}]{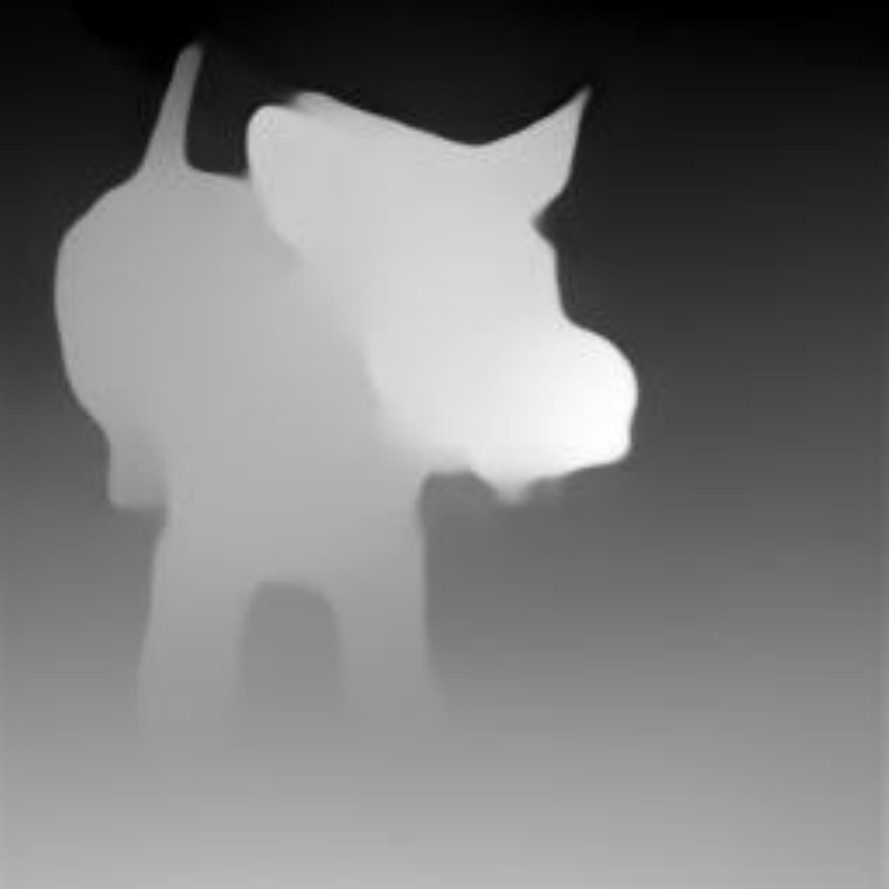} &
        \adjincludegraphics[clip,width=0.11\textwidth,trim={0 0 0 0}]{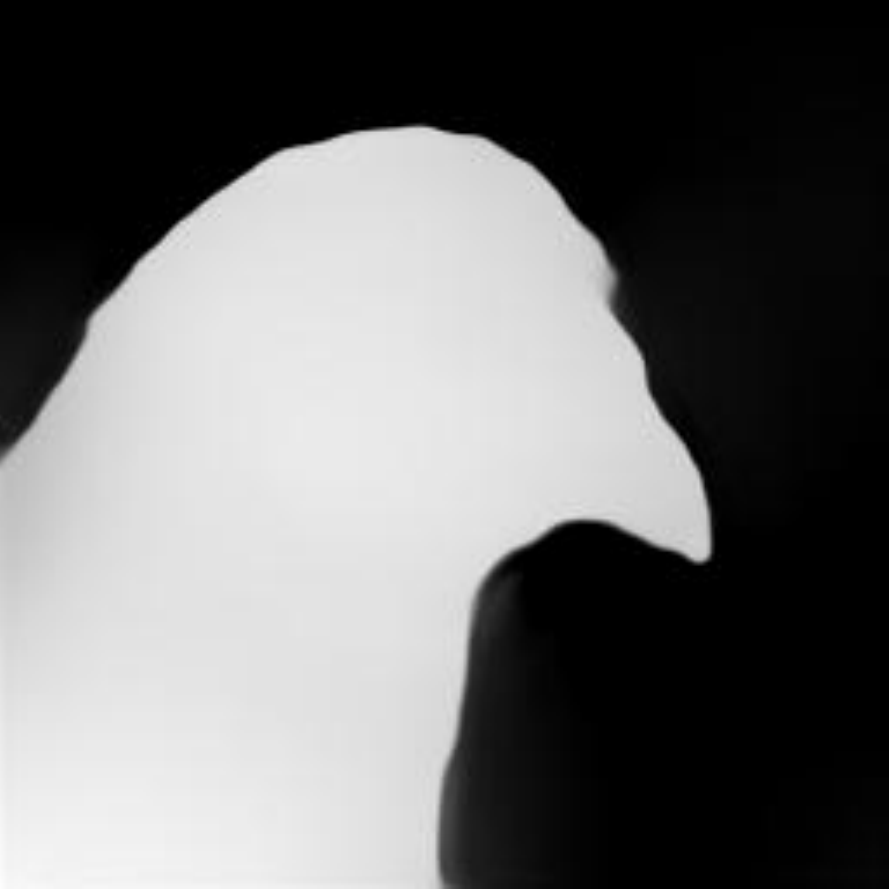} &
        \adjincludegraphics[clip,width=0.11\textwidth,trim={0 0 0 0}]{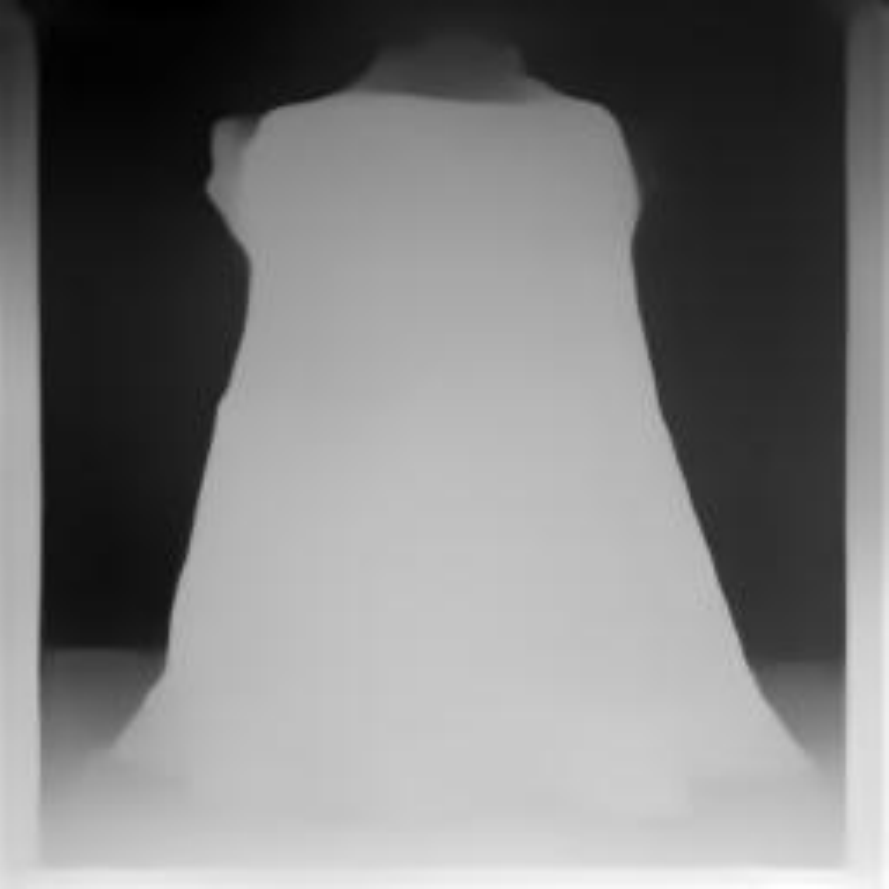} &
        \adjincludegraphics[clip,width=0.11\textwidth,trim={0 0 0 0}]{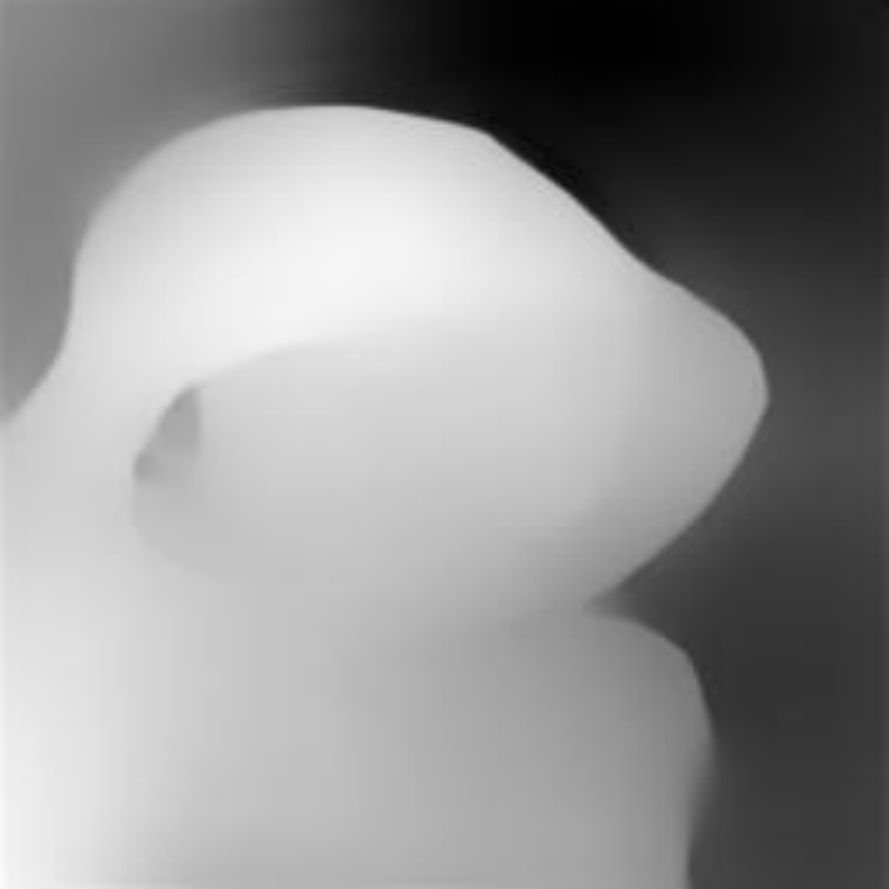} &
        \adjincludegraphics[clip,width=0.11\textwidth,trim={0 0 0 0}]{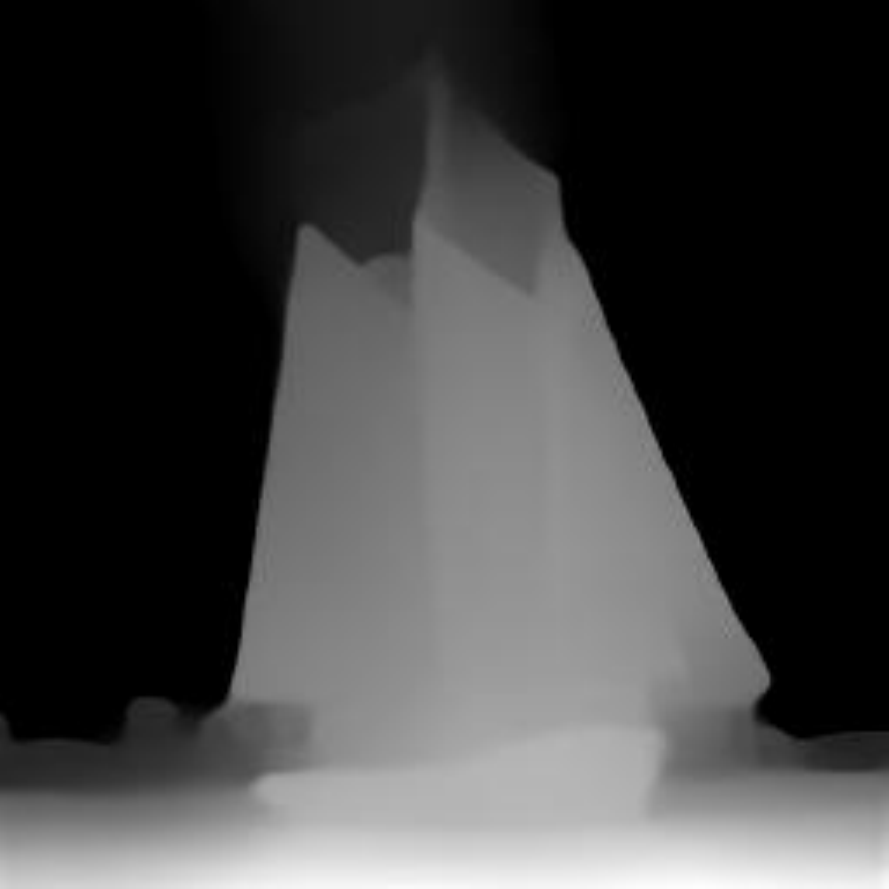} &
        \adjincludegraphics[clip,width=0.11\textwidth,trim={0 0 0 0}]{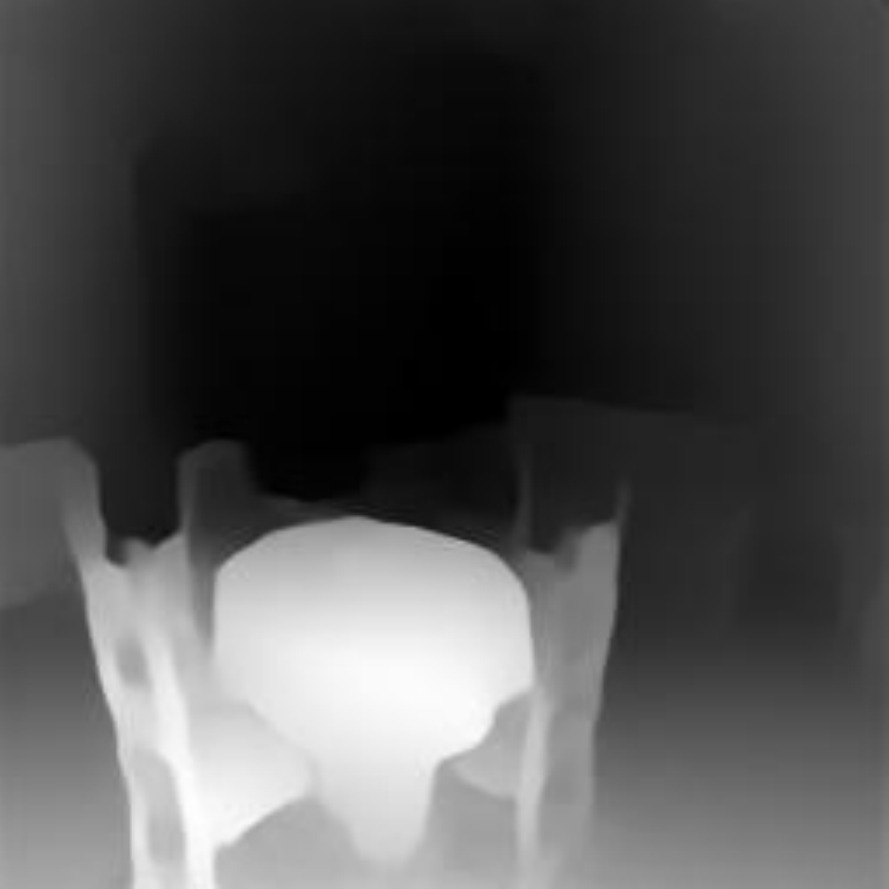} &
        \adjincludegraphics[clip,width=0.11\textwidth,trim={0 0 0 0}]{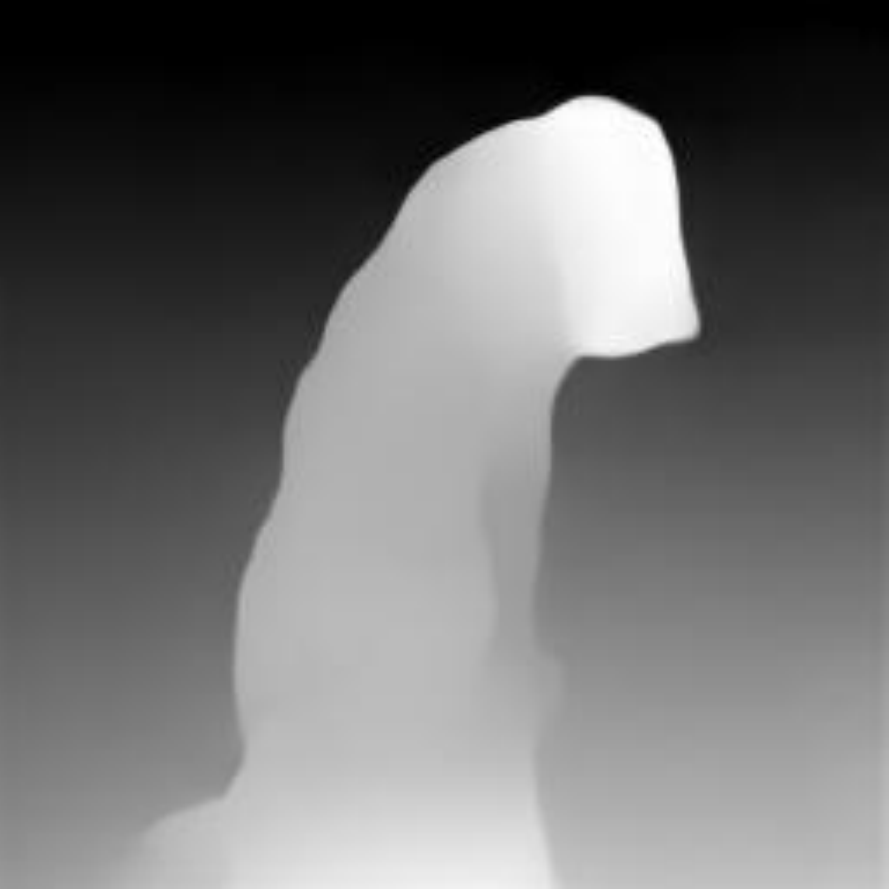} \\

        \raisebox{0.2\height}{\rotatebox{90}{\scriptsize PPAP (Ours)}} 
        \adjincludegraphics[clip,width=0.11\textwidth,trim={0 0 0 0}]{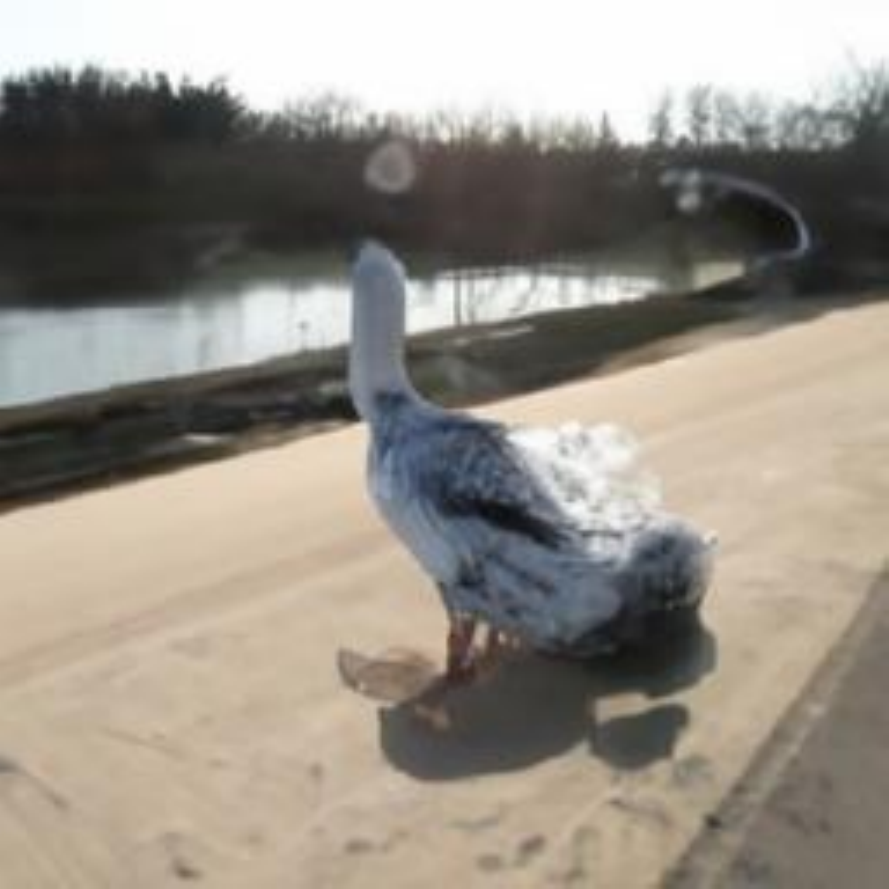} &
        \adjincludegraphics[clip,width=0.11\textwidth,trim={0 0 0 0}]{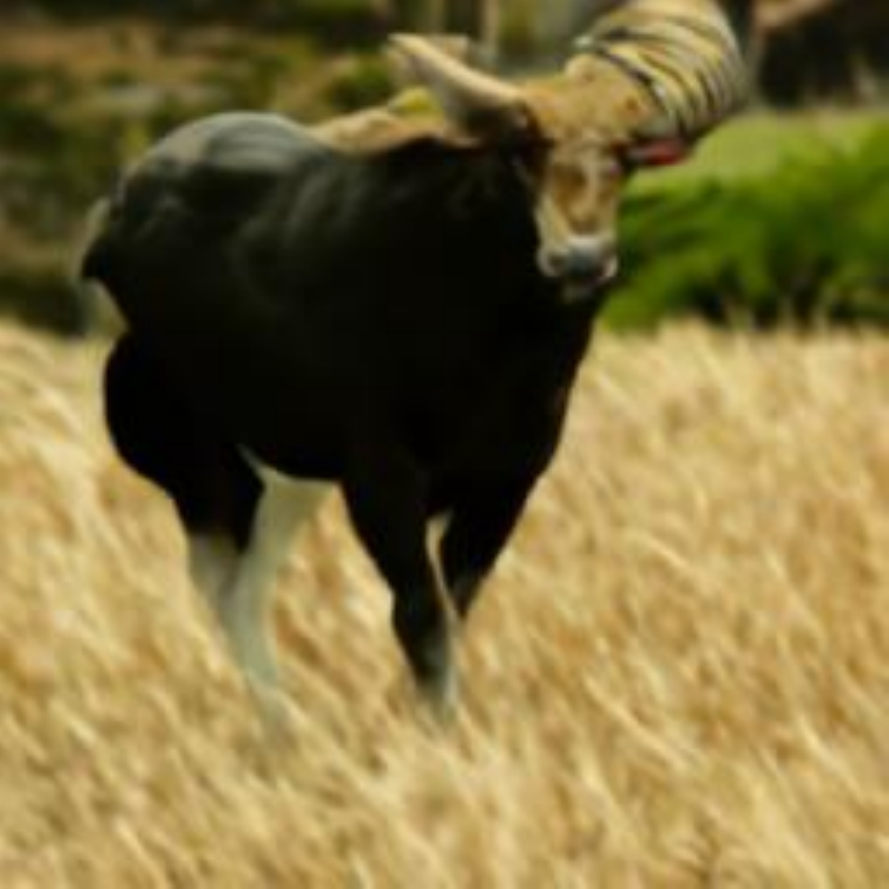} &
        \adjincludegraphics[clip,width=0.11\textwidth,trim={0 0 0 0}]{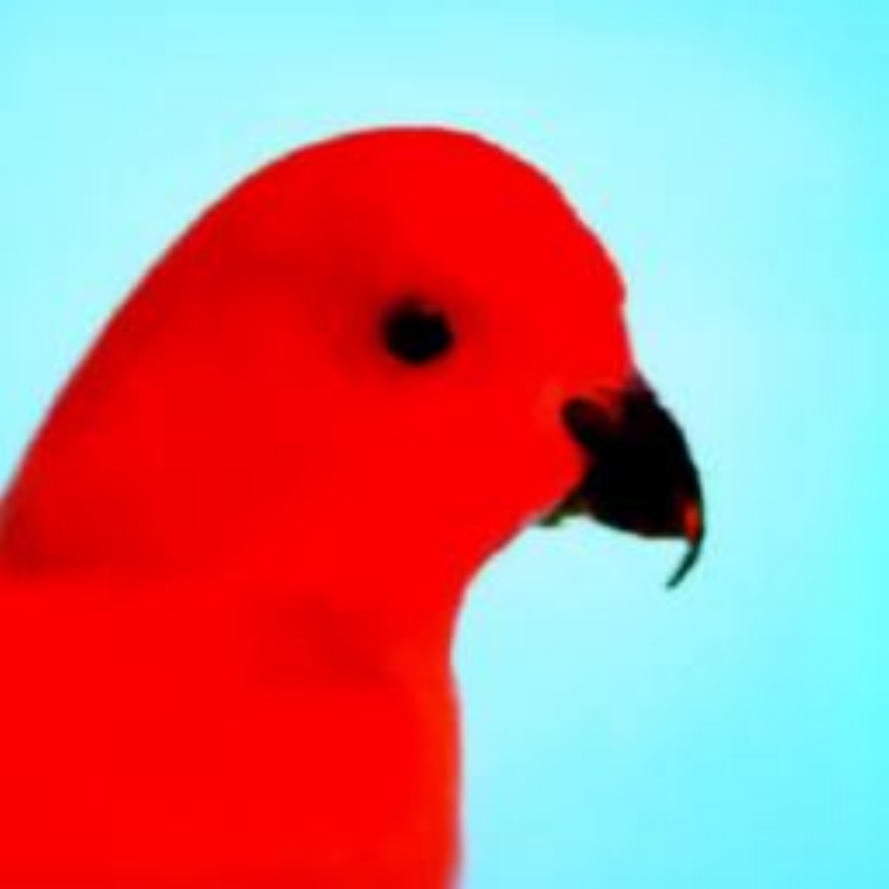} &
        \adjincludegraphics[clip,width=0.11\textwidth,trim={0 0 0 0}]{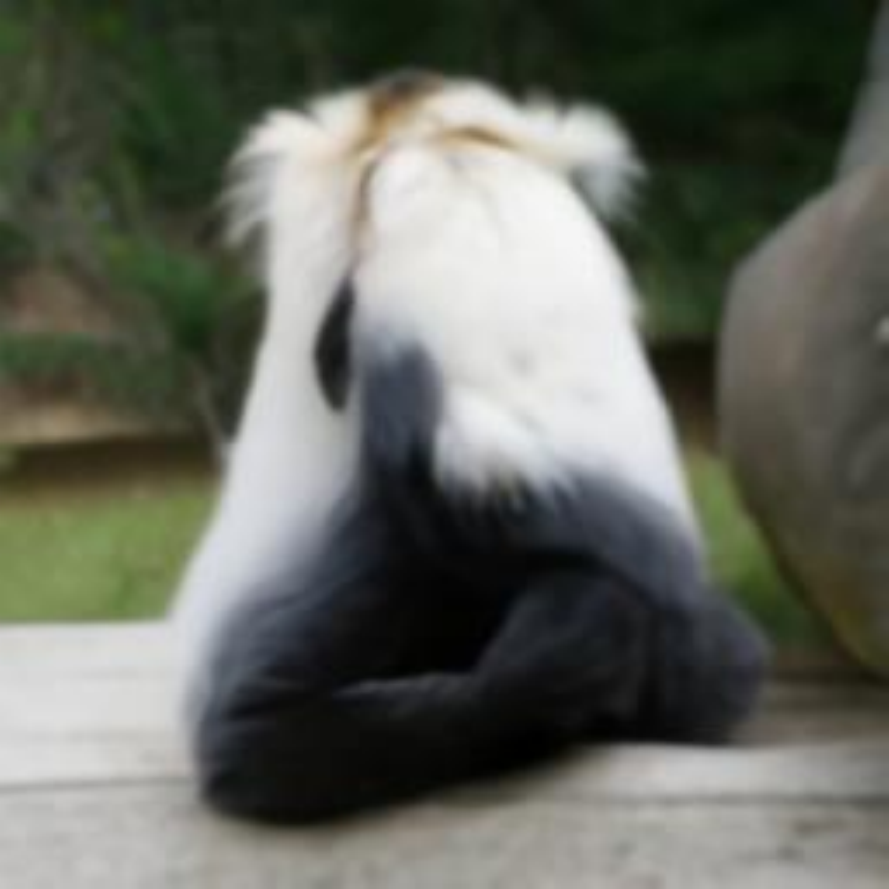} &
        \adjincludegraphics[clip,width=0.11\textwidth,trim={0 0 0 0}]{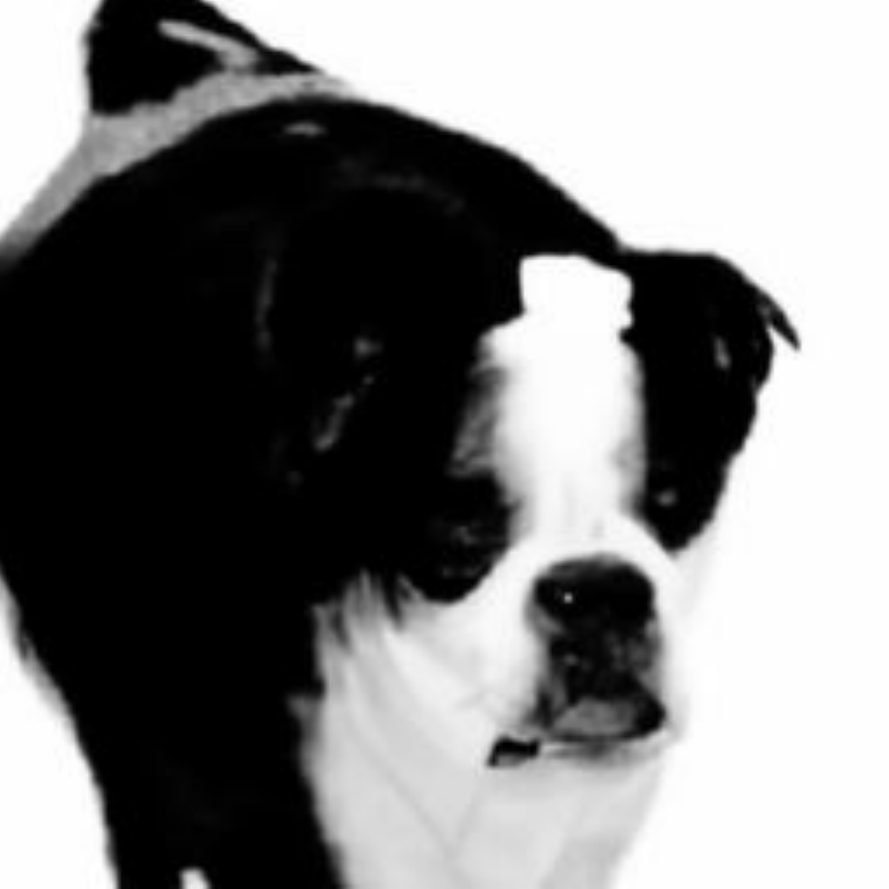} &
        \adjincludegraphics[clip,width=0.11\textwidth,trim={0 0 0 0}]{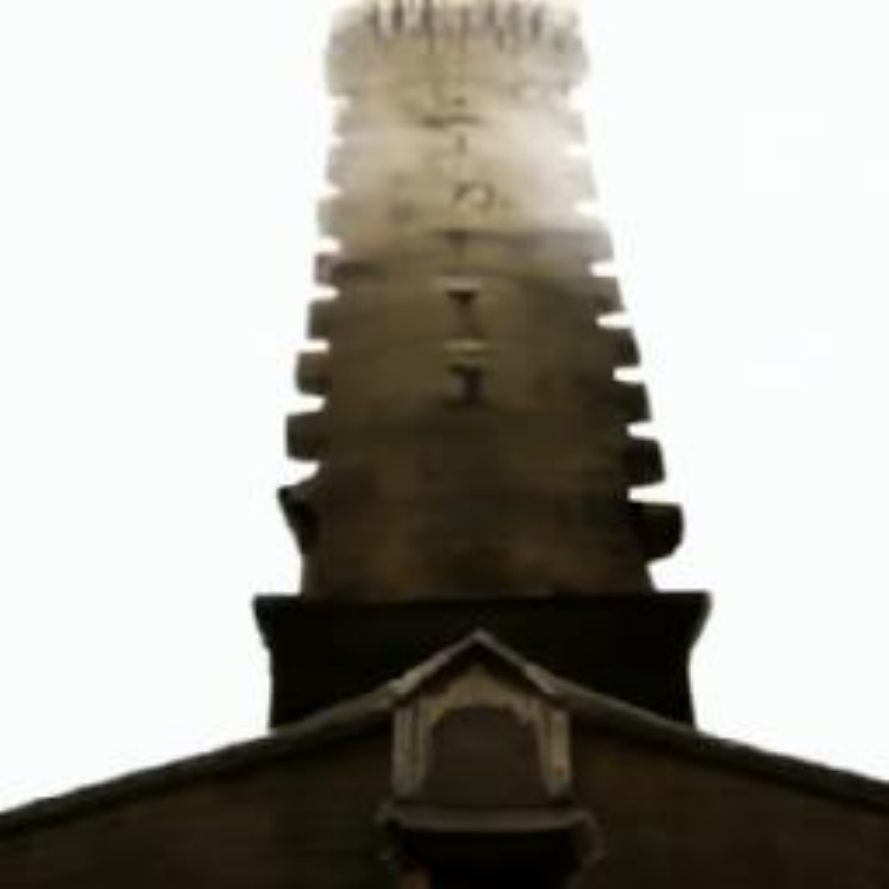} &
        \adjincludegraphics[clip,width=0.11\textwidth,trim={0 0 0 0}]{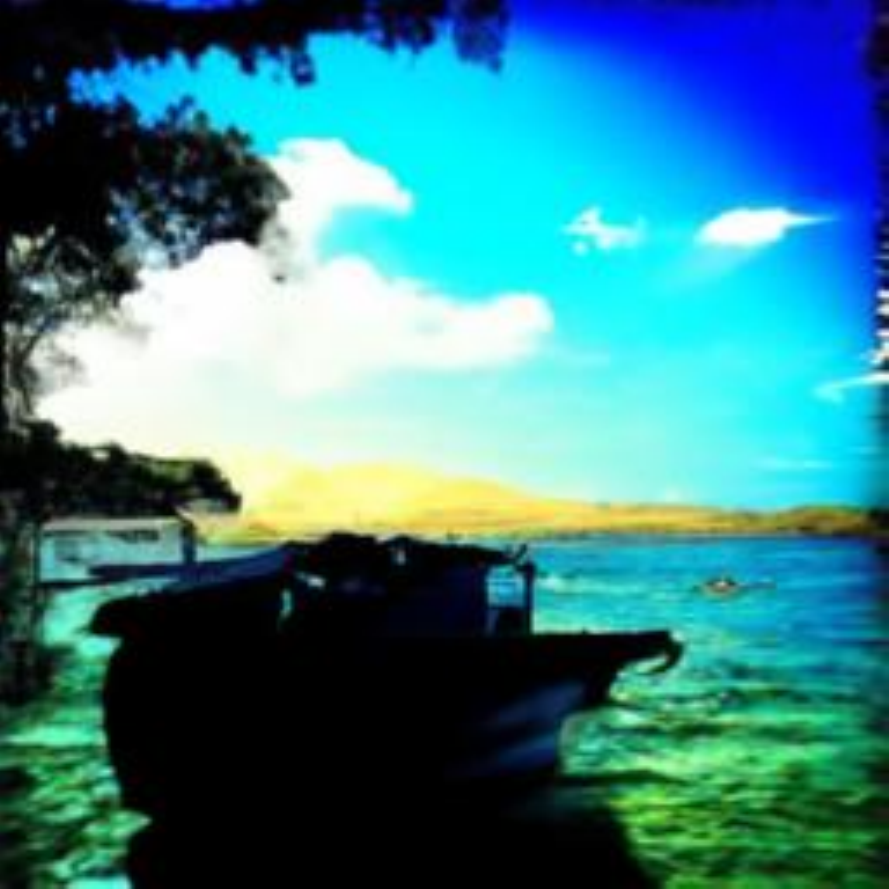} &
        \adjincludegraphics[clip,width=0.11\textwidth,trim={0 0 0 0}]{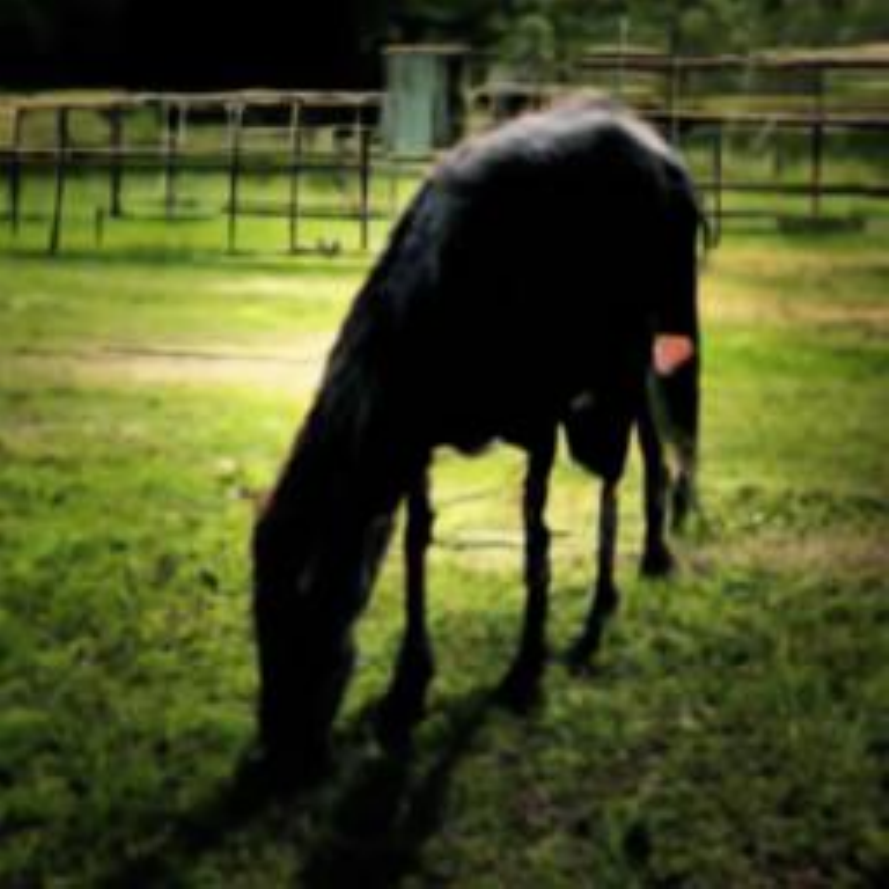} \\

        \raisebox{1.2\height}{\rotatebox{90}{\scriptsize Na\"ive}} 
        \adjincludegraphics[clip,width=0.11\textwidth,trim={0 0 0 0}]{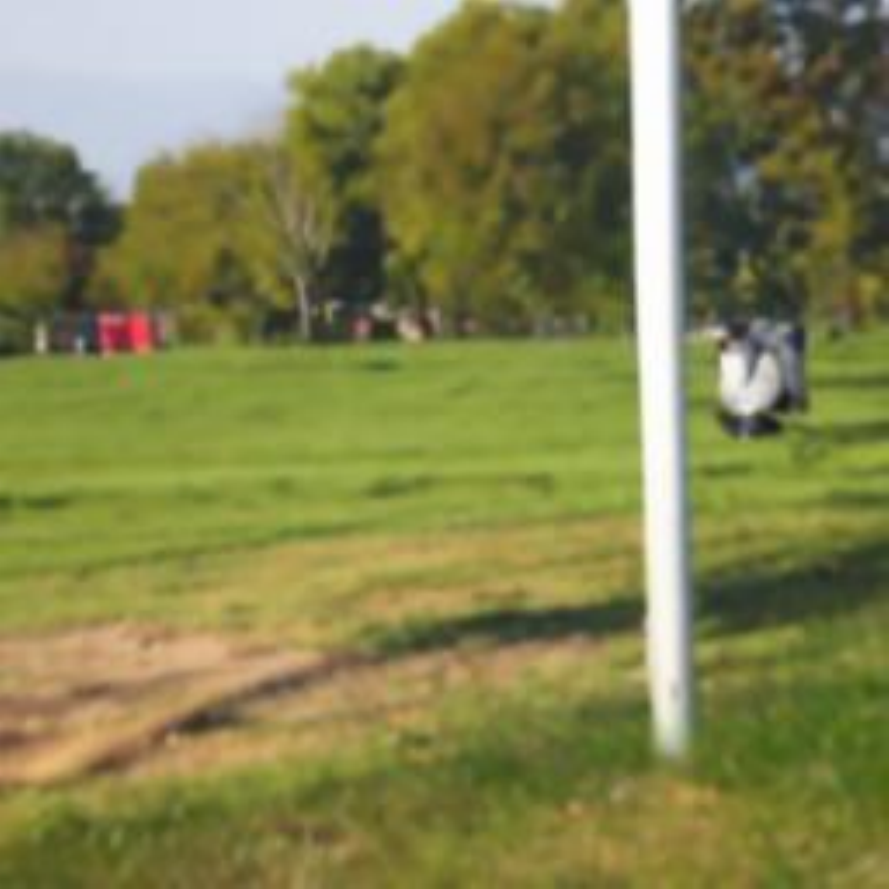} &
        \adjincludegraphics[clip,width=0.11\textwidth,trim={0 0 0 0}]{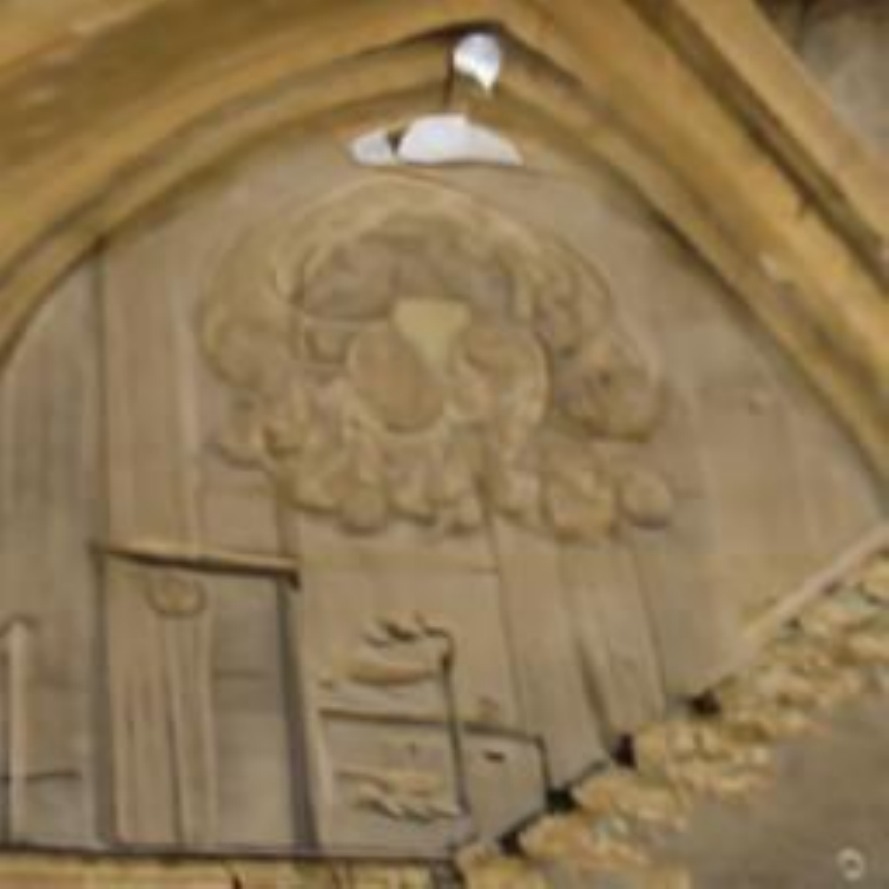} &
        \adjincludegraphics[clip,width=0.11\textwidth,trim={0 0 0 0}]{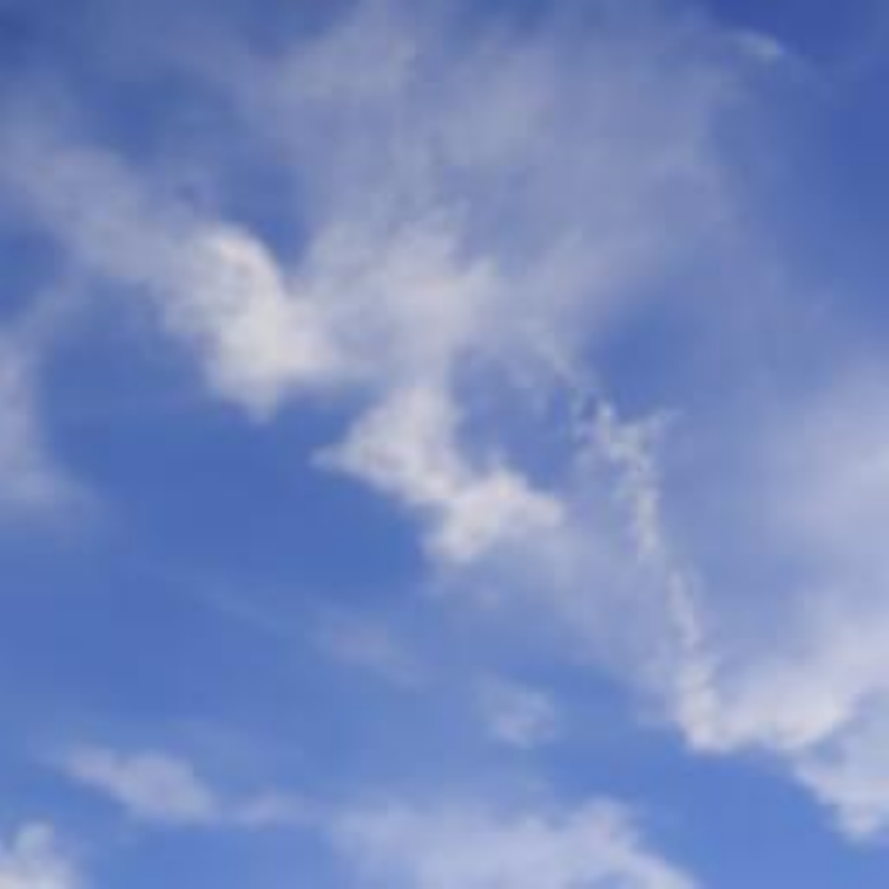} &
        \adjincludegraphics[clip,width=0.11\textwidth,trim={0 0 0 0}]{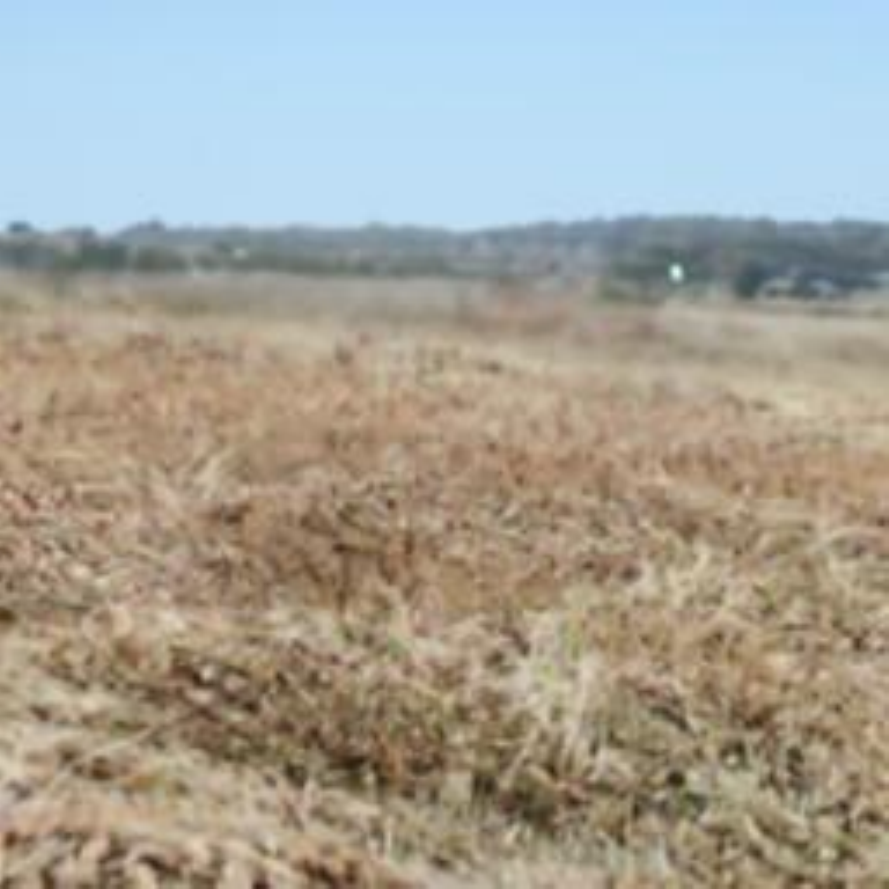} &
        \adjincludegraphics[clip,width=0.11\textwidth,trim={0 0 0 0}]{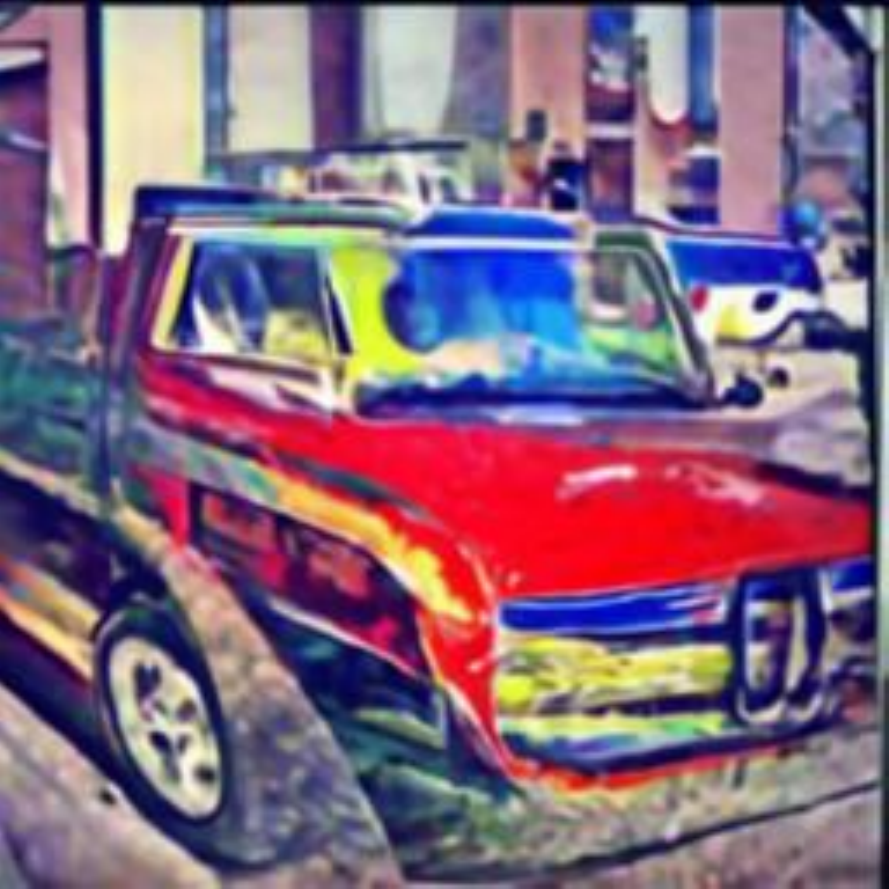} &
        \adjincludegraphics[clip,width=0.11\textwidth,trim={0 0 0 0}]{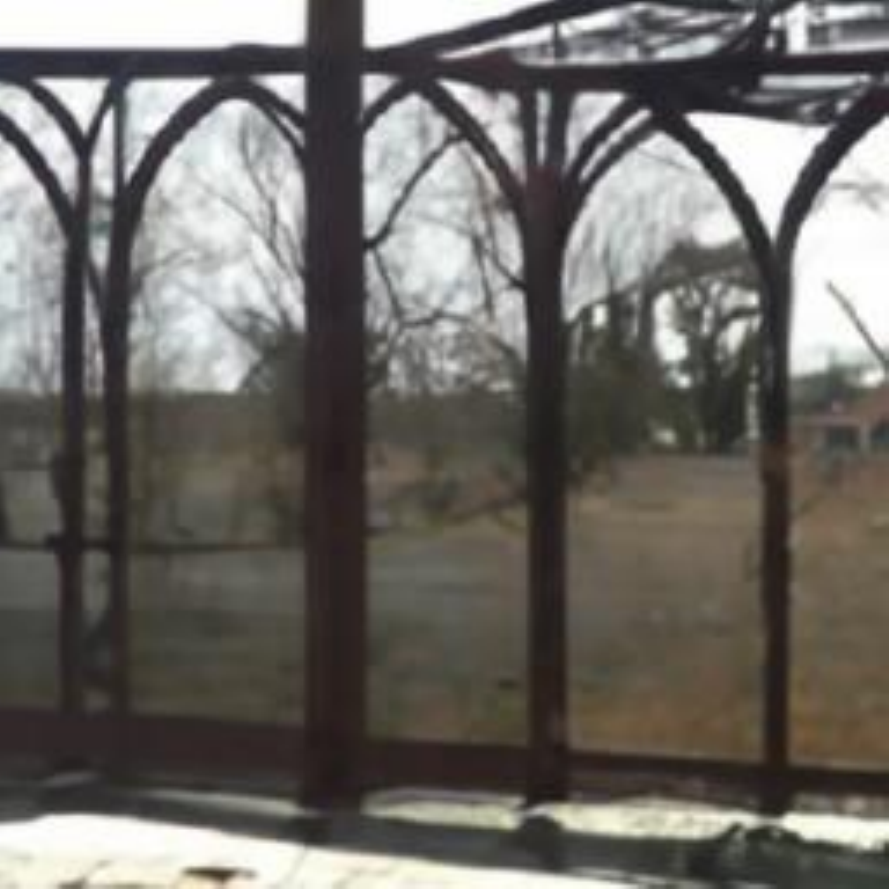} &
        \adjincludegraphics[clip,width=0.11\textwidth,trim={0 0 0 0}]{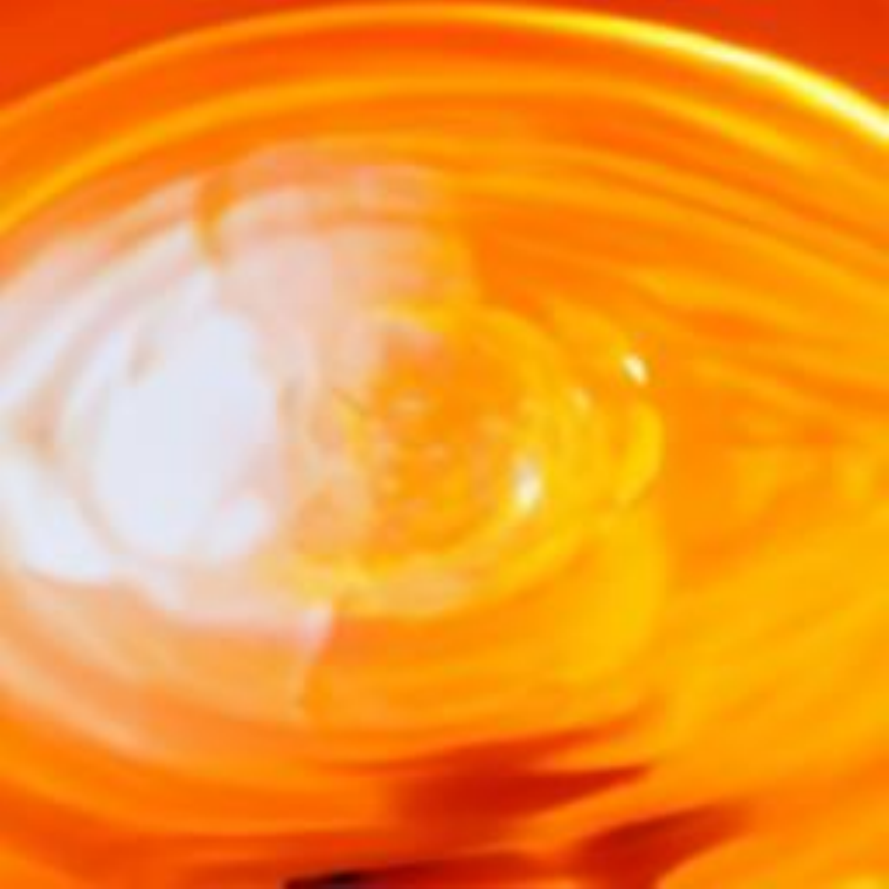} &
        \adjincludegraphics[clip,width=0.11\textwidth,trim={0 0 0 0}]{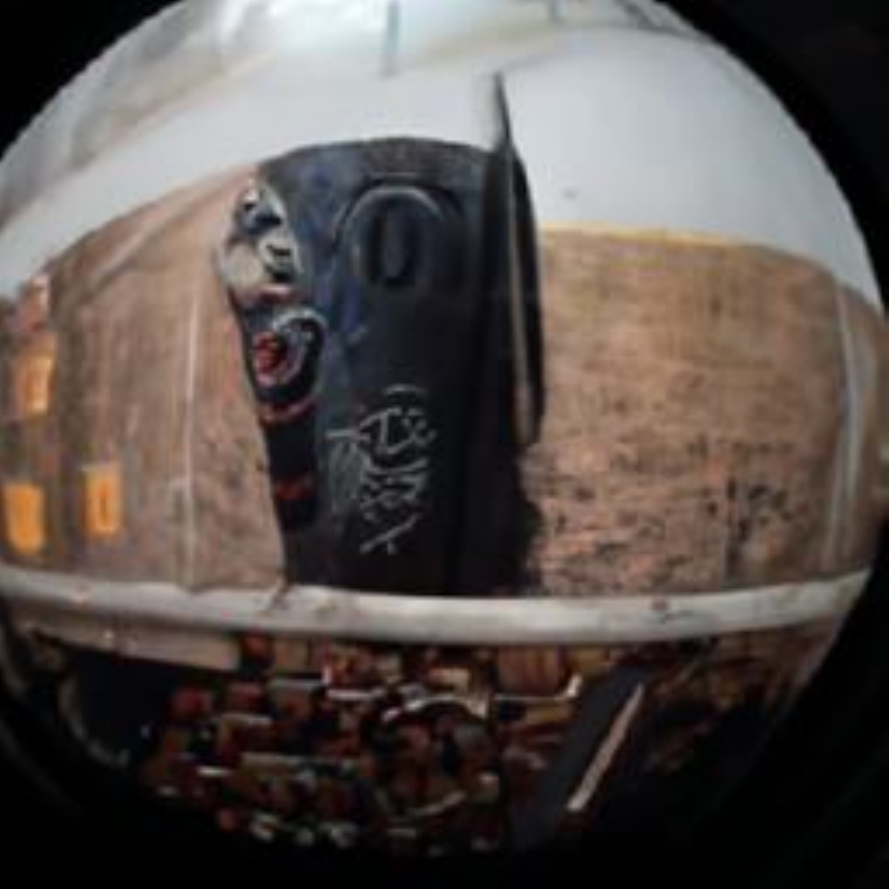} 
    \end{tabular}
    \caption{
    More qualitative results by guiding GLIDE with MiDaS depth estimation. Na\"ive off-the-shelf generates images that do not reflect the guidance with depth, whereas the proposed method succeeds. As shown in the third column, through the proposed framework, the generative model creates similar objects in the original image, but as we can see in the first and second columns, the images are generated by inferring the object only with depth.
    }
    \label{apx:fig_glide_depth}
\end{figure*}

\begin{figure*}[t]

    \centering
    \begin{tabularx}{0.99\textwidth} {>{\centering\arraybackslash}X >{\centering\arraybackslash}X >{\centering\arraybackslash}X >{\centering\arraybackslash}X >{\centering\arraybackslash}X >{\centering\arraybackslash}X >{\centering\arraybackslash}X >{\centering\arraybackslash}X >{\centering\arraybackslash}X >{\centering\arraybackslash}X }
        \scriptsize Original & \scriptsize Depth & \scriptsize PPAP$_1$ & \scriptsize PPAP$_2$ & \scriptsize PPAP$_3$ & \scriptsize PPAP$_4$ & \scriptsize Naive$_1$ & \scriptsize Naive$_2$ & \scriptsize Naive$_3$ & \scriptsize Naive$_4$ \\
        \multicolumn{10}{c}{\adjincludegraphics[clip,width=0.97\textwidth,trim={0 0 0 0}]{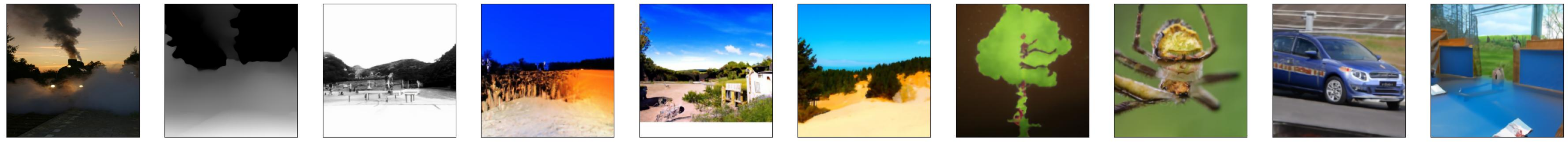} \vspace{-1mm}} \\
        \multicolumn{10}{c}{\adjincludegraphics[clip,width=0.97\textwidth,trim={0 0 0 0}]{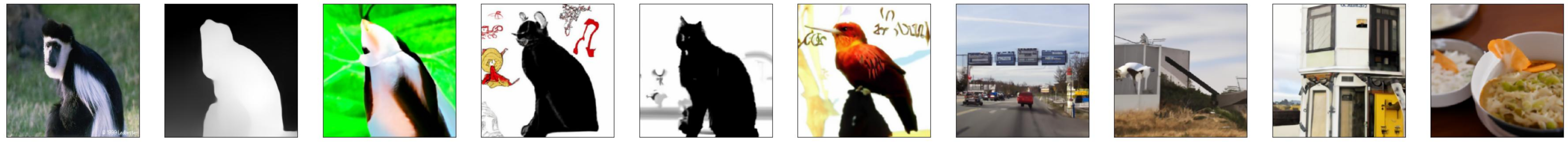} \vspace{-1mm}} \\
        \multicolumn{10}{c}{\adjincludegraphics[clip,width=0.97\textwidth,trim={0 0 0 0}]{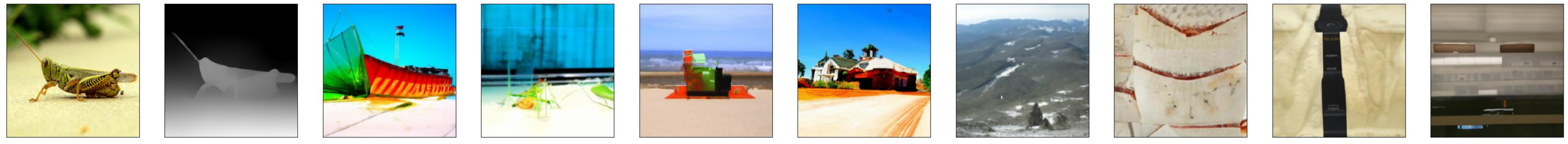} \vspace{-1mm}} \\
        \multicolumn{10}{c}{\adjincludegraphics[clip,width=0.97\textwidth,trim={0 0 0 0}]{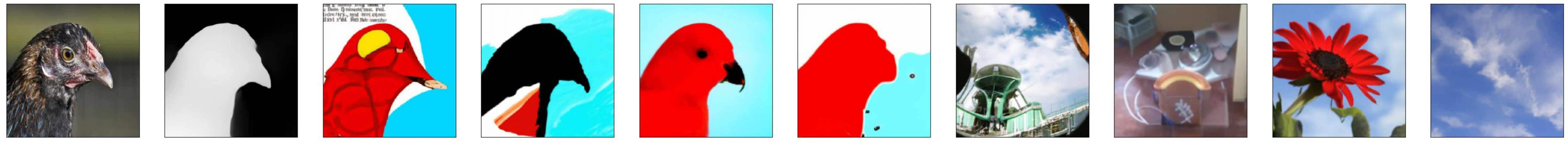} \vspace{-1mm}} \\
        \multicolumn{10}{c}{\adjincludegraphics[clip,width=0.97\textwidth,trim={0 0 0 0}]{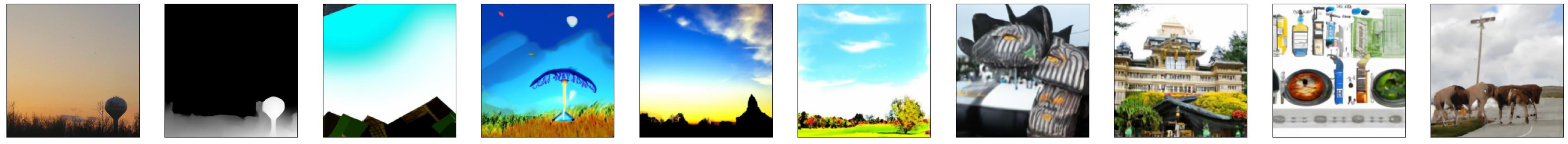} \vspace{-1mm}} \\
        \multicolumn{10}{c}{\adjincludegraphics[clip,width=0.97\textwidth,trim={0 0 0 0}]{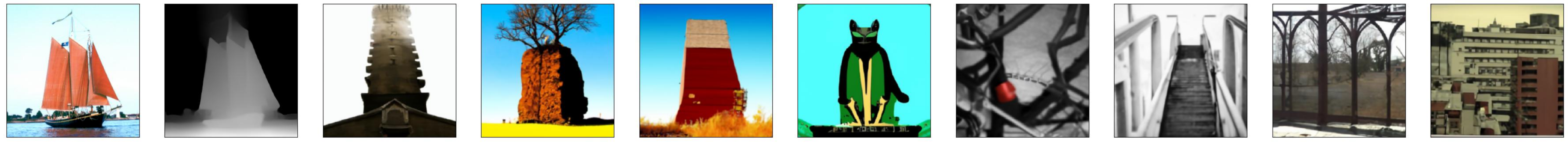} \vspace{-1mm}}
    \end{tabularx}
    
    \caption{
    Qualitative results of in-batch variations by guiding GLIDE with MiDaS depth estimation. Our PPAP framework can generate proper images corresponding to the given depth, but off-the-shelf fails. PPAP generates not only various views of objects with a given depth as in the fourth row, but also diverse objects shown in the sixth row.
    }
    \label{apx:fig_glide_depth2}

\end{figure*}

We provide more qualitative results in guiding GLIDE with depth estimator in Fig.~\ref{apx:fig_glide_depth} and Fig.~\ref{apx:fig_glide_depth2}. As illustrated in Fig.~\ref{apx:fig_glide_depth}, compared to off-the-shelf that does not reflect the given depth at all, it is confirmed that the proposed PPAP framework works well as a guide. We also provide multiple results from the same depth map, shown in Fig.~\ref{apx:fig_glide_depth2}. Interestingly, our framework not only generates the proper images corresponding to the given guidance with depth but also infers the diverse objects that suit the given depth.

\subsection{GLIDE + Segmentation}
\label{apx:subsec:glide+seg}

\begin{figure*}[t]

    \centering
    \begin{tabular}{@{\hspace{0.5mm}}c@{\hspace{0.5mm}}c@{\hspace{0.5mm}}c@{\hspace{0.5mm}}c@{\hspace{0.5mm}}c@{\hspace{0.5mm}}c@{\hspace{0.5mm}}c@{\hspace{0.5mm}}c}
        \raisebox{0.5\height}{\rotatebox{90}{\scriptsize Original}} 
        \adjincludegraphics[clip,width=0.11\textwidth,trim={0 0 0 0}]{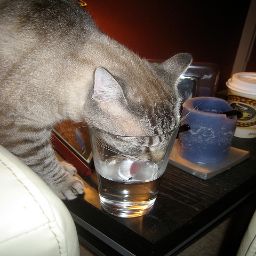} &
        \adjincludegraphics[clip,width=0.11\textwidth,trim={0 0 0 0}]{figure_files/apx_glide_seg/2/base_image.jpg} &
        \adjincludegraphics[clip,width=0.11\textwidth,trim={0 0 0 0}]{figure_files/apx_glide_seg/4/base_image.jpg} & 
        \adjincludegraphics[clip,width=0.11\textwidth,trim={0 0 0 0}]{figure_files/apx_glide_seg/5/base_image.jpg} &
        \adjincludegraphics[clip,width=0.11\textwidth,trim={0 0 0 0}]{figure_files/apx_glide_seg/6/base_image.jpg} &
        \adjincludegraphics[clip,width=0.11\textwidth,trim={0 0 0 0}]{figure_files/apx_glide_seg/7/base_image.jpg} &
        \adjincludegraphics[clip,width=0.11\textwidth,trim={0 0 0 0}]{figure_files/apx_glide_seg/8/base_image.jpg} &
        \adjincludegraphics[clip,width=0.11\textwidth,trim={0 0 0 0}]{figure_files/apx_glide_seg/9/base_image.jpg} \\

        \raisebox{0.2\height}{\rotatebox{90}{\scriptsize Segmentation}} 
        \adjincludegraphics[clip,width=0.11\textwidth,trim={0 0 0 0}]{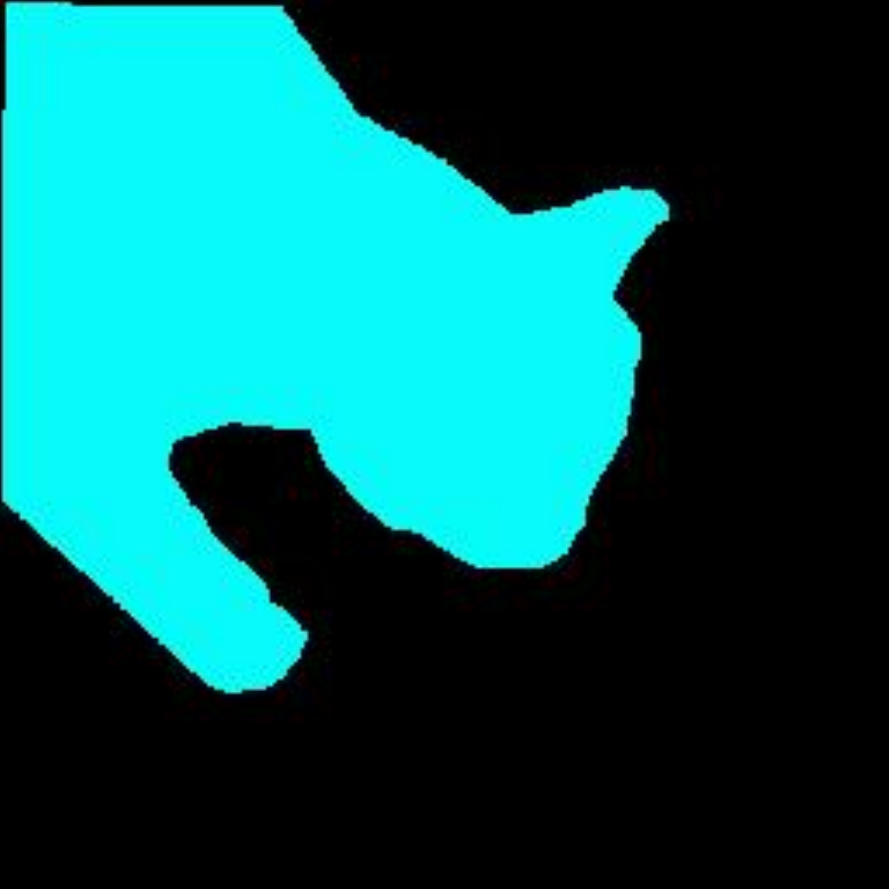} &
        \adjincludegraphics[clip,width=0.11\textwidth,trim={0 0 0 0}]{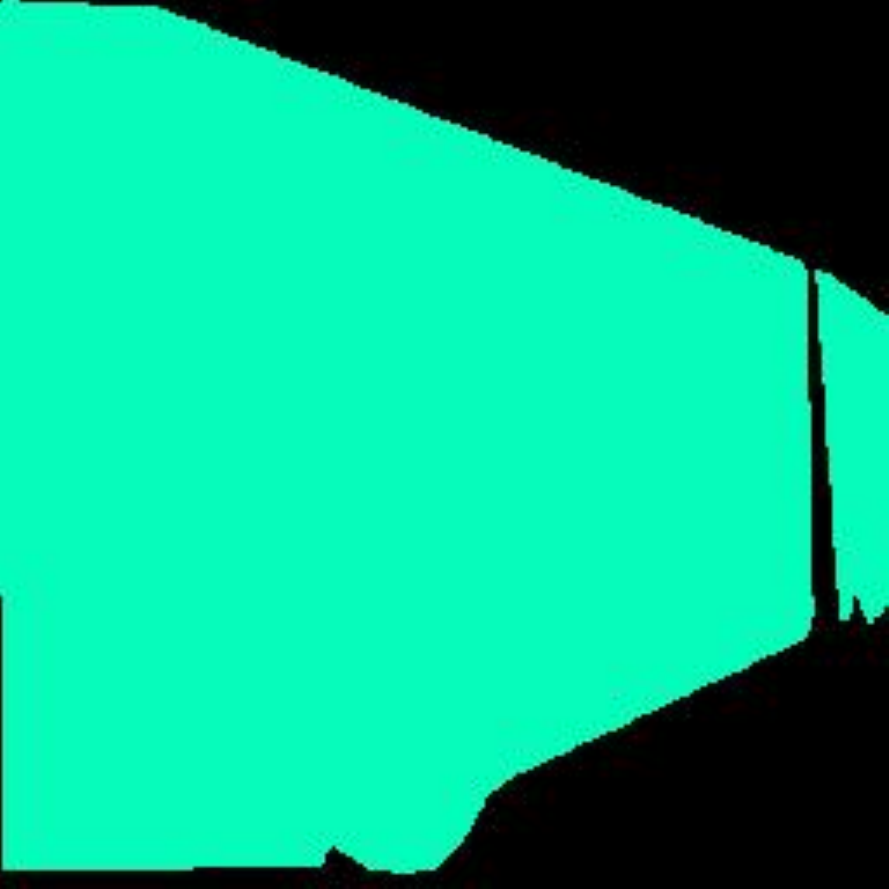} &
        \adjincludegraphics[clip,width=0.11\textwidth,trim={0 0 0 0}]{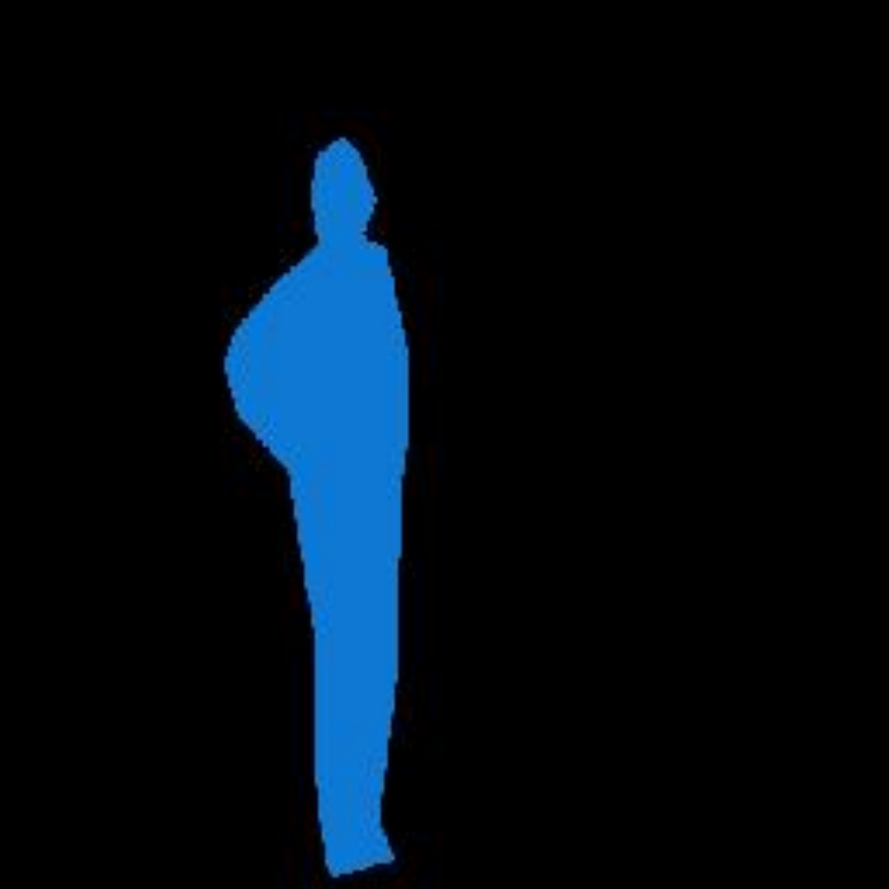} & 
        \adjincludegraphics[clip,width=0.11\textwidth,trim={0 0 0 0}]{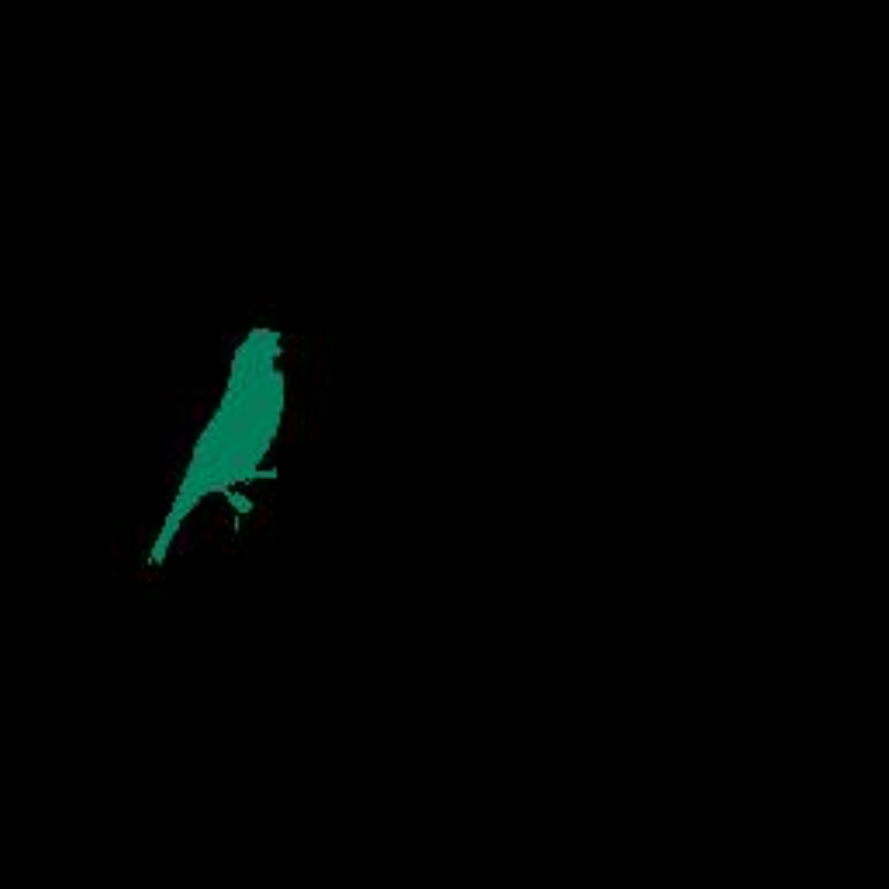} &
        \adjincludegraphics[clip,width=0.11\textwidth,trim={0 0 0 0}]{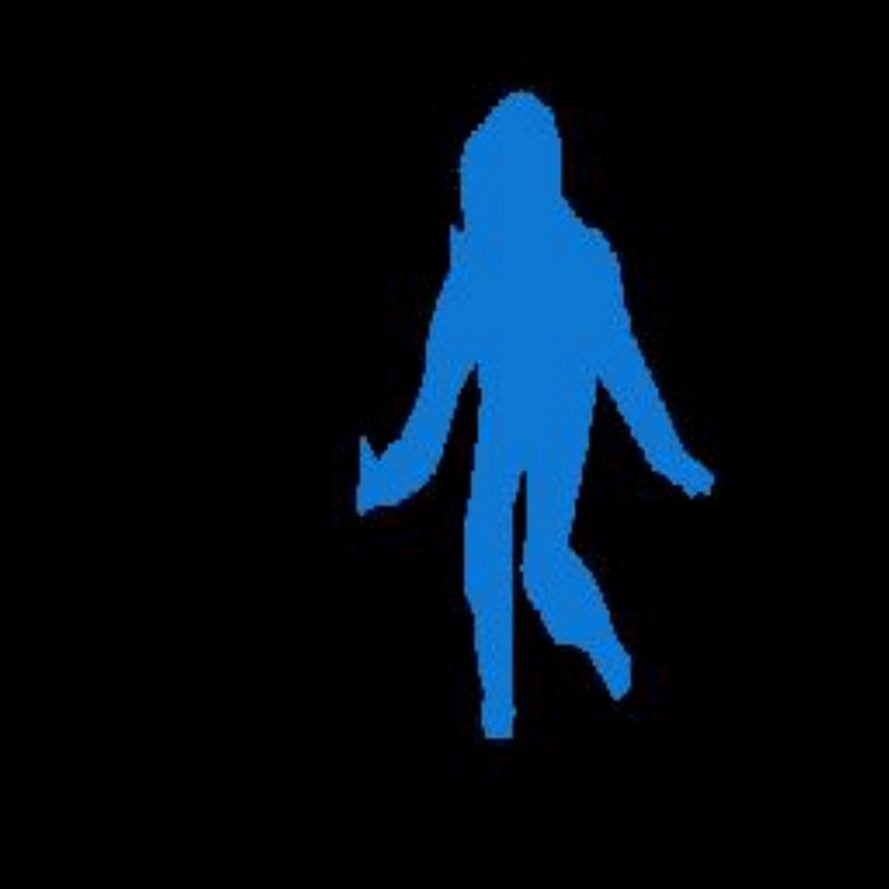} &
        \adjincludegraphics[clip,width=0.11\textwidth,trim={0 0 0 0}]{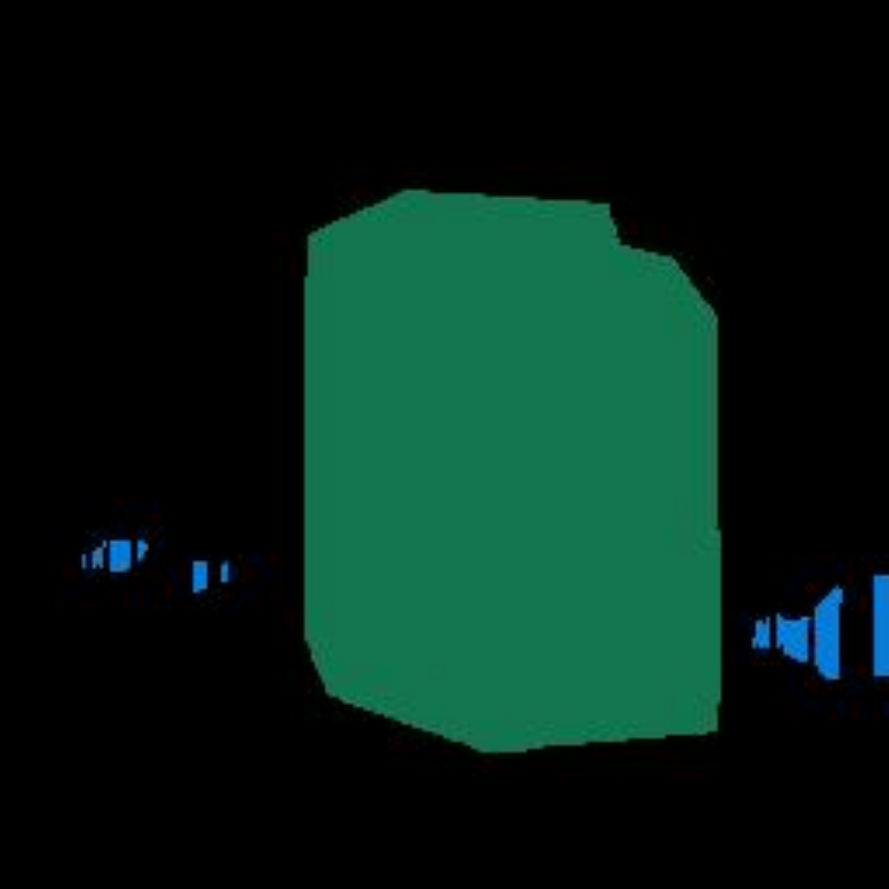} &
        \adjincludegraphics[clip,width=0.11\textwidth,trim={0 0 0 0}]{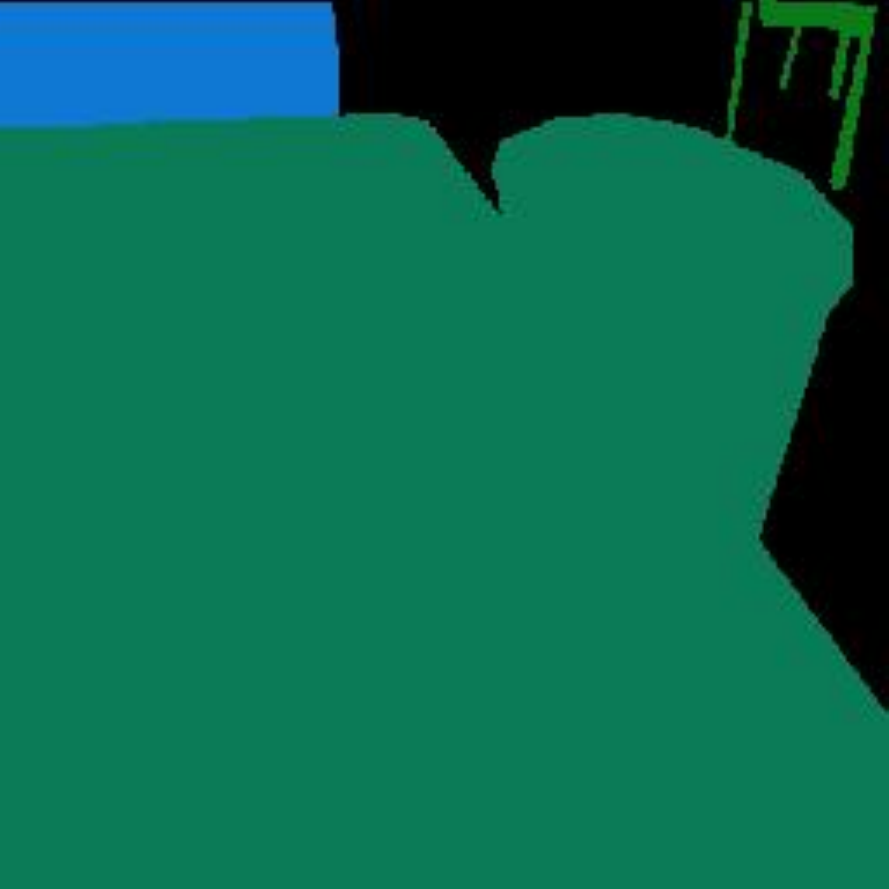} &
        \adjincludegraphics[clip,width=0.11\textwidth,trim={0 0 0 0}]{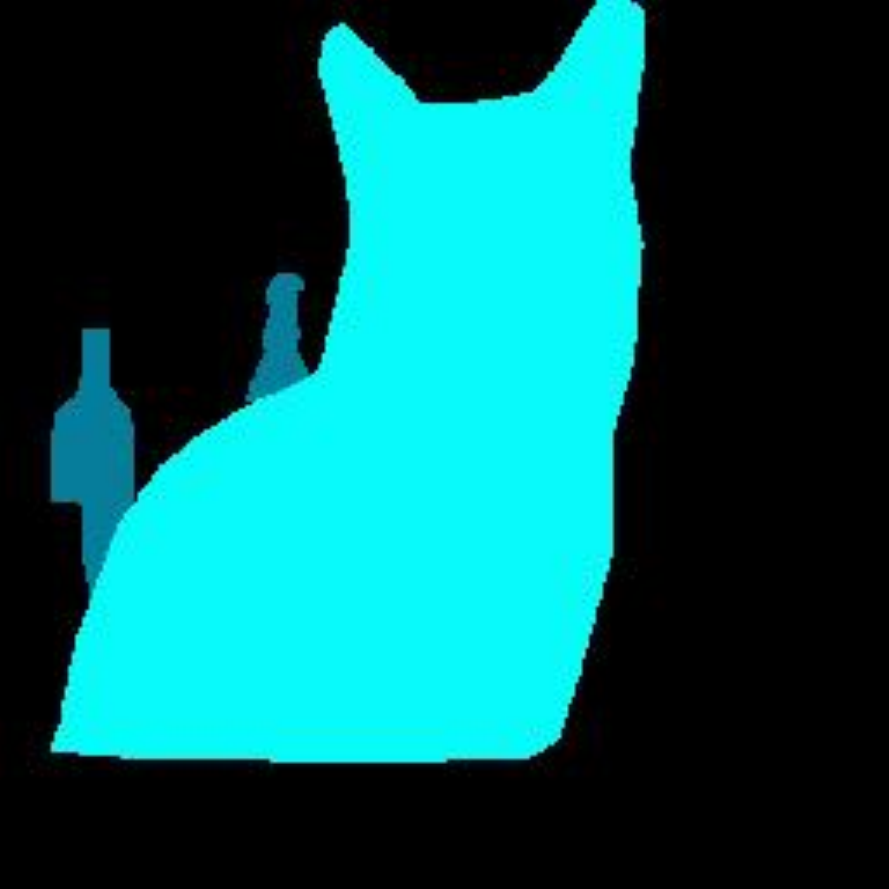} \\

        \raisebox{0.3\height}{\rotatebox{90}{\scriptsize PPAP (Ours)}} 
        \adjincludegraphics[clip,width=0.11\textwidth,trim={0 0 0 0}]{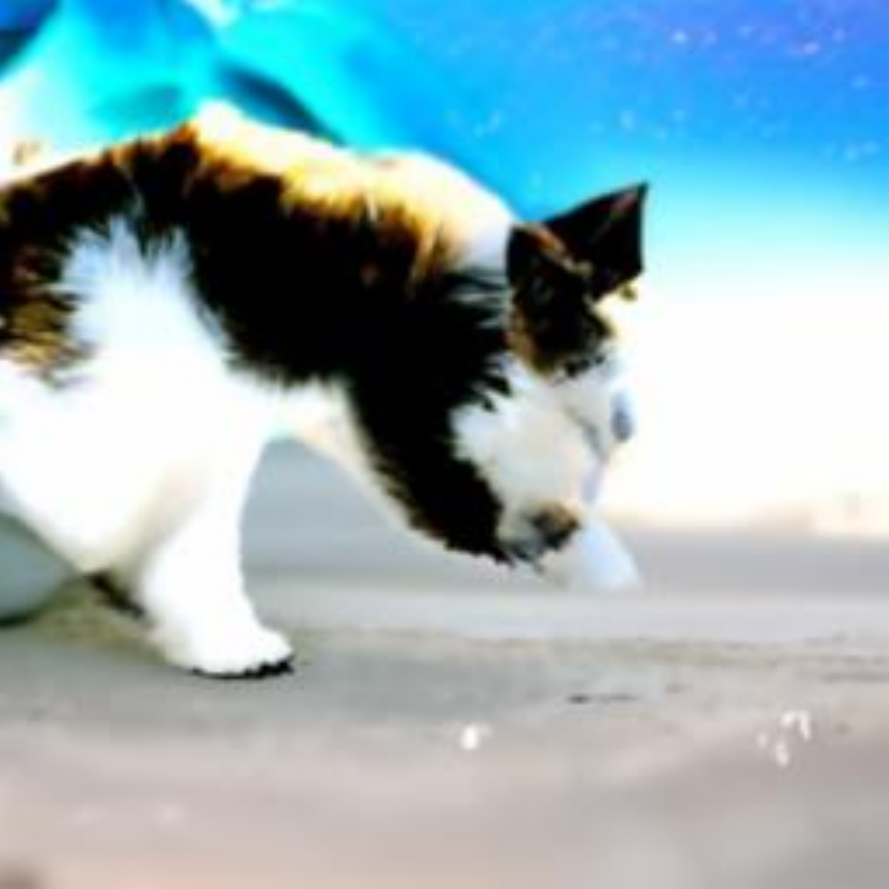} &
        \adjincludegraphics[clip,width=0.11\textwidth,trim={0 0 0 0}]{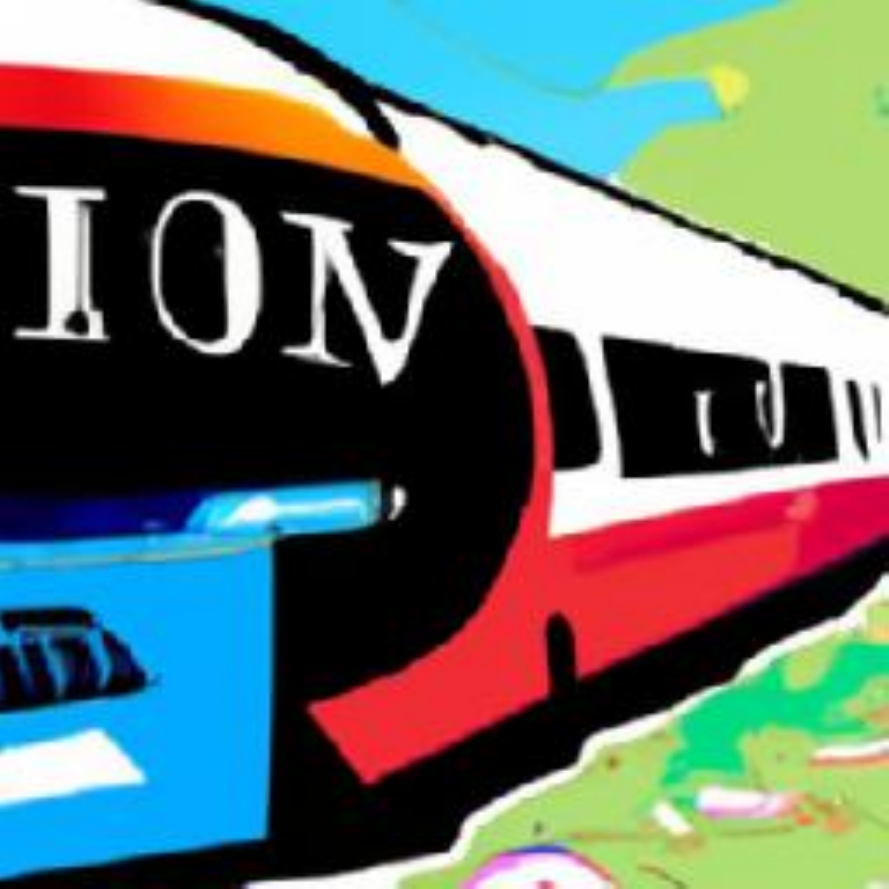} &
        \adjincludegraphics[clip,width=0.11\textwidth,trim={0 0 0 0}]{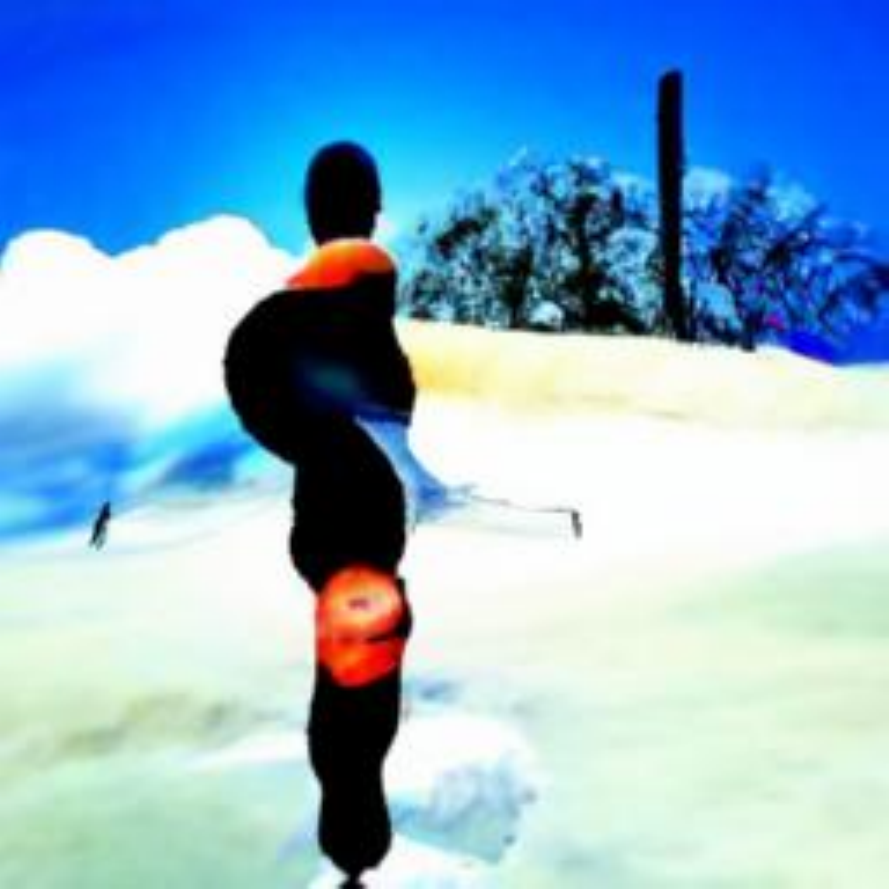} & 
        \adjincludegraphics[clip,width=0.11\textwidth,trim={0 0 0 0}]{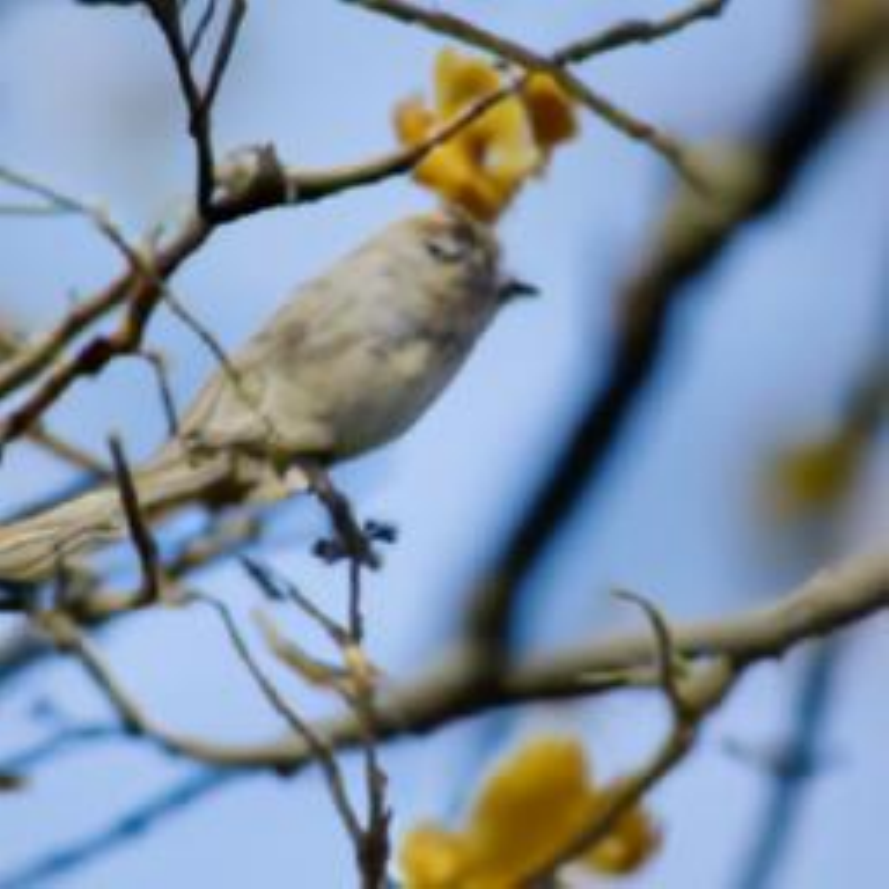} &
        \adjincludegraphics[clip,width=0.11\textwidth,trim={0 0 0 0}]{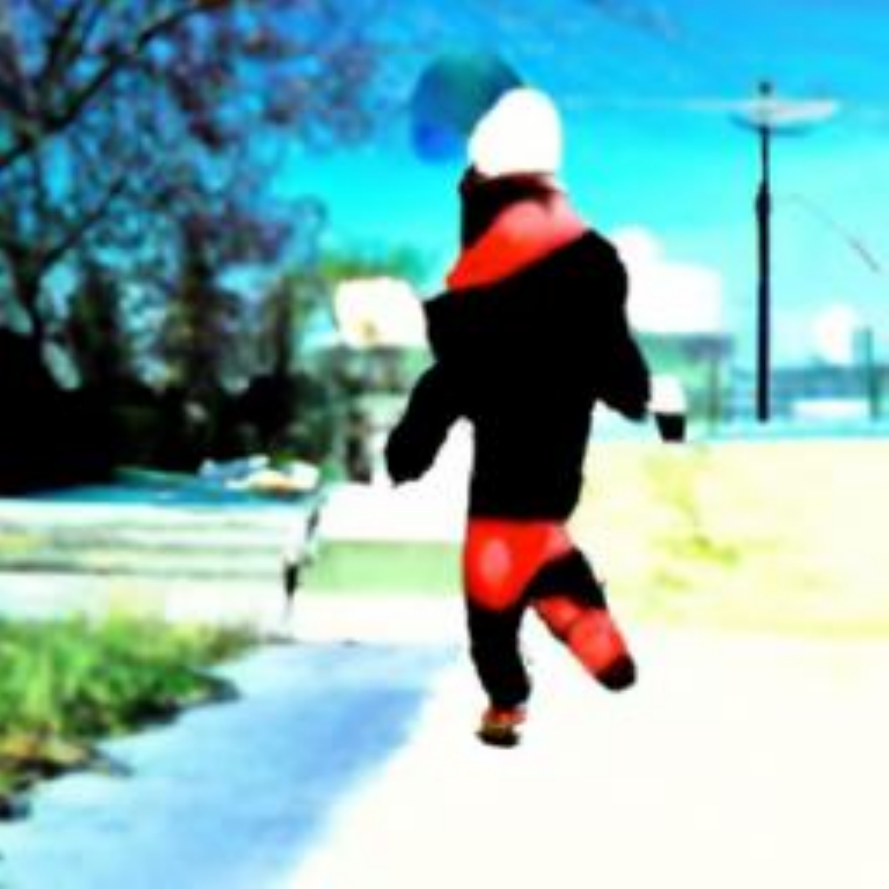} &
        \adjincludegraphics[clip,width=0.11\textwidth,trim={0 0 0 0}]{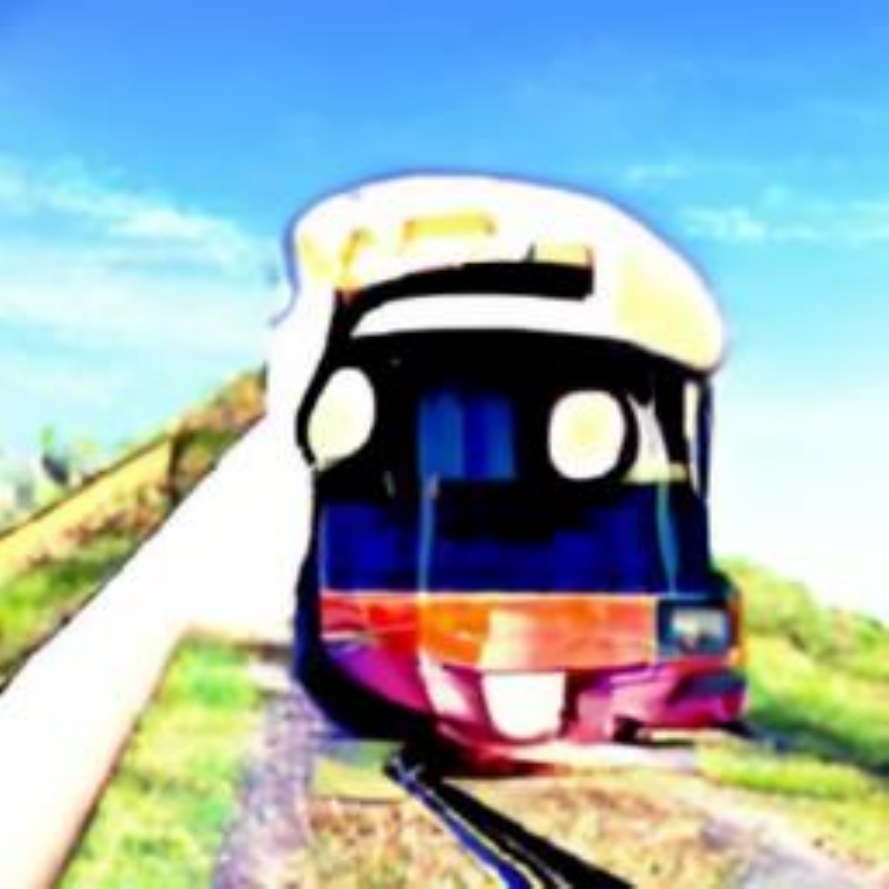} &
        \adjincludegraphics[clip,width=0.11\textwidth,trim={0 0 0 0}]{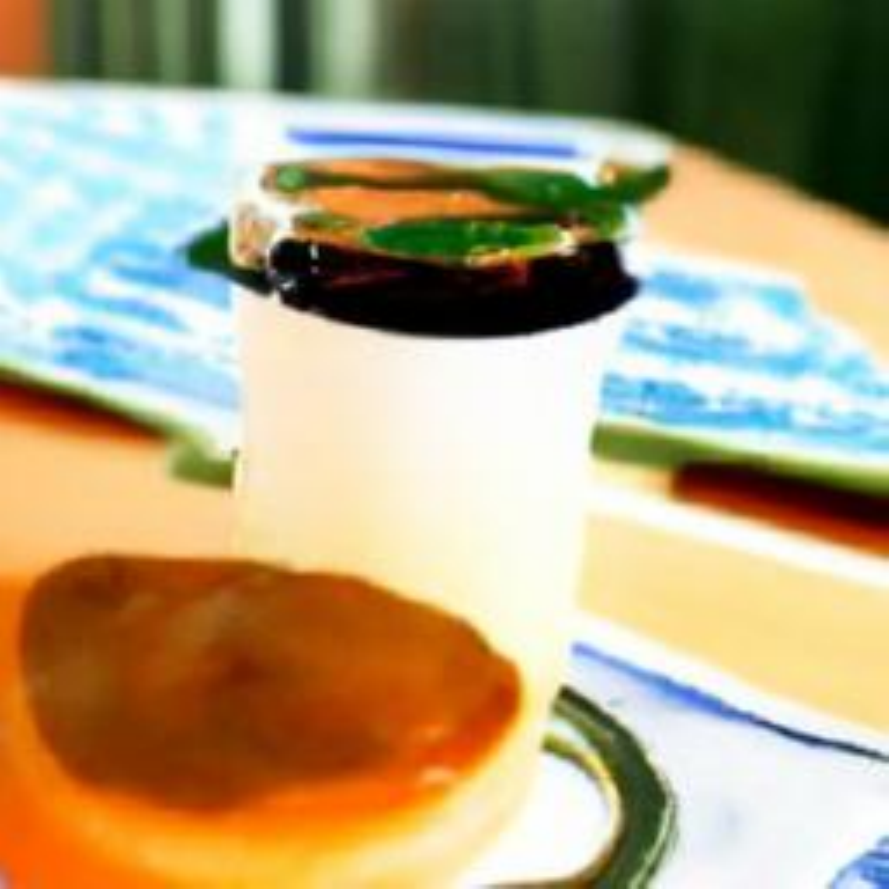} &
        \adjincludegraphics[clip,width=0.11\textwidth,trim={0 0 0 0}]{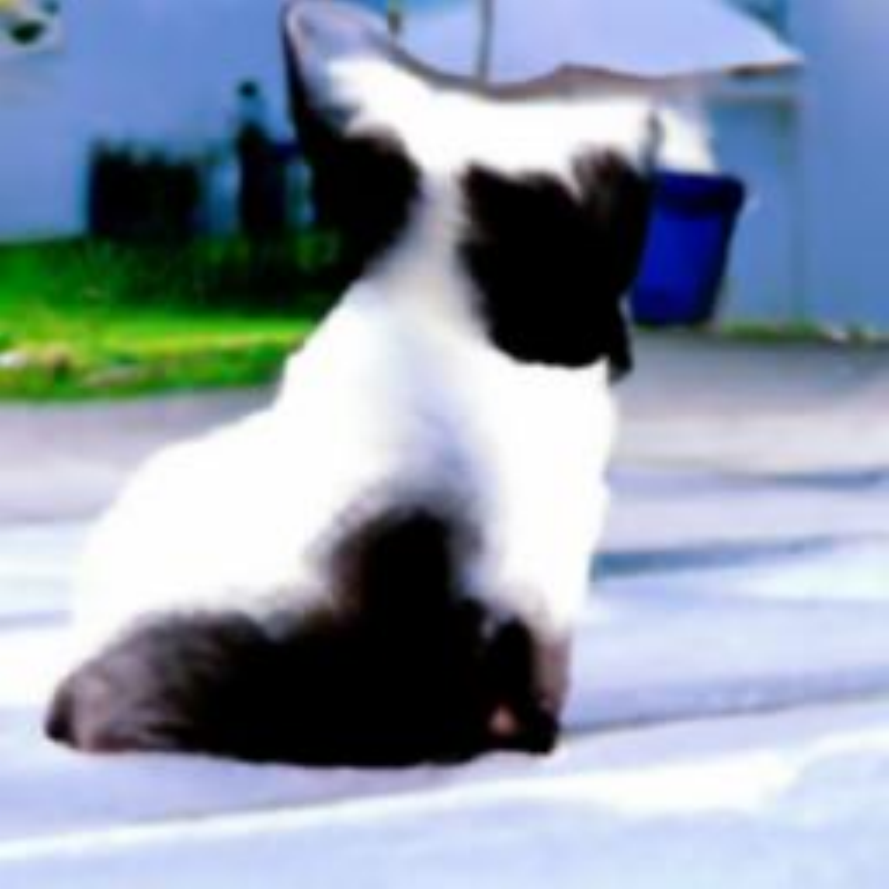} \\

        \raisebox{1.1\height}{\rotatebox{90}{\scriptsize Na\"ive}} 
        \adjincludegraphics[clip,width=0.11\textwidth,trim={0 0 0 0}]{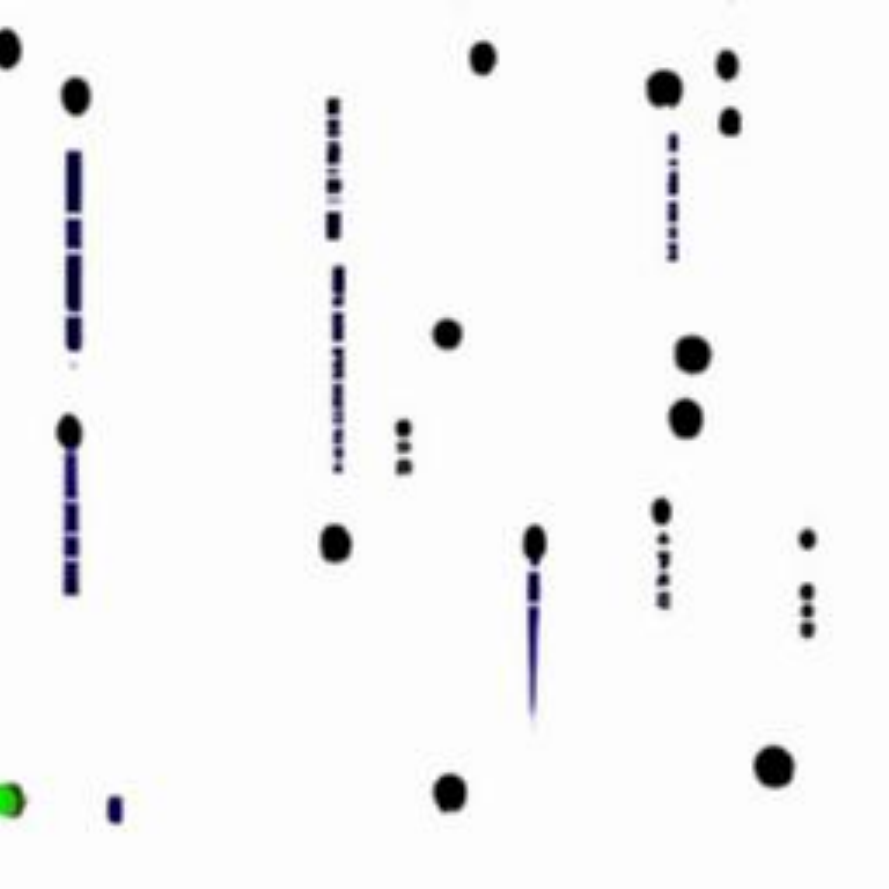} &
        \adjincludegraphics[clip,width=0.11\textwidth,trim={0 0 0 0}]{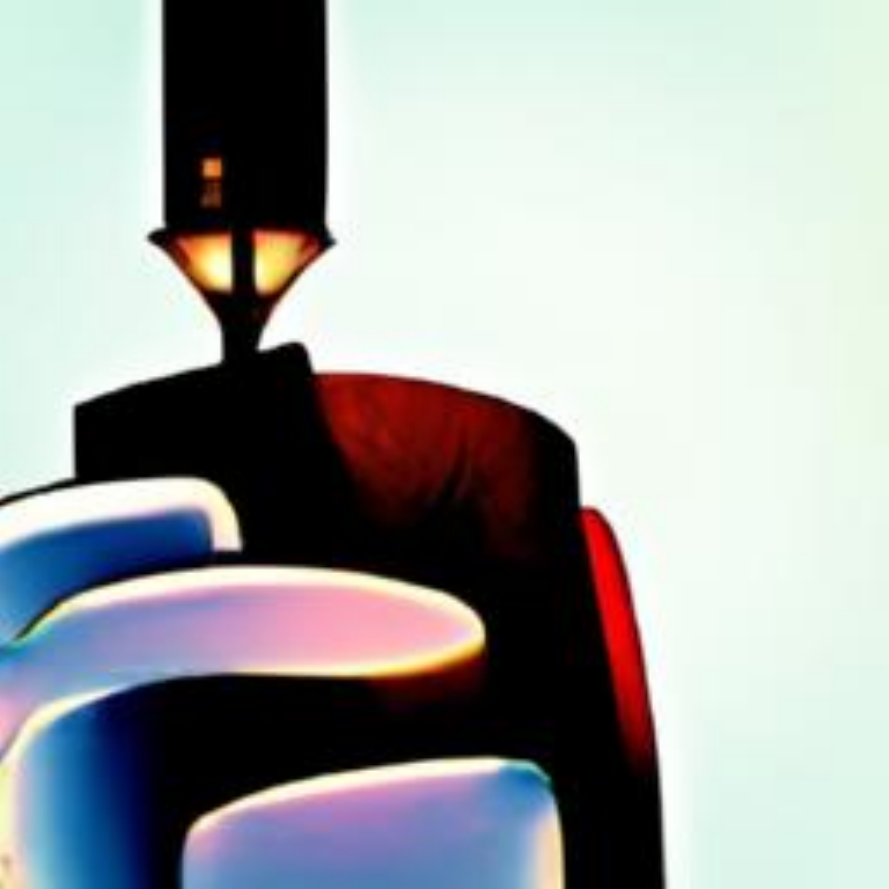} &
        \adjincludegraphics[clip,width=0.11\textwidth,trim={0 0 0 0}]{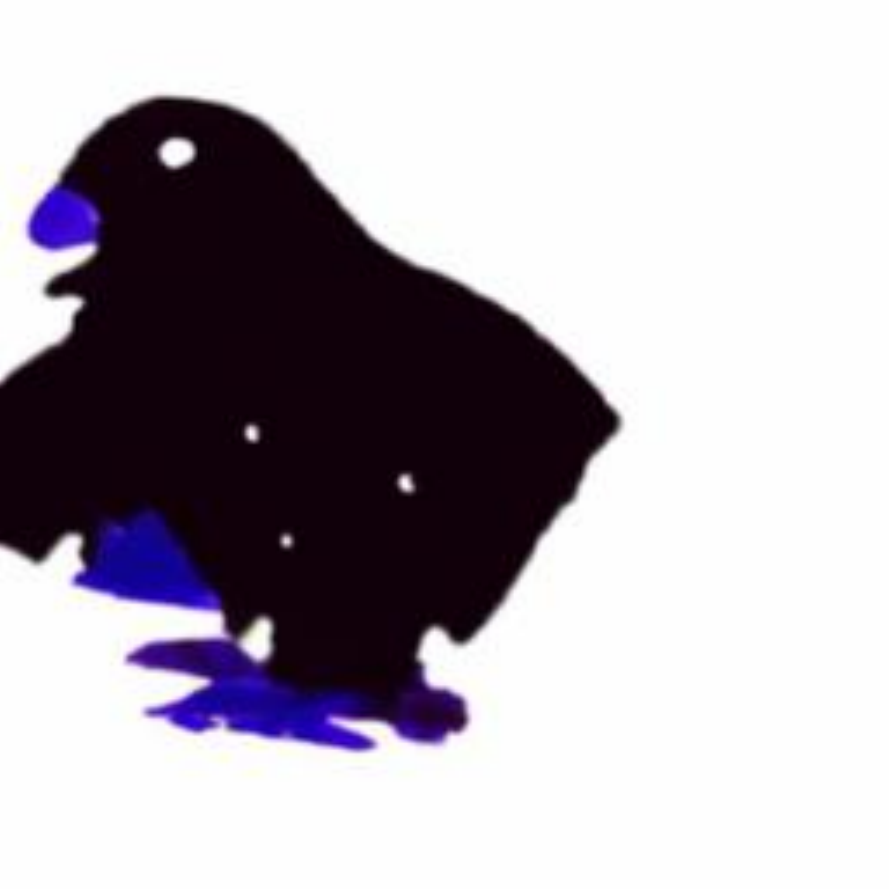} & 
        \adjincludegraphics[clip,width=0.11\textwidth,trim={0 0 0 0}]{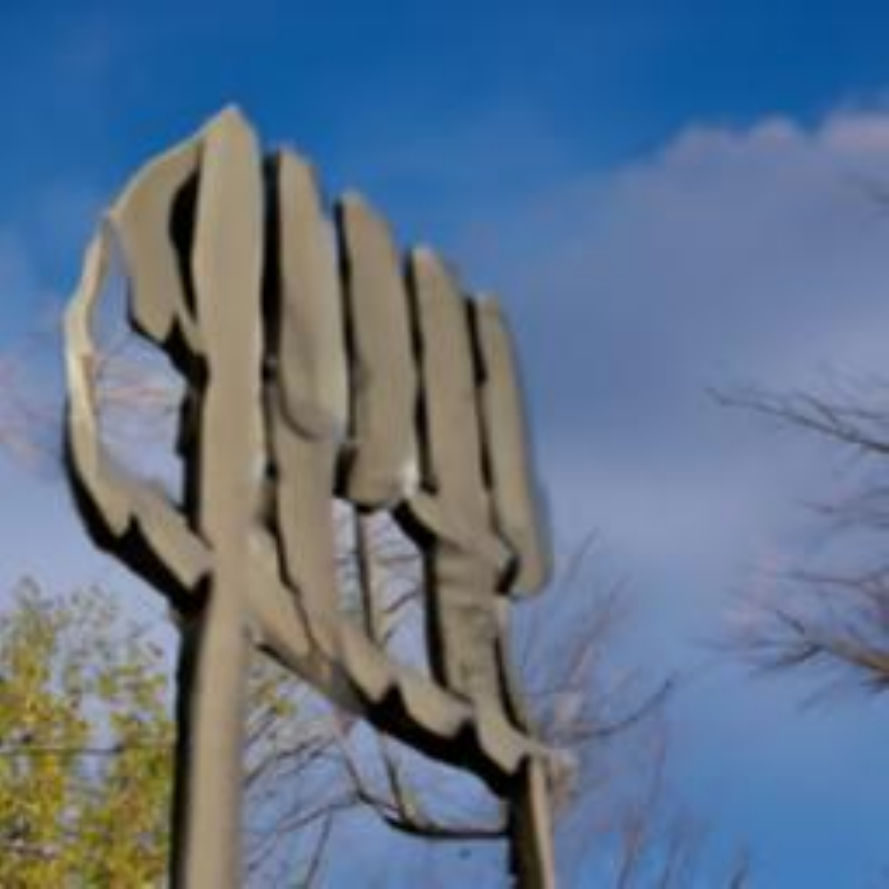} &
        \adjincludegraphics[clip,width=0.11\textwidth,trim={0 0 0 0}]{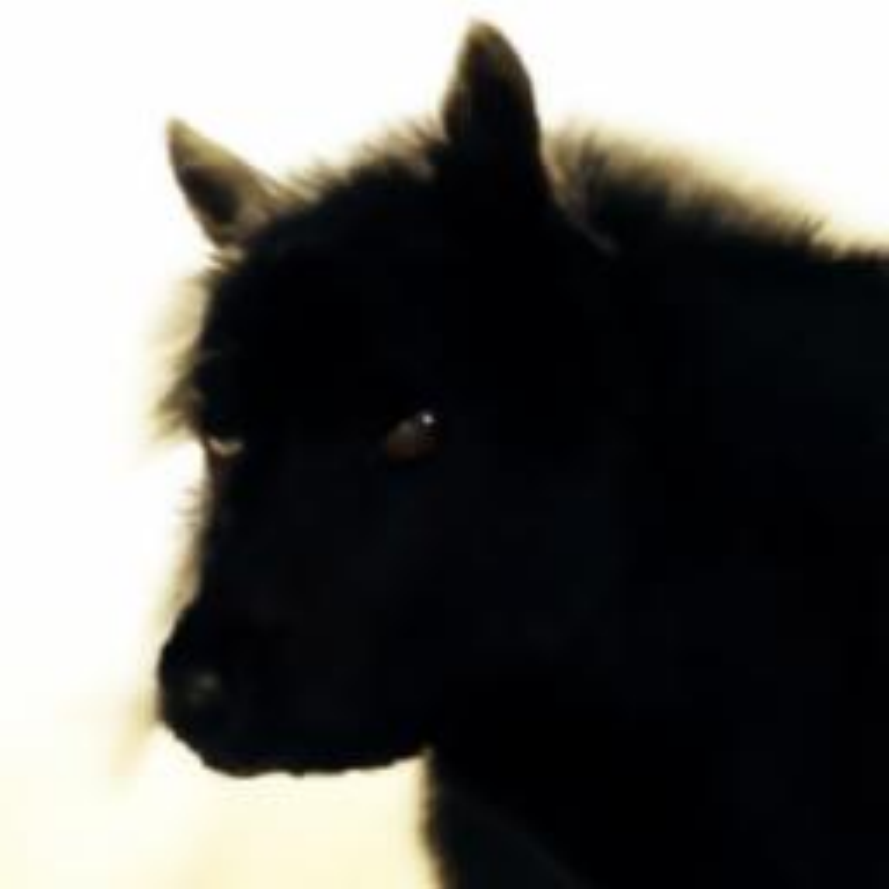} &
        \adjincludegraphics[clip,width=0.11\textwidth,trim={0 0 0 0}]{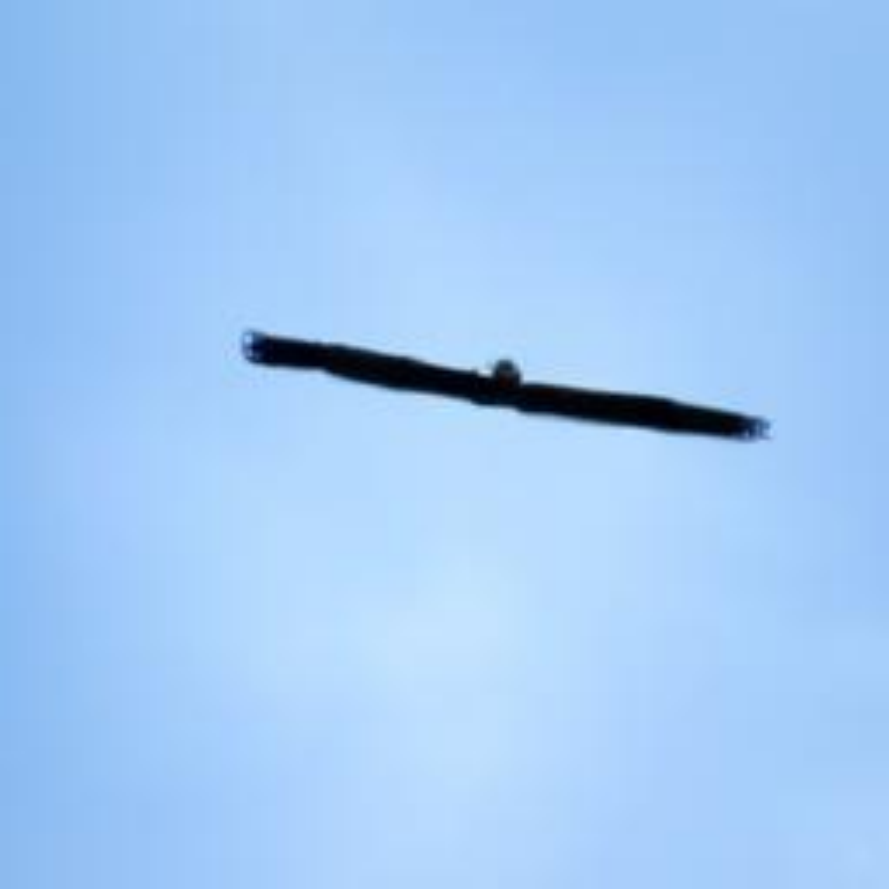} &
        \adjincludegraphics[clip,width=0.11\textwidth,trim={0 0 0 0}]{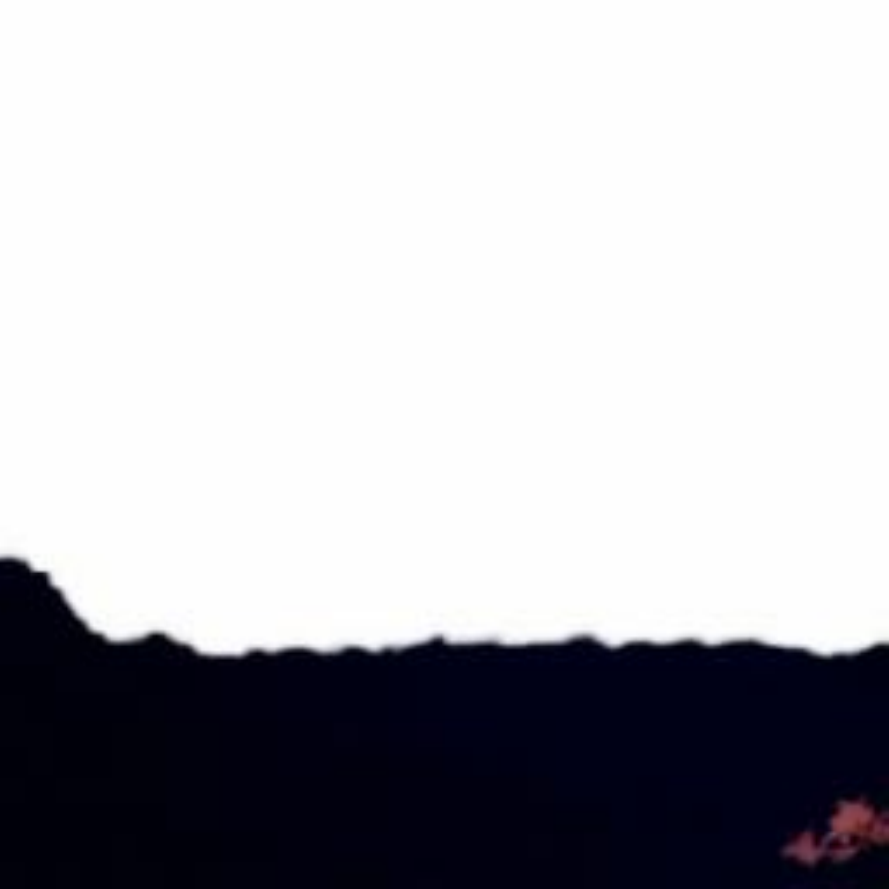} &
        \adjincludegraphics[clip,width=0.11\textwidth,trim={0 0 0 0}]{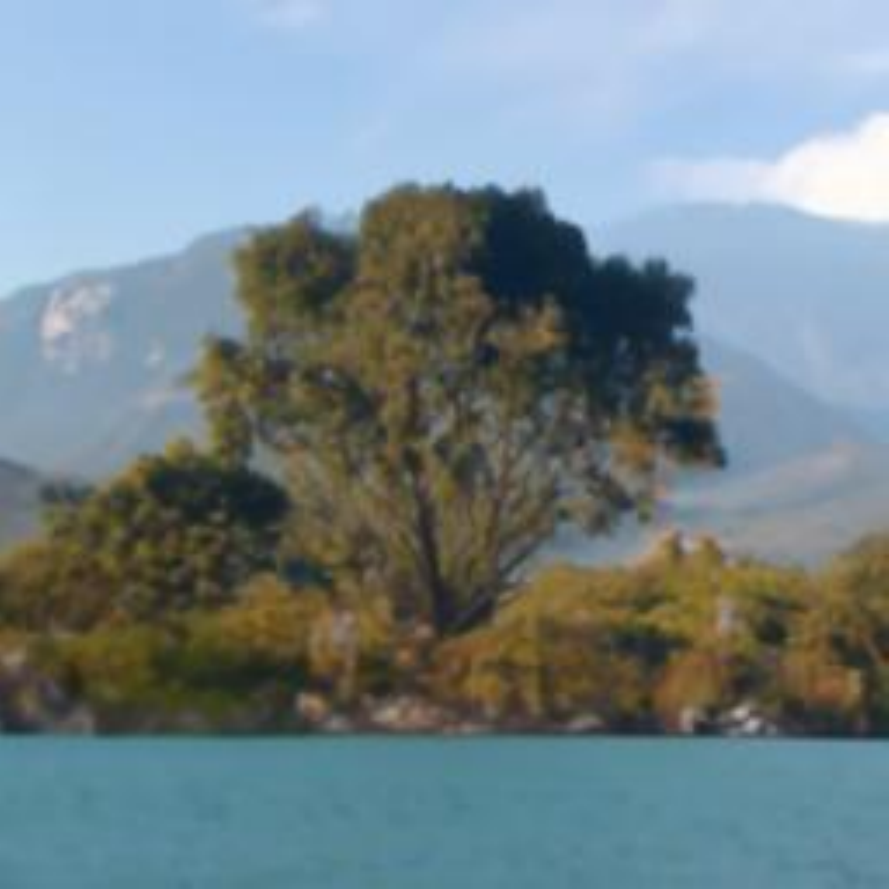} 
    \end{tabular}
    
    \caption{
    More qualitative results by guiding GLIDE with DeepLabv3 semantic segmentation.
    Our framework PPAP succeeds in the guidance of diffusion, but the na\"ive off-the-shelf model fails.
    }
    \label{apx:fig_glide_seg}

\end{figure*}

\begin{figure*}[t]

    \centering
    \begin{tabular}{@{\hspace{0.5mm}}c@{\hspace{0.5mm}}c@{\hspace{0.5mm}}c|c@{\hspace{0.5mm}}c@{\hspace{0.5mm}}c|c@{\hspace{0.5mm}}c@{\hspace{0.5mm}}c}
        \raisebox{1.4\height}{\rotatebox{90}{\scriptsize Bird}} 
        \adjincludegraphics[clip,width=0.1\textwidth,trim={0 0 0 0}]{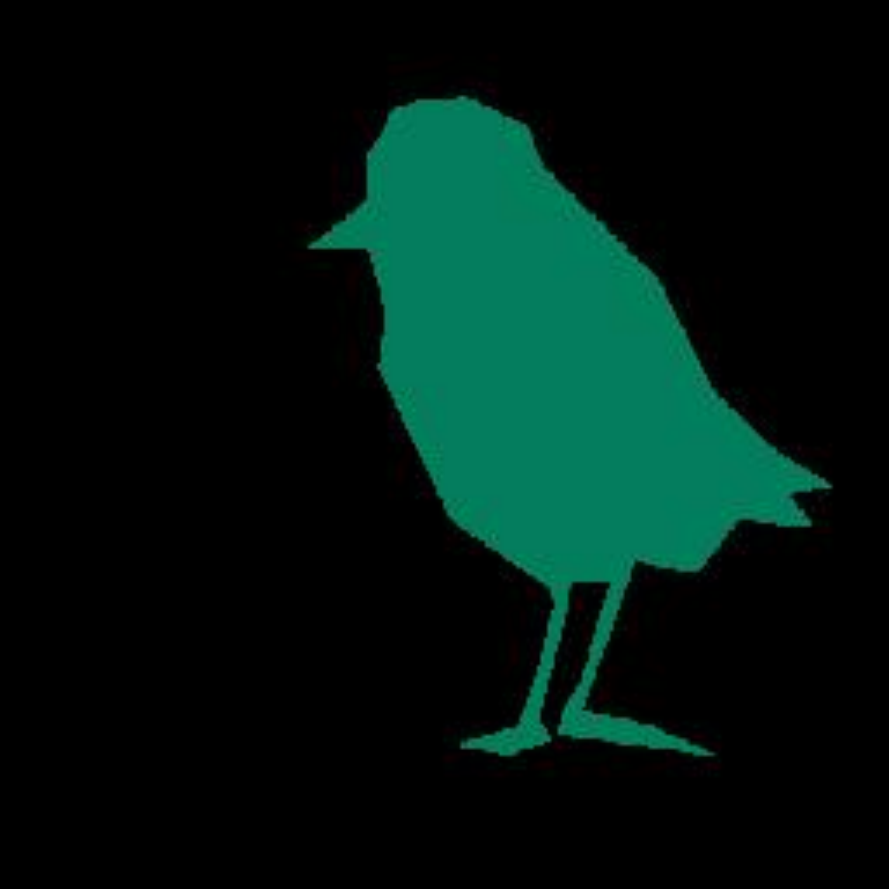} &
        \adjincludegraphics[clip,width=0.1\textwidth,trim={0 0 0 0}]{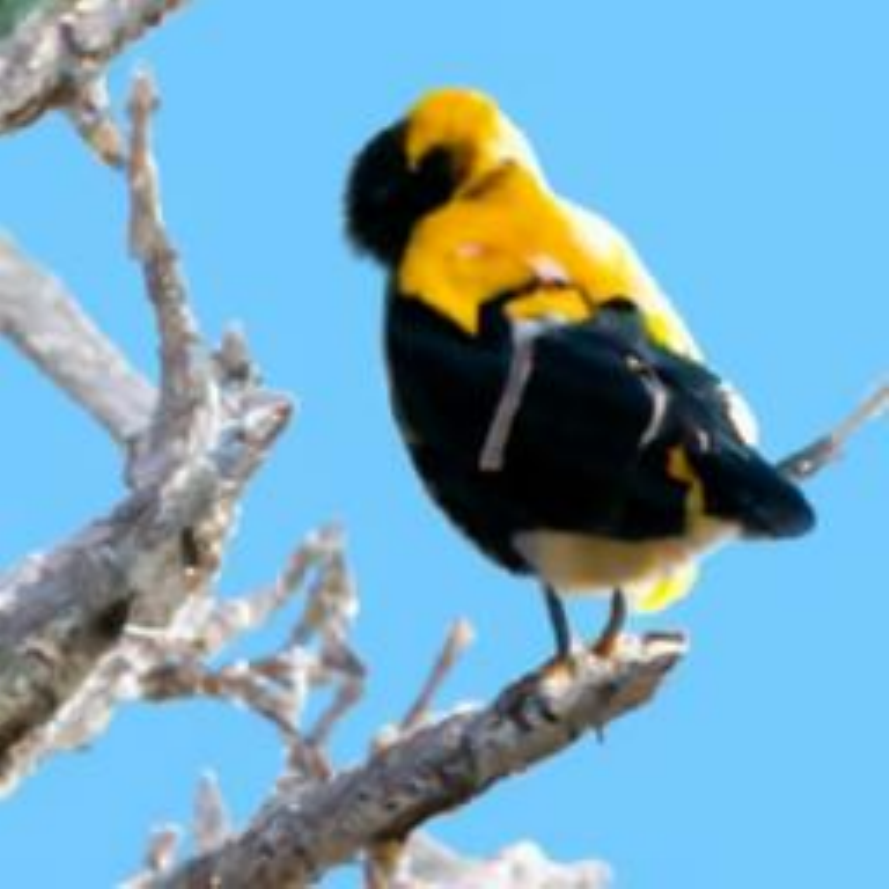} &
        \adjincludegraphics[clip,width=0.1\textwidth,trim={0 0 0 0}]{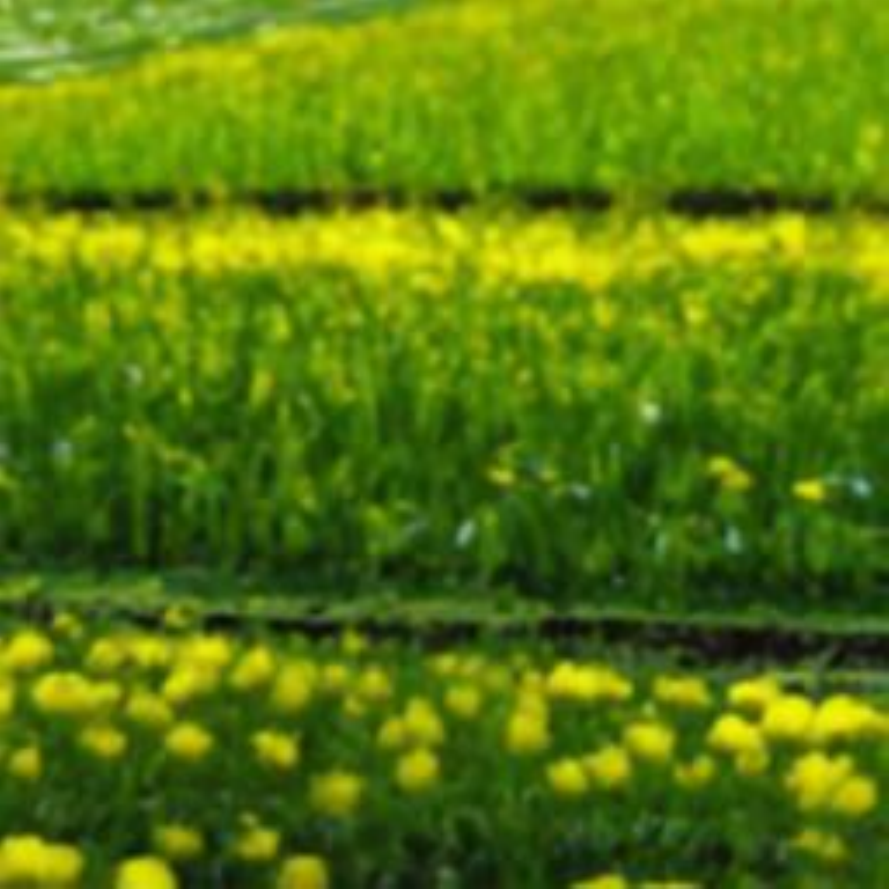} &
        \adjincludegraphics[clip,width=0.1\textwidth,trim={0 0 0 0}]{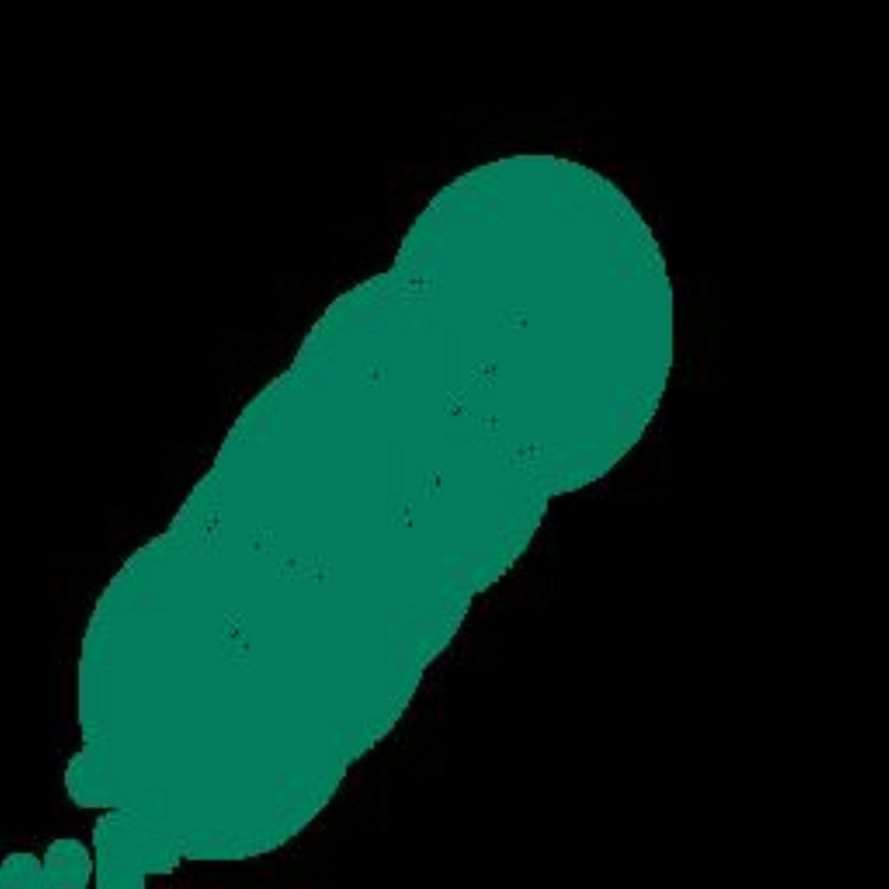} &
        \adjincludegraphics[clip,width=0.1\textwidth,trim={0 0 0 0}]{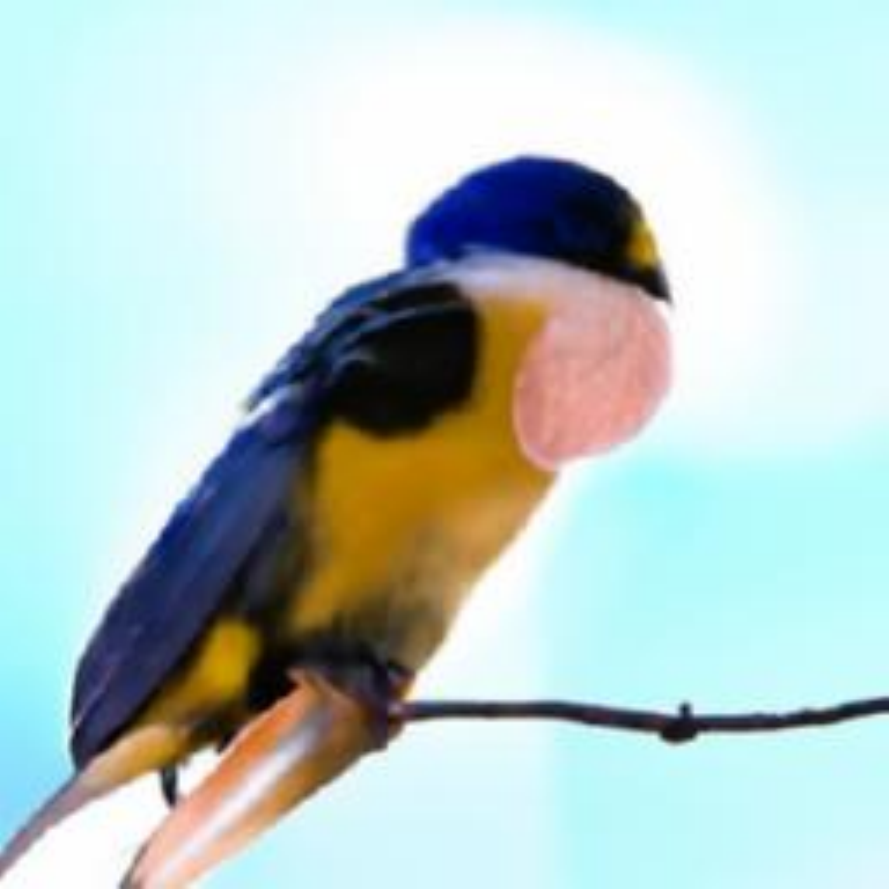} &
        \adjincludegraphics[clip,width=0.1\textwidth,trim={0 0 0 0}]{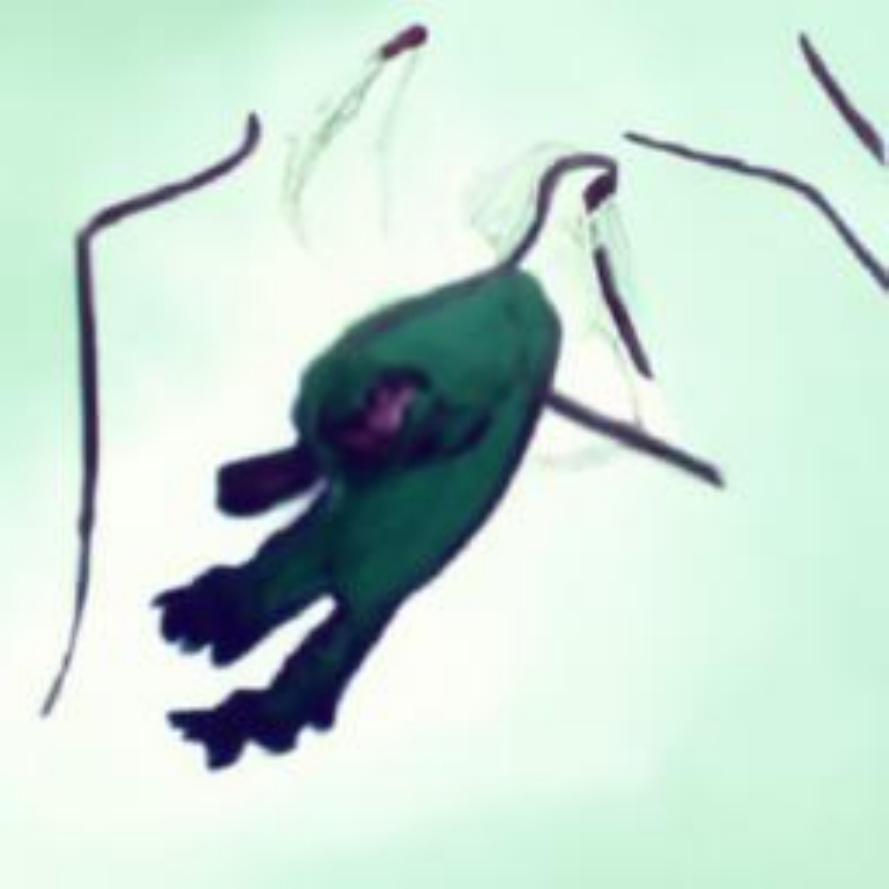} &
        \adjincludegraphics[clip,width=0.1\textwidth,trim={0 0 0 0}]{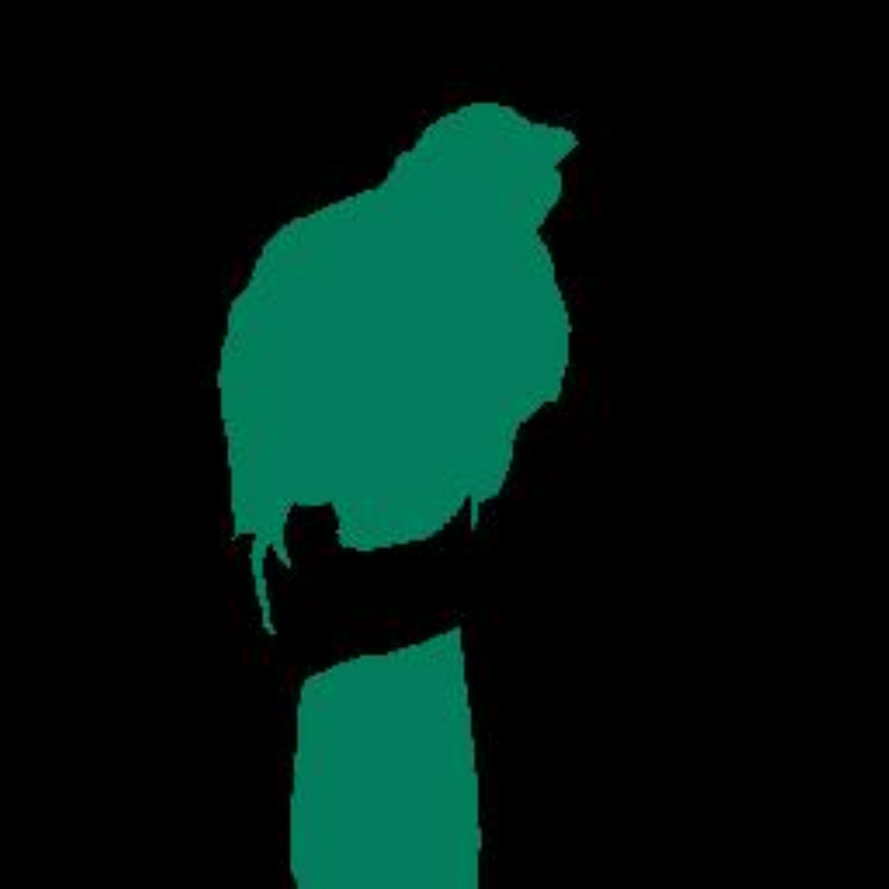} &
        \adjincludegraphics[clip,width=0.1\textwidth,trim={0 0 0 0}]{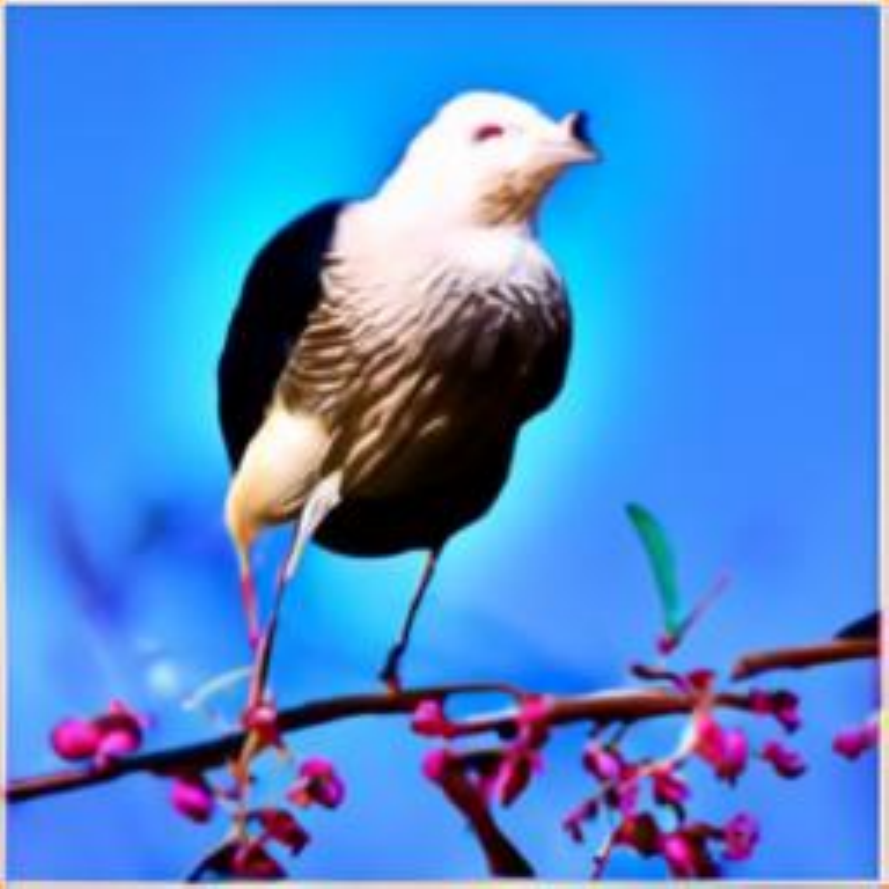} &
        \adjincludegraphics[clip,width=0.1\textwidth,trim={0 0 0 0}]{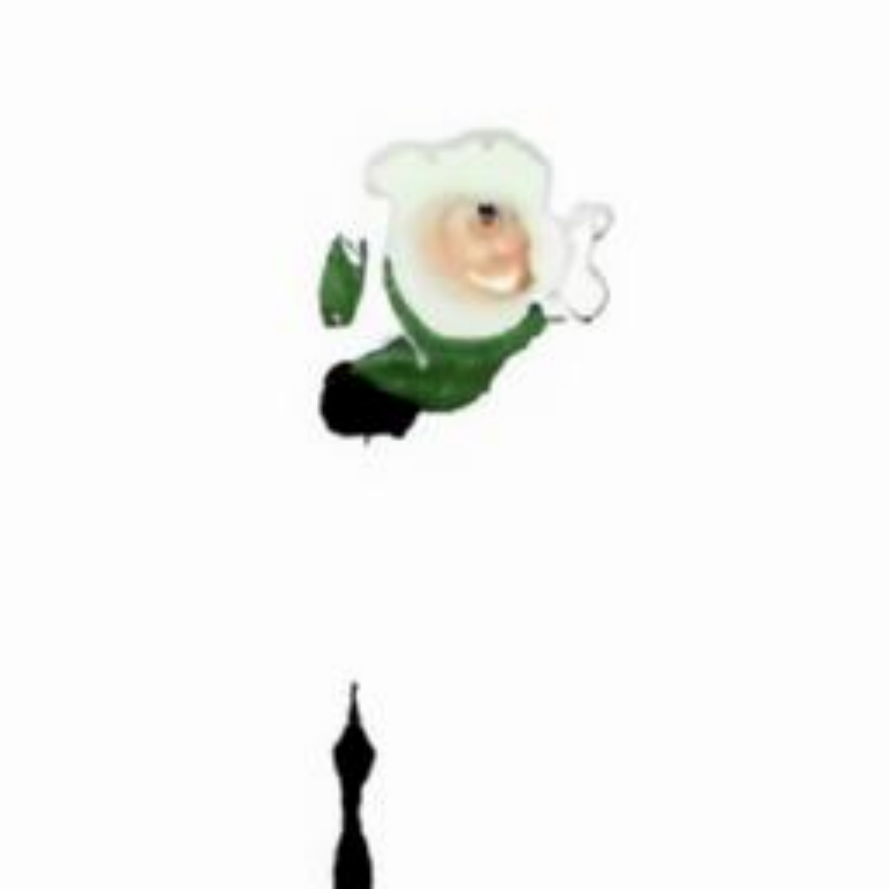}
        \\
        
        \raisebox{1.5\height}{\rotatebox{90}{\scriptsize Bus}} 
        \adjincludegraphics[clip,width=0.1\textwidth,trim={0 0 0 0}]{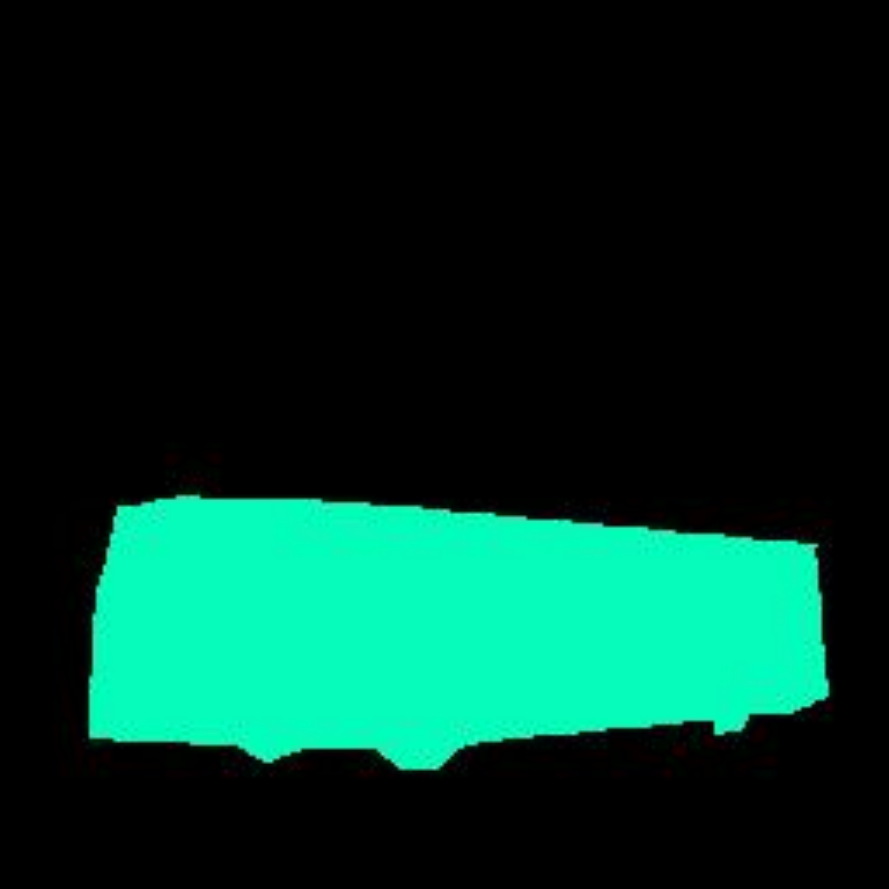} &
        \adjincludegraphics[clip,width=0.1\textwidth,trim={0 0 0 0}]{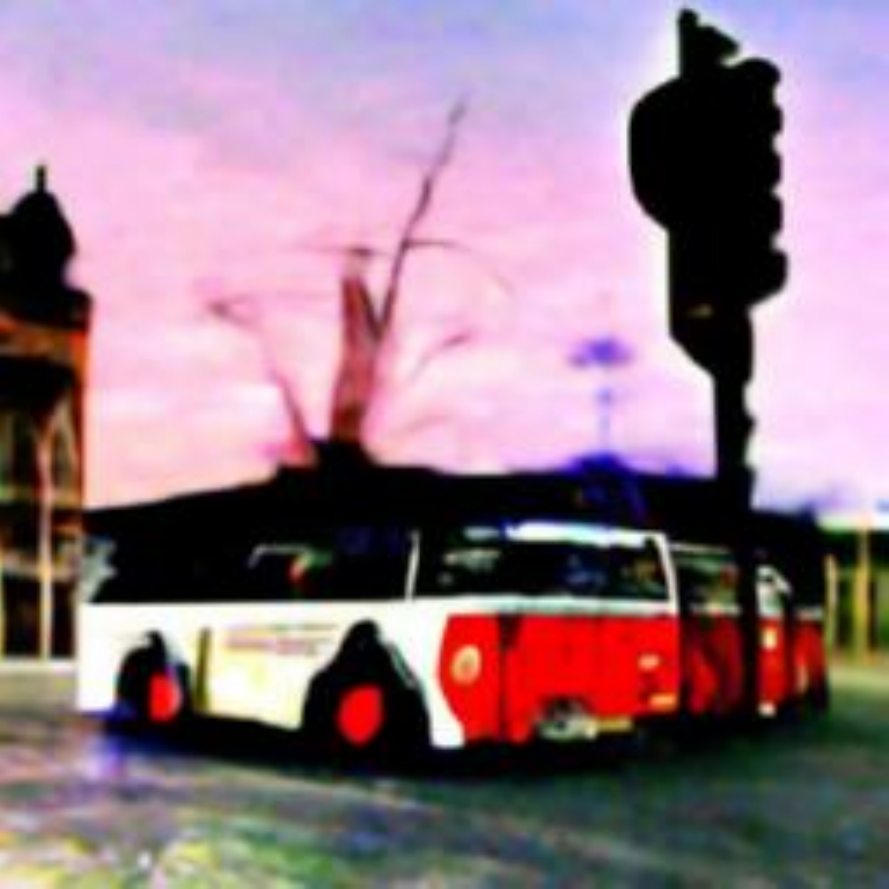} &
        \adjincludegraphics[clip,width=0.1\textwidth,trim={0 0 0 0}]{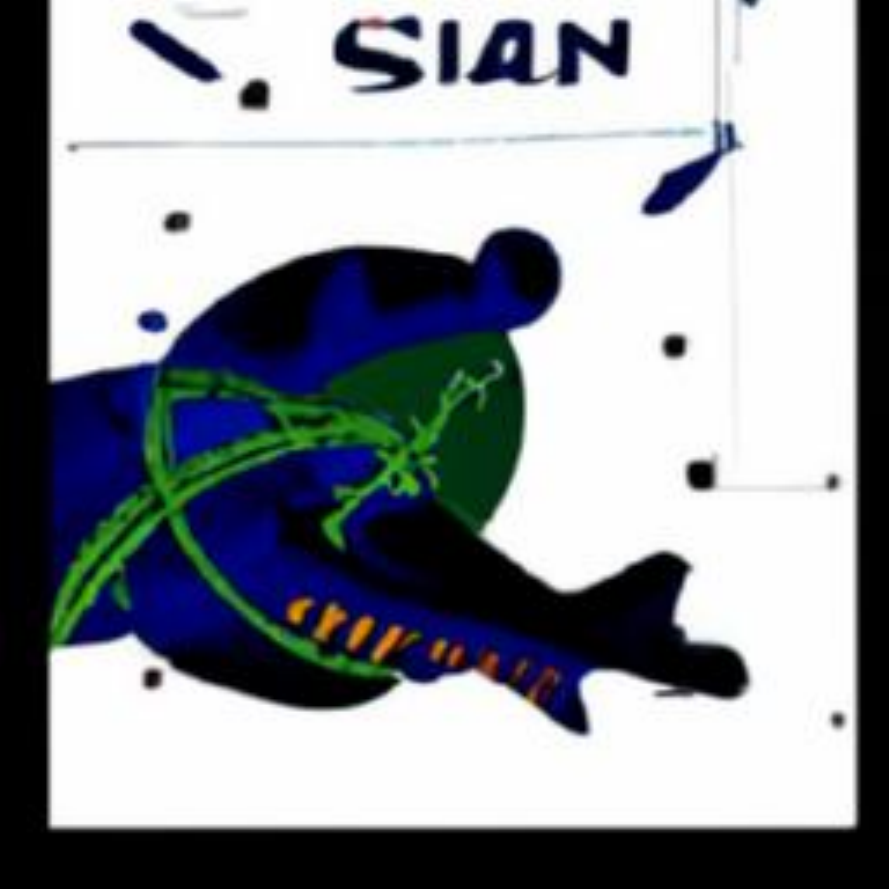} &
        \adjincludegraphics[clip,width=0.1\textwidth,trim={0 0 0 0}]{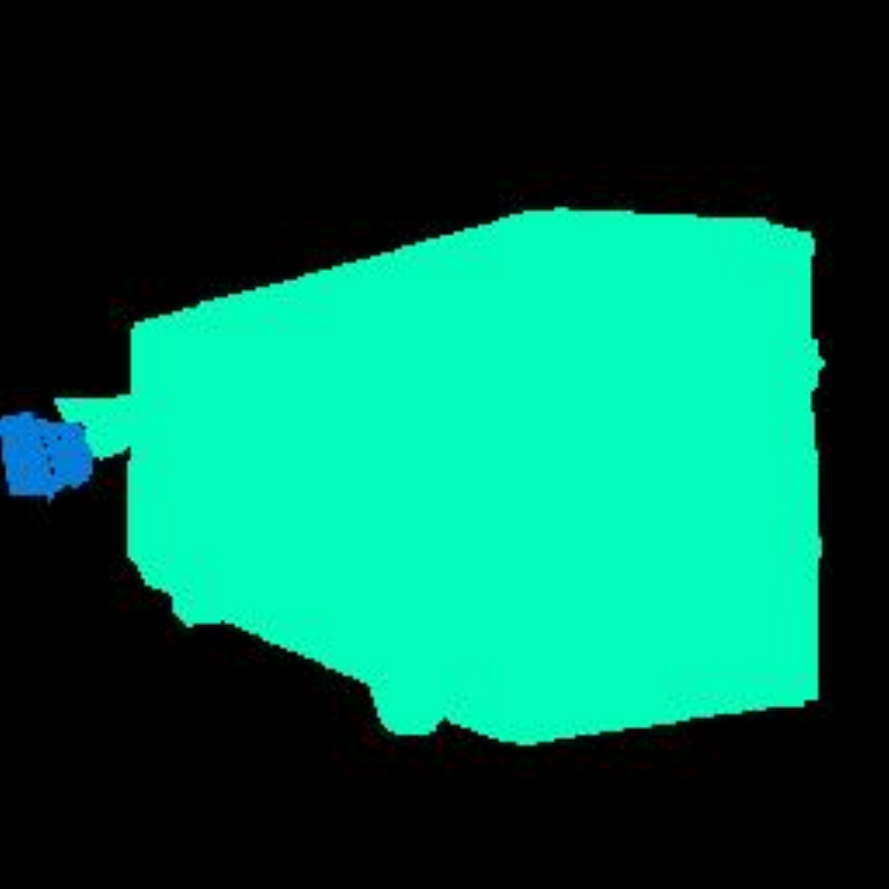} &
        \adjincludegraphics[clip,width=0.1\textwidth,trim={0 0 0 0}]{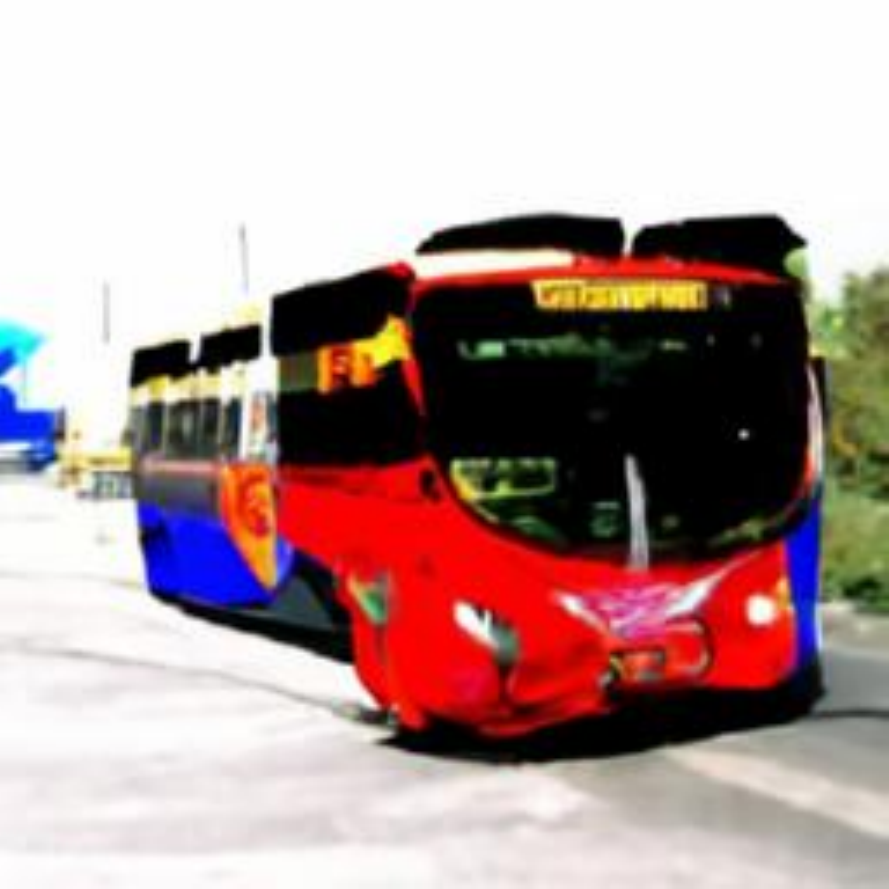} &
        \adjincludegraphics[clip,width=0.1\textwidth,trim={0 0 0 0}]{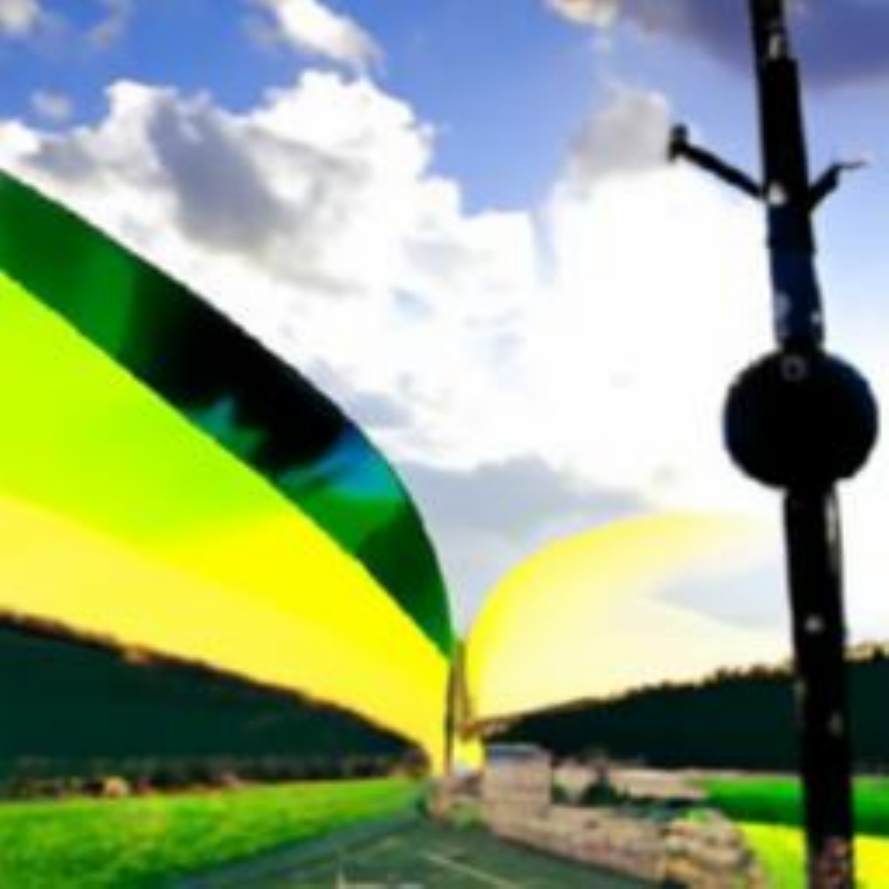} &
        \adjincludegraphics[clip,width=0.1\textwidth,trim={0 0 0 0}]{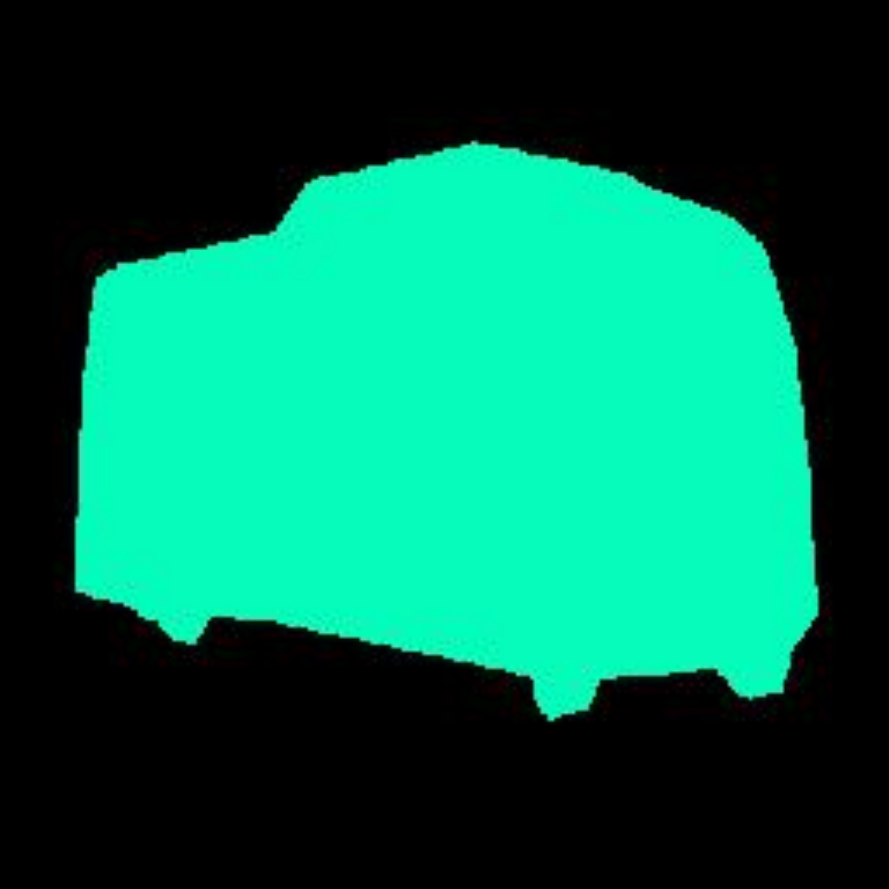} &
        \adjincludegraphics[clip,width=0.1\textwidth,trim={0 0 0 0}]{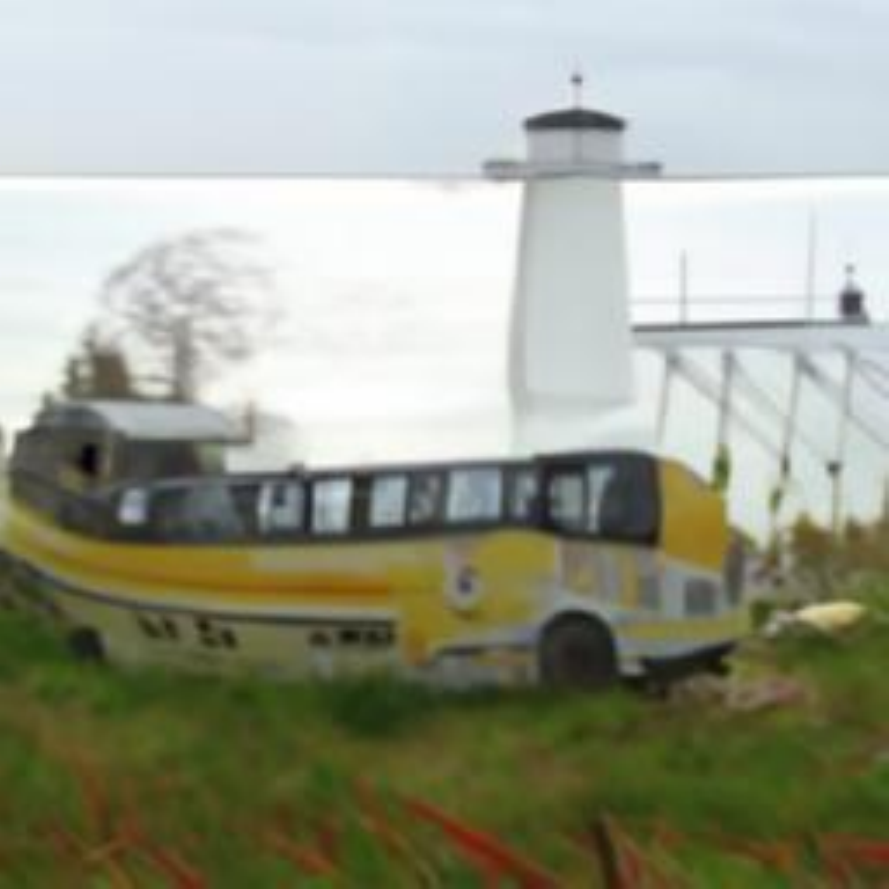} &
        \adjincludegraphics[clip,width=0.1\textwidth,trim={0 0 0 0}]{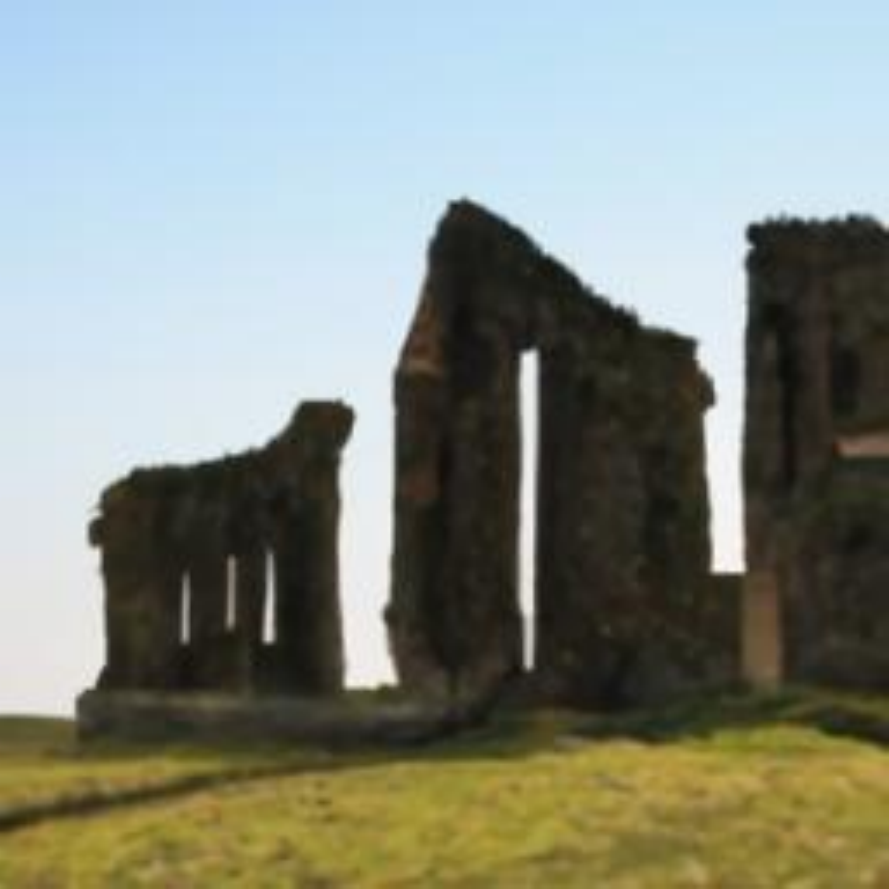}
        \\
        
        \raisebox{1.9\height}{\rotatebox{90}{\scriptsize Cat}} 
        \adjincludegraphics[clip,width=0.1\textwidth,trim={0 0 0 0}]{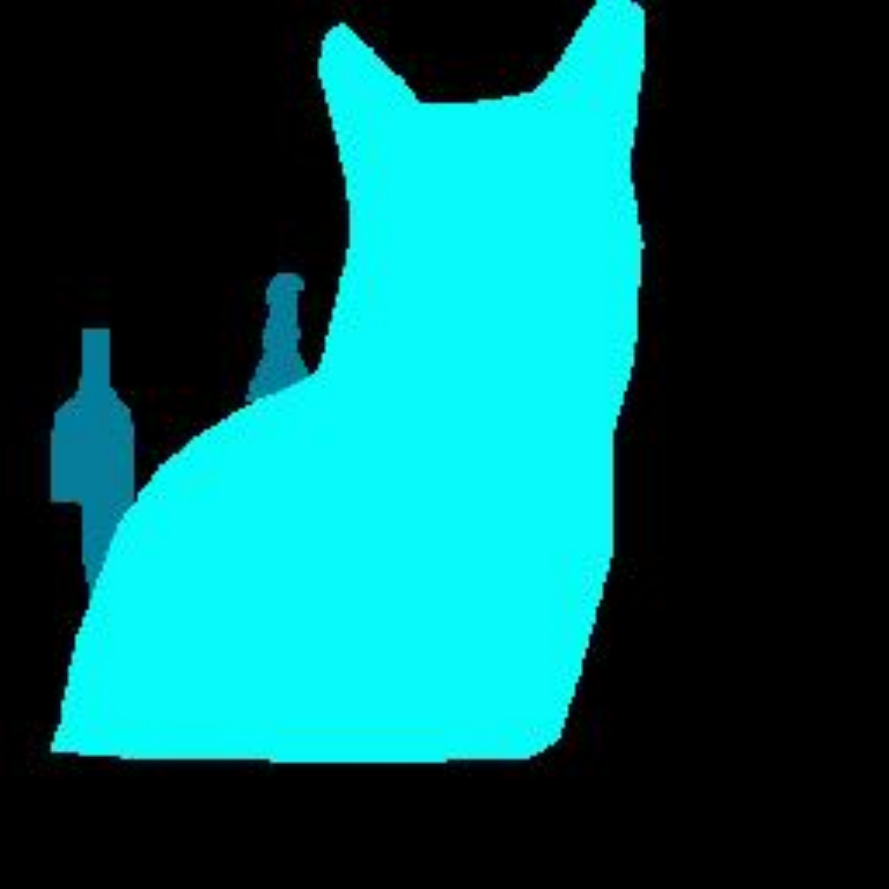} &
        \adjincludegraphics[clip,width=0.1\textwidth,trim={0 0 0 0}]{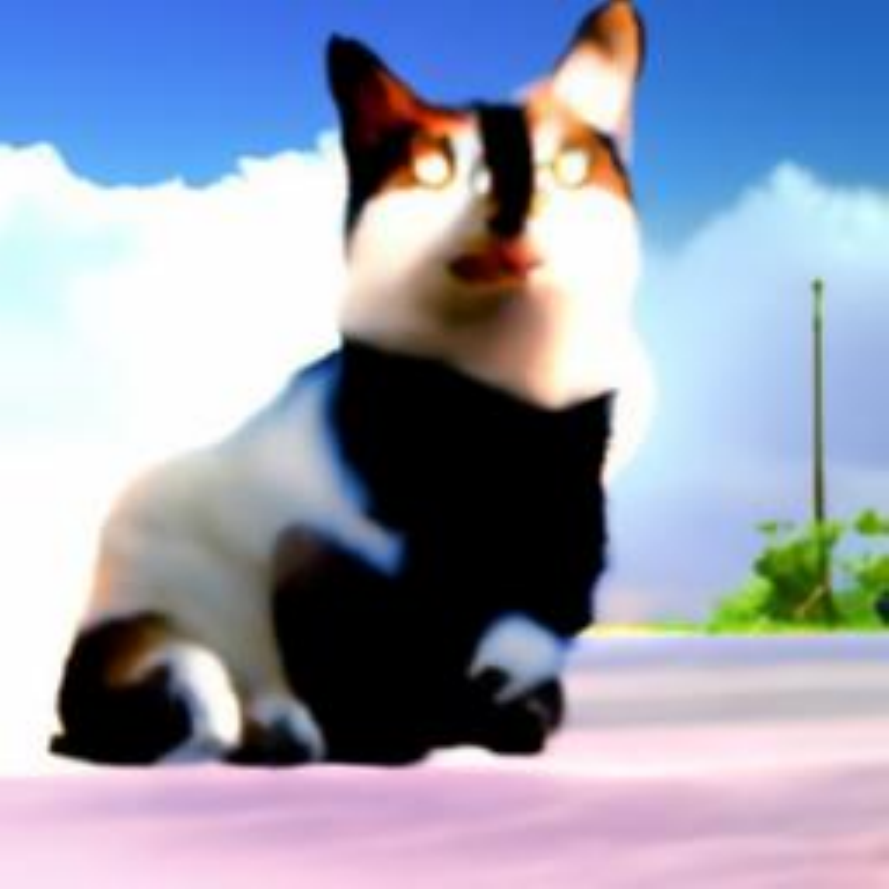} &
        \adjincludegraphics[clip,width=0.1\textwidth,trim={0 0 0 0}]{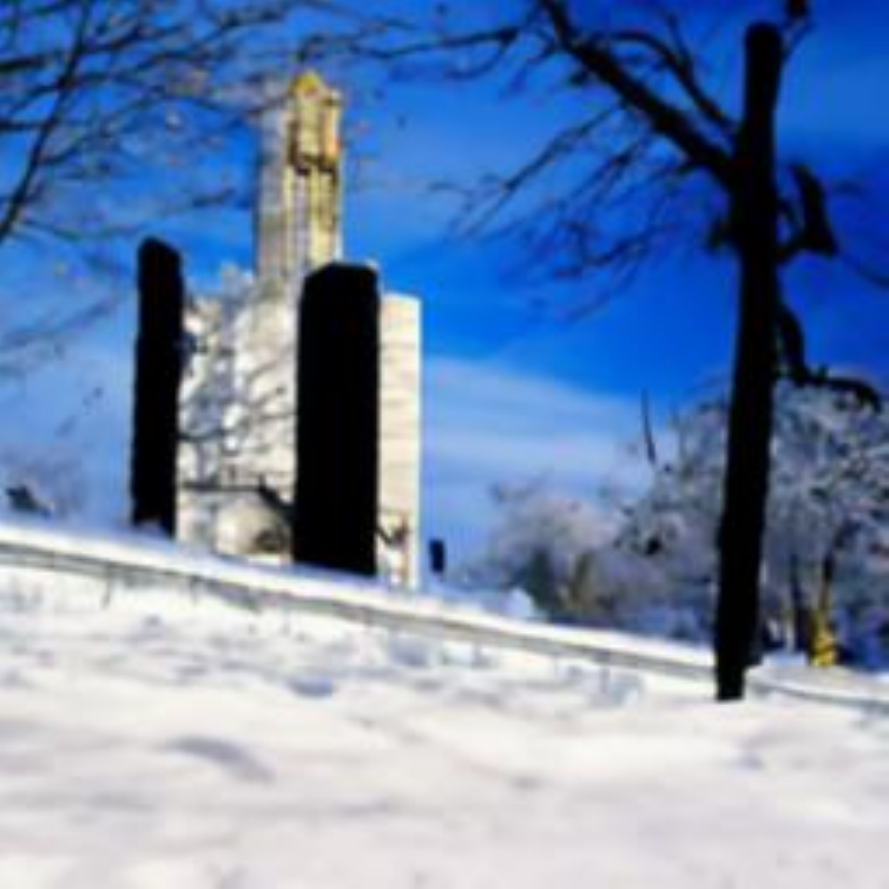} &
        \adjincludegraphics[clip,width=0.1\textwidth,trim={0 0 0 0}]{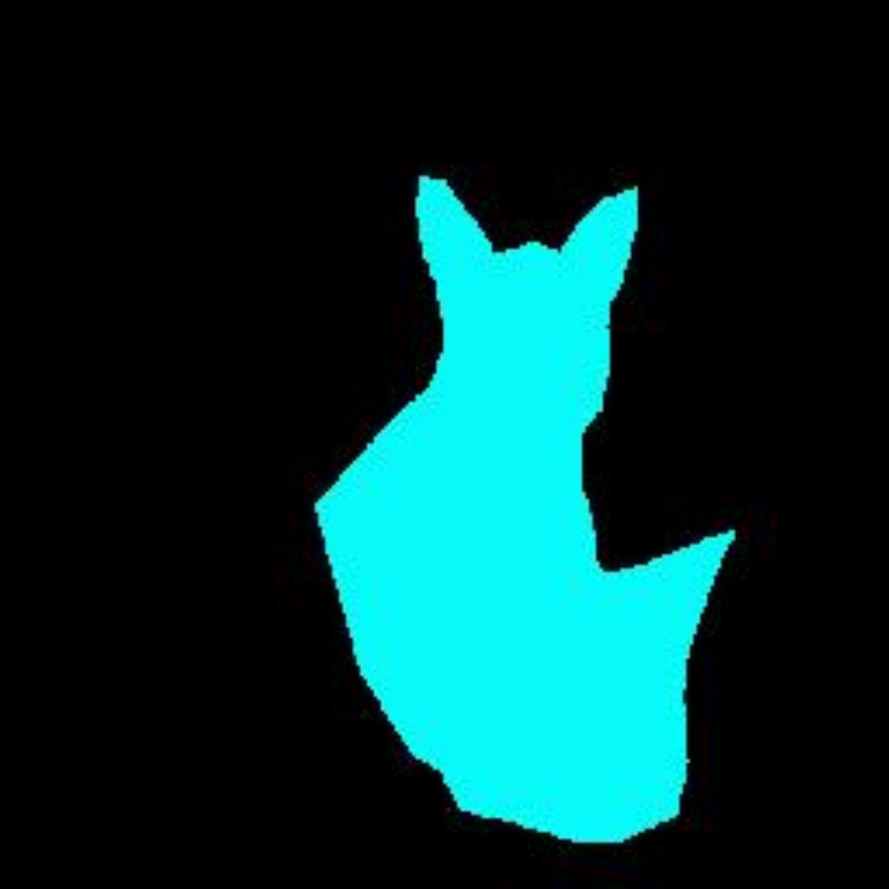} &
        \adjincludegraphics[clip,width=0.1\textwidth,trim={0 0 0 0}]{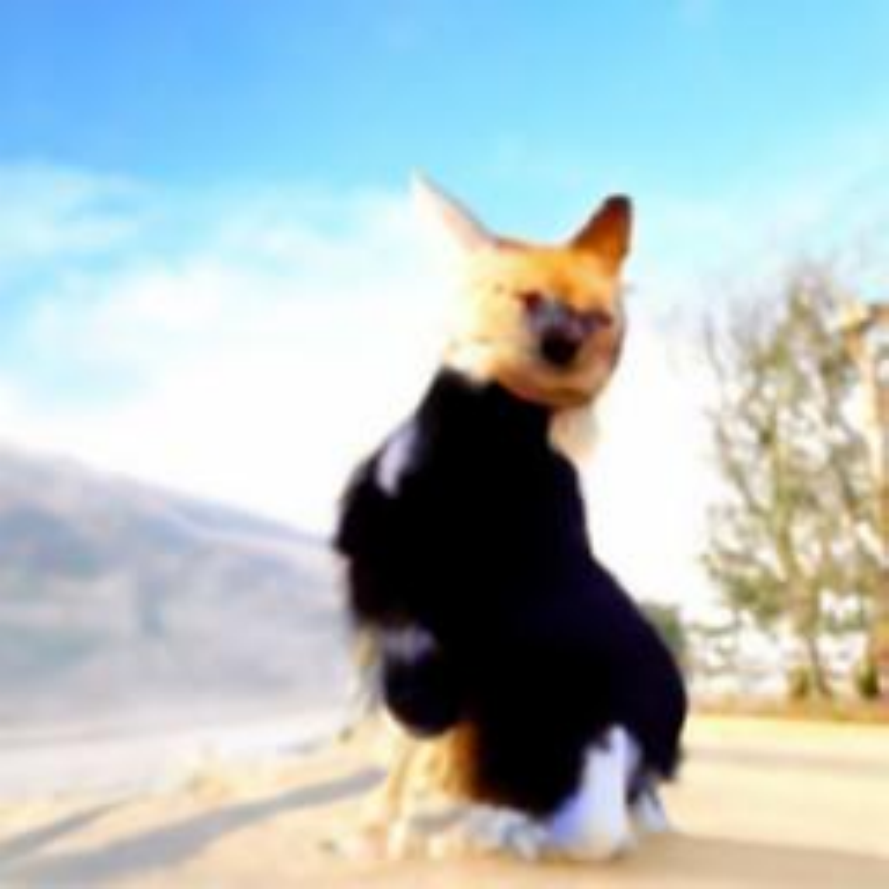} &
        \adjincludegraphics[clip,width=0.1\textwidth,trim={0 0 0 0}]{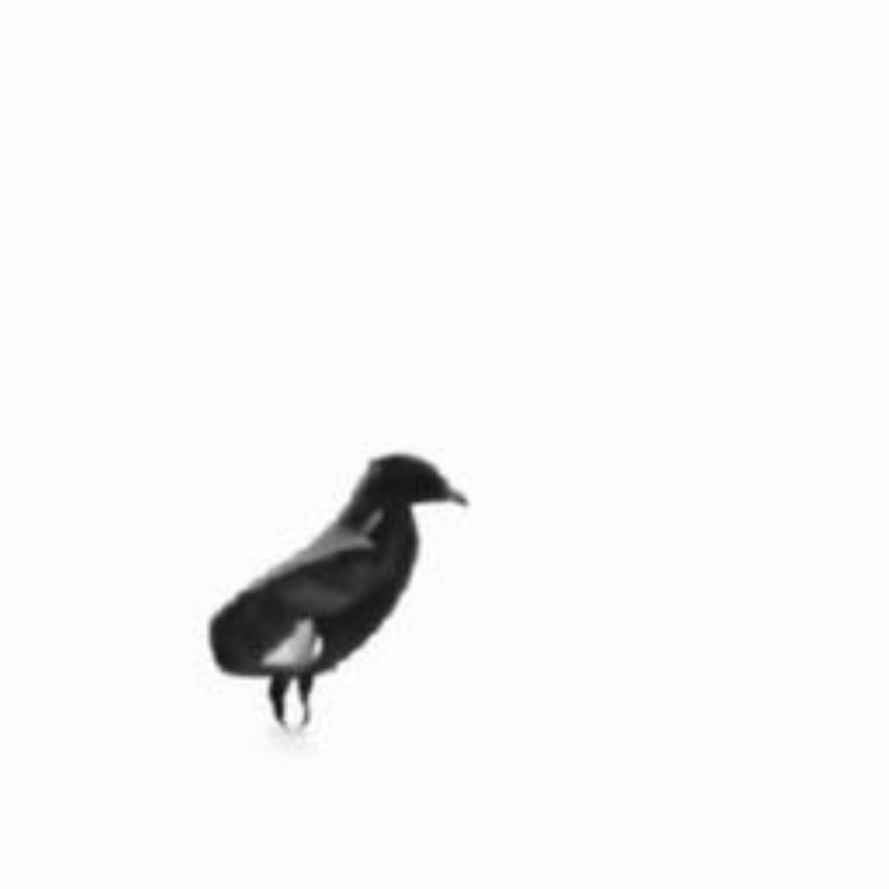} &
        \adjincludegraphics[clip,width=0.1\textwidth,trim={0 0 0 0}]{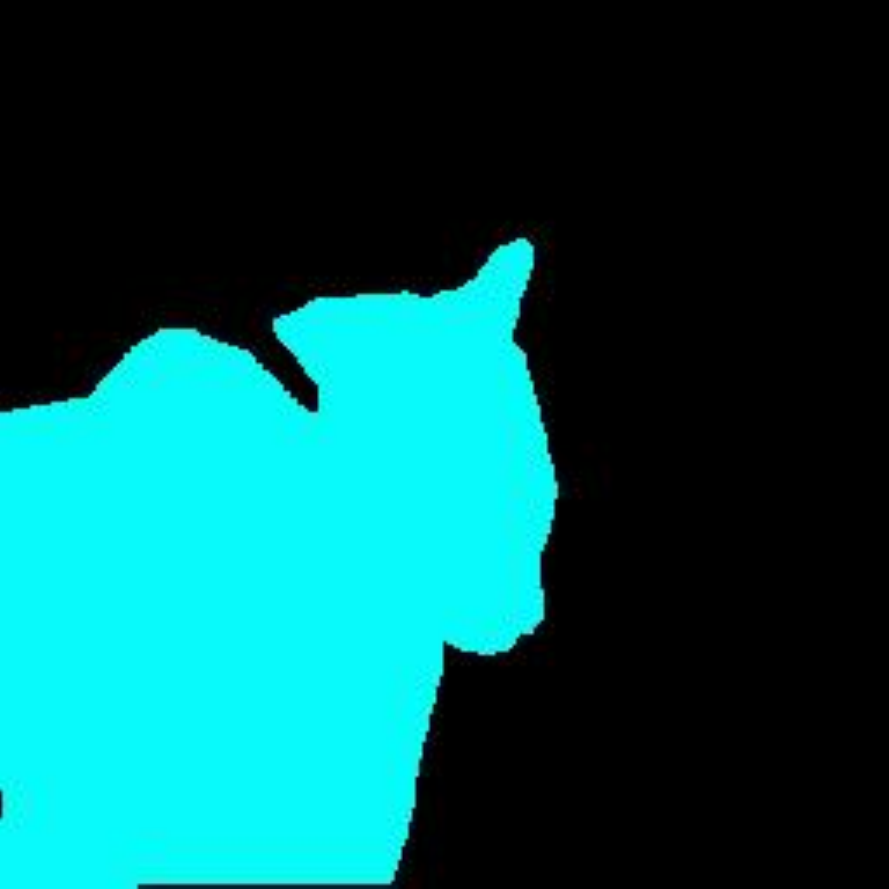} &
        \adjincludegraphics[clip,width=0.1\textwidth,trim={0 0 0 0}]{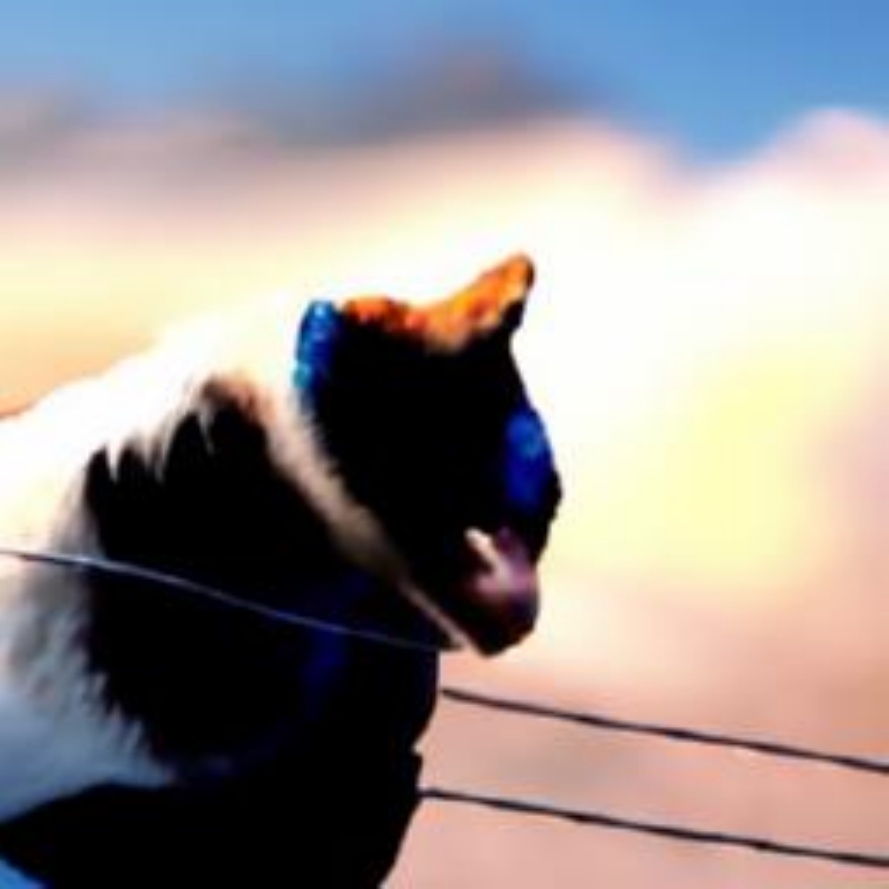} &
        \adjincludegraphics[clip,width=0.1\textwidth,trim={0 0 0 0}]{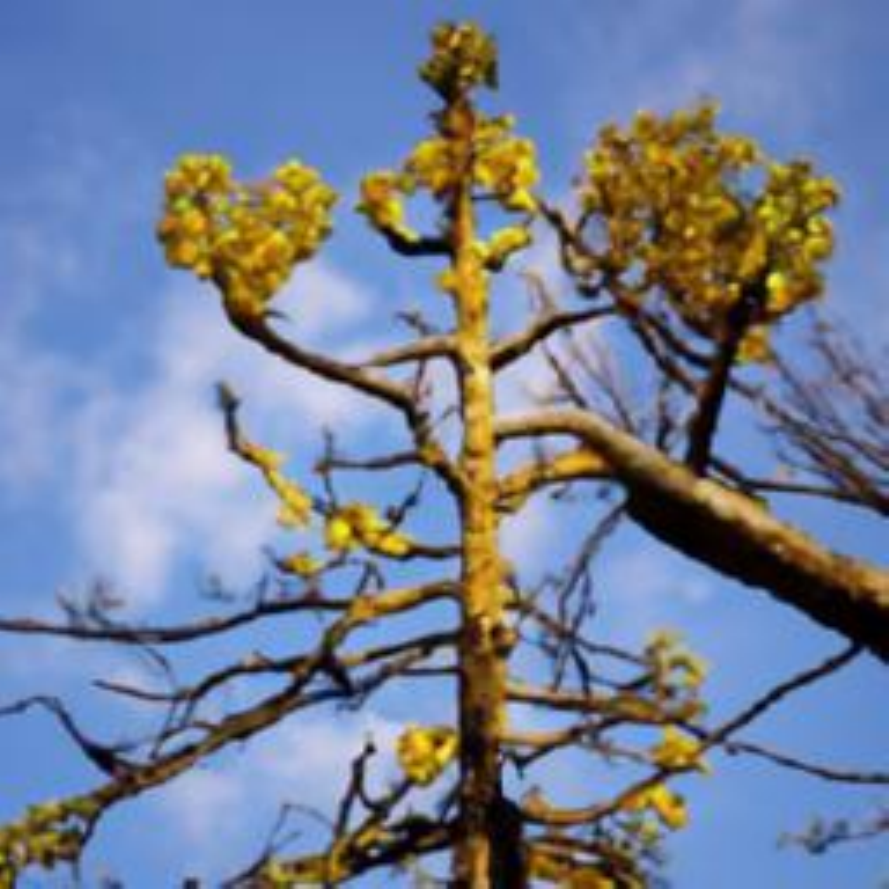}
        \\
        
        \raisebox{0.2\height}{\rotatebox{90}{\scriptsize Dining table}} 
        \adjincludegraphics[clip,width=0.1\textwidth,trim={0 0 0 0}]{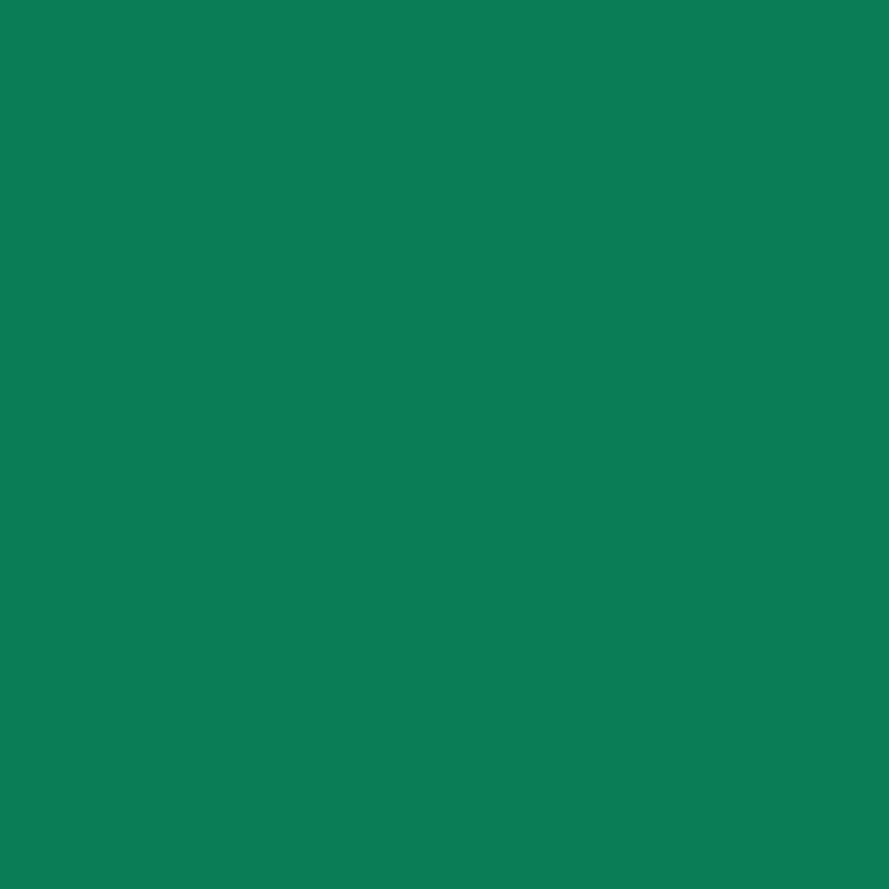} &
        \adjincludegraphics[clip,width=0.1\textwidth,trim={0 0 0 0}]{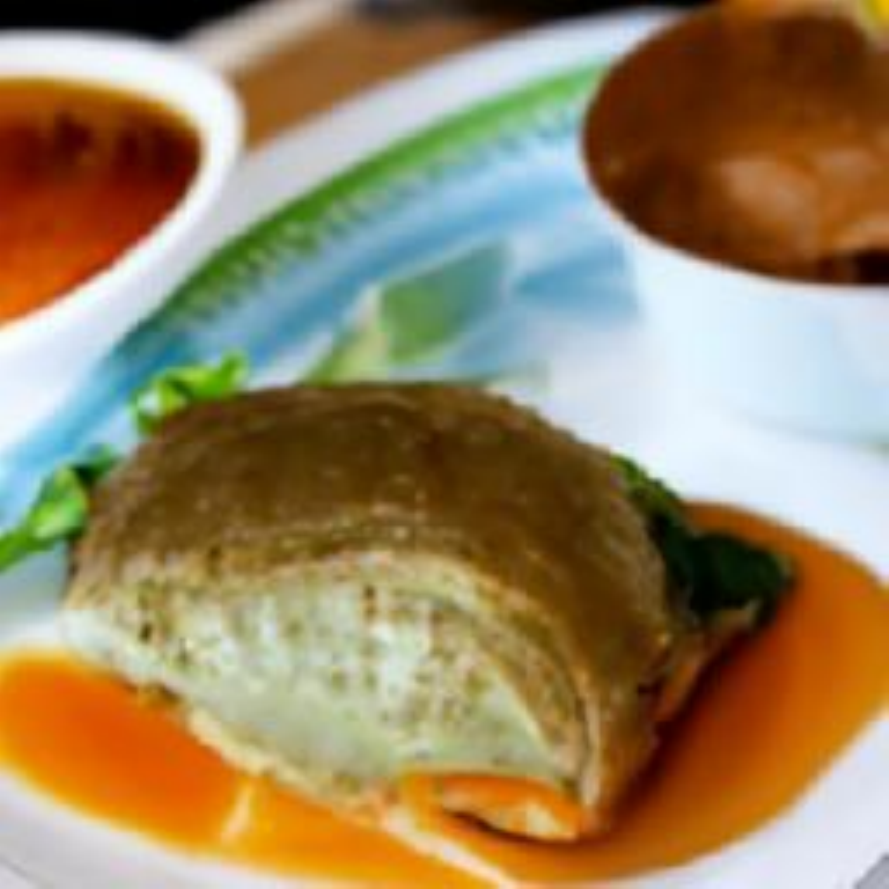} &
        \adjincludegraphics[clip,width=0.1\textwidth,trim={0 0 0 0}]{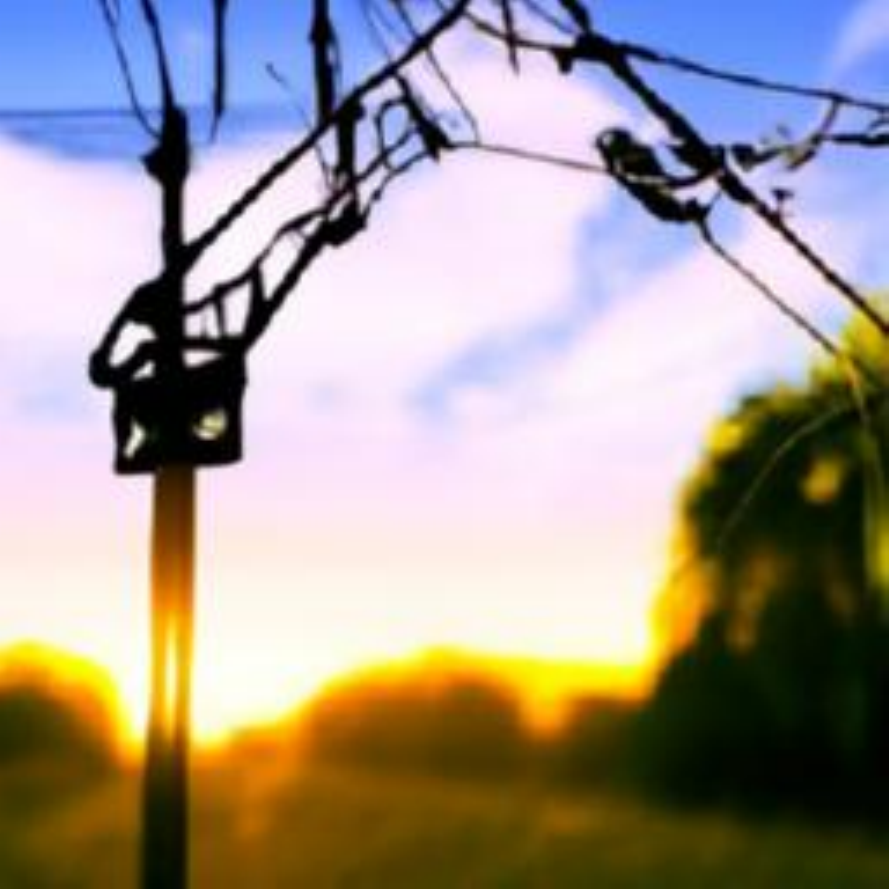} &
        \adjincludegraphics[clip,width=0.1\textwidth,trim={0 0 0 0}]{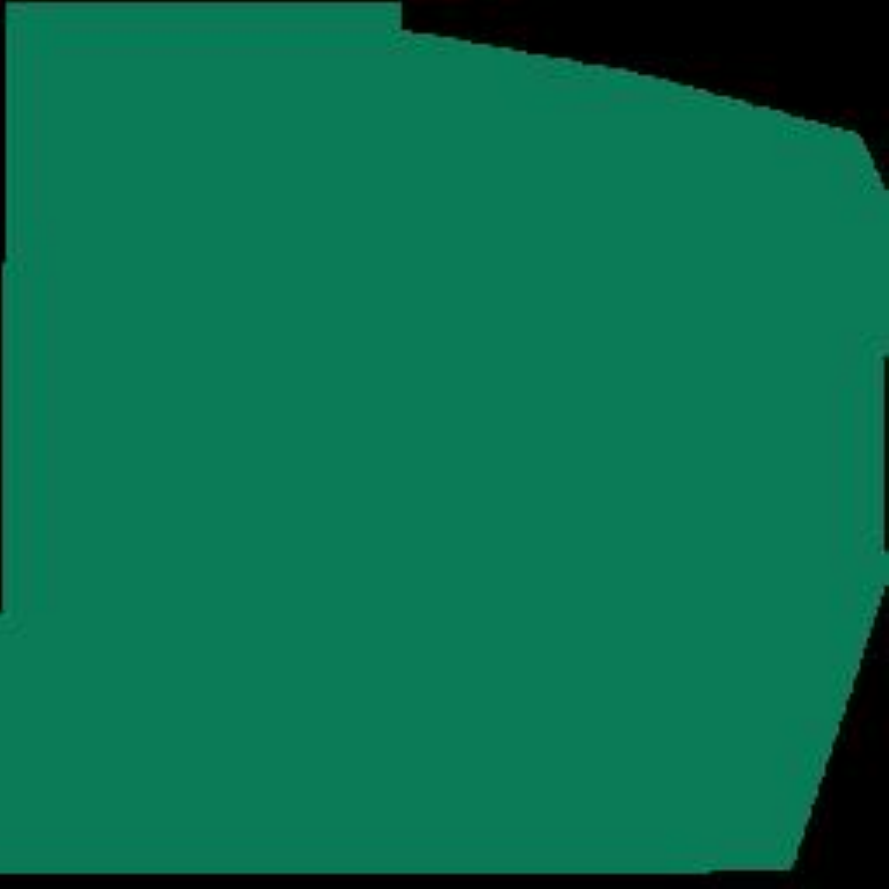} &
        \adjincludegraphics[clip,width=0.1\textwidth,trim={0 0 0 0}]{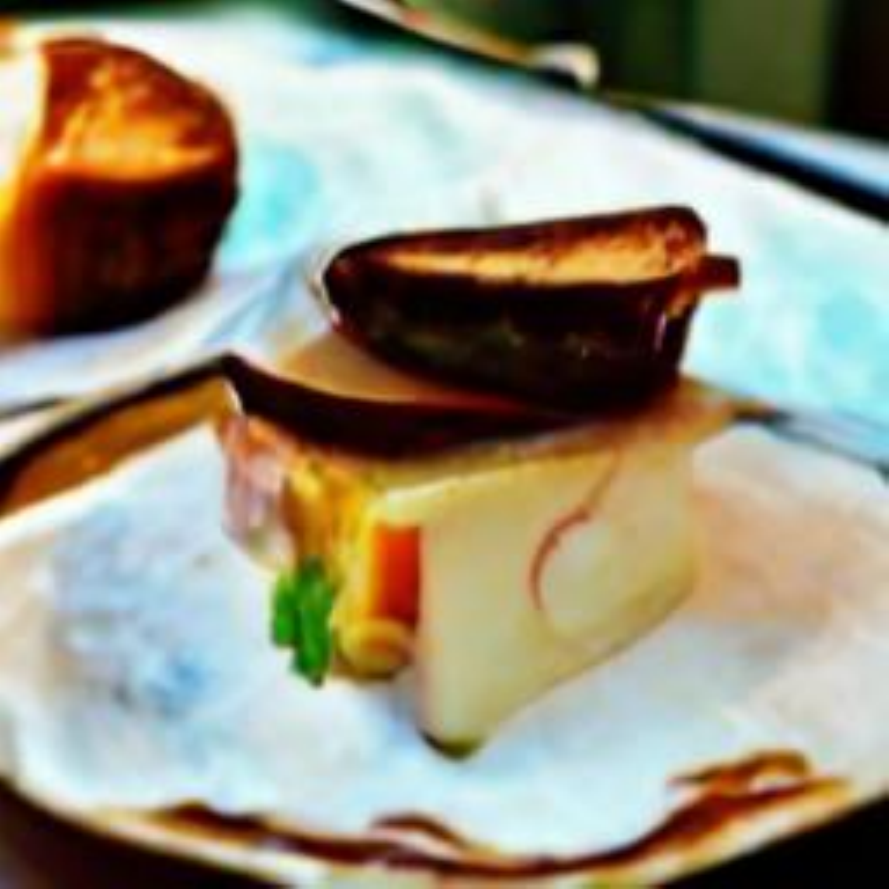} &
        \adjincludegraphics[clip,width=0.1\textwidth,trim={0 0 0 0}]{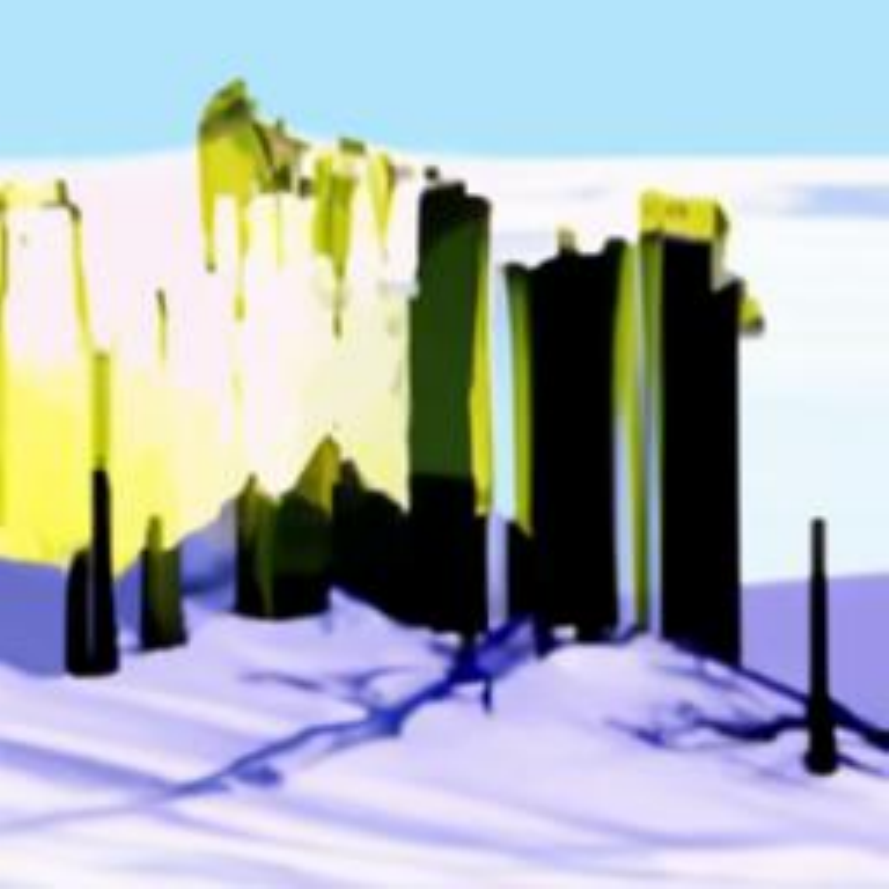} &
        \adjincludegraphics[clip,width=0.1\textwidth,trim={0 0 0 0}]{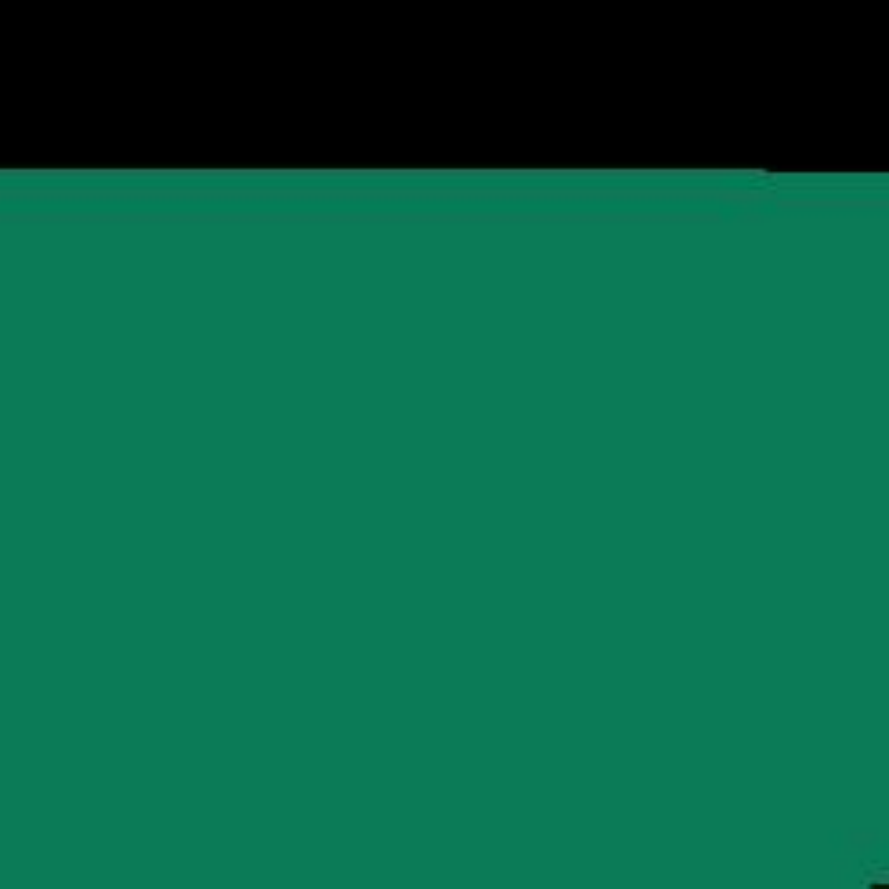} &
        \adjincludegraphics[clip,width=0.1\textwidth,trim={0 0 0 0}]{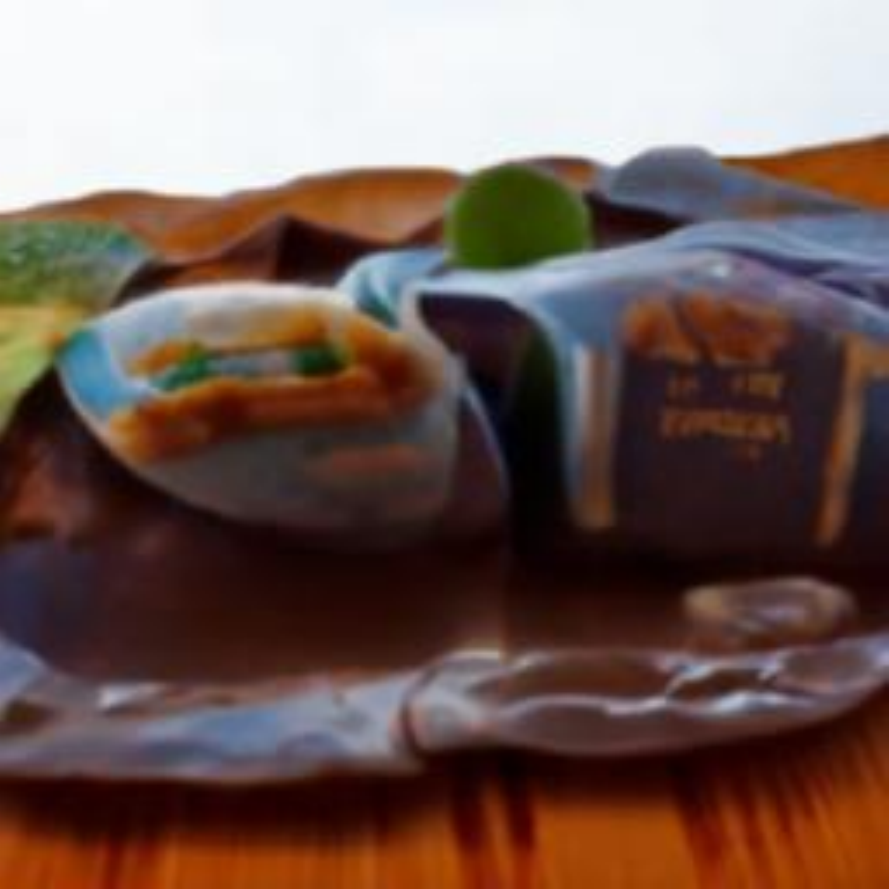} &
        \adjincludegraphics[clip,width=0.1\textwidth,trim={0 0 0 0}]{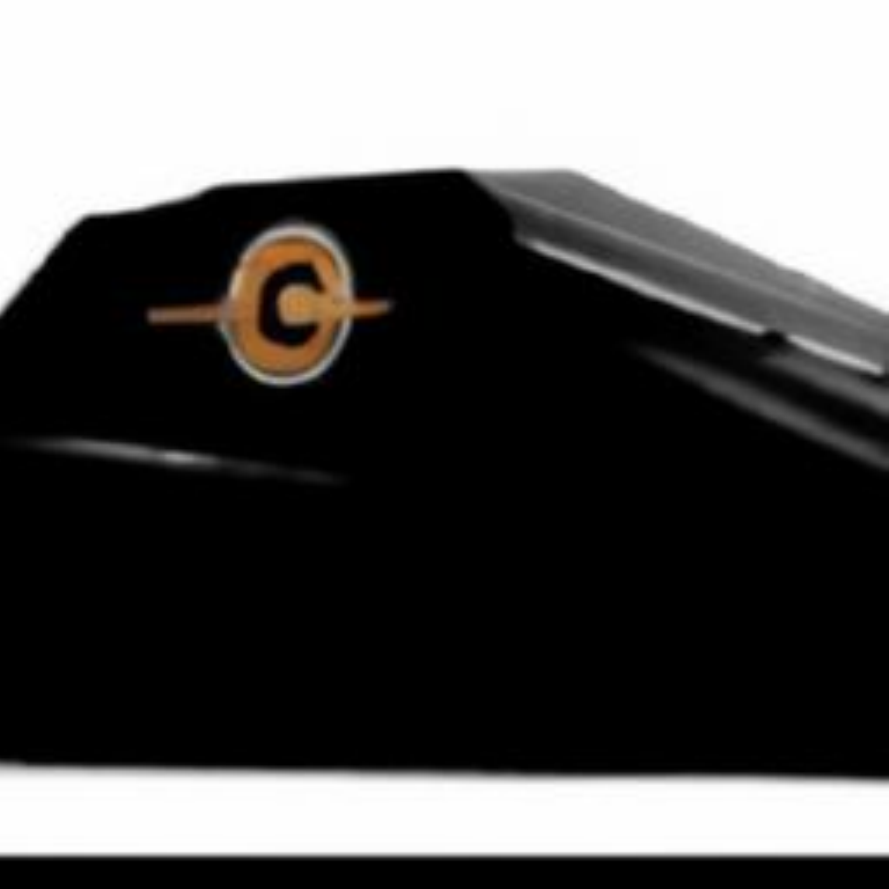}
        \\
        
        \scriptsize ~~~\begin{tabular}{@{}c@{}}Segmentation \\ map\end{tabular} & \scriptsize  PPAP (ours) & \scriptsize  na\"ive & \scriptsize  \begin{tabular}{@{}c@{}}Segmentation \\ map\end{tabular} & \scriptsize PPAP (ours) & \scriptsize na\"ive &
        \scriptsize  \begin{tabular}{@{}c@{}}Segmentation \\ map\end{tabular} & \scriptsize PPAP (ours) & \scriptsize na\"ive 
    \end{tabular}
    \caption{Qualitative results of in-class variations by guiding GLIDE with DeepLabv3 semantic segmentation.
    Our framework PPAP succeeds in the guidance of diffusion, but the na\"ive off-the-shelf model fails.
    }
    \label{apx:fig_glide_seg_class}
\end{figure*}

More qualitative results in guiding GLIDE with semantic segmentation are illustrated in Fig.~\ref{apx:fig_glide_seg} and Fig.~\ref{apx:fig_glide_seg_class}. As illustrated in Fig.~\ref{apx:fig_glide_seg}, our PPAP framework can generate images suited to given segmentation maps. It shows that our PPAP framework is capable of both semantic-level guidance and pixel-level guidance at once. We also provide in-class multiple images in Fig.~\ref{apx:fig_glide_seg_class}.

\section{Limitation and Discussion}
\label{apx:sec:limitation}

Guiding GLIDE with our PPAP often generates images with the style of data that trains the diffusion model but not the guidance model.
On the one hand, it means that the guidance model can leverage various data covered by diffusion, but on the other hand, it can be interpreted that the guidance model cannot perfectly guide the diffusion model to the data it has learned.
Considering this, addressing the train dataset mismatch between the diffusion model and the off-the-shelf model can be the future direction of this work.

Also, we only use the guidance models that take a single image as input.
There are several publicly available off-the-shelf models which take not only the image but also other inputs.
With designing suitable knowledge transfer loss and guidance loss, collaborating with these off-the-shelf models will produce various applications.
Therefore, applying it to various applications can be future work.



\section{Acknowledgement}
We thank Minsam Kim, Minhyuk Choi, Eunsam Lee, and Jaeyeon Park for the inspiration for this work.

\clearpage
\newpage
\onecolumn
\twocolumn

\end{document}